\documentclass{article} 
\usepackage{iclr2025_conference,times}

\usepackage{amsmath,amsfonts,bm}

\def\eqref#1{equation~\ref{#1}}

\def\1{\bm{1}}

\DeclareMathAlphabet{\mathsfit}{\encodingdefault}{\sfdefault}{m}{sl}
\SetMathAlphabet{\mathsfit}{bold}{\encodingdefault}{\sfdefault}{bx}{n}

\usepackage{booktabs}
\usepackage{hyperref}
\usepackage{url}
\usepackage{graphicx}
\usepackage[capitalize,noabbrev]{cleveref}
\usepackage{hyperref}

\usepackage{algorithm}
\usepackage{algorithmic}
\usepackage{bbm}
\usepackage{array}
\usepackage{amsmath}
\usepackage{amssymb}
\usepackage{mathtools}
\usepackage{amsthm}

\iclrfinalcopy
\usepackage[textsize=tiny]{todonotes}

\title{Uncertainty of Vision Medical Foundation Models}

\author{Haoxu Huang\textsuperscript{1}\thanks{Correspondence author}\quad\& Narges Razavian\textsuperscript{2,3} \\
\textsuperscript{1}Center for Data Science,
New York University \\
\textsuperscript{2}Department of Radiology, NYU Grossman School of Medicine\\
\textsuperscript{3}Department of Population Health, NYU Grossman School of Medicine\\
\texttt{hh2740@nyu.edu, Narges.Razavian@nyulangone.org}\\
}

\begin{document}

\maketitle

\begin{abstract}
Accurate uncertainty estimation is essential for machine learning systems deployed in high-stakes domains such as medicine. Traditional approaches primarily rely on probability outputs from trained models ({\it point predictions}), which provide no formal guarantees on prediction coverage and often require additional calibration techniques to improve reliability. In contrast, conformal prediction ({\it region prediction}) offers a principled alternative by generating prediction sets with finite-sample validity guarantees, ensuring that the ground truth is contained within the set at a specified confidence level.

In this study, we explore the impact of pre-training approach, dataset scale and domain on both point and region-level uncertainty quantification, by studying domain-specific vision medical foundation models vs. general domain vision foundation models. We conduct a comprehensive evaluation across foundation models trained on retinal, histopathological, and Chest X-Rays data, applying various calibration techniques. Our results demonstrate that (1) pre-training on higher-quality domain-specific datasets along with self-supervised learning leads to better-calibrated point predictions than general domain pre-training, (2) standard re-calibration methods alone cannot fully mitigate uncertainty discrepancies across models trained on different data sources, (3) domain-specific foundation model can lead to more efficient conformal prediction.

These findings highlight the importance of careful model selection and the integration of both point and region prediction to enhance the reliability and trustworthiness of medical AI systems. Our work underscores the need for a holistic approach to uncertainty quantification in recent development of medical vision foundation model, ensuring robust and interpretable AI-driven decision-making.
\end{abstract}

\section{Introduction}
A fundamental question in machine learning is how well a model can quantify its confidence in predictions, particularly in high-stakes applications where accurate uncertainty estimation is critical. Traditionally, this question is answered by interpreting the probability outputs of learning algorithms. However, such approaches often lead to mis-calibration of the machine learning system, where the predicted probabilities deviate significantly from the true likelihood of correctness on unseen data~\citep{pmlr-v70-guo17a}. This mis-calibration undermines the reliability of confidence estimates and offers no formal guarantees for prediction coverage --- an essential requirement for robust decision-making in real-world scenarios.

Conformal prediction is an alternative method for evaluating uncertainty with exact coverage guarantee with no need to re-train the model. Specifically, through a post-hoc approach, it generates a $(1-\alpha)$ prediction region --- a set $\mathcal{C}^{\alpha}$ that contains ground truth prediction $y$ with probability at least $(1-\alpha)$. Unlike traditional point predictions $\hat{y}$, these prediction regions ensure formal coverage guarantees, making them applicable across a wide range of learning tasks. For instance, in regression, the prediction region can be an interval around $\hat{y}$ that includes the true value; in classification, it can be a set of possible classes containing the ground truth; and in segmentation tasks, it can identify a region of pixels encompassing the true segmentation.

\begin{figure*}[!ht]
\centering
    \includegraphics[width=0.8\linewidth]{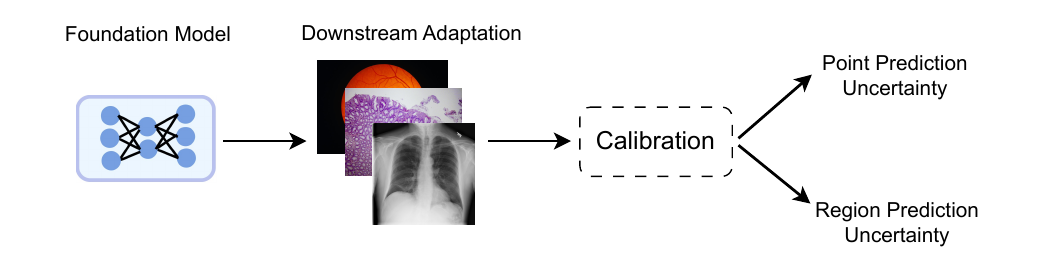}
    \caption{{\bf Uncertainty Evaluation}: initially, a foundation model is trained to adapt to downstream tasks, either with a calibration method or without. Subsequently, the model’s uncertainty is evaluated using both point predictions and region predictions. \label{fig:pipeline}}
\end{figure*}

While the concept of {\it prediction regions} is initially rooted in theoretical and heuristic frameworks suggesting that the world cannot always be represented by single-point predictions~\citep{shafer08conformal}, recent work ~\citep{cresswell2024conformal} have demonstrated its practical value. Notably, region-based AI systems, when combined with human decision-makers, have shown promising potential to improve outcomes in randomized controlled trials. This highlights the transformative potential of incorporating region predictions into critical decision-making pipelines.

In this work, we study the growing importance of accurately understanding model predictions under uncertainty, particularly in the context of high-stakes decision-making. With recent emphasis on building domain specific visual foundation model in medicine \citep{wang_pathology_2024, huang_visuallanguage_2023, chen_towards_2024, Vorontsov2024, zhou2023foundation, dong2024brainjepa, codella2024medimageinsight}, we investigate how foundation models trained on diverse data sources influence predictive uncertainty. Specifically, we explore both point and region prediction under varying conditions and evaluate the potential of model calibration on improving uncertainty estimates across these models. This study aims to shed light on the interplay between foundation model and uncertainty quantification, paving the way for understanding trustworthiness of AI systems in critical domains.

\section{Related Works}
In the case of deep learning model prediction with softmax output, the softmax scores are often used as a proxy of model uncertainty. However, there are many studies~\citep{pmlr-v70-guo17a,minderer2021revisiting,pmlr-v139-bai21c} show that softmax scores are not well calibrated and different calibration techniques are proposed to re-calibrate them. 

While there have been many previous works~\citep{pmlr-v70-guo17a,minderer2021revisiting,pmlr-v97-hendrycks19a} studying uncertainty of deep learning models, their studies are mostly constrained on evaluation with limited data types, model trained on small scale and point prediction uncertainty, with model uncertainty on region prediction under-explored. Additionally, their studies present contradictory results where \citep{pmlr-v70-guo17a} observes that larger neural networks are worse calibrated, \citep{minderer2021revisiting} shows that model architecture families matter more on model uncertainty than model size or pre-training amount, \citep{pmlr-v97-hendrycks19a} claims that better model pre-training improves uncertainty upon model trained from scratch. 

\citet{charles22,angelopoulos2021uncertainty,angelopoulos2024conformal} explored how conformal prediction can be more adaptive on image classification. \citet{angelopoulos2024conformal,quach2024conformal} studied how to leverage conformal prediction concept to segmentation and language model problems with main focus on algorithmic design. However, their work primarily focused on methodological development, with limited exploration of models trained using different pre-training sources and methods.

\section{Preliminary}
\paragraph{Uncertainty Quantification and Calibration}
Building a real-world applicable model under high-stake environment is not only about high model performance on standard benchmarks, the model should also confidently represent its uncertainty of prediction outcomes with human decision-makers in the loop. Yet, it is easy to have a model that achieves high performance and does a poor job on representing its uncertainty~\citep{pmlr-v70-guo17a}. Therefore, previous studies come up with different ways of calibrating the model uncertainty, where they can be categorized by post-hoc method such as Temperature Scaling~\citep{pmlr-v70-guo17a}, regularization method such as Entropy Regularization~\citep{pereyra2017regularizing}, ensembling method such as Deep Ensembling~\citep{Lakshminarayanan2017simple} and bayesian method such as MC Dropout~\citep{pmlr-v48-gal16}. 

Formally, given input $X\in\mathcal{X}$ from input space $\mathcal{X}$, ground truth prediction $Y\in\mathcal{Y}=\{1,...,K\}$ from label space $\mathcal{Y}$, a model with $f(X)=(\hat{Y},\hat{P})$, where $\hat{Y}$ represents model class prediction, $\hat{P}$ represents model probability prediction (confidence), a {\it perfect calibrated} model should fulfill the condition ~\cite{pmlr-v70-guo17a}.
\begin{align}
    \mathbb{P}(\hat{Y}=Y|\hat{P}=p)=p,\quad\forall p\in[0,1]
\end{align}
where it means the model probability prediction should accurately corresponds to its accuracy. While it is impossible to achieve perfect calibration in real world, achieving better calibration means closing the gap between model probability and accuracy. 
\paragraph{Prediction Sets and Conformal Prediction}
While many machine learning problems are framed as single output prediction, there are many cases in real world that giving a set of predictions with correctness guarantee can be more sensible. For example, forecasting weather changes with only one possible outcome is un-informative, a disease progression prediction can potentially have multiple outcomes, etc. Conformal prediction~\citep{vovk05algorithm} is a general framework rather than a specific algorithm, and it is designed to provides prediction sets with {\it coverage guarantee}.

Formally, consider the problem setup where the input $X\in\mathcal{X}$ comes from the input space $\mathcal{X}$, and the ground truth label $Y\in\mathcal{Y}=\{1,...,K\}$ belongs to the label space $\mathcal{Y}$. Our goal is to construct a prediction set $\mathcal{C}(X)$ that satisfies the coverage guarantee:
\begin{align}
    P(Y\in\mathcal{C}(X))\geq1-\alpha
\end{align}
To construct $\mathcal{C}(X)$, we first define a conformal score function $s(x,y)$, which quantifies the uncertainty of a label $y$ for a given input $x$. The conformal score is computed using a calibration dataset, a held-out set of labeled examples that help determine the threshold for uncertainty quantification. Specifically, given a threshold $\hat{q}$, estimated from the calibration dataset, the prediction set $\mathcal{C}(X)$ is formed as:
\begin{align}
    \mathcal{C}_{\hat{q}}(x)=\{y:s(x,y)\leq\hat{q}\}
\end{align}
Here, $\mathcal{C}(X)$ maps each input $X$ to a subset of possible labels, ensuring that the probability of the true label included in the set meets the desired confidence level $1-\alpha$.

The parameter $\alpha$ acts as a risk control factor by adjusting the prediction set size, thereby influencing model uncertainty. A key advantage of conformal prediction is that its coverage guarantee holds under the simple assumption of input exchangeability—without requiring i.i.d. data~\citep{vovk05algorithm}. This independence from assumptions about the underlying model $f$ or data distribution~\cite{anastasios21gentle} makes it especially suited for black-box uncertainty quantification in deep learning.

\paragraph{Efficiency of Conformal Prediction} 
In the point prediction uncertainty calculation method, usually a separate quantitative measure - such as Expected Calibration Error (ECE) ~\citep{naeini15ece}, Brier Score (BS) ~\citep{brier50verf} and Negative Log-Likelihood (NLL) - needs to be calculated. However, in conformal prediction, uncertainty is directly measured by the size of its prediction set: Smaller prediction set represents more informative conformal predictor. Additionally, the empirical coverage from the validation data is calculated to verify that the expected coverage is achieved under no violation of exchangeability. e.g. $\alpha=0.05$ should ideally produce empirical coverage with $\geq95\%$ correctness prediction set. The ideal conformal predictor should provide a prediction set that is small with easy examples, and relatively larger with harder examples, such that it represents uncertainty of its prediction.

\section{Method}
\paragraph{Problem Setting}
The main objective of this study is to understand the impact of domain specific pre-training in uncertainty estimation for vision foundation models in medicine. This study mainly focus on disease classification and our experimental setups can be easily extended to other use-cases such as segmentation~\citep{MedSAM}, or report generation~\citep{quach2024conformal}, among others. Additionally, since this study focuses on evaluating the uncertainty of foundation models based on their pre-trained weights, we use linear probing as our primary evaluation protocol.

We focus on linear probing the foundation models, rather than full end-to-end finetuning for two reasons: only training the linear classification layer reduces the risk of over-fitting on the likelihood by cross-entropy loss, which is the known cause of overconfidence on model uncertainty~\citep{wei2022logitnorm,wang2021rethinking}. Further, recent released medical foundation models often restrict access to only the generated features, withholding model weights to prevent the privacy issue in medical data~\citep{yang2024advancingmultimodalmedicalcapabilities,googlehealth_imaging_research_zenodo}. 

After model downstream training, we evaluate uncertainty with point-prediction metrics (ECE, BS, NLL --- \Cref{apd:metrics}) and region-prediction metrics (Empirical Coverage, Set Size) via Least Ambiguous set-valued Classifier (LAC) \citep{Sadinle2016LeastAS} and Regularized Adaptive Prediction Sets (RAPS) \citep{angelopoulos2021uncertainty}. We further show standard performance metrics (accuracy, balanced accuracy, AUROC, AUPRC) in \Cref{apd:model_perf}. Finally, we assess whether re-calibration techniques can resolve calibration gaps after a foundation model has been pre-training. The evaluation pipeline is shown in \Cref{fig:pipeline}, and further details are provided in \Cref{apd:metrics}.

\begin{figure*}
    \centering
    \makebox[\textwidth][l]{%
        \hspace{0.10\textwidth}
        \textbf{Retina (Retina)} \hspace{0.10\textwidth} \textbf{IDRiD (Retina)} \hspace{0.09\textwidth} \textbf{APTOS2019 (Retina)}
    } \\[0.2cm]
    \includegraphics[trim={0 0 0 0},clip,height=0.14\textwidth, width=0.28\textwidth]{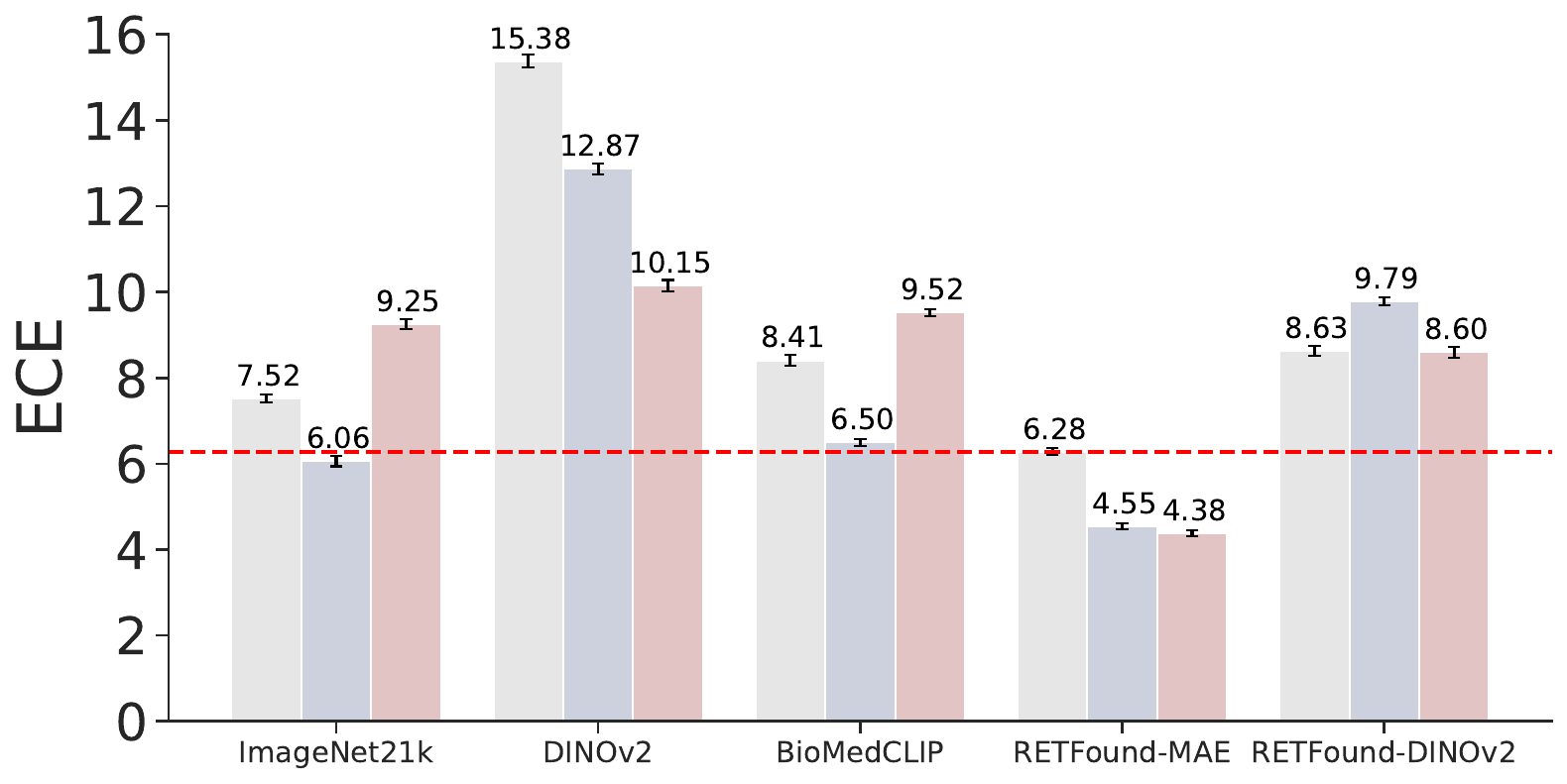}
    \includegraphics[trim={0 0 0 0},clip,height=0.14\textwidth, width=0.28\textwidth]{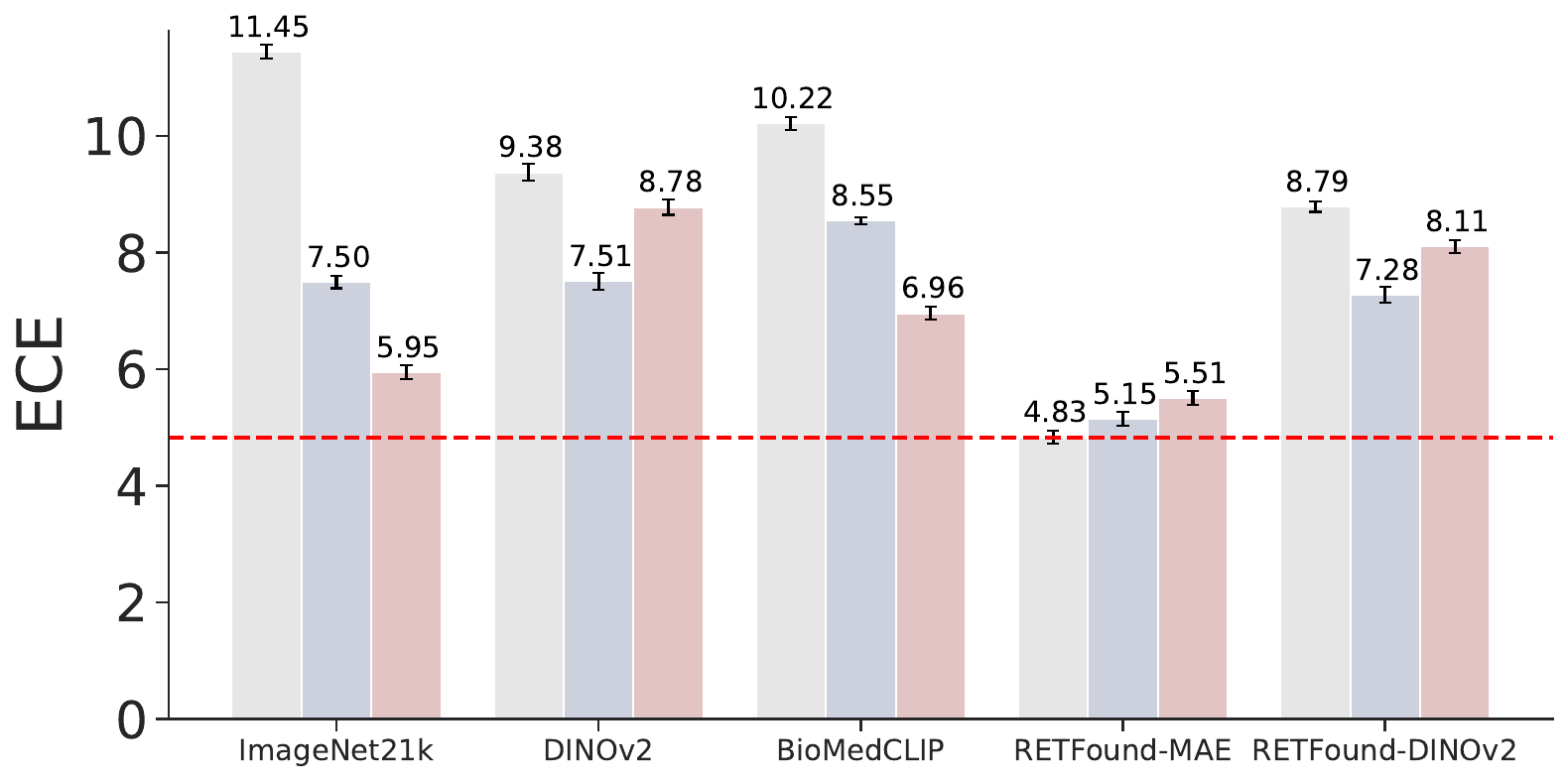}
    \includegraphics[trim={0 0 0 0},clip,height=0.14\textwidth, width=0.38\textwidth]{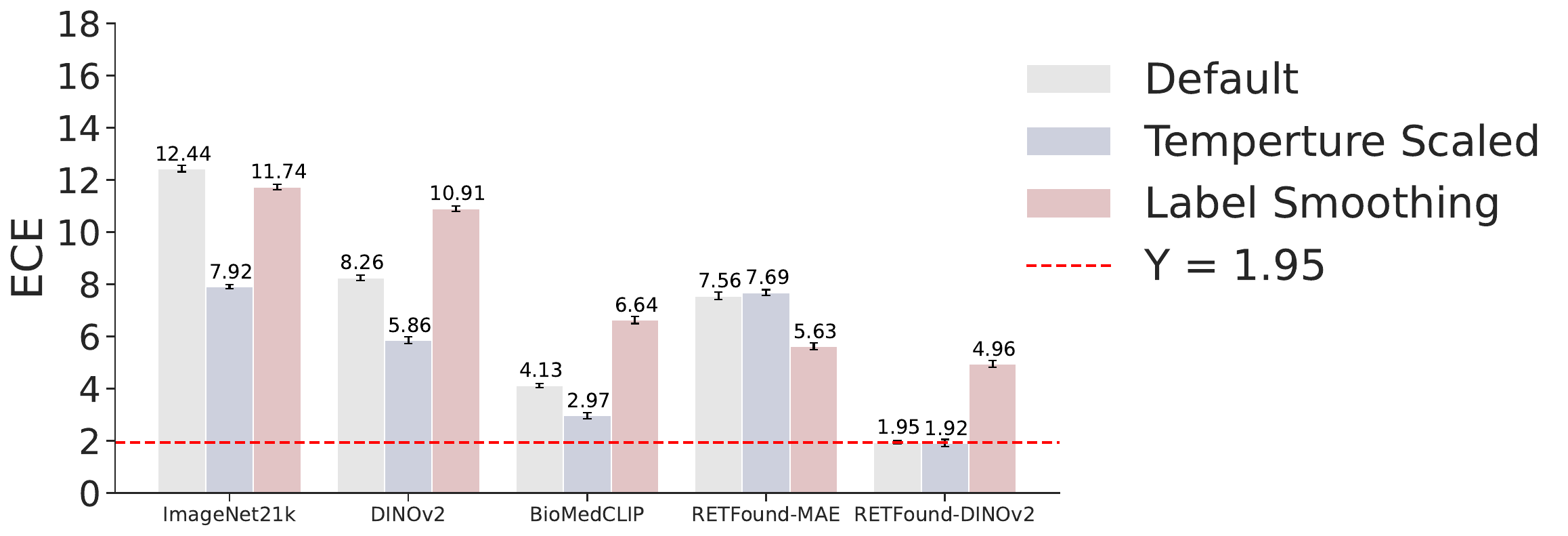}
    \\
    \makebox[\textwidth][l]{%
        \hspace{0.02\textwidth}
        \textbf{CRC100K (Histopathology)} \hspace{0.01\textwidth} \textbf{TCGA (Histopathology)} \hspace{0.01\textwidth} \textbf{BraTS (Histopathology)}
    } \\[0.2cm]
    \includegraphics[trim={0 0 0 0},clip,height=0.14\textwidth, width=0.28\textwidth]{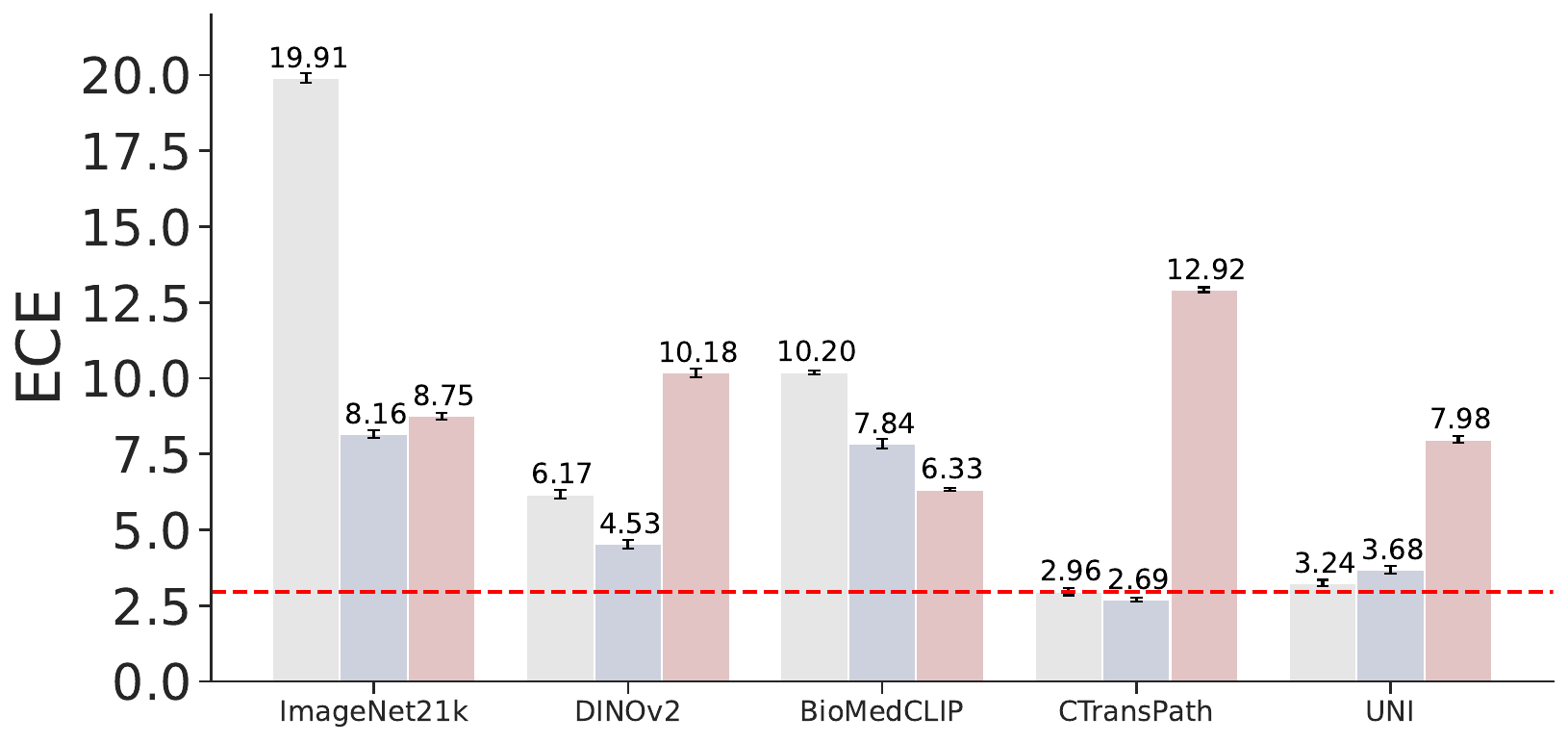}
    \includegraphics[trim={0 0 0 0},clip,height=0.14\textwidth, width=0.28\textwidth]{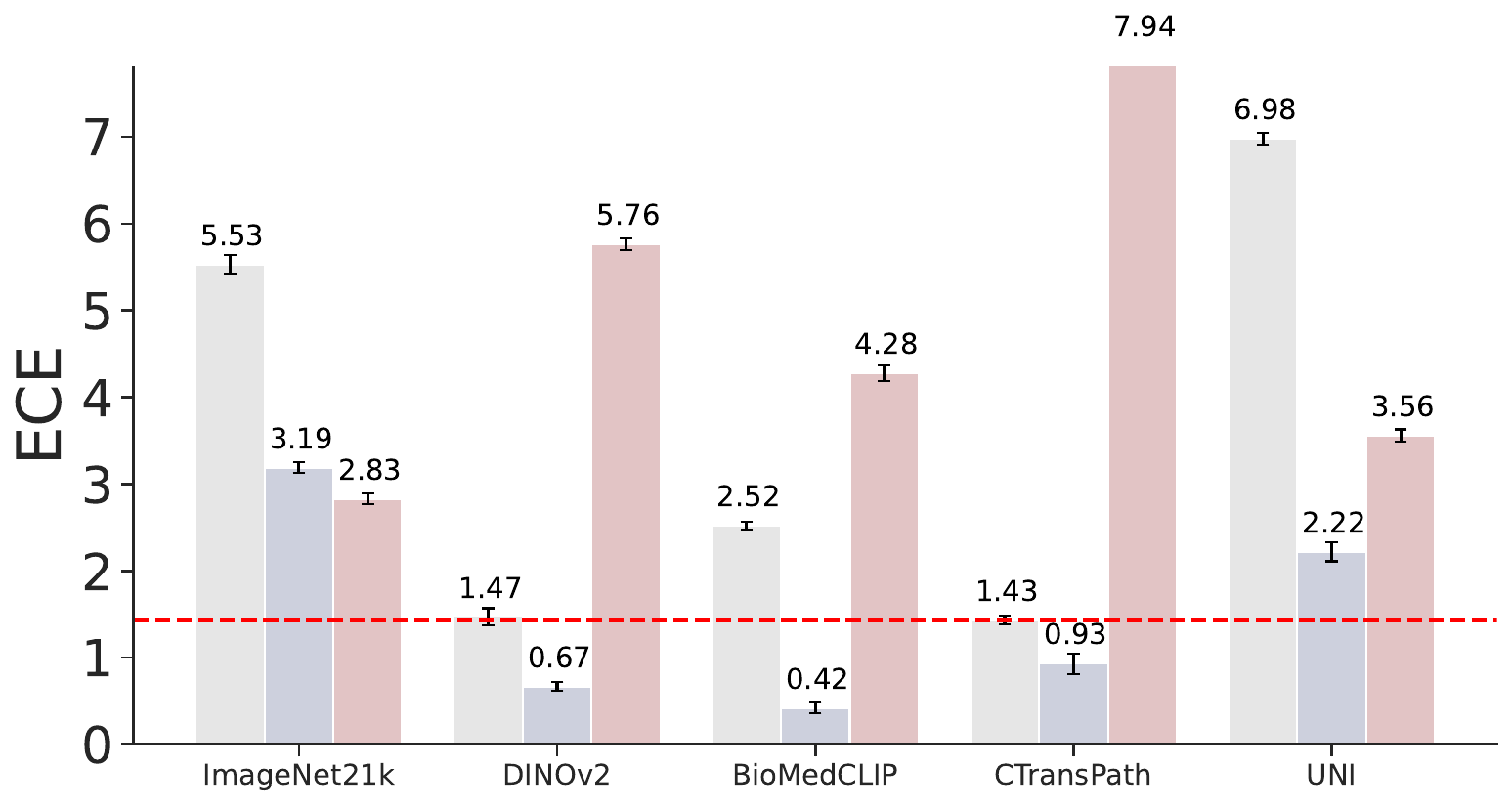}
    \includegraphics[trim={0 0 0 0},clip,height=0.14\textwidth, width=0.38\textwidth]{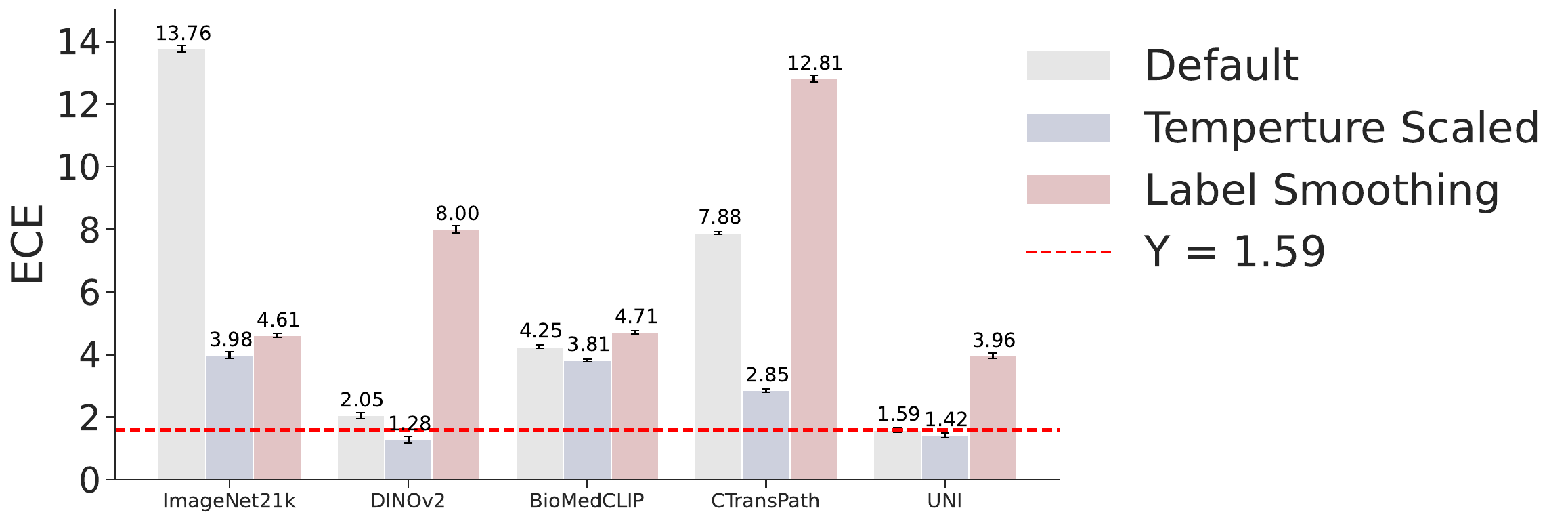}
    \\
    \makebox[\textwidth][l]{%
        \hspace{0.10\textwidth}
        \textbf{RSNA (X-Rays)} \hspace{0.06\textwidth} \textbf{POLCOVID (X-Rays)} \hspace{0.04\textwidth} \textbf{COVID-Rad (X-Rays)}
    } \\[0.2cm]
    \includegraphics[trim={0 0 0 0},clip,height=0.14\textwidth, width=0.28\textwidth]{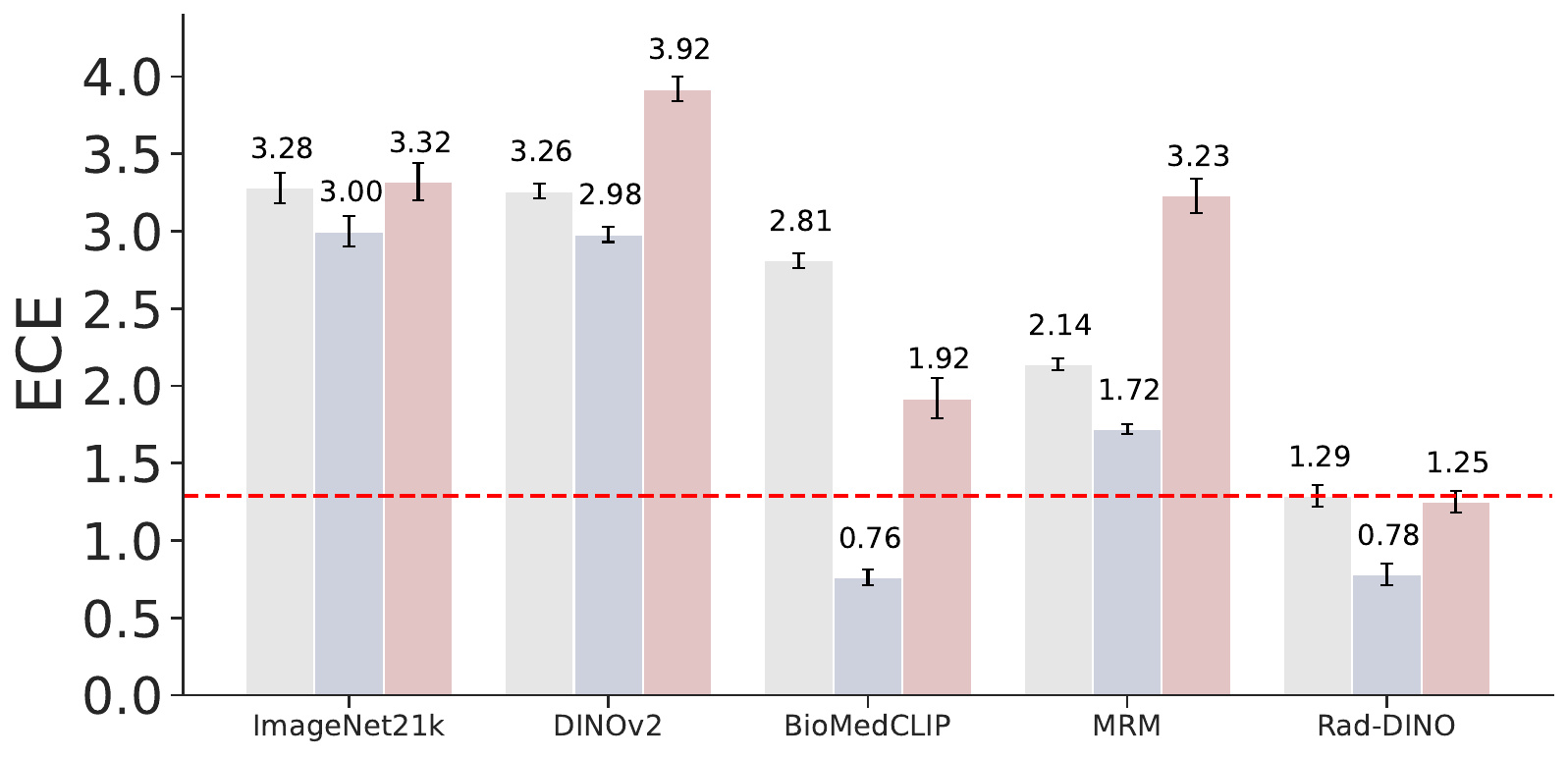}
    \includegraphics[trim={0 0 0 0},clip,height=0.14\textwidth, width=0.28\textwidth]{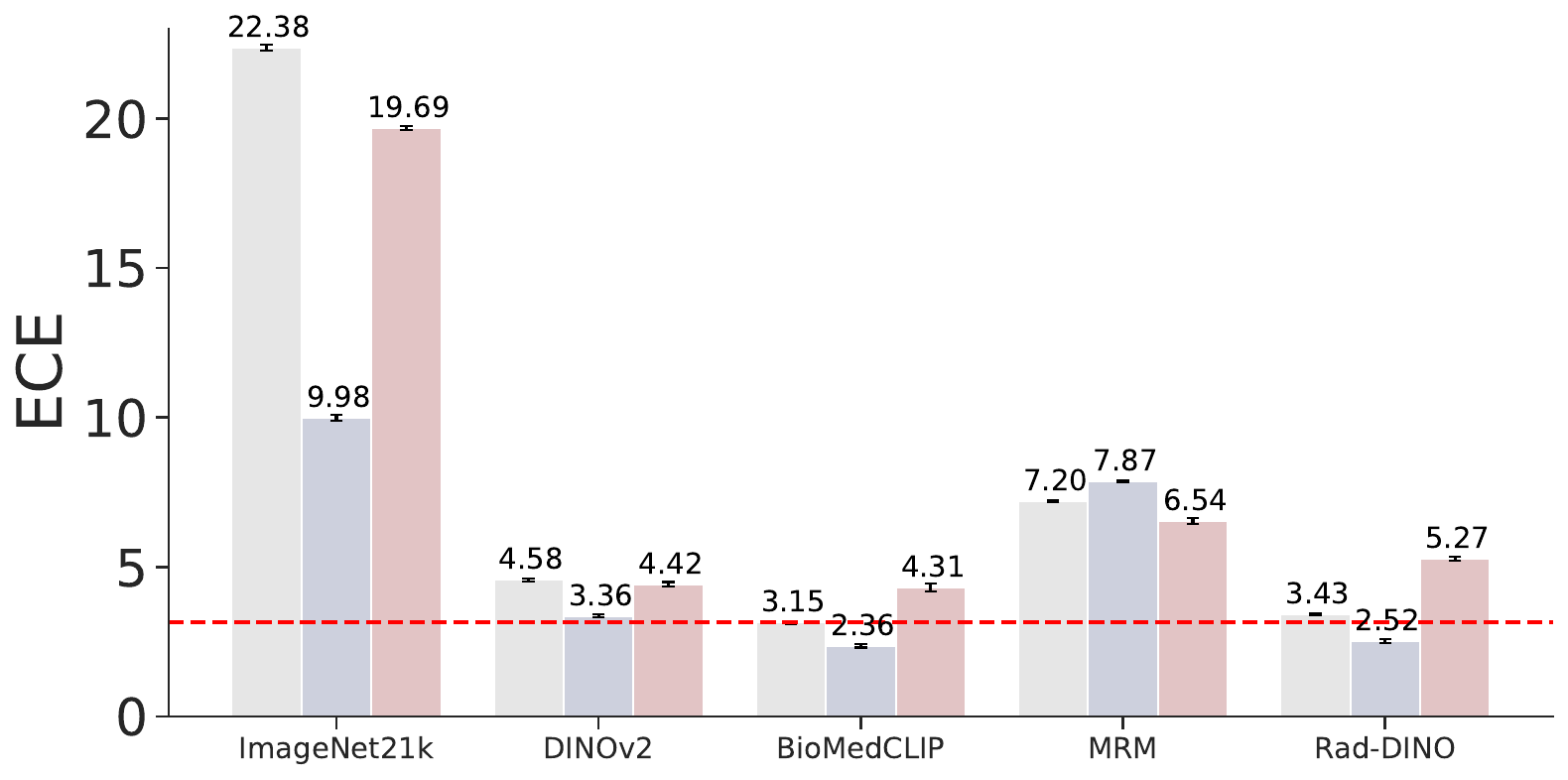}
    \includegraphics[trim={0 0 0 0},clip,height=0.14\textwidth, width=0.38\textwidth]{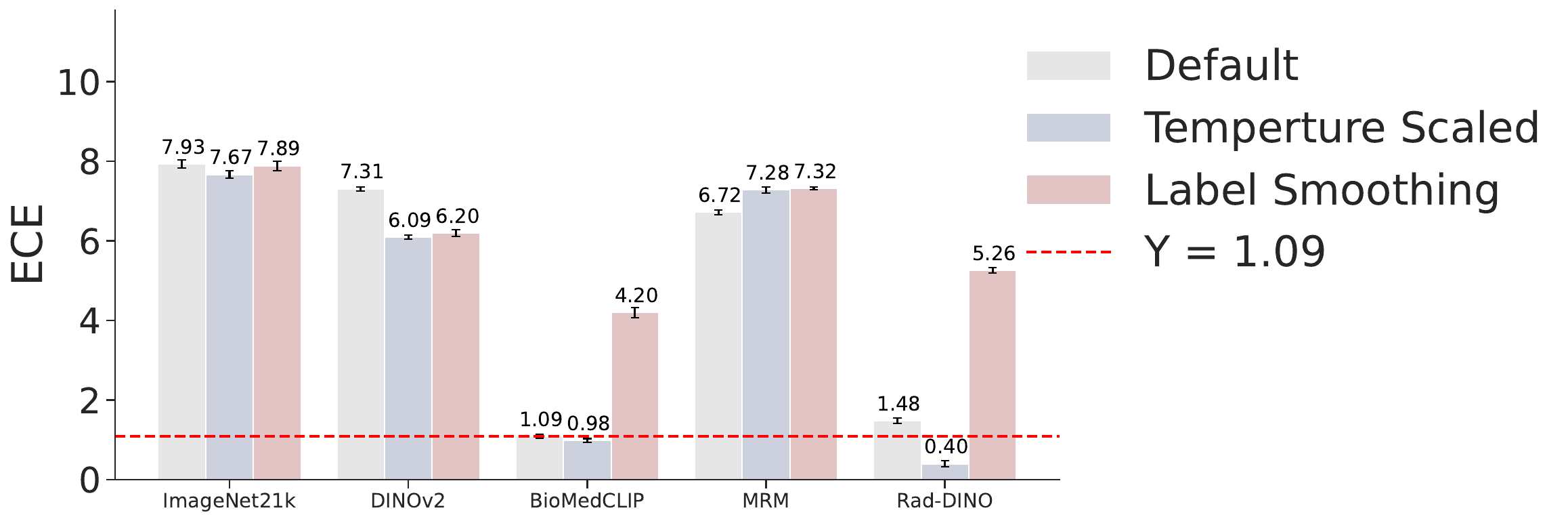}
    \caption{\textbf{Comparison of Default vs. Temperature Scaling vs. Label Smoothing Model}: the red dot line indicates best performing default model. The plot shows that uncertainty raises from different pre-training sources and methods cannot be fully addressed by re-calibrating the model. Further evaluation with Brier Score and NLL are shown in Appendix \Cref{tab:retina_uncertainty,tab:histopathology_uncertainty,tab:xray_uncertainty}.}
    \label{fig:point_pred_eval}
\end{figure*}

\section{Data and Models}
This study evaluates model uncertainty on three different widely used medical imaging modalities (Retina fundus imaging, Histopathology images (H\&E), and Chest X-Rays). Within  these modalities, we explore seven foundation models trained on different sources (four domain-specific foundation models and three general domain foundation models). We provide explanation on datasets and models in the following sections with detailed label naming and distribution for each dataset in \Cref{apd:dataset}.

\subsection{Datasets}
\label{sec:dataset}
{\bf Retina}~\citep{cataract_dataset} is a cataract and normal eye retina fundus image dataset for cataract detection with 4 labels of normal, glaucoma, cataract and retina disease.

{\bf IDRiD}~\citep{PORWAL2020101561} is a fundus retina dataset for diabetic retinopathy diagnosis. The labels for diabetic retinopathy are derived from the International Clinical Diabetic Retinopathy Severity Scale, which categorizes the condition into 5 stages, ranging from no diabetic retinopathy to proliferative diabetic retinopathy.

{\bf APTOS2019}~\citep{aptos2019_dataset} is a dataset with retina images for blindness assessment with 5 grading of No, Mild, Moderate, Severe and Proliferative.

{\bf CRC100k}~\citep{kather_2018_1214456} is a histological non-overlapping image patches dataset from hematoxylin \& eosin (H\&E) stained histological images of human colorectal cancer (CRC) and normal tissue. The tissue is separated to 9 sub-typing labels.

{\bf TCGA-Lymph}~\citep{BALANIS201917} is a histological images data of tumor-infiltrating lymphocyte maps for cancer sub-typing with total of 32 different labels.

{\bf BraTS-Path}~\citep{bakas2024bratspath} is histological images data of H\&E-stained FFPE digitized tissue sections from The Cancer Imaging Archive's TCGA-GBM and TCGA-LGG collections. Tissue sections are re-classified using the latest WHO criteria, focusing on glioblastoma, where it contains in total 6 sub-typing labels. For BraTS-Path, we curated a subset containing 2,000 samples per label, since the full dataset’s is large enough to achieve close to optimal model performance.

{\bf RSNA-Pneumonia}~\citep{rsna-pneumonia} is a large scale chest X-Rays dataset on diagnosing pneumonia collected from National Institutes of Health. The goal is to distinguish normal vs. pneumonia for each X-Rays image.

{\bf POLCOVID}~\citep{Suwalska2023} is a large, multi-center chest X‐ray collection gathered from 15 Polish hospitals during 2020–2021. It contains 4809 images classified into COVID-19, other pneumonia, and normal cases, and provides not only the original and lung-focused preprocessed images but also corresponding lung masks— both model-generated and manually annotated.

{\bf COVID-Rad}~\citep{Chowdhury20,Rahman21} is a large, multi-stage collection of chest X-ray images curated by an international team. The dataset includes images of COVID-19 positive cases, normal lungs, and various lung infections such as viral pneumonia and lung opacity from non-COVID causes.

\subsection{Models}
\label{apd:model}
For fair comparison, all models are Vision Transformer Large (ViT-Large)~\citep{dosovitskiy2020vit} with the difference being their pre-training data.

{\bf ImageNet21k} pre-trained on ImageNet-21k~\citep{imagenet} dataset with supervised learning to classify around 21k categories.

{\bf DINOv2}~\citep{oquab2024dinov} pre-trained on large-scale curated data set from Internet with 142 million images by self-supervised learning of distillation.

{\bf BioMedCLIP}~\citep{Zhang2023BiomedCLIPAM} pre-trained on large scale 15 million biomedical image-text pairs collected from scientific articles by contrastive multi-modal learning~\citep{radford2021learningtransferablevisualmodels}.

{\bf RETFound}~\citep{zhou2023foundation} pre-trained on 1.6 million retinal images collected from various unannotated public dataset and Moorfields Eye Hospital, London, UK by self-supervised learning.

{\bf CTransPath}~\citep{WANG2022102559} pre-trained on large-scale public available 15 million unlabeled histopathology imaging patches with contrastive self-supervised learning

{\bf UNI}~\citep{chen2024uni} pre-trained on large scale and high-quality 100 million images from over 100,000 diagnostic H\&E-stained WSIs histopathology images collected from Massachusett General Hospitals and Brigham and Women's Hospital, Boston, USA with distillation self-supervised learning

{\bf MRM}~\citep{zhou2023advancing} pre-trained on MIMIC-CXR~\citep{Johnson2019-mu} by learning to reconstruct both masked image patches from chest X-rays and masked tokens from associated radiology reports, effectively incorporating both invariant visual semantics and expert domain knowledge.

{\bf Rad-DINO}~\citep{Pérez-García2025} challenges the current reliance on text supervision for training biomedical image encoders. Instead, it introduces RAD-DINO — an image encoder pretrained solely on large-scale, uni-modal Chest X-Rays imaging data collected from multiple public datasets with DINOv2~\citep{oquab2024dinov} — which achieves comparable or superior performance to text-supervised models on tasks like classification, semantic segmentation, and report generation.

\begin{figure*}[!ht]
    \centering
    \makebox[\textwidth][l]{%
        \hspace{0.26\textwidth}
        \textbf{Retina} \hspace{0.11\textwidth} \textbf{IDRiD} \hspace{0.11\textwidth} \textbf{APTOS2019}
    } \\[0.2cm]
    \includegraphics[height=0.16\textwidth, width=0.20\textwidth]{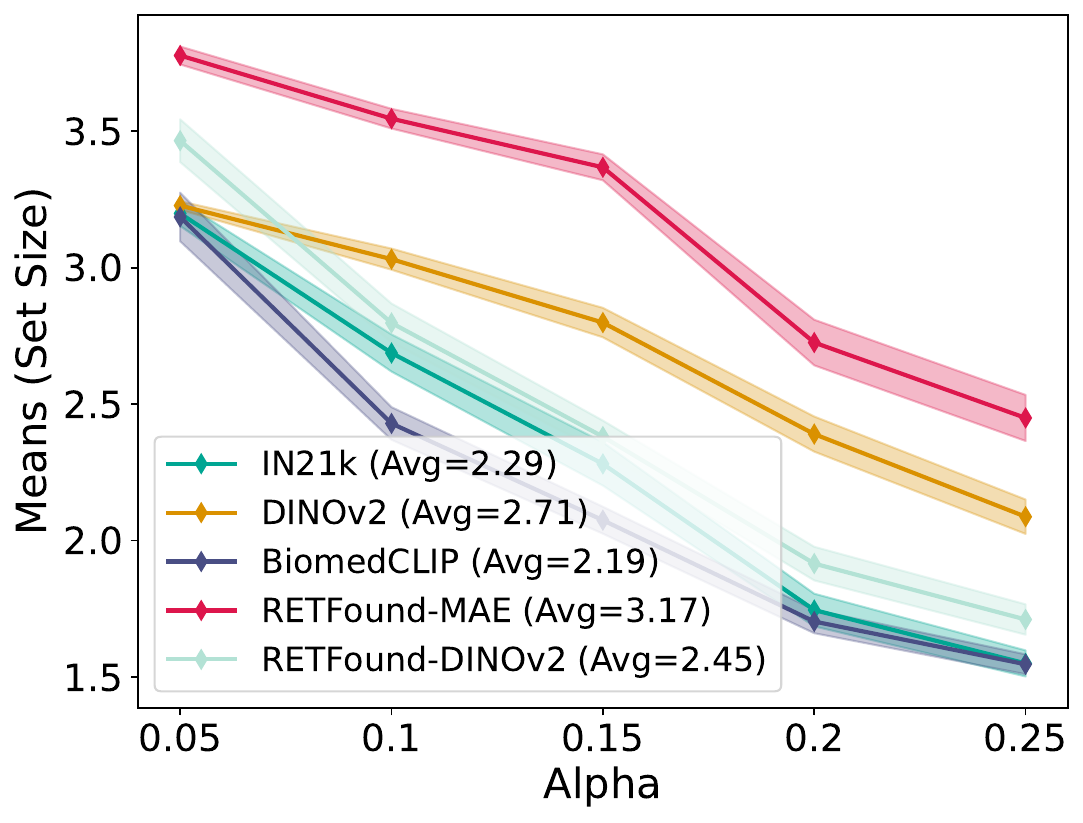}
    \includegraphics[height=0.16\textwidth, width=0.20\textwidth]{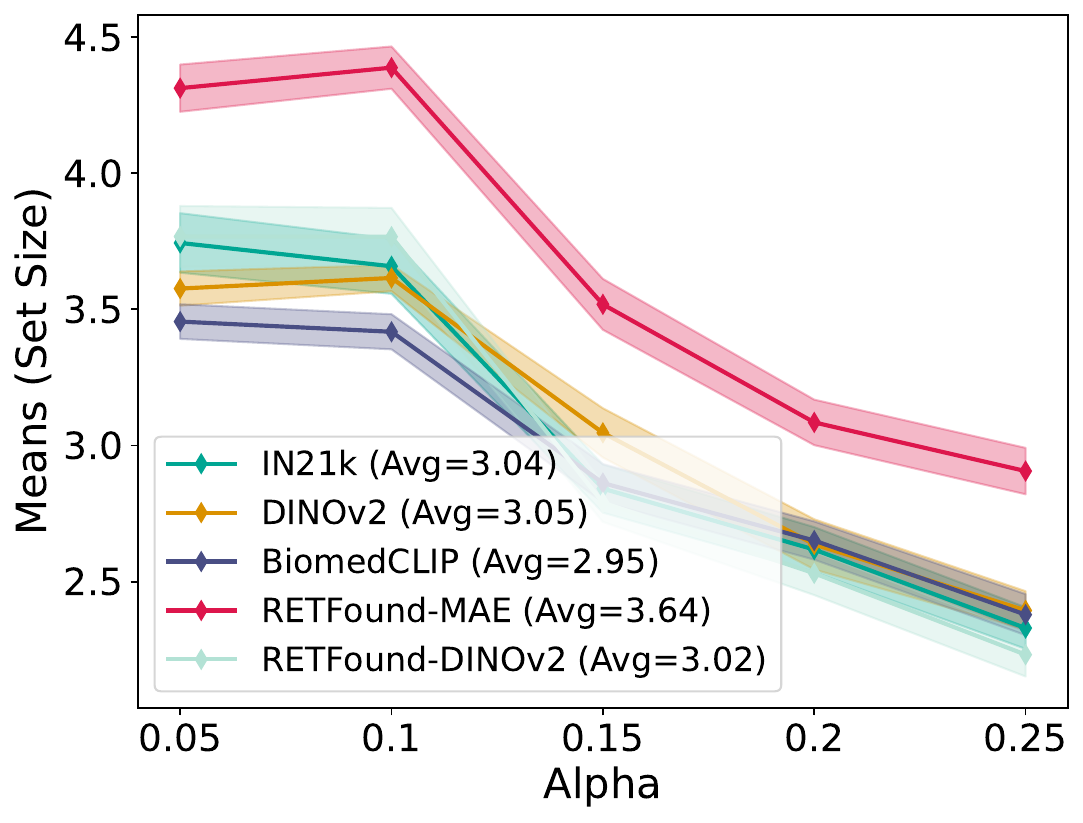}
    \includegraphics[height=0.16\textwidth, width=0.20\textwidth]{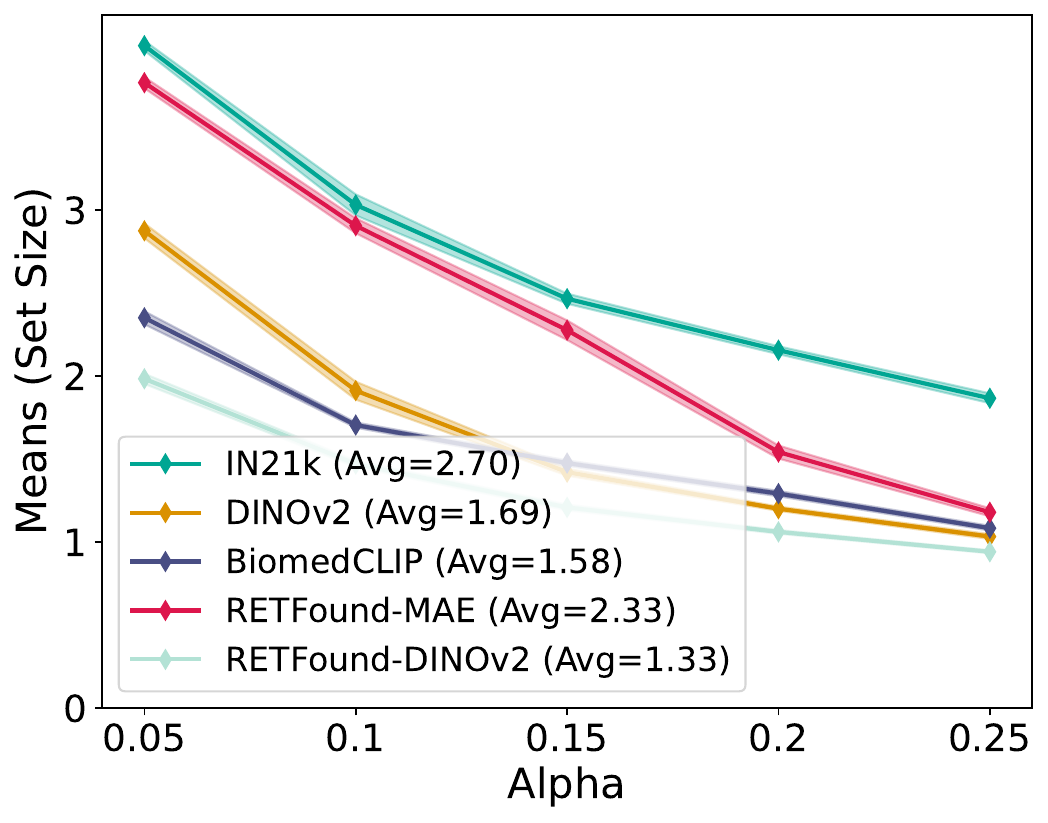}\\[0.2cm]
    \makebox[\textwidth][l]{%
        \hspace{0.25\textwidth}
        \textbf{Retina (T)} \hspace{0.07\textwidth} \textbf{IDRiD (T)} \hspace{0.07\textwidth} \textbf{APTOS2019 (T)}
    } \\[0.2cm]
    \includegraphics[height=0.16\textwidth, width=0.20\textwidth]{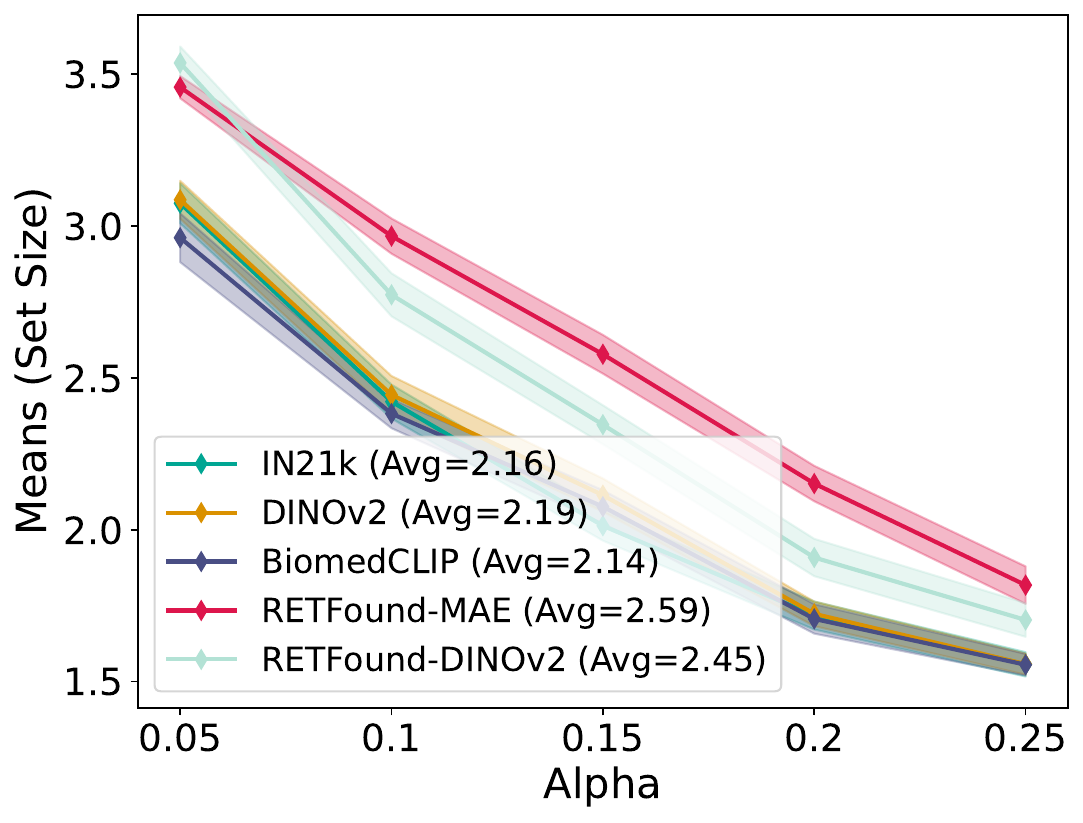}
    \includegraphics[height=0.16\textwidth, width=0.20\textwidth]{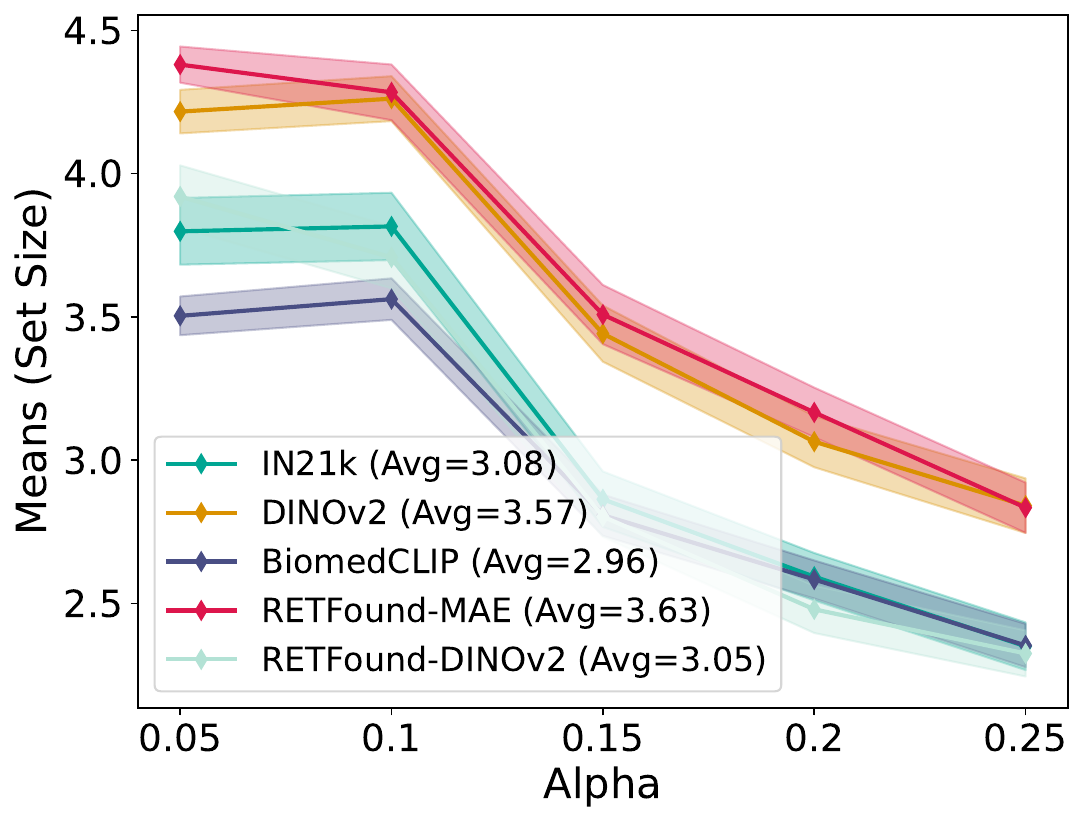}
    \includegraphics[height=0.16\textwidth, width=0.20\textwidth]{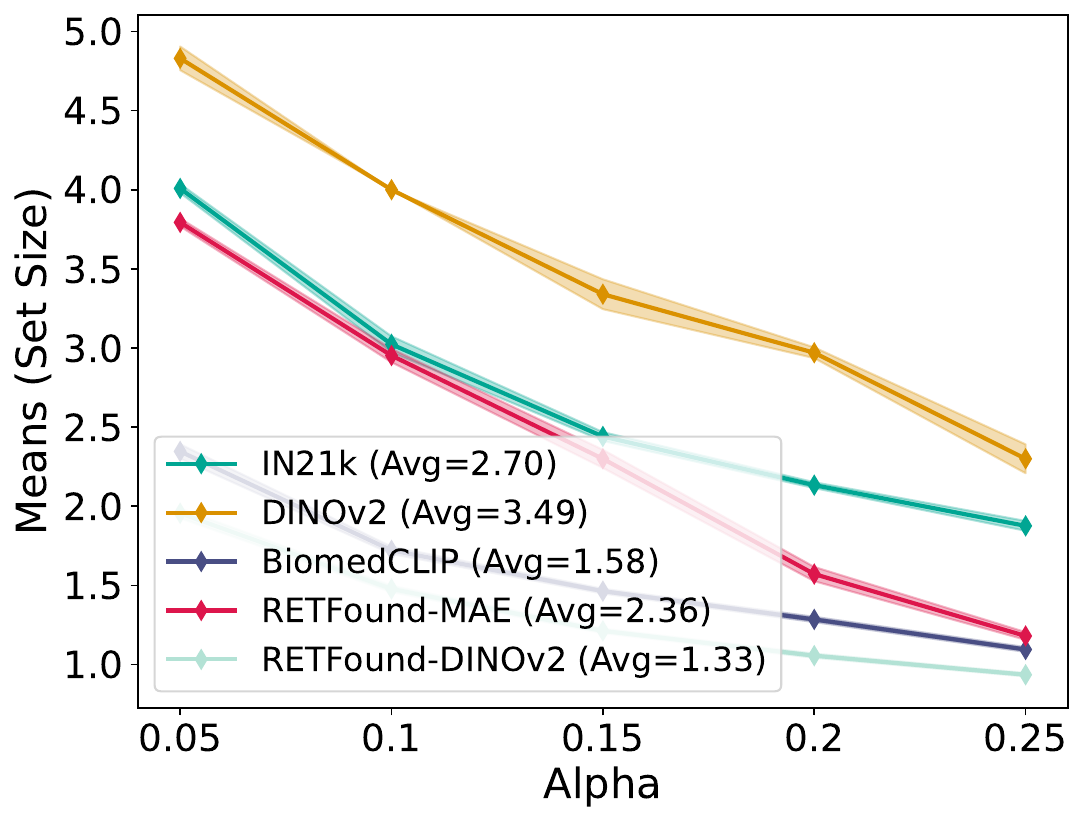}\\[0.2cm]
    \makebox[\textwidth][l]{%
        \hspace{0.24\textwidth}
        \textbf{Retina (LS)} \hspace{0.06\textwidth} \textbf{IDRiD (LS)} \hspace{0.05\textwidth} \textbf{APTOS2019 (LS)}
    } \\[0.2cm]
    \includegraphics[height=0.16\textwidth, width=0.20\textwidth]{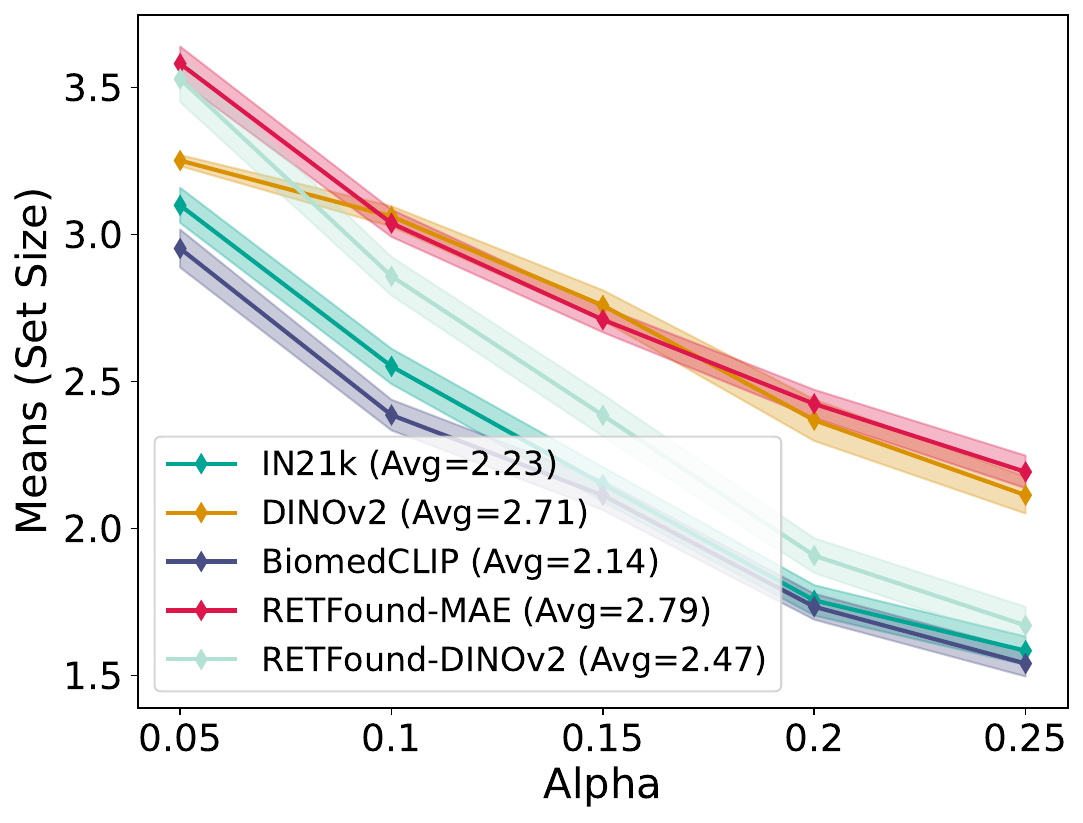}
    \includegraphics[height=0.16\textwidth, width=0.20\textwidth]{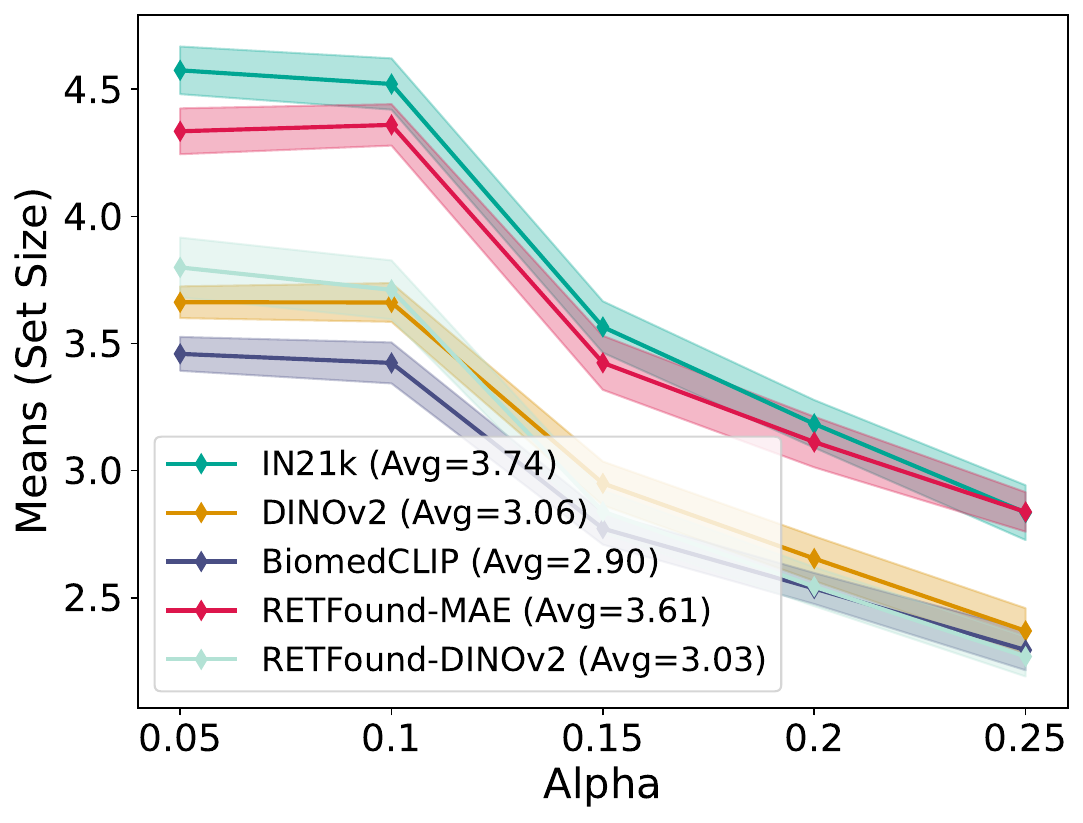}
    \includegraphics[height=0.16\textwidth, width=0.20\textwidth]{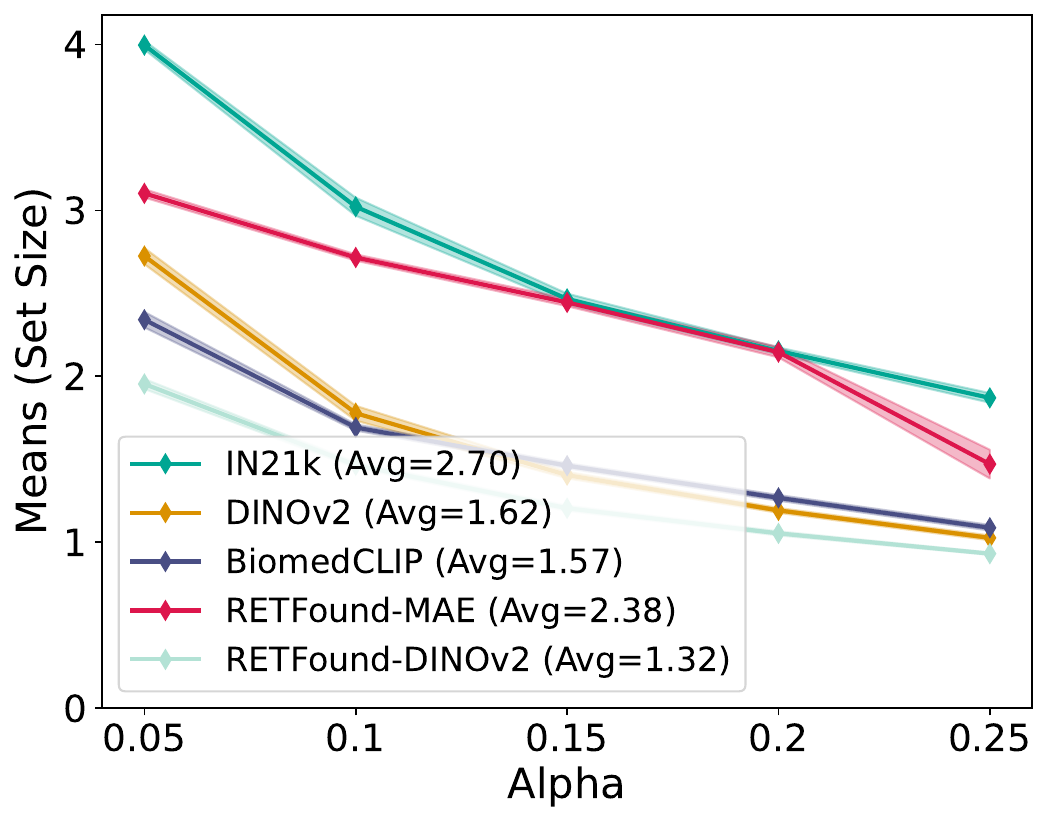}
    \caption{\textbf{Conformal prediction Set Size (Retina)}: the average conformal prediction Set Size across different $\alpha$ thresholds for retina data does not exhibit a clear trend based on point prediction uncertainty, pre-training source, or method.}
    \label{fig:cp_retina_setsize}
\end{figure*}

\begin{figure*}[!ht]
    \centering
    \makebox[\textwidth][l]{%
        \hspace{0.24\textwidth}
        \textbf{CRC100K} \hspace{0.10\textwidth} \textbf{TCGA} \hspace{0.12\textwidth} \textbf{BraTS}
    } \\[0.2cm]
    \includegraphics[height=0.16\textwidth, width=0.20\textwidth]{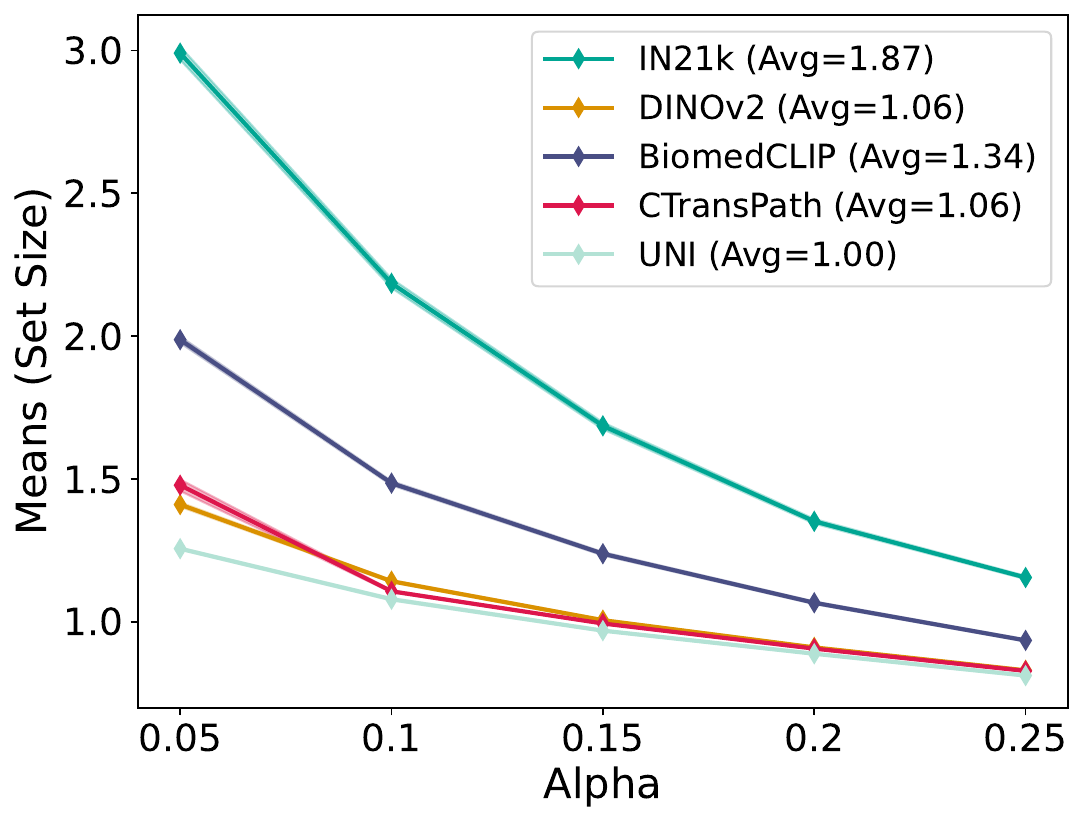}
    \includegraphics[height=0.16\textwidth, width=0.20\textwidth]{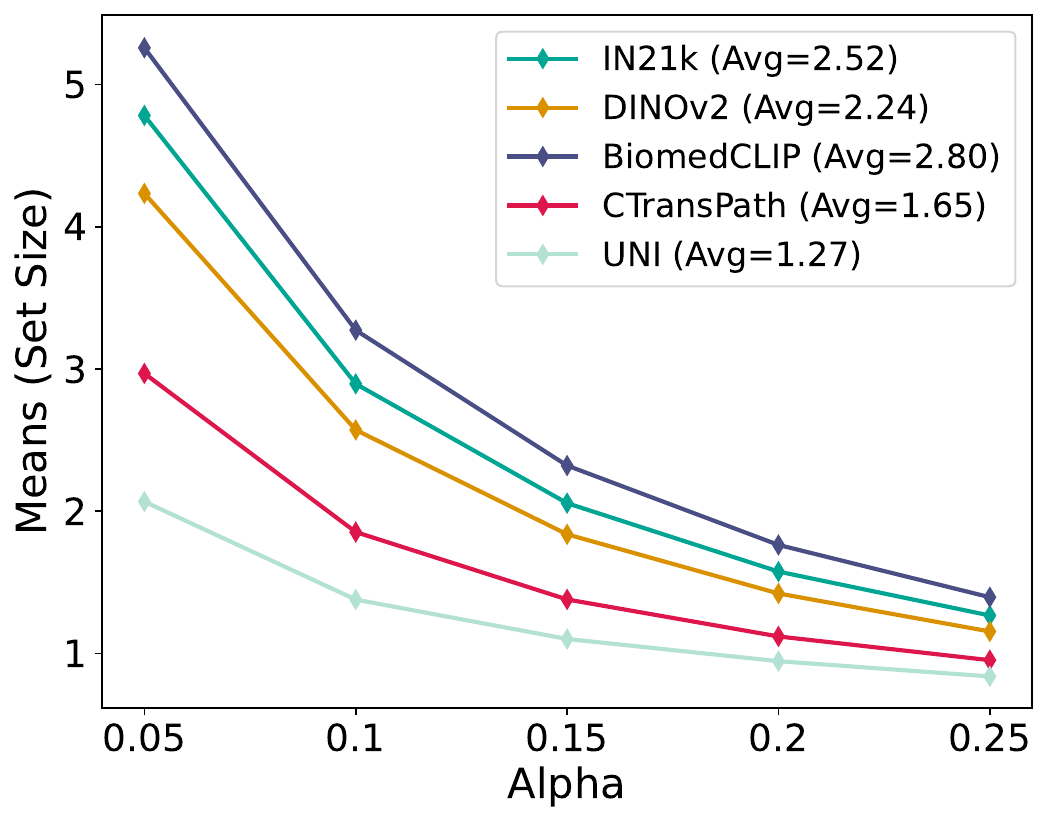}
    \includegraphics[height=0.16\textwidth, width=0.20\textwidth]{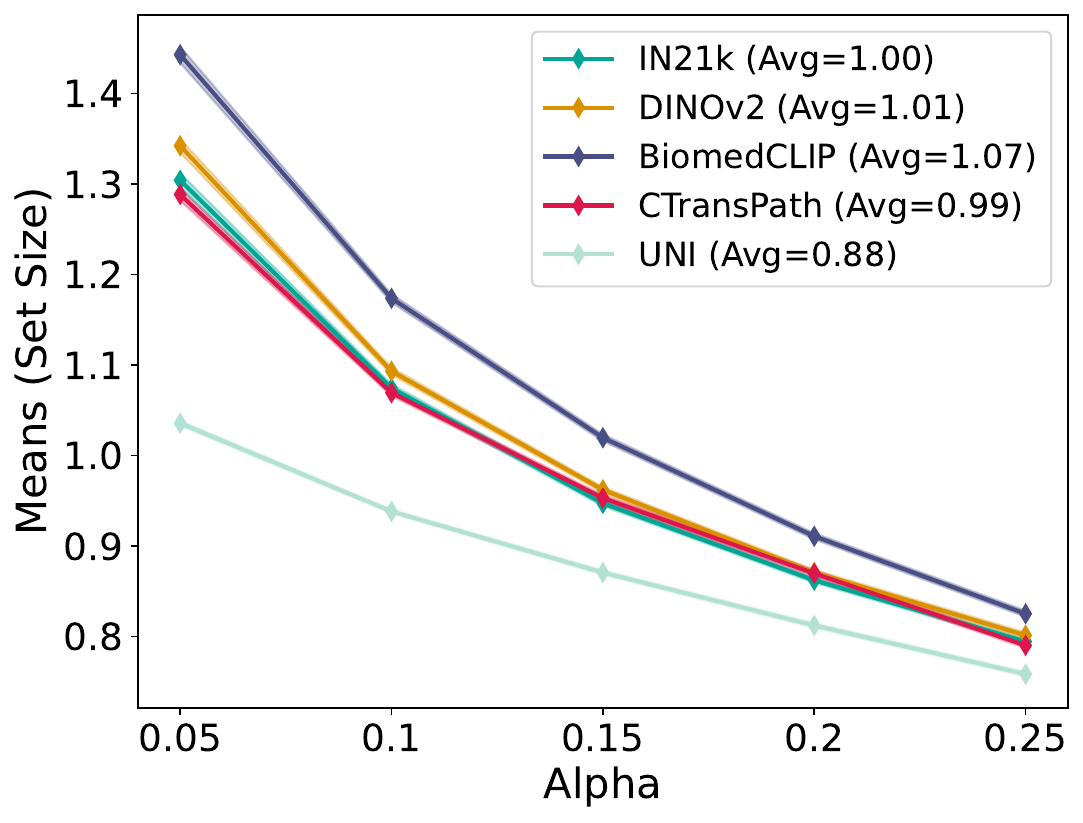}
    \makebox[\textwidth][l]{%
        \hspace{0.23\textwidth}
        \textbf{CRC100K (T)} \hspace{0.05\textwidth} \textbf{TCGA (T)} \hspace{0.08\textwidth} \textbf{BraTS (T)}
    } \\[0.2cm]
    \includegraphics[height=0.16\textwidth, width=0.20\textwidth]{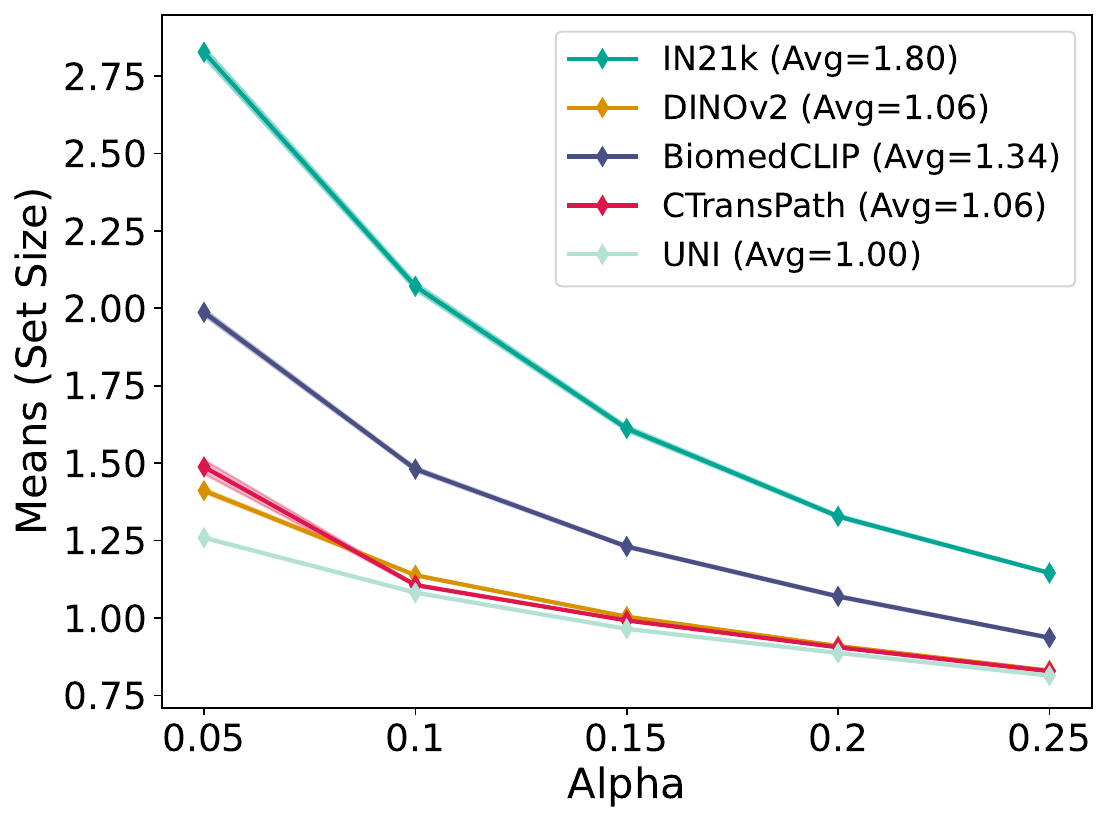}
    \includegraphics[height=0.16\textwidth, width=0.20\textwidth]{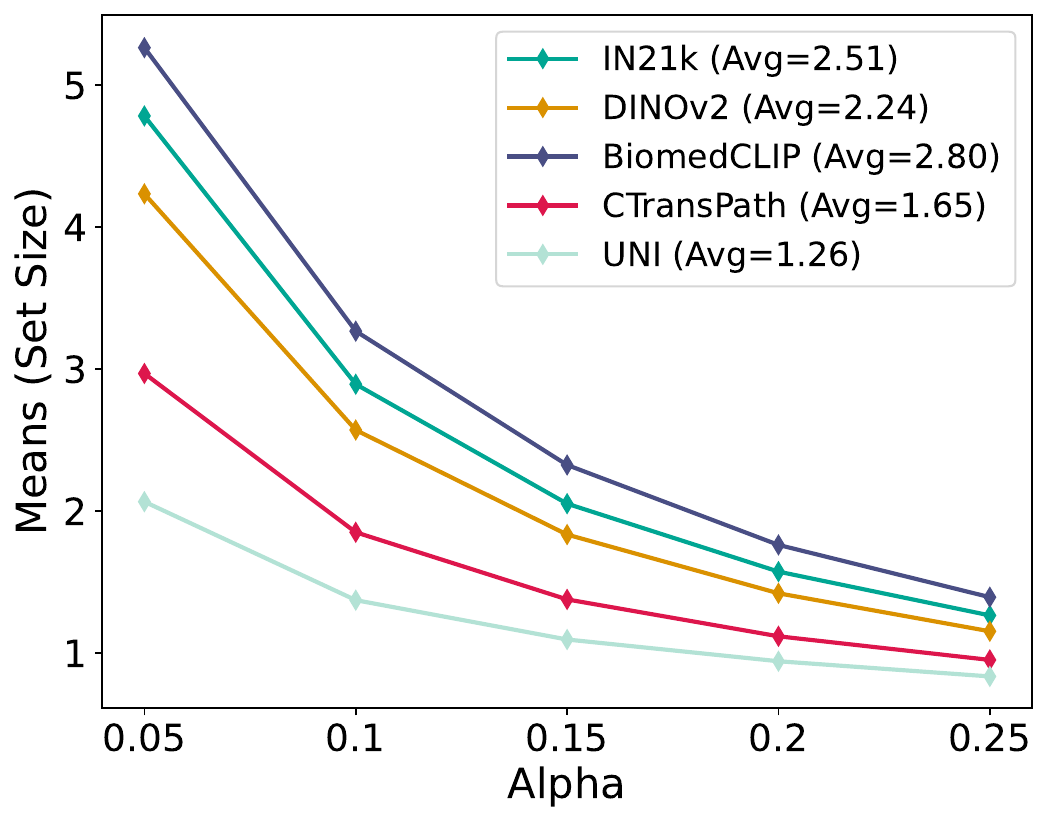}
    \includegraphics[height=0.16\textwidth, width=0.20\textwidth]{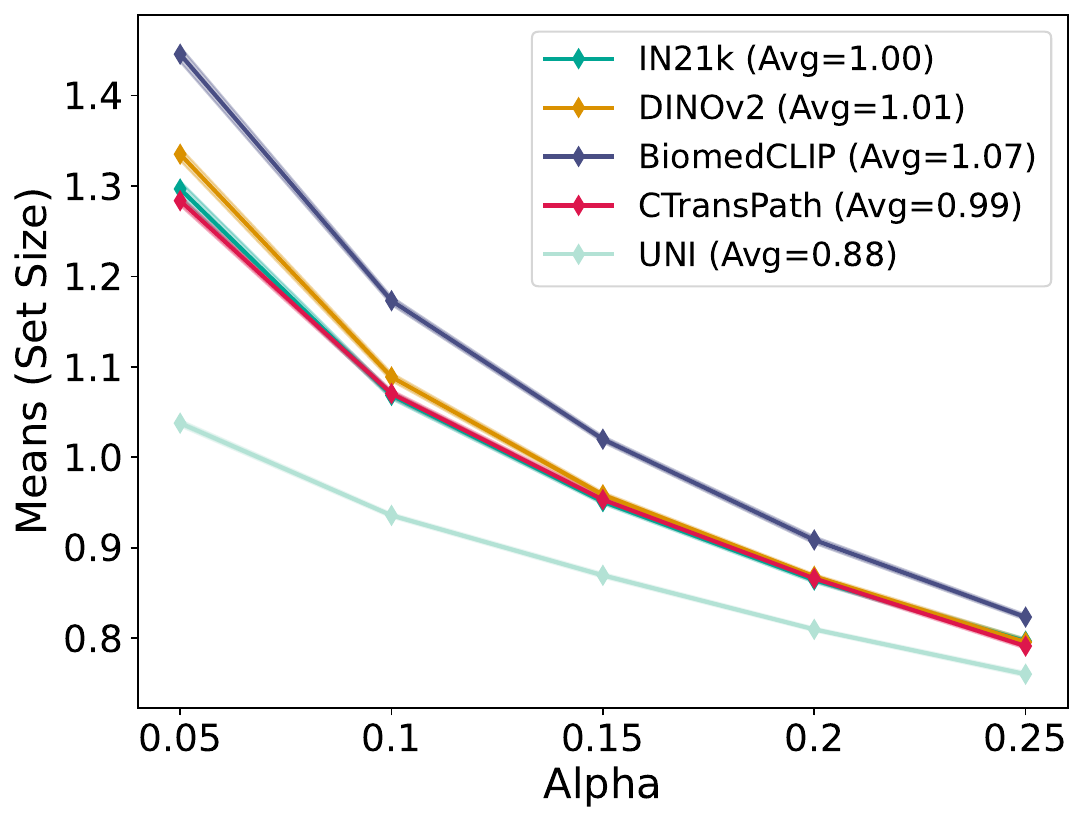}    \makebox[\textwidth][l]{%
        \hspace{0.22\textwidth}
        \textbf{CRC100K (LS)} \hspace{0.04\textwidth} \textbf{TCGA (LS)} \hspace{0.07\textwidth} \textbf{BraTS (LS)}
    } \\[0.2cm]
    \includegraphics[height=0.16\textwidth, width=0.20\textwidth]{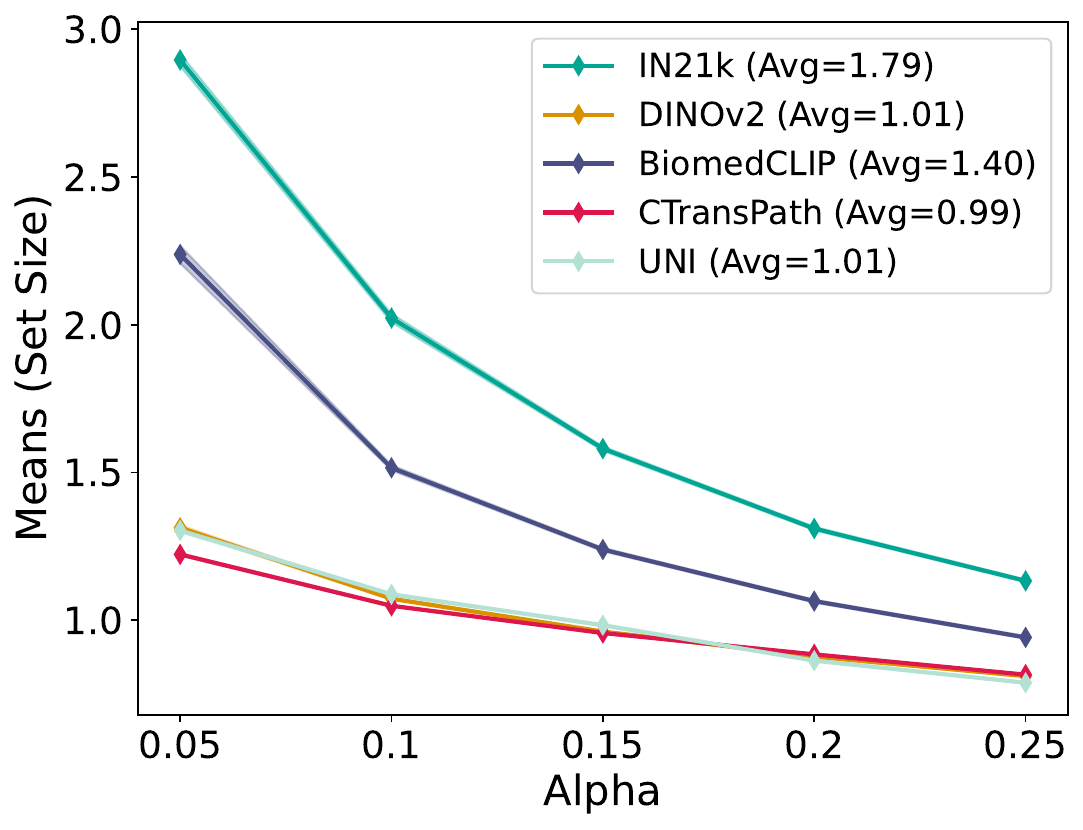}
    \includegraphics[height=0.16\textwidth, width=0.20\textwidth]{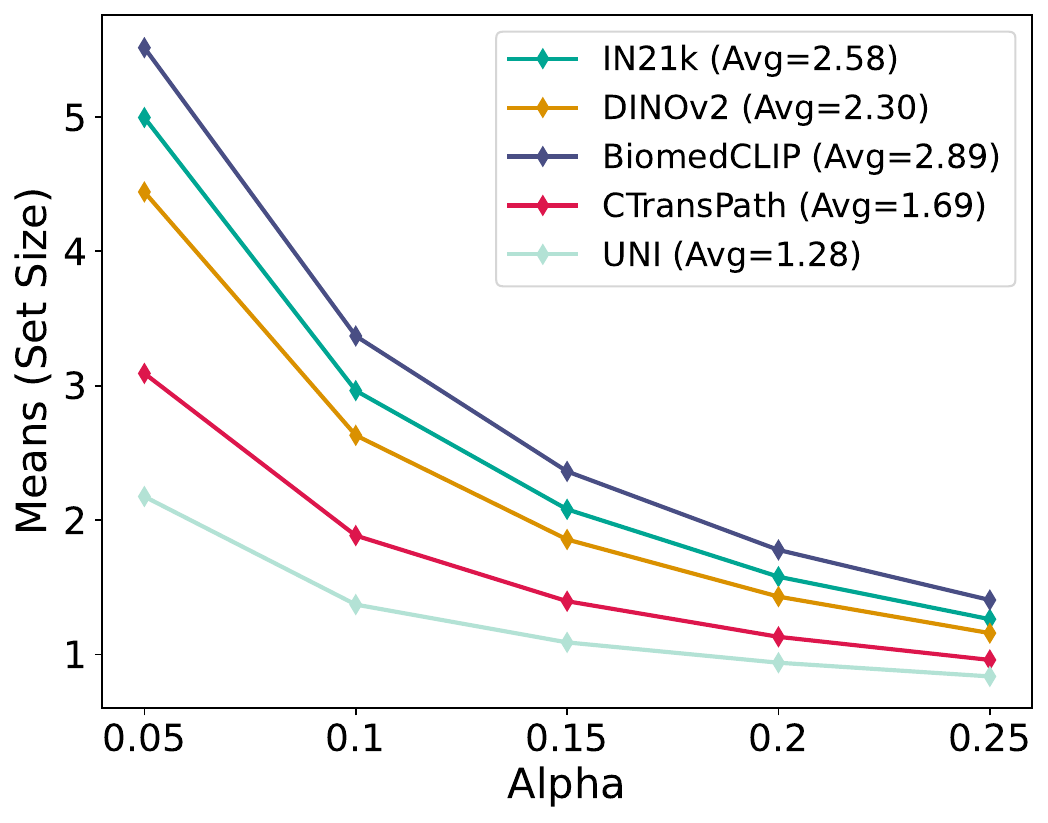}
    \includegraphics[height=0.16\textwidth, width=0.20\textwidth]{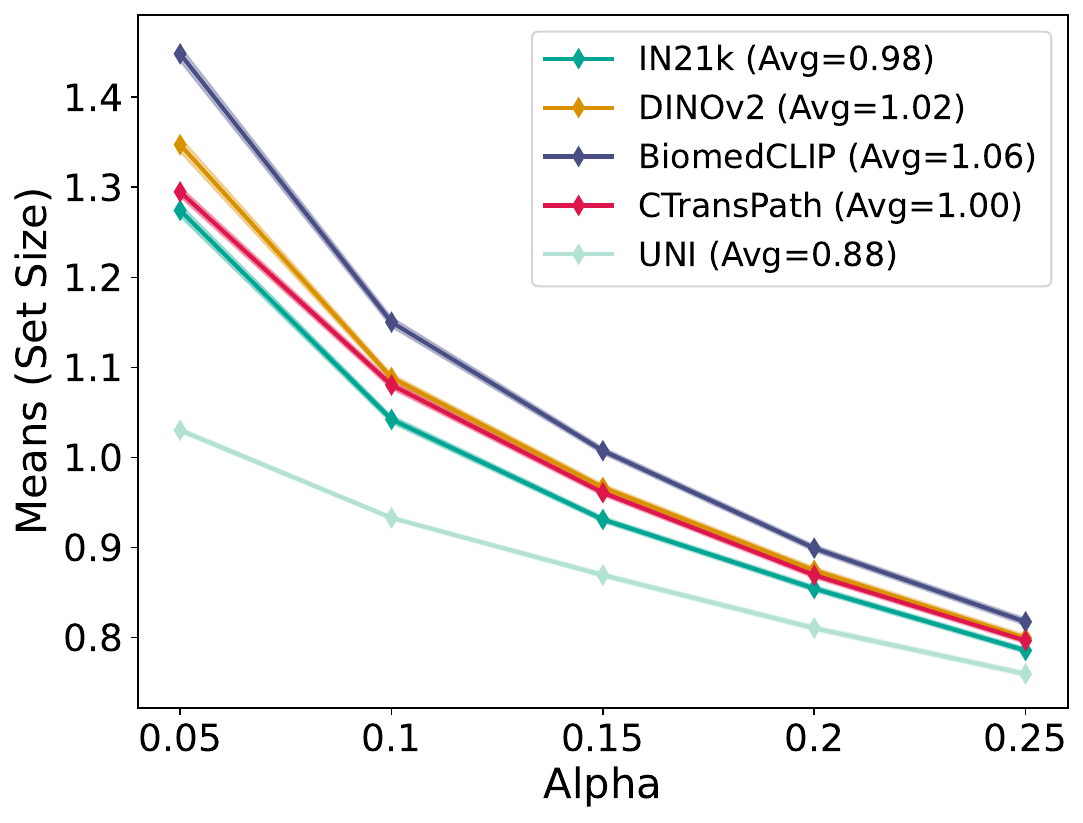}
    \caption{\textbf{Conformal prediction Set Size (Histopathology)}: the average conformal prediction Set Size across different $\alpha$ thresholds for histopathology data does not exhibit a clear trend based on point prediction uncertainty, pre-training source, or method. While UNI performs among the best (more accurate model), DINOv2 falls short on many datasets.}
    \label{fig:cp_histopath_setsize}
\end{figure*}

\begin{figure*}[!ht]
    \centering
    \makebox[\textwidth][l]{%
        \hspace{0.26\textwidth}
        \textbf{RSNA} \hspace{0.09\textwidth} \textbf{POLCOVID} \hspace{0.07\textwidth} \textbf{COVID-Rad}
    } \\[0.2cm]
    \includegraphics[height=0.16\textwidth, width=0.20\textwidth]{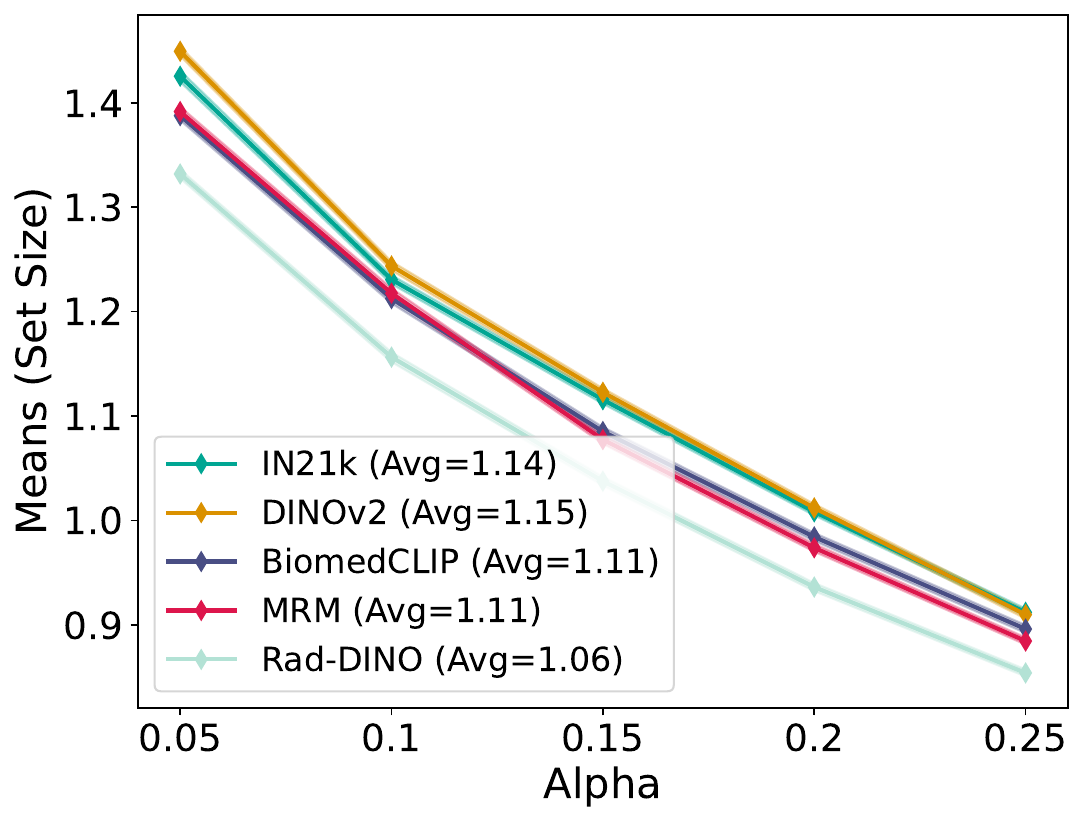}
    \includegraphics[height=0.16\textwidth, width=0.20\textwidth]{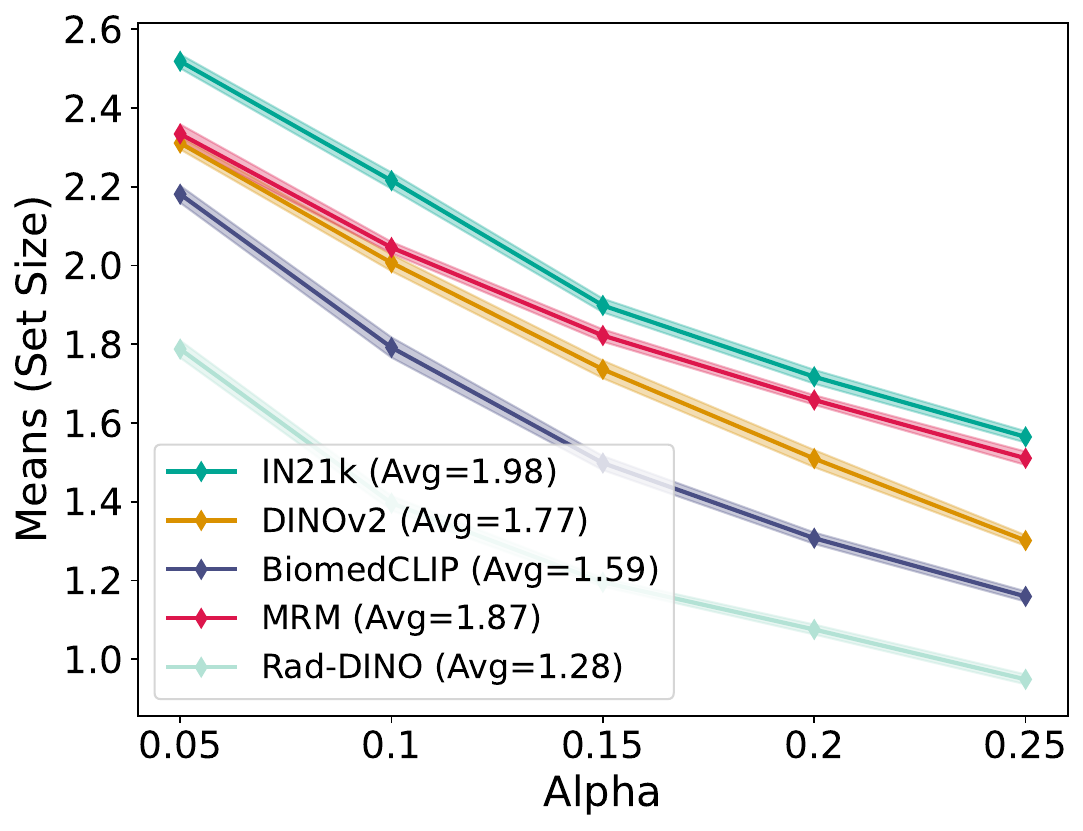}
    \includegraphics[height=0.16\textwidth, width=0.20\textwidth]{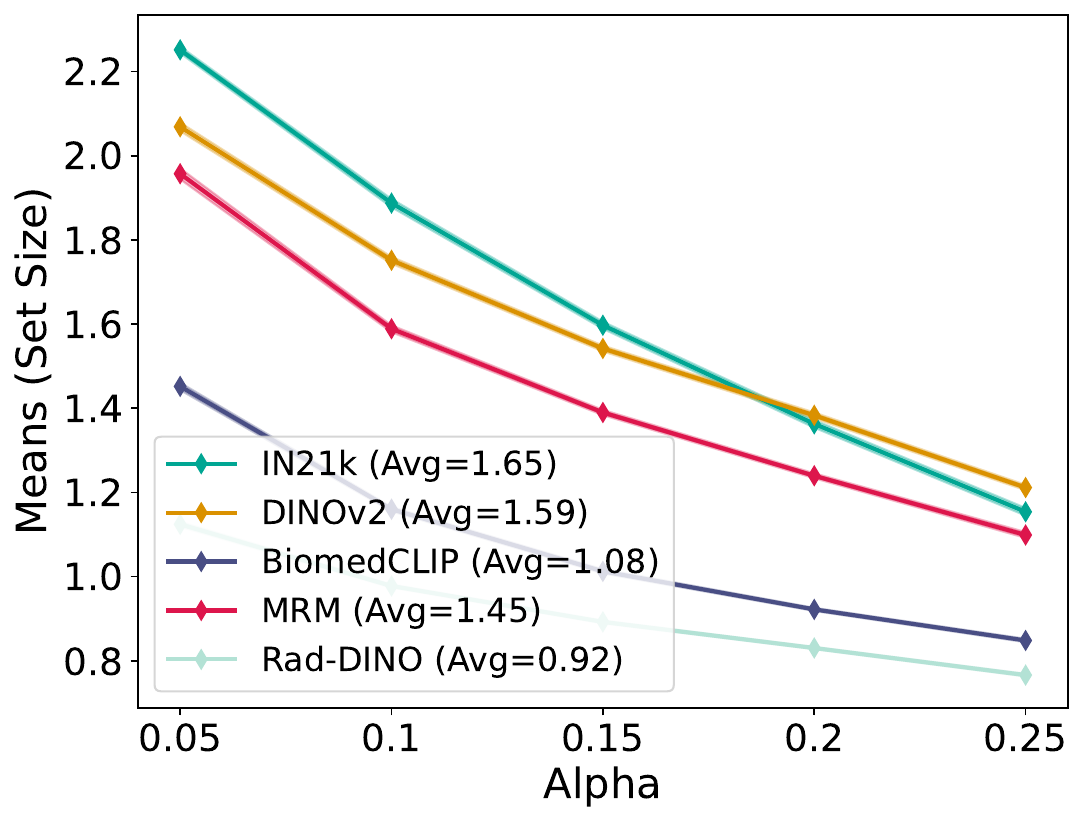}
    \makebox[\textwidth][l]{%
        \hspace{0.25\textwidth}
        \textbf{RSNA (T)} \hspace{0.05\textwidth} \textbf{POLCOVID (T)} \hspace{0.02\textwidth} \textbf{COVID-Rad (T)}
    } \\[0.2cm]
    \includegraphics[height=0.16\textwidth, width=0.20\textwidth]{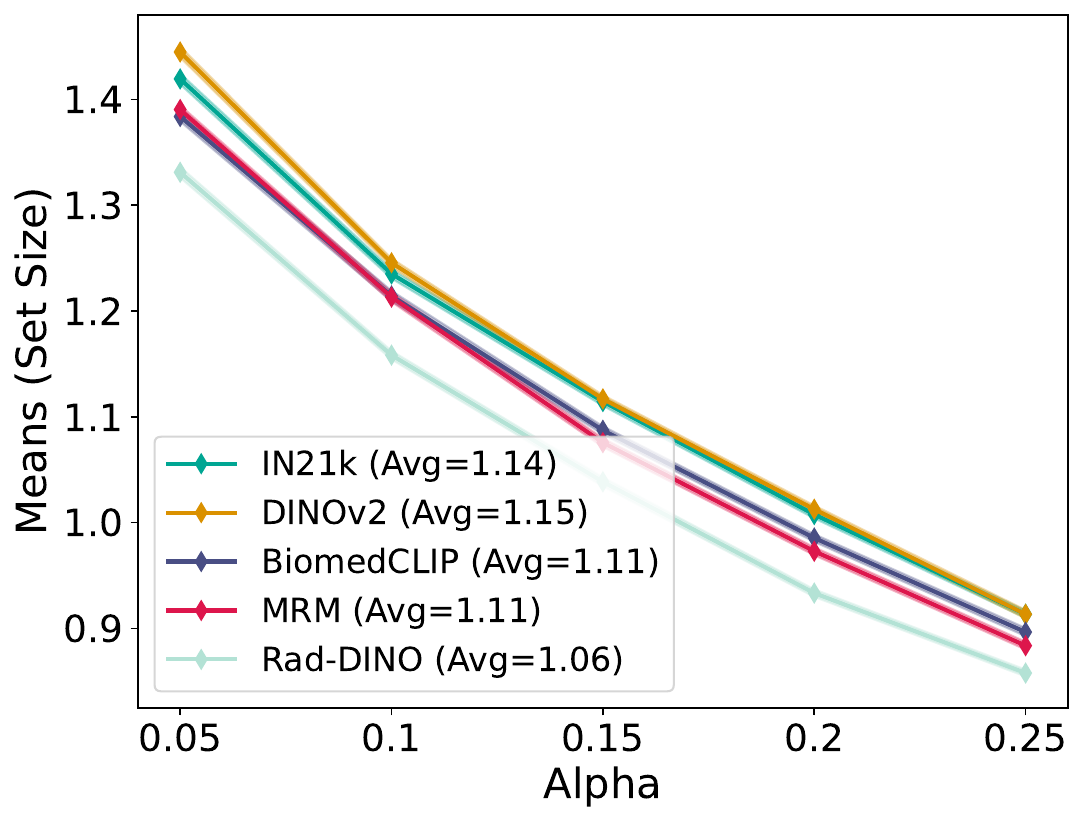}
    \includegraphics[height=0.16\textwidth, width=0.20\textwidth]{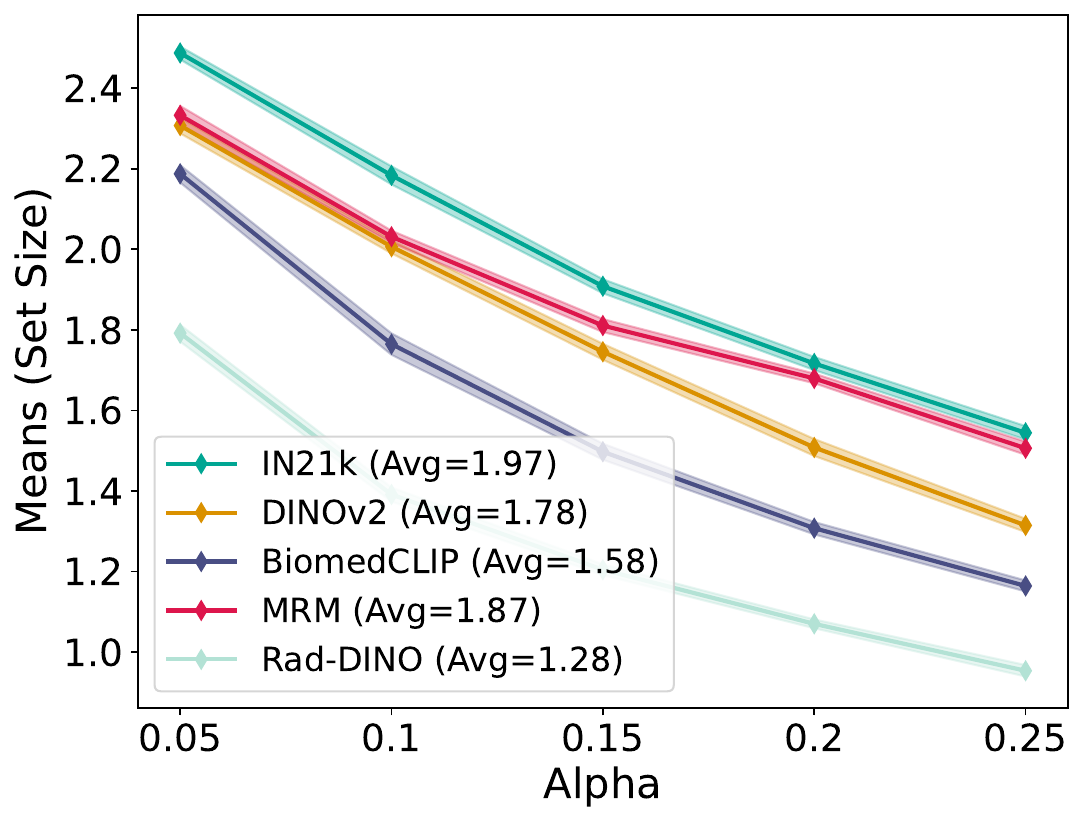}
    \includegraphics[height=0.16\textwidth, width=0.20\textwidth]{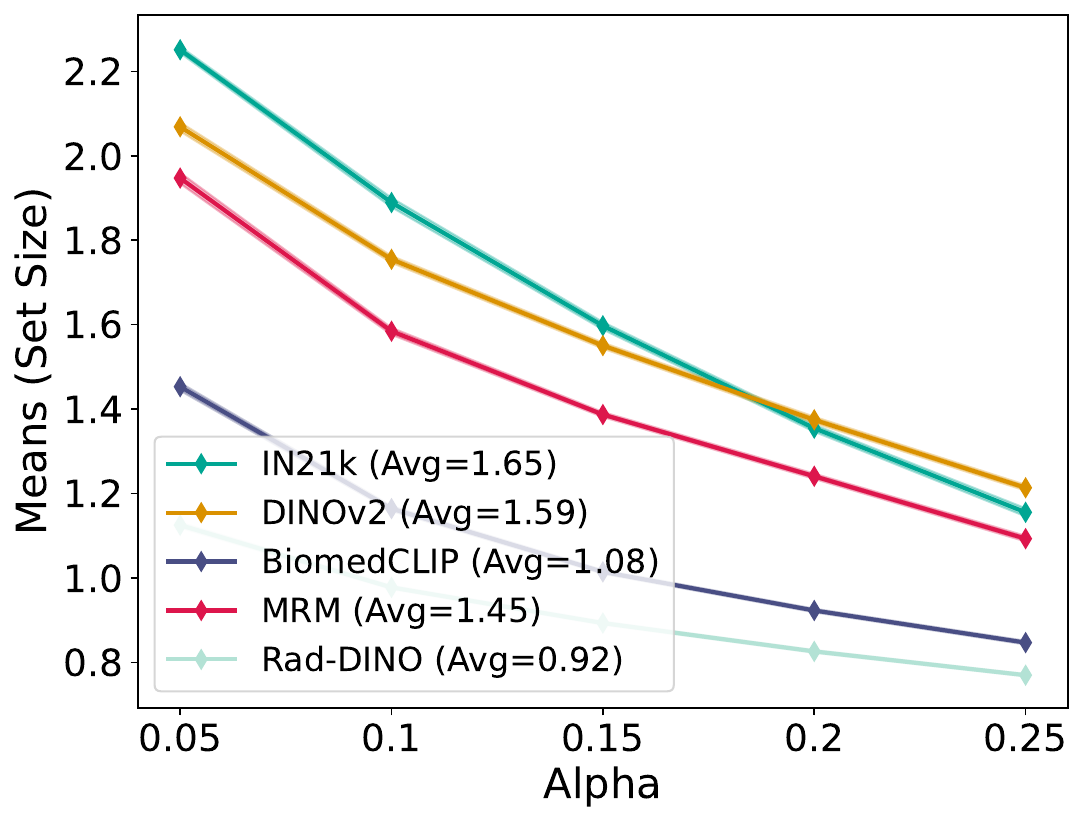}
    \makebox[\textwidth][l]{%
        \hspace{0.24\textwidth}
        \textbf{RSNA (LS)} \hspace{0.04\textwidth} \textbf{POLCOVID (LS)} \hspace{0.01\textwidth} \textbf{COVID-Rad (LS)}
    } \\[0.2cm]
    \includegraphics[height=0.16\textwidth, width=0.20\textwidth]{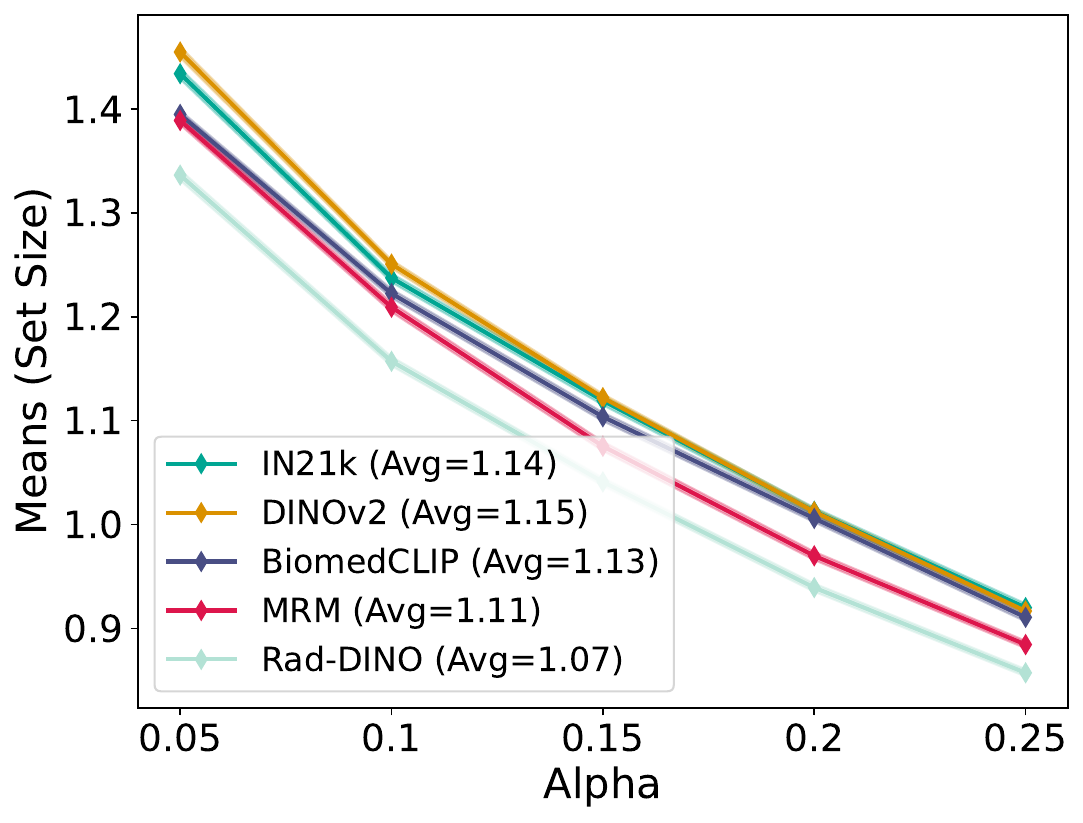}
    \includegraphics[height=0.16\textwidth, width=0.20\textwidth]{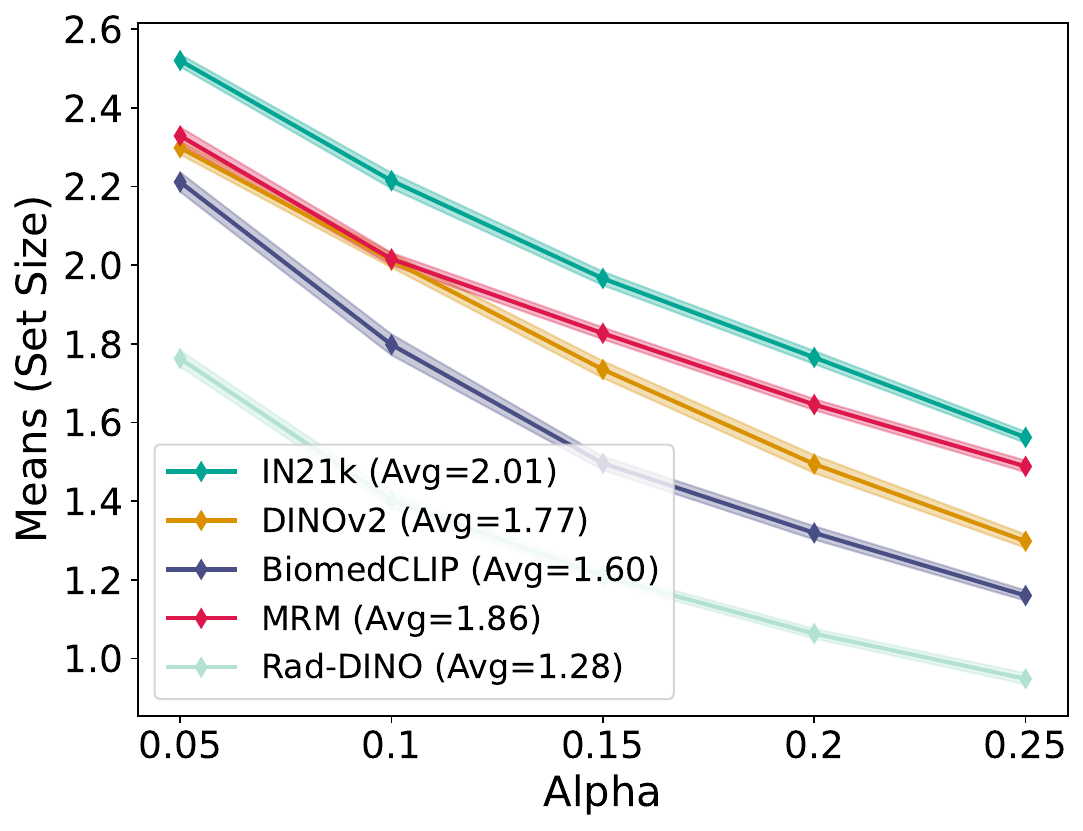}
    \includegraphics[height=0.16\textwidth, width=0.20\textwidth]{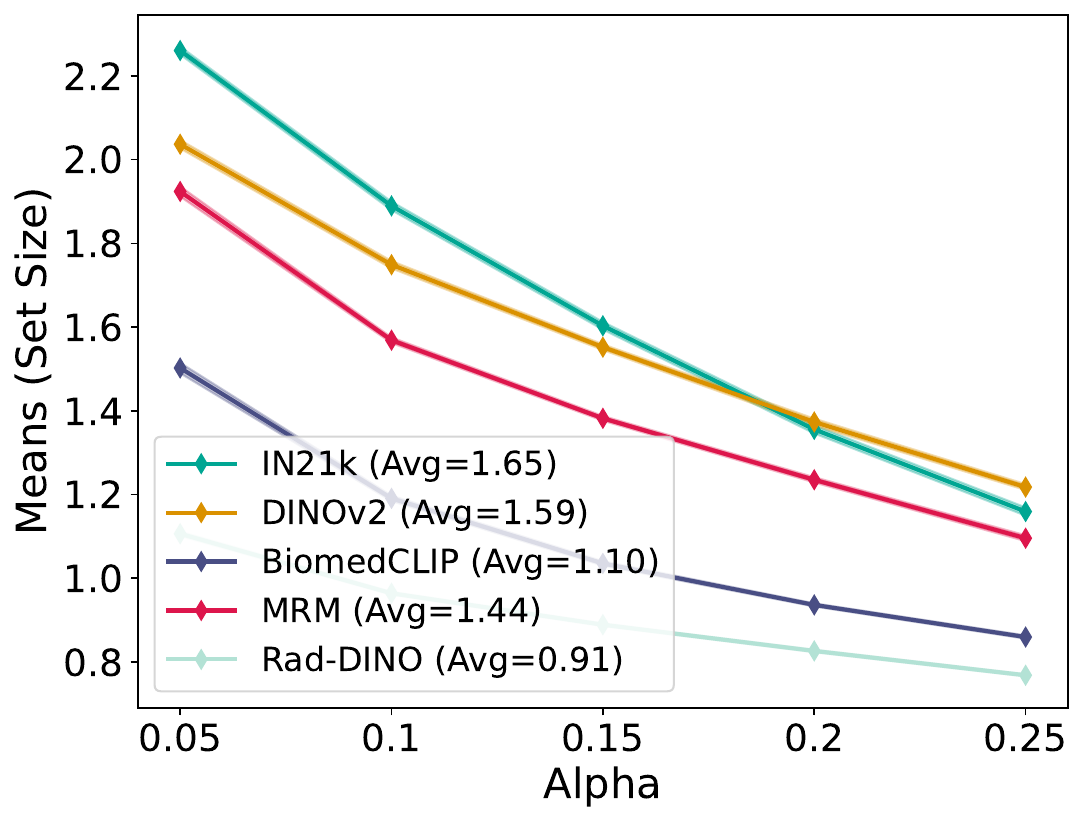}
    \caption{\textbf{Conformal prediction Set Size (X-Rays)}: the average conformal prediction Set Size across different $\alpha$ thresholds for MRI data does not exhibit a clear trend based on point prediction uncertainty, pre-training source, or method. RadImageNet generally shows smaller Set Size with more accurate model performance.}
    \label{fig:cp_mri_setsize}
\end{figure*}

\section{Result}
\subsection{Point Prediction Uncertainty}
We present the result for point prediction with ECE with linear probing, linear probing follows by temperature scaling and linear probing follows by label smoothing on \Cref{fig:point_pred_eval}. We further present the complete results for Brier Score and NLL in the Appendix (\Cref{tab:retina_uncertainty,tab:histopathology_uncertainty,tab:xray_uncertainty}). The observations for point prediction uncertainty experimental results are concluded as follows:

\paragraph{Higher Quality Domain-Specific Pre-training Reduce Uncertainty} Across all datasets, domain-specific pre-training consistently show low uncertainty on ECE before re-calibration compared to other methods with few exceptions (e.g. UNI from TCGA, CTransPath for BraTS), where the discrepancy may come from mismatch on pre-training and downstream data distribution.

\paragraph{Re-calibrating Models Does Not Close the Uncertainty Gap} Our results indicate that while uncertainty calibration techniques such as temperature scaling and label smoothing can reduce model uncertainty in certain cases, they do not consistently bridge the gap between models with inherently different levels of uncertainty. Specifically, models exhibiting higher uncertainty prior to calibration generally remain more uncertain after re-calibration, compared to models that originally had lower uncertainty. This underscores the importance of selecting an appropriate foundation model for domain-specific tasks, as post-hoc calibration and regularization techniques alone cannot fully compensate for suboptimal model choices in pre-training.

\paragraph{Supervised Pre-trained Models Present Higher Uncertainty} Supervised pre-trained models, such as model trained on ImageNet21k generally exhibit higher uncertainty compared to models trained using self-supervised approaches. Previous study~\citep{wang2021rethinking} has attributed this to the overconfidence introduced by updating model parameters with cross-entropy on overfitting specific specific task labels. However, the result shows that even when the backbone model remains entirely unchanged during linear probing for domain-specific tasks --- such as adapting an ImageNet pre-trained model to Retin, Histopathology and X-Rays tasks --- supervised pre-trained models still demonstrate high uncertainty. This increased uncertainty persists even when the downstream tasks differ significantly from the pre-training tasks.

\subsection{Region Prediction Uncertainty}
We present the result for region prediction with LAC (\Cref{alg:1}) and report prediction set size with different $\alpha$ on \Cref{fig:cp_retina_setsize,fig:cp_histopath_setsize,fig:cp_xray_coverage}. We further present the empirical coverage analysis in the Appendix \Cref{fig:cp_retina_coverage,fig:cp_histopath_coverage,fig:cp_xray_coverage} to show that the algorithm is a valid $(1-\alpha)$ conformal predictor in our experimental settings. We additionally show the same experimental settings with an alternative conformal prediction algorithm (RAPS by \citet{angelopoulos2021uncertainty}) in \Cref{apd:raps} to further verify our conclusion. The finding for region prediction uncertainty is concluded as follows:

\paragraph{Better Calibration Does Not Indicate Conformal Prediction Efficiency} Better point prediction calibration after temperature scaling or label smoothing does not translate to more efficient conformal predictor (e.g. lower point prediction uncertainty after calibration does not decrease the prediction set size). This can also be seen from \Cref{alg:1} that even if a model is well-calibrated, it can still assign similar $(1-\alpha)$ conformal quantile threshold, hence not reducing the prediction set size.

\paragraph{Pre-trained Domain-Specific Models Lead to Smaller Conformal Prediction Set} Domain-specific pre-training combined with self-supervised learning improves point prediction uncertainty — but this benefit doesn't always extend to conformal prediction. For example, while a foundation model pre-trained on retinal data still produces relatively large prediction sets, those trained on Histopathology and X-Rays consistently yield much smaller conformal prediction sets. Our experiments show that Histopathology and X-Rays foundation models (e.g., UNI and Rad-DINO) outperform other pre-trained models in large margin while retina foundation model (RETFound) only marginally improves performance in most cases, highlighting that domain-specific pre-training can lead to tighter region prediction under the case of high quality pre-training (see \Cref{apd:model_perf}).

\section{Conclusion}
This work investigates both point- and region-level uncertainty quantification in foundation models for medical image classification. Our findings demonstrate that leveraging domain-specific pre-training data in conjunction with self-supervised learning generally lead to reduced point and region prediction uncertainty. While calibration techniques such as temperature scaling and label smoothing can improve uncertainty calibration, they fall short in fully addressing the discrepancies in uncertainty across foundation models pre-trained on different data sources with appropriate methodologies.

Notably, the efficiency of conformal prediction cannot be directly inferred from point prediction uncertainty calibration. However, conformal prediction remains a robust framework for uncertainty quantification, offering formal coverage guarantees on prediction sets. This capability makes it remain an important tool for human-in-the-loop decision-making, providing interpretable and reliable measures of uncertainty that can enhance clinical workflows~\cite{cresswell2024conformal}.

We underscore the importance of a holistic approach to uncertainty quantification, where the selection of appropriate foundation models and the implementation of uncertainty-aware methods work in tandem to improve the reliability of medical AI systems. While domain-specific models often achieve superior accuracy, they do not trade the model performance with higher uncertainty. By combining careful model selection with rigorous uncertainty quantification techniques, we can foster greater trust in AI-driven medical decisions, ultimately supporting safer and more informed clinical practice.

\subsubsection*{Acknowledgments}
H.H. and N.R. were supported by the National Institute On Aging of the National Institutes of Health under Award R01AG085617. H.H. received partial support from NSF Award 1922658. N.R. were also partially supported by the National Institute On Aging of the National Institutes of Health under Awards R01AG079175 and P30AG066512.

\bibliography{iclr2025_conference}
\bibliographystyle{iclr2025_conference}

\newpage
\appendix
\onecolumn

\section{Dataset}
\label{apd:dataset}
The present the detailed label distribution for train set of each dataset below:

Retina: Normal (N = 168), Glaucoma (N = 56), Cataract (N = 56), Retina disease (N = 56)

IDRiD: No (N = 107), Mild (N = 16), Moderate (N = 108), Severe (N = 59), Proliferative (N = 39)

APTOS2019: No (N = 1805, Mild (N = 370), Moderate (N = 1039), Severe (N = 193), Proliferative (N = 295)

CRC100K: Adipose (N = 10407), Background (N = 10566), Debris (N = 11512), Lymphocytes (N = 11557), Mucus (N = 8896), Smooth muscle (N = 13536), Normal colon mucosa (N = 8763), Cancer-associated stroma (N = 10446),  Colorectal adenocarcinoma epithelium (N = 14317)

TCGA-Lymph: Adrenocortical carcinoma (N = 24290), Bladder Urothelial Carcinoma (N = 48790), Brain Lower Grade Glioma (N = 111990), Breast invasive carcinoma (N = 111550), Cervical squamous cell carcinoma and endocervical adenocarcinoma (N = 29930), Cholangiocarcinoma (N = 3780), Colon adenocarcinoma (N = 41360), Esophageal carcinoma (N = 15840), Glioblastoma multiforme (N = 109520), Head and Neck squamous cell carcinoma (N = 56130), Kidney Chromophobe (N = 12420), Kidney renal clear cell carcinoma (N = 57560), Kidney renal papillary cell carcinoma (N = 32490), Liver hepatocellular carcinoma (N = 38280), Lung adenocarcinoma (N = 76210), Lung squamous cell carcinoma (N = 78380), Lymphoid Neoplasm Diffuse Large B-cell Lymphoma (N = 3390), Mesothelioma (N = 10830), Ovarian serous cystadenocarcinoma (N = 12380), Pancreatic adenocarcinoma (N = 20220), Pheochromocytoma and Paraganglioma (N = 6620), Prostate adenocarcinoma (N = 45120), Rectum adenocarcinoma (N = 8050), Sarcoma (N = 61310), Skin Cutaneous Melanoma (N = 46090), Stomach adenocarcinoma (N = 47010), Testicular Germ Cell Tumors (N = 27890), Thymoma (N = 16860), Thyroid carcinoma (N = 54470), Uterine Carcinosarcoma (N = 10050), Uterine Corpus Endometrial Carcinoma (N = 58800), Uveal Melanoma (N = 8200)

BraTS-Path: Cellular tumor (N = 2000), Pseudopalisading necrosis (N = 2000), Areas abundant
in microvascular proliferation (N = 2000), Geographic necrosis (N =2 000), Infiltration into the cortex (N = 2000), Penetration into white matter (N = 2000)

RSNA-Pneumonia: Normal (N = 20672), Pneumonia (N = 6012)

POLCOVID: Normal (N = 2426), Covid (N = 1236), Pneumonia (N = 1147)

COVID-Rad: Normal (N = 10192), Covid (N = 3616), Viral Pneumonia (N = 1345), Lung Opacity (N = 6012)

\section{Metrics and Algorithms}
\label{apd:metrics}
\paragraph{Expected Calibration Error (ECE)}
Given the notion that mis-calibration is defined as the difference in expectation between model confidence and accurcay for point prediction, ECE approximates the expectation by partitioning the predictions into $M$ bins baesed on their predicted probabilities and then aggregating the discrepancies with each bin. Given a set of prediction $\{(\hat{p}_i,y_i)\}^{n}_{i=1}$, where $\hat{p}_{i}\in[0,1]$ is the predicted probability of positive label and $y_i\in\{0,1\}$ is the true label, ECE is formally defined as
\begin{align}
    ECE=\sum_{m=1}^{M}\frac{|B_m|}{n}|\text{acc}(B_m)-\text{conf}(B_m)|
\end{align}
where $B_m$ is the number of instances in bin $m$, $n$ is the total number of instances, $\text{acc}(B_m)=\frac{1}{|B_m|}\sum_{i\in B_{m}}y_i$ is the accuracy in bin $m$, $\text{conf}(B_m)=\frac{1}{B_m}\sum_{i\in B_m}\hat{p}_{i}$ is the average predicted confidence in bin $m$.

While ECE provides a direct measure of model calibration, it suffers from sensitivity of bin sizes choice and loss of information within bins by aggregating discrepancies within each bin.

\paragraph{Brier Score (BS)}
Brier Score measures the mean squared difference between predicted probabilities and the actual outcomes. It is commonly used for evaluating model uncertainty and sharpness of probabilistic predictions. Given the set of predictions $\{\hat{p}_i,y_i\}_{i=1}^{n}$ with same definition as before, BS is defined as
\begin{align}
    BS=\frac{1}{n}\sum_{i=1}^{n}(\hat{p}_i-y_i)^2
\end{align}
While Brier Score can provide measure on both calibration and refinement, it also makes it less suitable for scenario where calibration and refinement evaluation need to be separated. 

\paragraph{Negative Log-Likelihood (NLL)}
Negative Log-Likelihood accesses the probability assigned to the true class labels, where it penalizes incorrect and uncertain predictions more severely than correct and confidence ones. Given the set of predictions $\{\hat{p}_i,y_i\}_{i=1}^{n}$ with same definition as before, NLL is defined as
\begin{align}
    NLL=-\frac{1}{n}\sum_{i=1}^{n}(y_i\log(\hat{p}_i)+(1-y_i)\log(1-\hat{p}_i))
\end{align}
While NLL directly measure the likelihood of the true labels with predicted probability distribution, probing a direct probabilistic assessment, it focus on penalizing the confident errors, which may cause misleading calibration if the model is mis-specified or probabilities are misaligned.

\paragraph{Temperature Scaling}
Temperature Scaling~\cite{pmlr-v70-guo17a} smooth the model probability distribution by dividing model logits output with a single scalar parameter $T$. Given calibrated probabilities $\mathbf{p}=(p_1,...,p_K)$ and logits output $\mathbf{z}=(z_1,...,z_K)$, the calibrated probability for each class can be represented as 
\begin{align}
    p_i=\frac{\exp(\frac{z_i}{T})}{\sum_{j=1}^{K}\exp(\frac{z_j}{T})},\quad\forall i=1,...,K
\end{align}

While temperature is a post-hoc calibration method that is simple and with small computation overhead, it heavily relies on the choice of $T$ to achieve correct calibration with $T$ to be chosen from a held-out validation set by minimizing some evaluation metric (e.g. Negative Log-Likelihood). This can cause mis-calibration when the validation set does not well represent the data distribution of test set. 

In this study, a separate held-out set is created for each dataset and the optimal temperatures are computed by minimizing Negative Log-Likelihood on this held-out set.

\paragraph{Label Smoothing}
Label smoothing~\cite{NEURIPS2019_f1748d6b} modifies the target class distribution to be a mixture of the one-hot encoded vector and a uniform distribution. Specifically, the modified target after label smoothing become
\begin{align}
    y'_j=
    \begin{cases} 
    1-\epsilon+\frac{\epsilon}{K} & \text{if }j=k \\
    \frac{\epsilon}{K} & \text{otherwise}
    \end{cases}
\end{align}
for a specified smoothing parameter $\epsilon\in[0,1]$, number of classes $K$ and class index $j$.

Label smoothing is equivalent to introduce a regularization term that encourage the model to distribute probability mass evenly across classes, where 
\begin{align}
    \mathcal{L}=-\sum_{j=1}^{K}y'_j\log p_j=\mathcal{L}_{CE}+\epsilon\mathbb{E}_\textbf{u}[-\log p_j]
\end{align}
for uniform distribution $\mathbf{u}$ with $u_j=\frac{1}{K}$. We clarify the derivation of this conclusion in \Cref{apd:label_smoothing}.

While label smoothing can be a simple method to mitigate model over-confidence, it is sensible to the choice of penalty term, where a poor choice of penalty term can lead to under-confidence. Additionally, as a regularization method, label smoothing requires model re-training, which introduces additional computational overhead.

In this study, different $\epsilon=\{0.05, 0.10, 0.15, 0.20\}$ values are experimented on each dataset and the optimal $\epsilon$ is chosen for the result.

\paragraph{Least Ambiguous set-valued Classifier (LAC)}
LAC (shown in \Cref{alg:1}) is a recent conformal prediction algorithm \citep{Sadinle2016LeastAS} that is proven to minimize the average set size with accurate input probabilities and ensures small sets even when probabilities are only approximately correct. The algorithm detail is provided in \Cref{apd:lac}

In this study, as an image classification task, the conformal score is designed as 1-softmax{\it (logits)} for true class of model {\it logits} output.

\paragraph{Regularized Adaptive Prediction Sets (RAPS)}
RAPS~\citep{angelopoulos2021uncertainty} is a method for constructing conformal prediction sets that typically produces smaller sets (on average) than simpler approaches like top-k classification at the same coverage level. It does this by defining a score function that balances three components:
\begin{itemize}
    \item Probability Mass of More Likely Labels: $\rho_x(y)=\sum_{y'}f(x)_{y'}\mathbbm{1}[f(x)_{y'}>f(x)_{y}]$
    which is how much probability mass is assigned to labels more likely than $y$.
    \item Randomly Weighted Probability of the Candidate Label $uf(x)$ for $u\sim Uniform(0,1)$, which helps break ties among labels that have similar probabilities.
    \item Set Size Regularizer: $\lambda(o_x(y)-k_{reg})_{+}$ where $o_x(y)$ is the ranking of $y$ by its (softmax) score, $k_{reg}$ is a desired baseline for how many labels to include, and $\lambda$ controls the penalization for exceeding the baseline.
\end{itemize}
given $f(x)$ represents model probability output for ground truth label $y$. We follow the original paper to choose optimal $k_{reg}$ and $\lambda$.

\section{Conformal Prediction Set Size and Coverage with RAPS}
\label{apd:raps}
\begin{figure*}[!ht]
    \centering
    \makebox[\textwidth][l]{%
        \hspace{0.23\textwidth}
        \textbf{Retina} \hspace{0.14\textwidth} \textbf{IDRiD} \hspace{0.14\textwidth} \textbf{APTOS2019}
    } \\[0.2cm]
    \includegraphics[height=0.19\textwidth, width=0.23\textwidth]{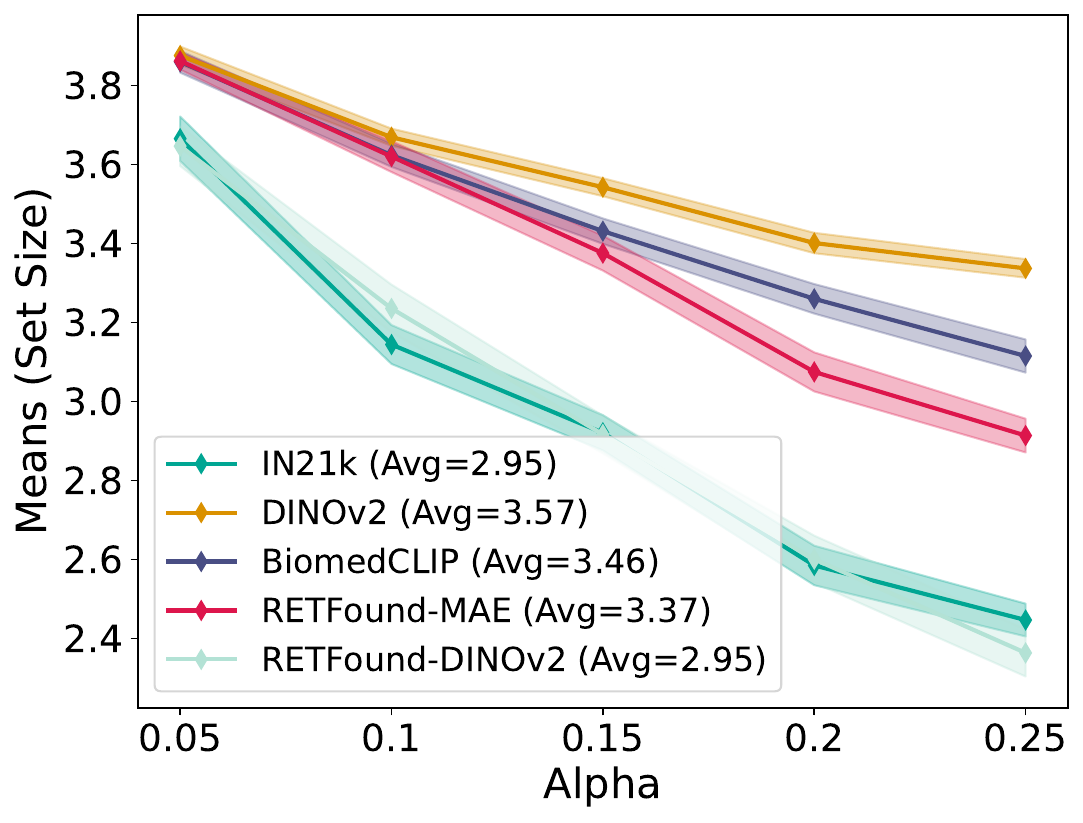}
    \includegraphics[height=0.19\textwidth, width=0.23\textwidth]{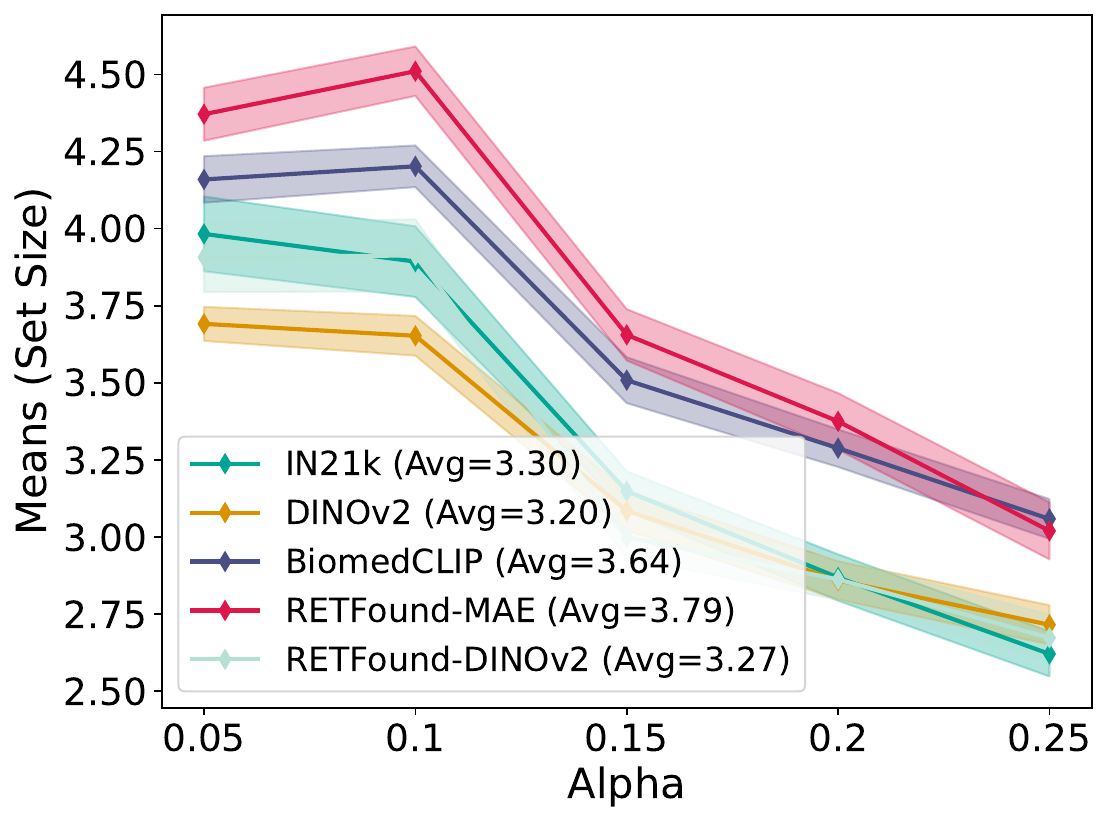}
    \includegraphics[height=0.19\textwidth, width=0.23\textwidth]{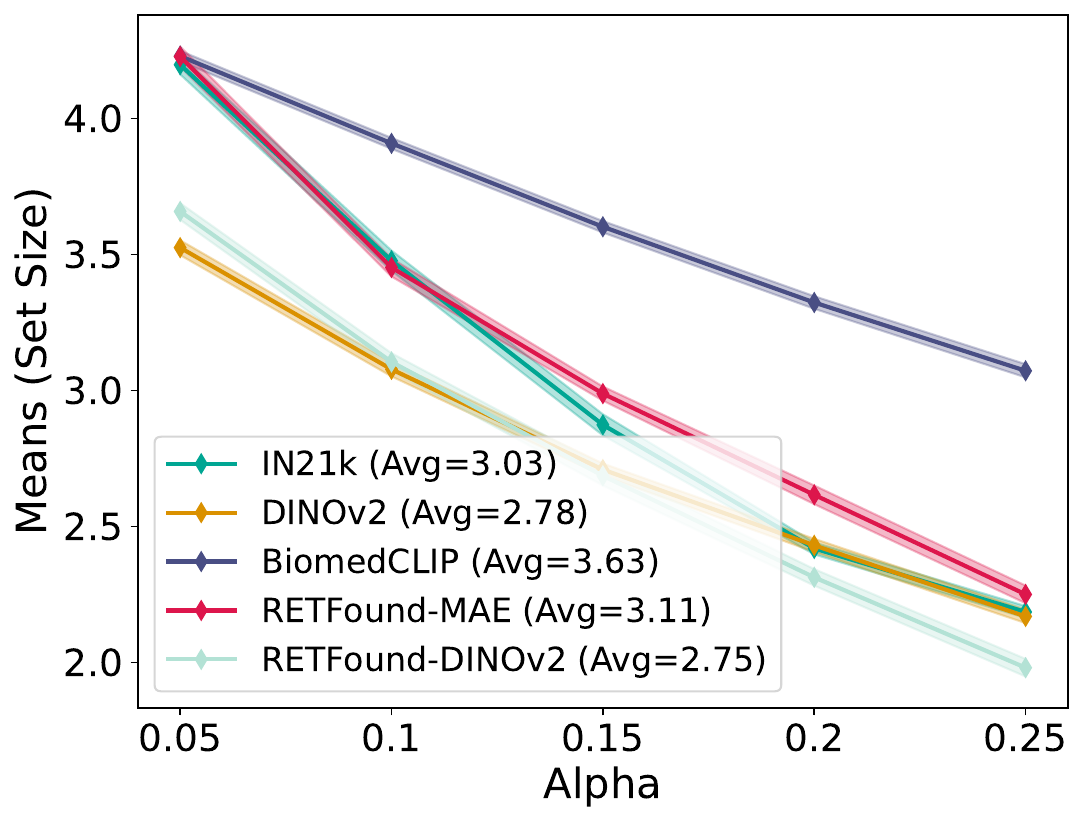}\\[0.2cm]
    \makebox[\textwidth][l]{%
        \hspace{0.21\textwidth}
        \textbf{Retina (T)} \hspace{0.11\textwidth} \textbf{IDRiD (T)} \hspace{0.10\textwidth} \textbf{APTOS2019 (T)}
    } \\[0.2cm]
    \includegraphics[height=0.19\textwidth, width=0.23\textwidth]{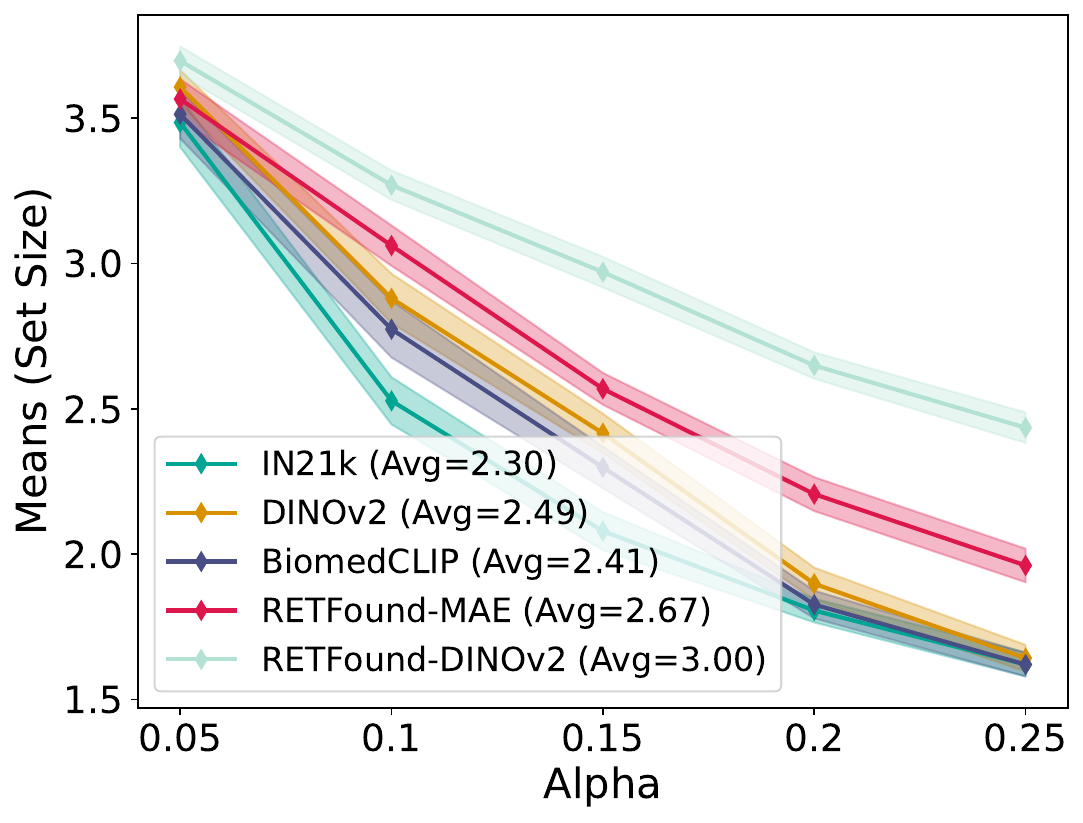}
    \includegraphics[height=0.19\textwidth, width=0.23\textwidth]{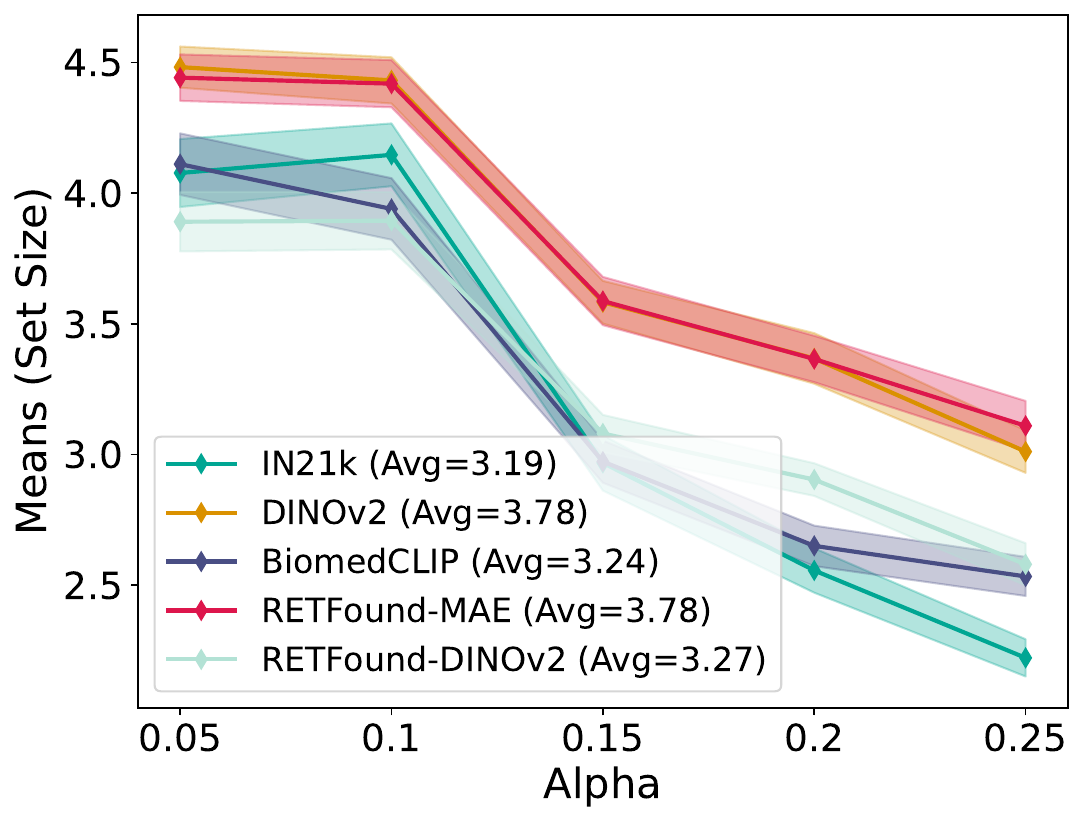}
    \includegraphics[height=0.19\textwidth, width=0.23\textwidth]{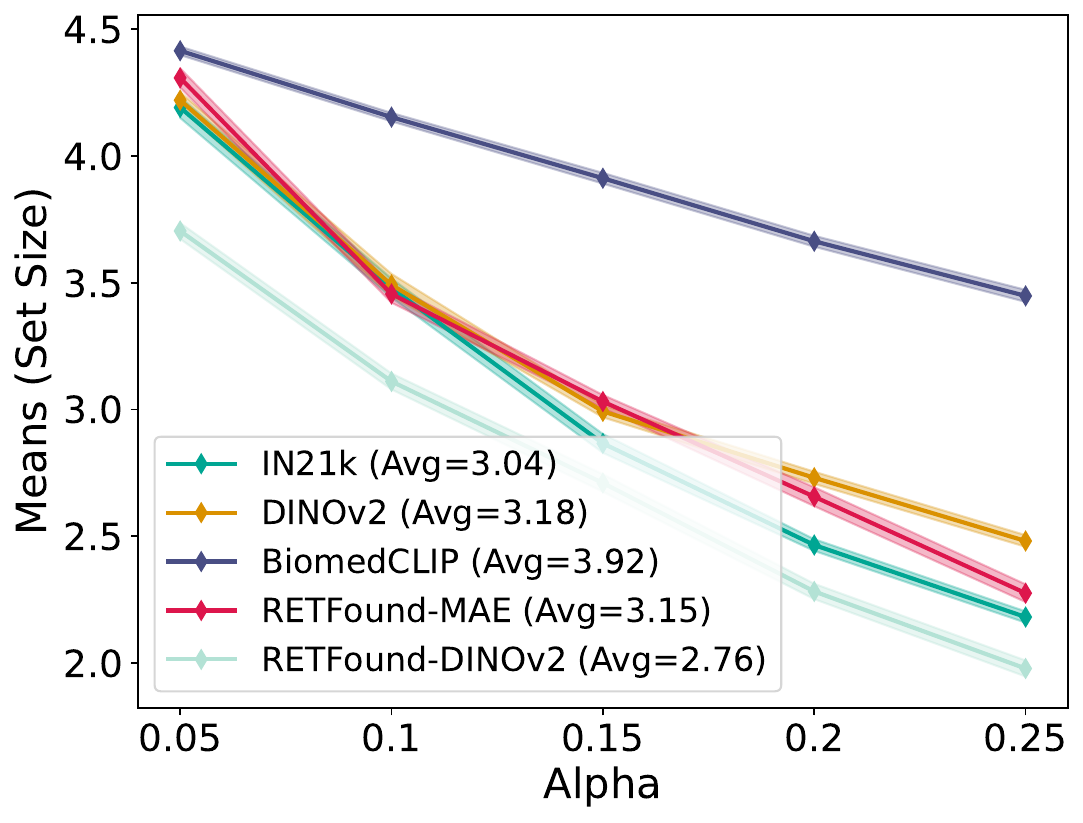}\\[0.2cm]
    \makebox[\textwidth][l]{%
        \hspace{0.20\textwidth}
        \textbf{Retina (LS)} \hspace{0.10\textwidth} \textbf{IDRiD (LS)} \hspace{0.08\textwidth} \textbf{APTOS2019 (LS)}
    } \\[0.2cm]
    \includegraphics[height=0.19\textwidth, width=0.23\textwidth]{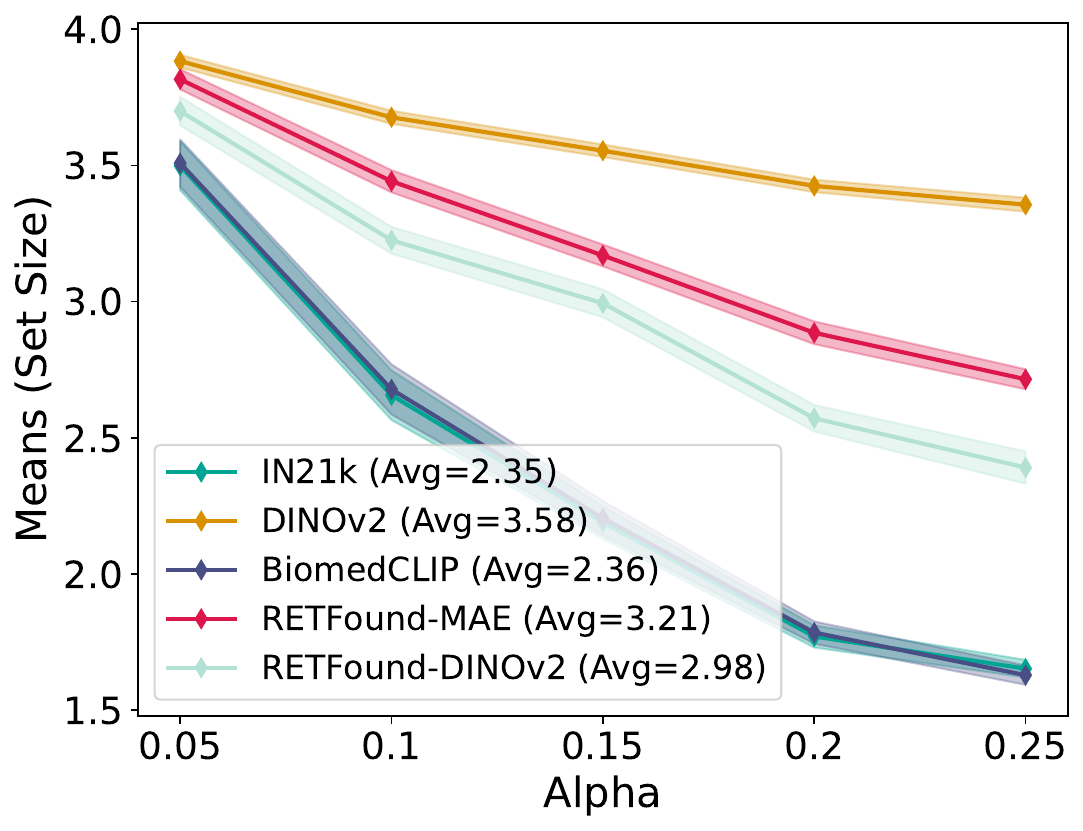}
    \includegraphics[height=0.19\textwidth, width=0.23\textwidth]{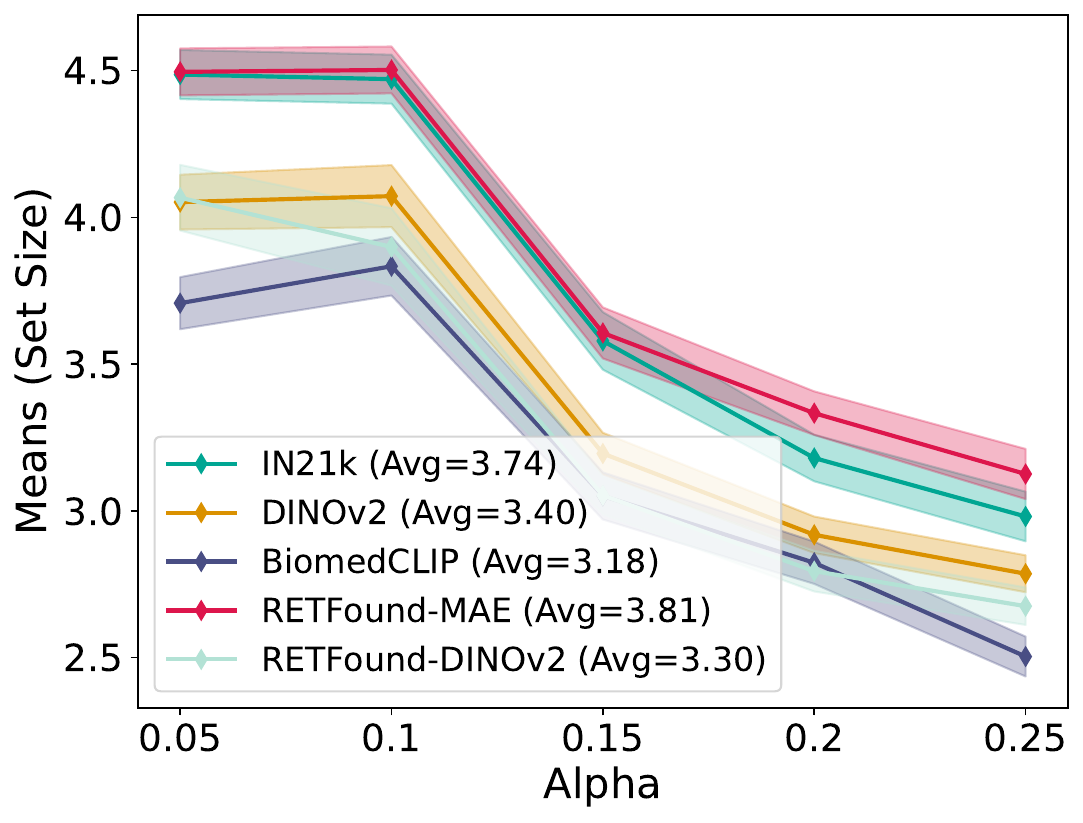}
    \includegraphics[height=0.19\textwidth, width=0.23\textwidth]{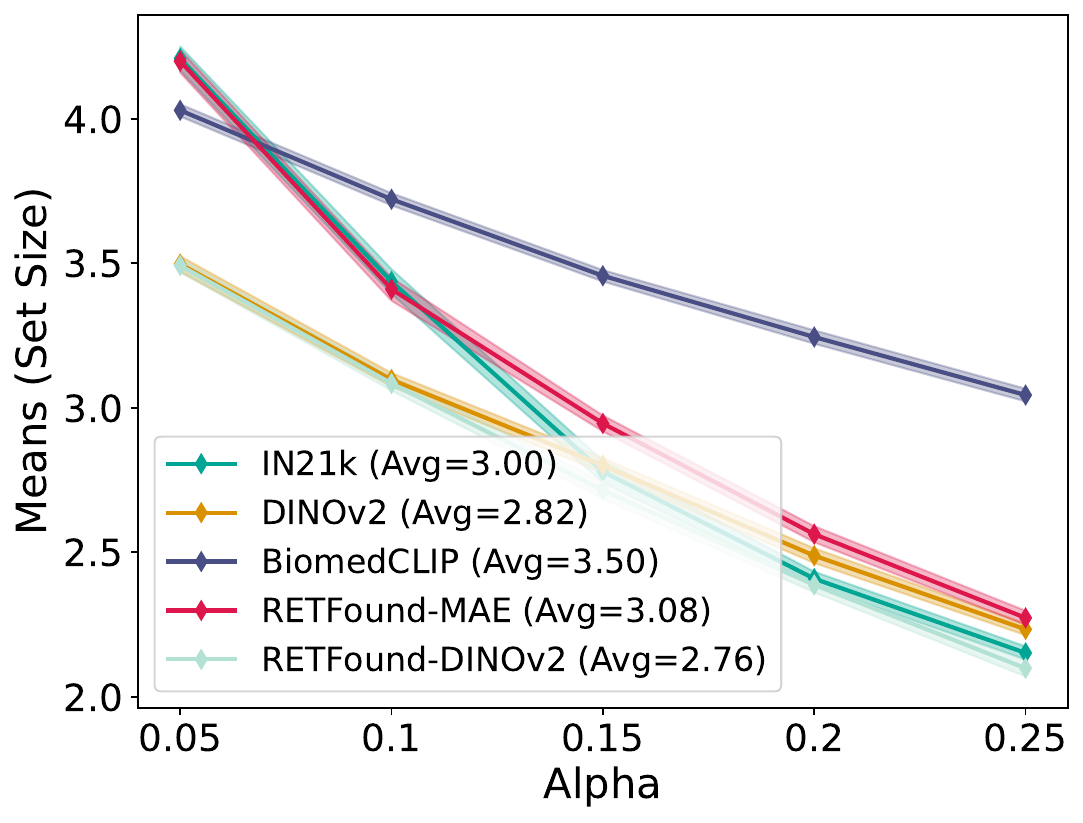}
    \caption{\textbf{Conformal prediction Set Size - RAPS (Retina)}}
    \label{fig:raps_cp_retina_setsize}
\end{figure*}

\begin{figure*}[!ht]
    \centering
    \makebox[\textwidth][l]{%
        \hspace{0.21\textwidth}
        \textbf{CRC100K} \hspace{0.13\textwidth} \textbf{TCGA} \hspace{0.15\textwidth} \textbf{BraTS}
    } \\[0.2cm]
    \includegraphics[height=0.19\textwidth, width=0.23\textwidth]{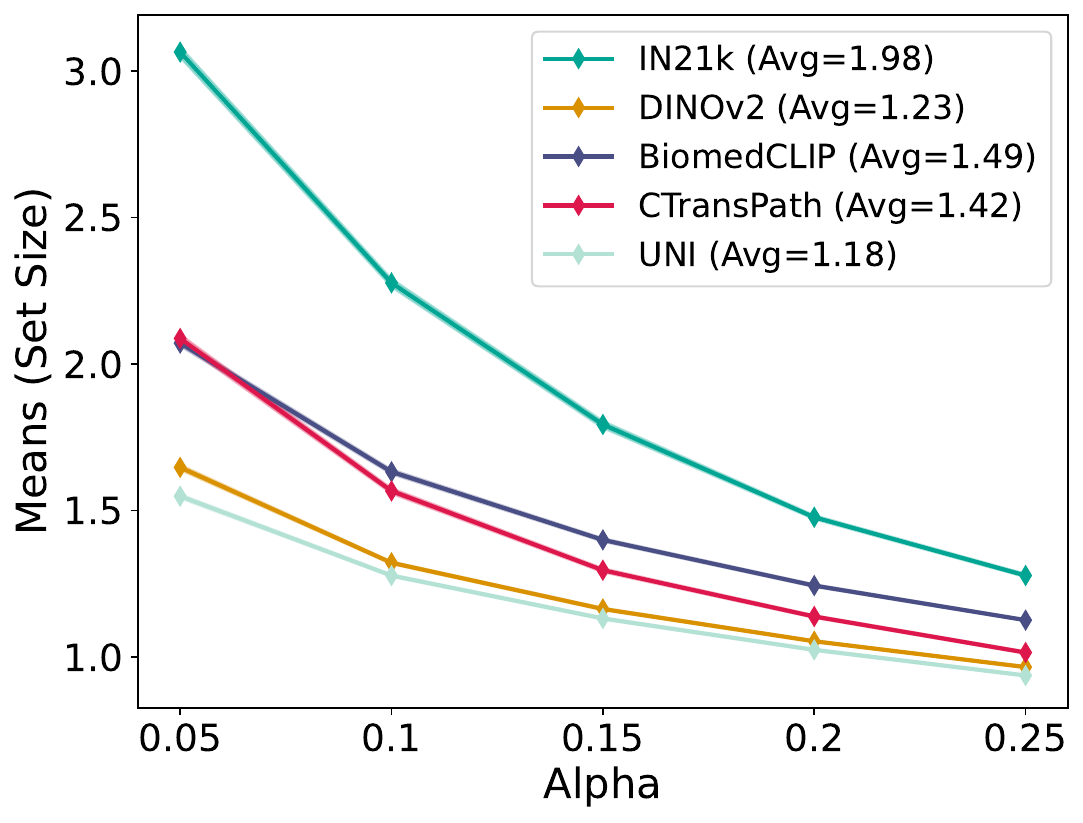}
    \includegraphics[height=0.19\textwidth, width=0.23\textwidth]{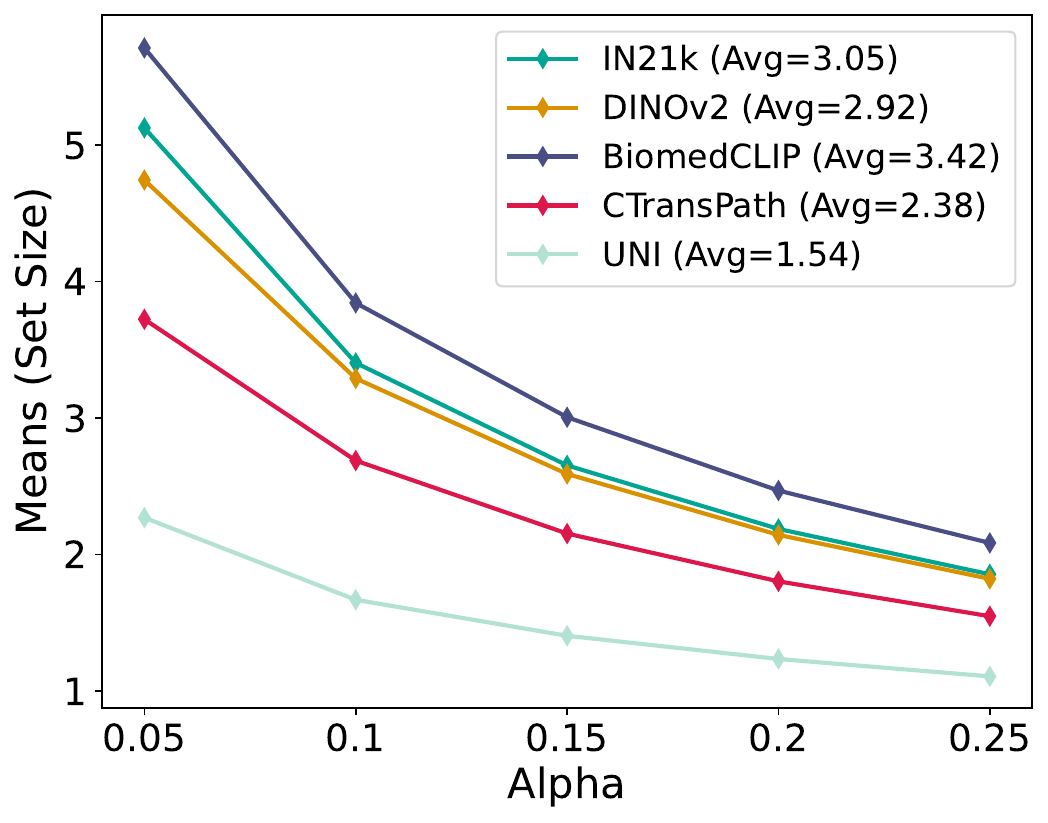}
    \includegraphics[height=0.19\textwidth, width=0.23\textwidth]{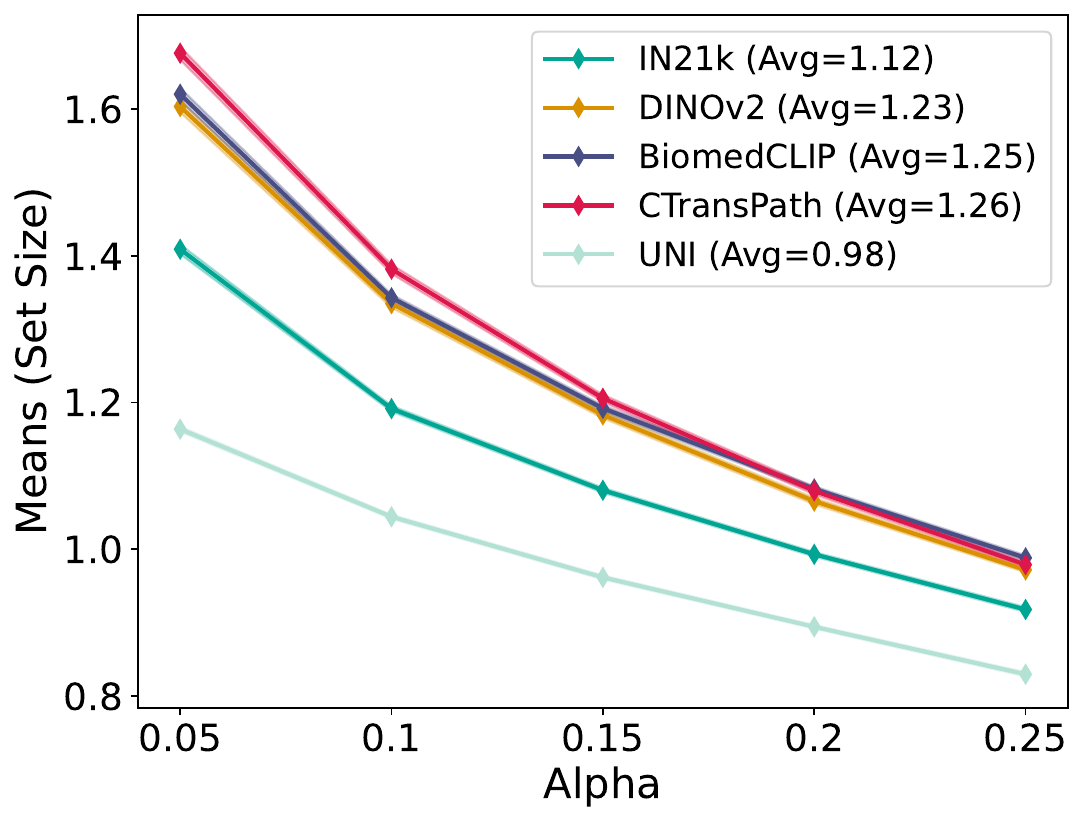}
    \makebox[\textwidth][l]{%
        \hspace{0.20\textwidth}
        \textbf{CRC100K (T)} \hspace{0.08\textwidth} \textbf{TCGA (T)} \hspace{0.11\textwidth} \textbf{BraTS (T)}
    } \\[0.2cm]
    \includegraphics[height=0.19\textwidth, width=0.23\textwidth]{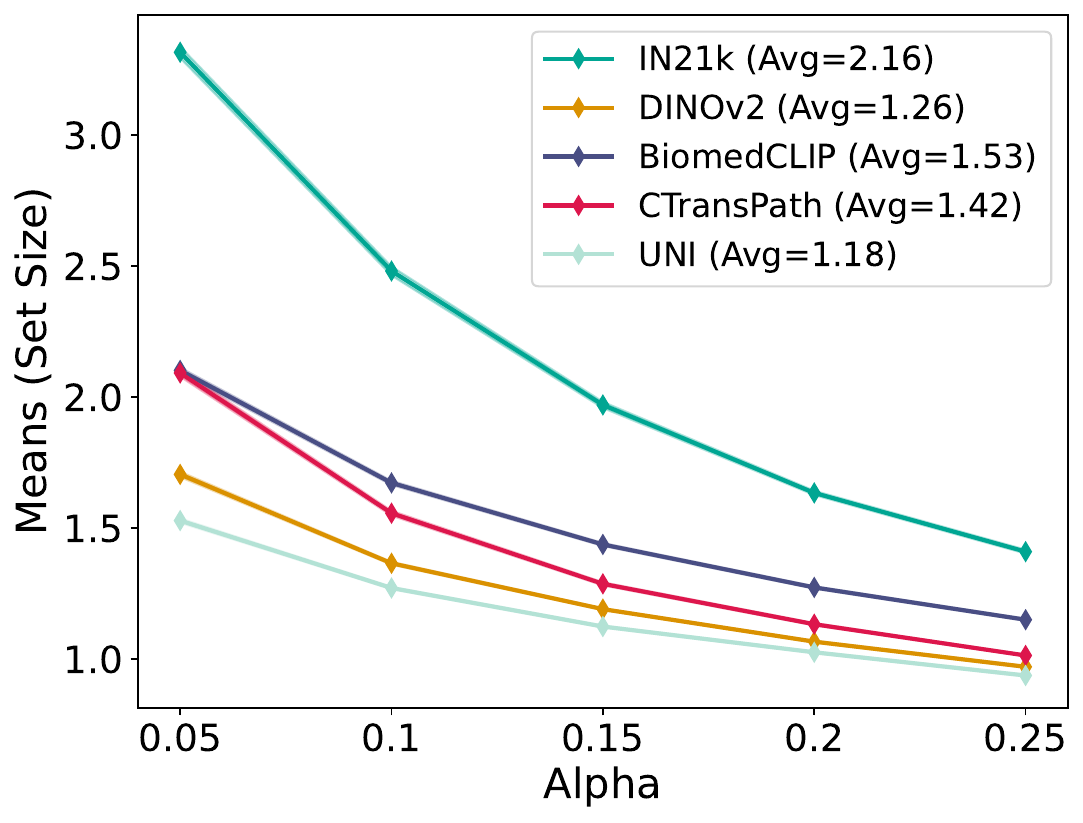}
    \includegraphics[height=0.19\textwidth, width=0.23\textwidth]{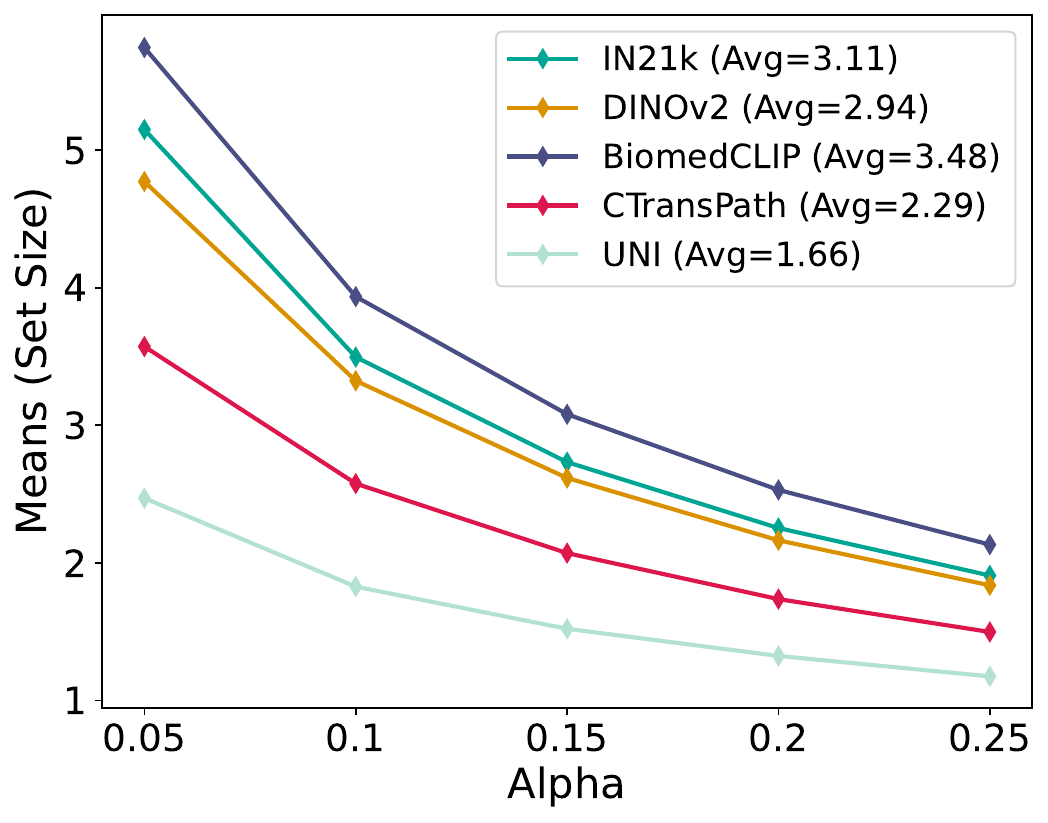}
    \includegraphics[height=0.19\textwidth, width=0.23\textwidth]{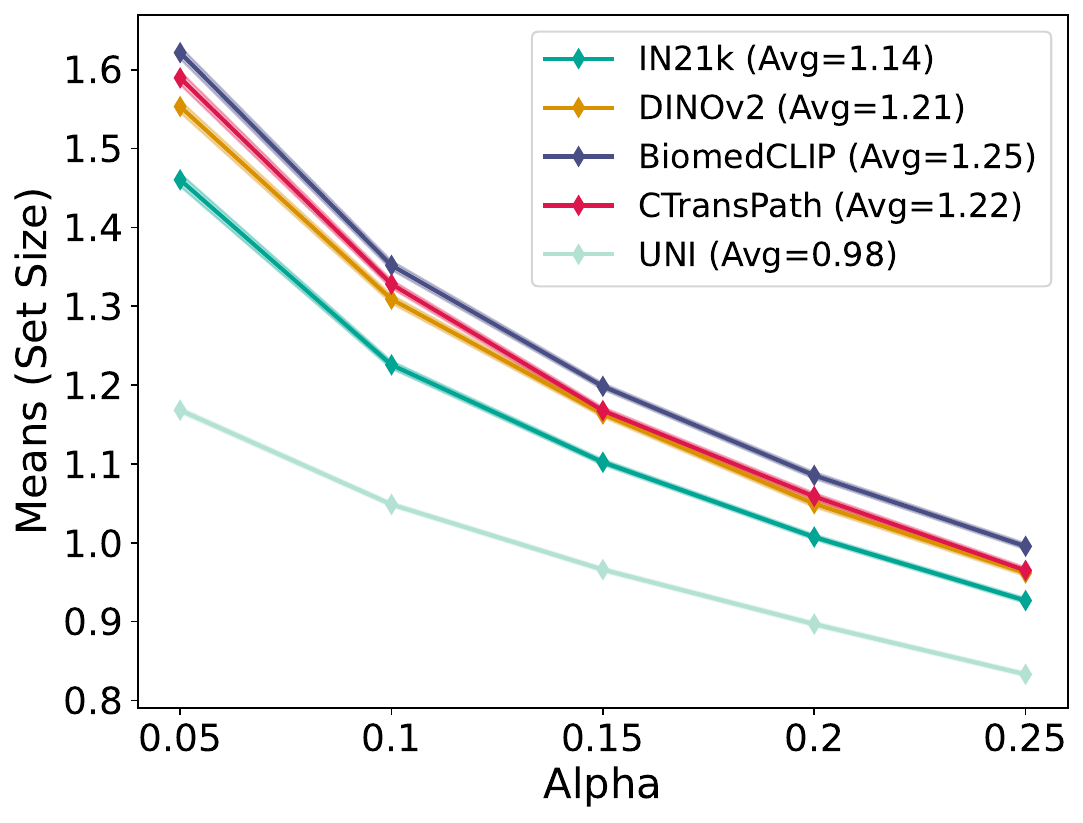}    \makebox[\textwidth][l]{%
        \hspace{0.19\textwidth}
        \textbf{CRC100K (LS)} \hspace{0.07\textwidth} \textbf{TCGA (LS)} \hspace{0.09\textwidth} \textbf{BraTS (LS)}
    } \\[0.2cm]
    \includegraphics[height=0.19\textwidth, width=0.23\textwidth]{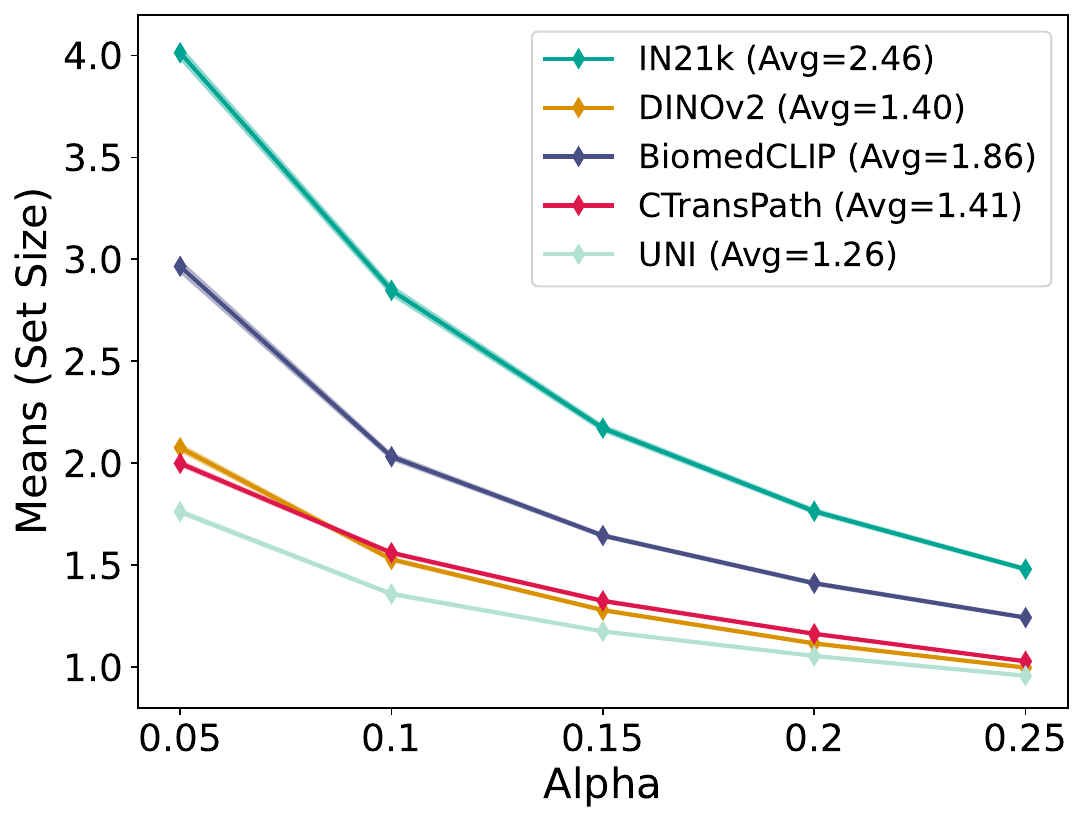}
    \includegraphics[height=0.19\textwidth, width=0.23\textwidth]{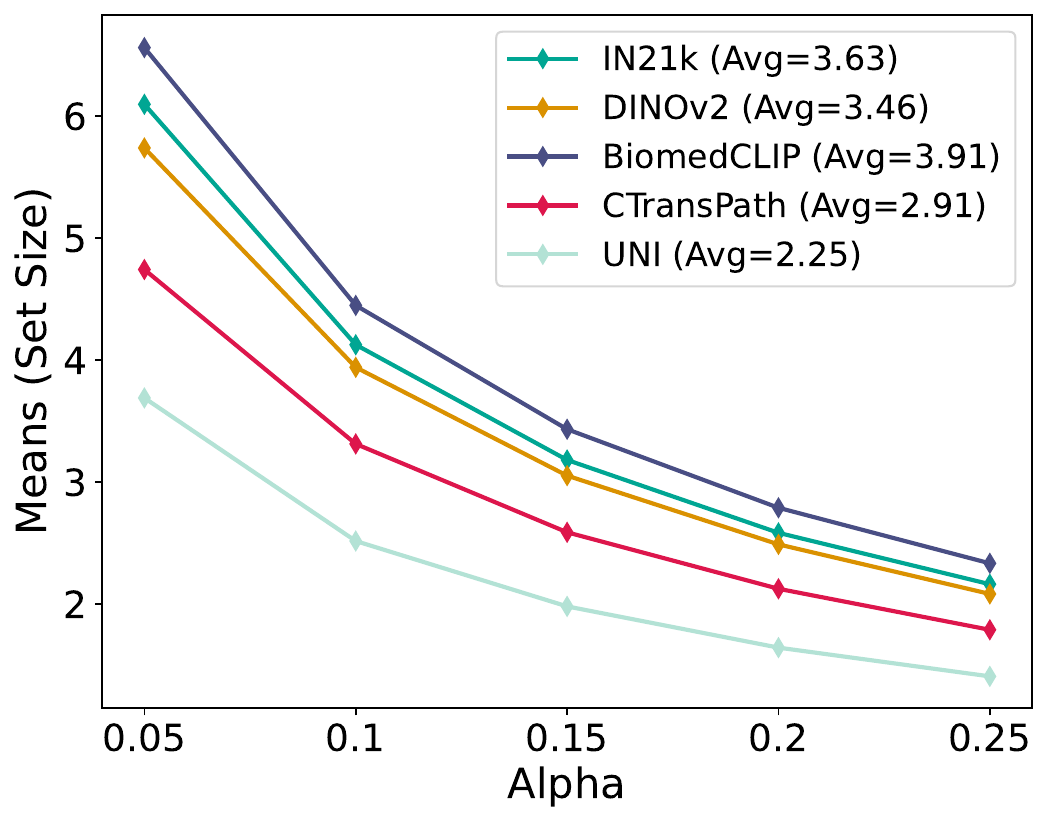}
    \includegraphics[height=0.19\textwidth, width=0.23\textwidth]{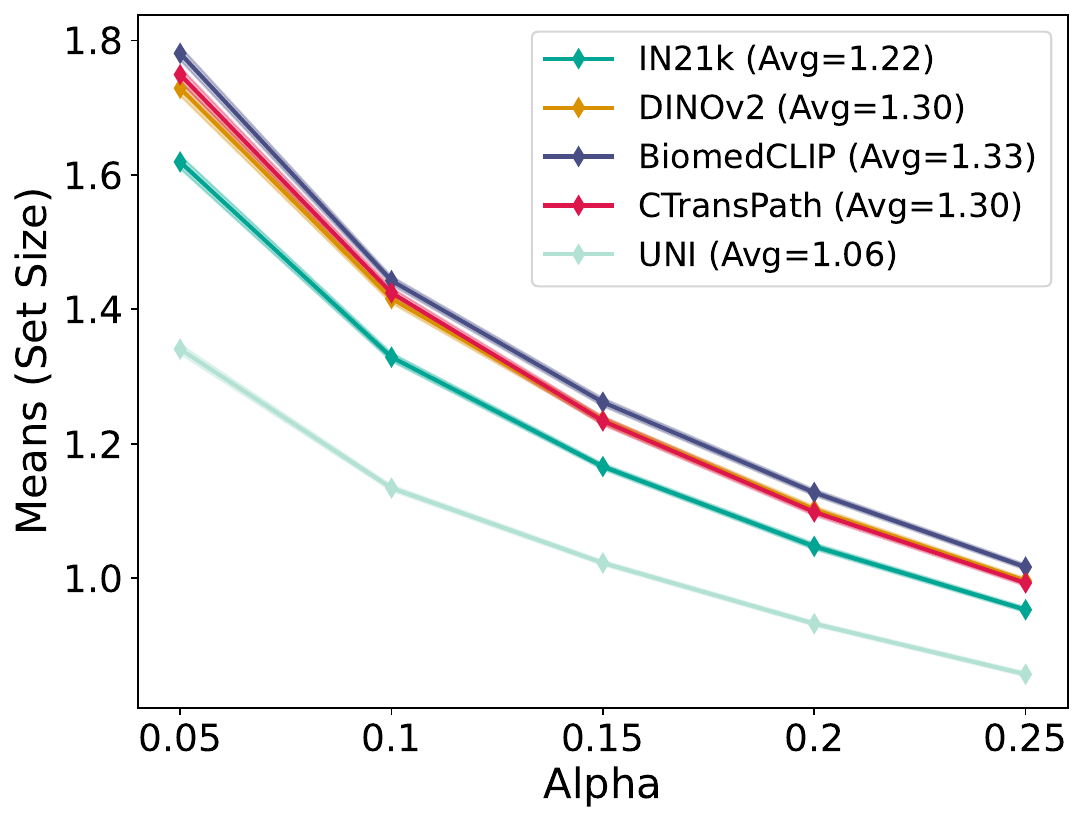}
    \caption{\textbf{Conformal prediction Set Size - RAPS (Histopathology)}}
    \label{fig:raps_cp_histopath_setsize}
\end{figure*}

\begin{figure*}[!ht]
    \centering
    \makebox[\textwidth][l]{%
        \hspace{0.24\textwidth}
        \textbf{RSNA} \hspace{0.12\textwidth} \textbf{POLCOVID} \hspace{0.10\textwidth} \textbf{COVID-Rad}
    } \\[0.2cm]
    \includegraphics[height=0.19\textwidth, width=0.23\textwidth]{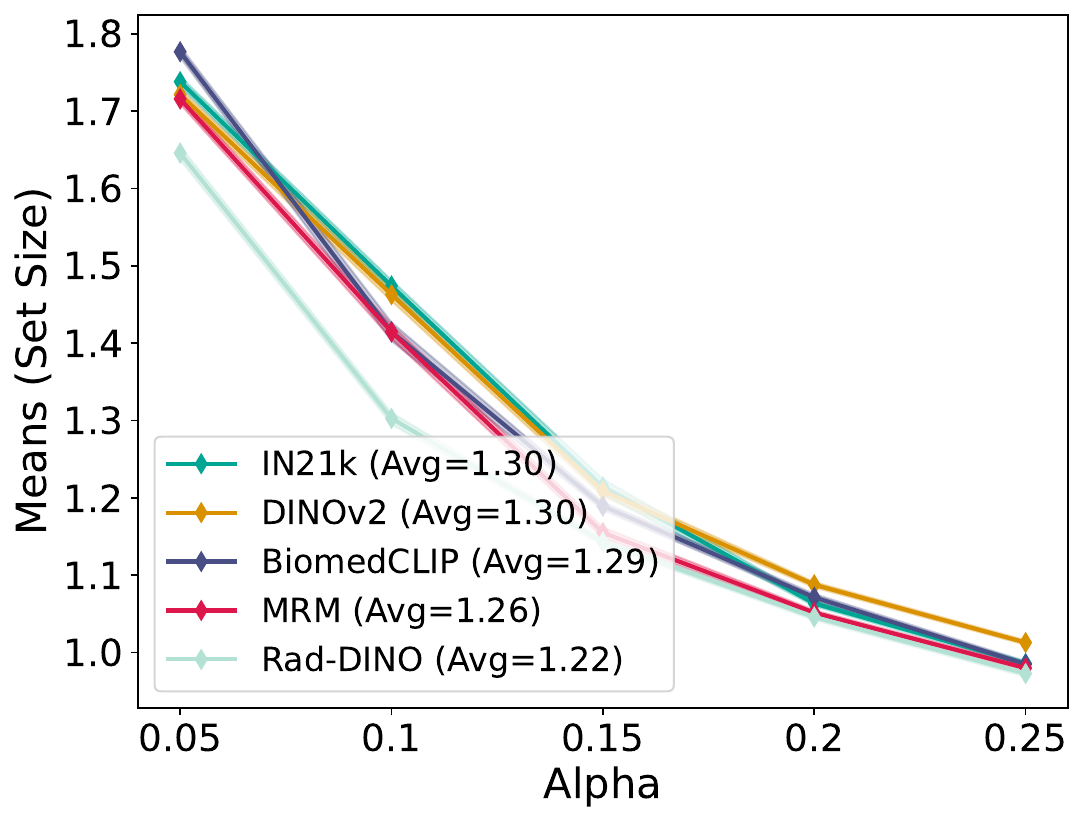}
    \includegraphics[height=0.19\textwidth, width=0.23\textwidth]{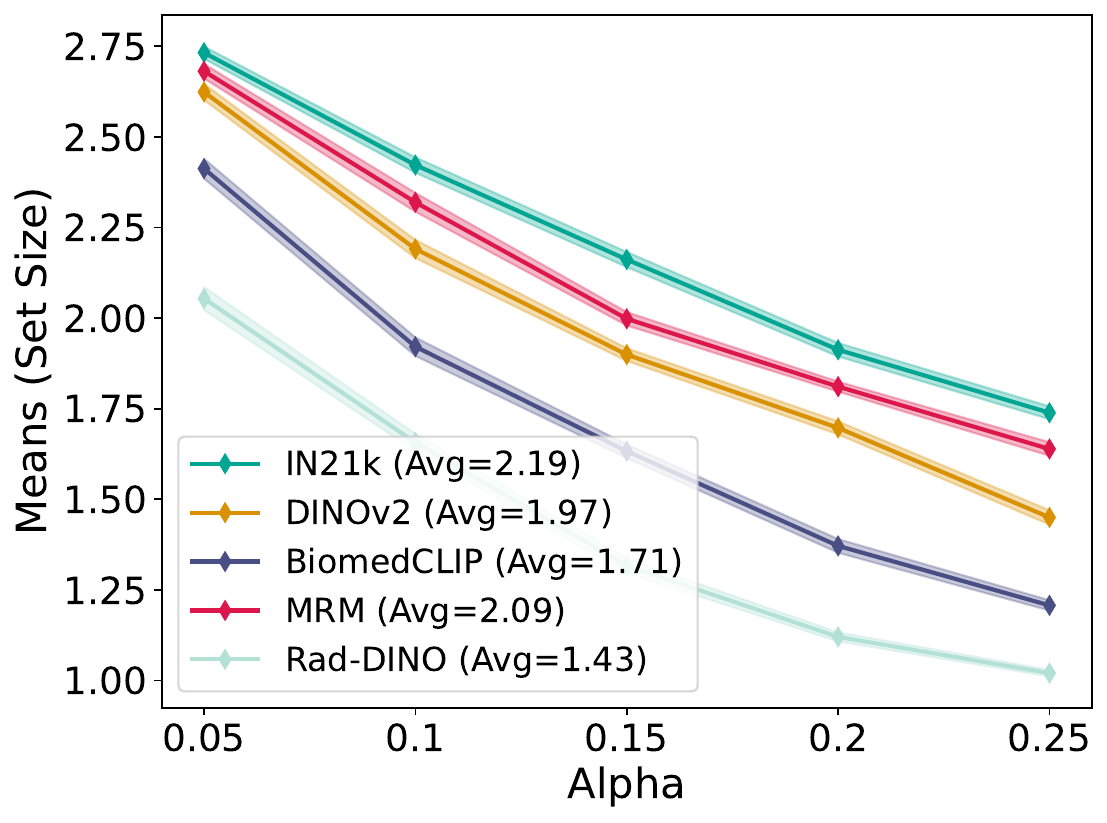}
    \includegraphics[height=0.19\textwidth, width=0.23\textwidth]{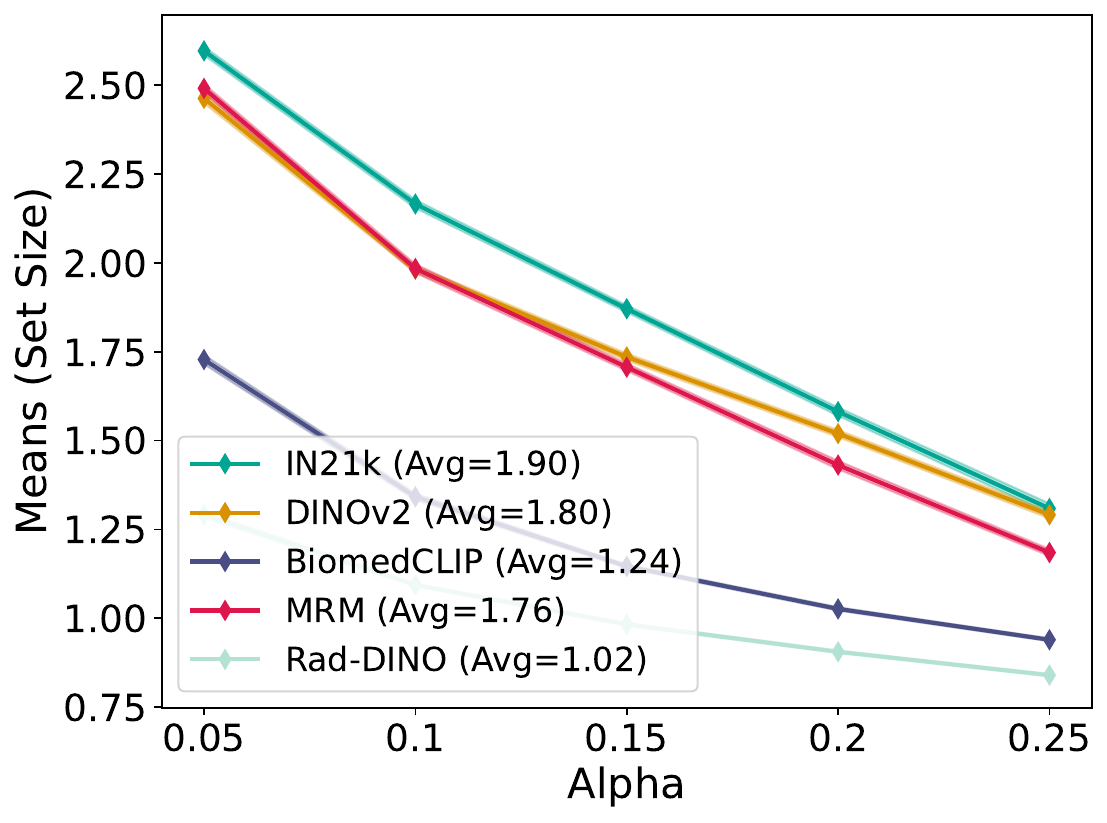}
    \makebox[\textwidth][l]{%
        \hspace{0.22\textwidth}
        \textbf{RSNA (T)} \hspace{0.09\textwidth} \textbf{POLCOVID (T)} \hspace{0.05\textwidth} \textbf{COVID-Rad (T)}
    } \\[0.2cm]
    \includegraphics[height=0.19\textwidth, width=0.23\textwidth]{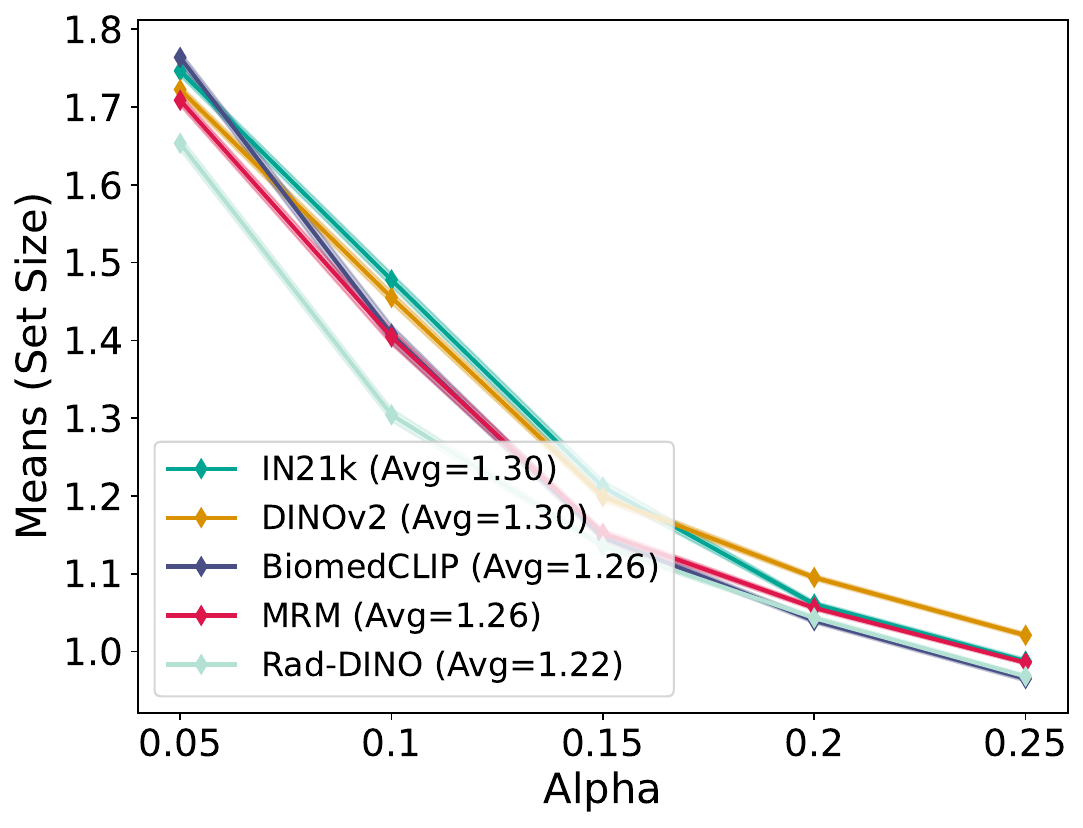}
    \includegraphics[height=0.19\textwidth, width=0.23\textwidth]{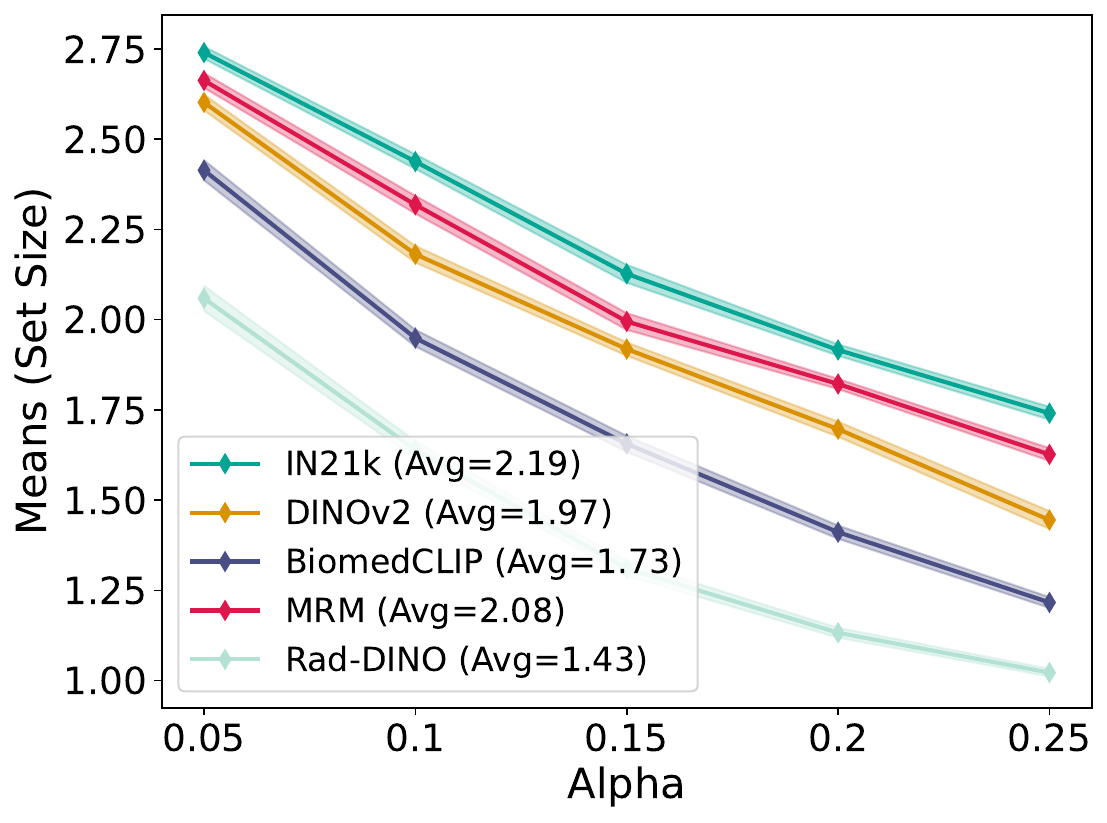}
    \includegraphics[height=0.19\textwidth, width=0.23\textwidth]{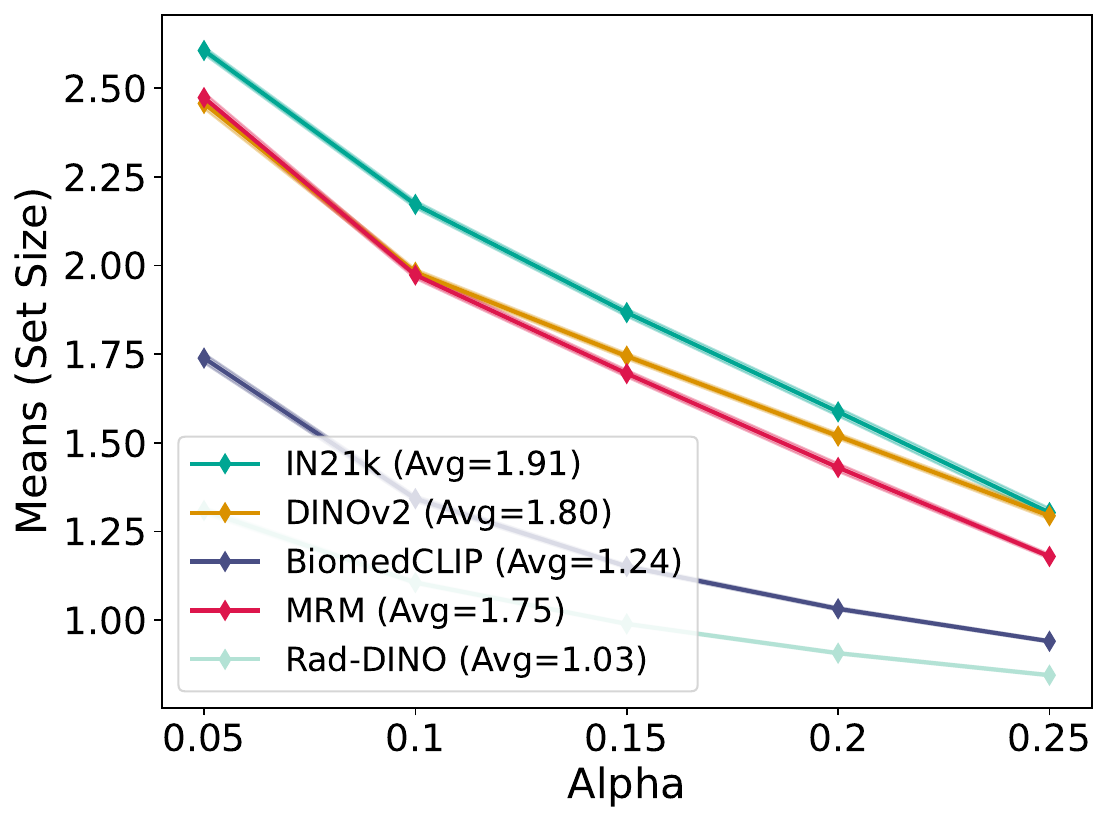}
    \makebox[\textwidth][l]{%
        \hspace{0.21\textwidth}
        \textbf{RSNA (LS)} \hspace{0.08\textwidth} \textbf{POLCOVID (LS)} \hspace{0.04\textwidth} \textbf{COVID-Rad (LS)}
    } \\[0.2cm]
    \includegraphics[height=0.19\textwidth, width=0.23\textwidth]{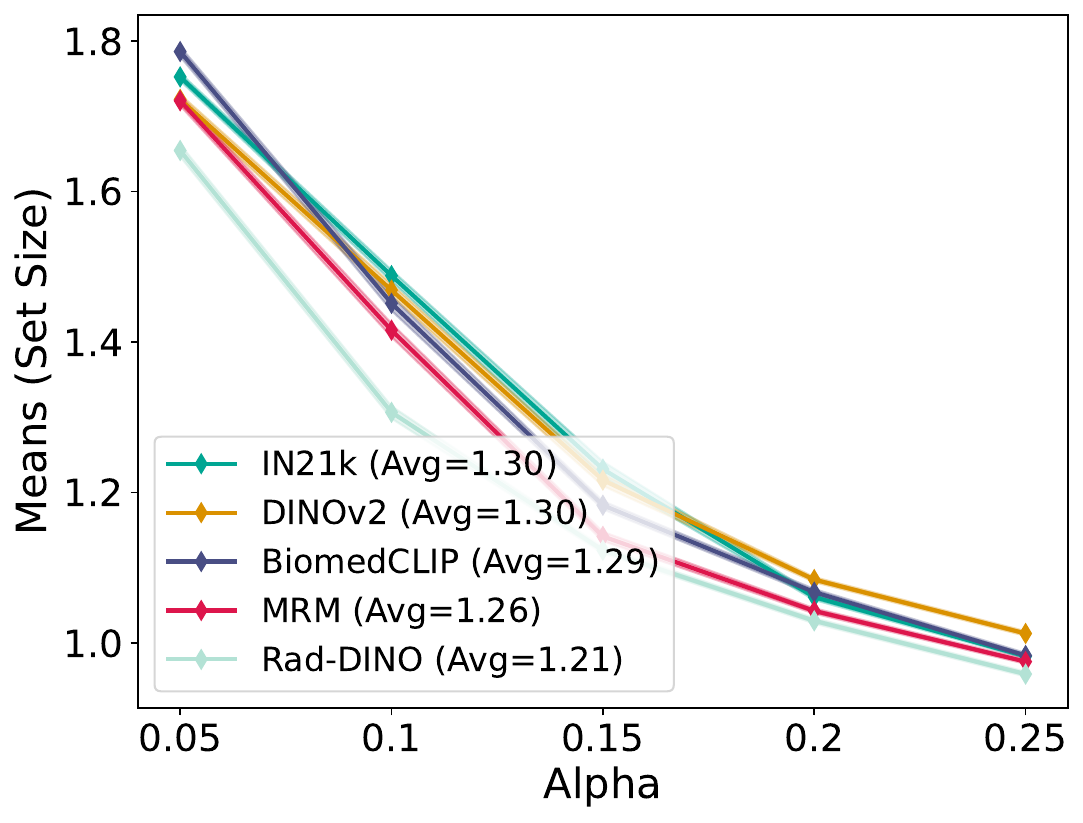}
    \includegraphics[height=0.19\textwidth, width=0.23\textwidth]{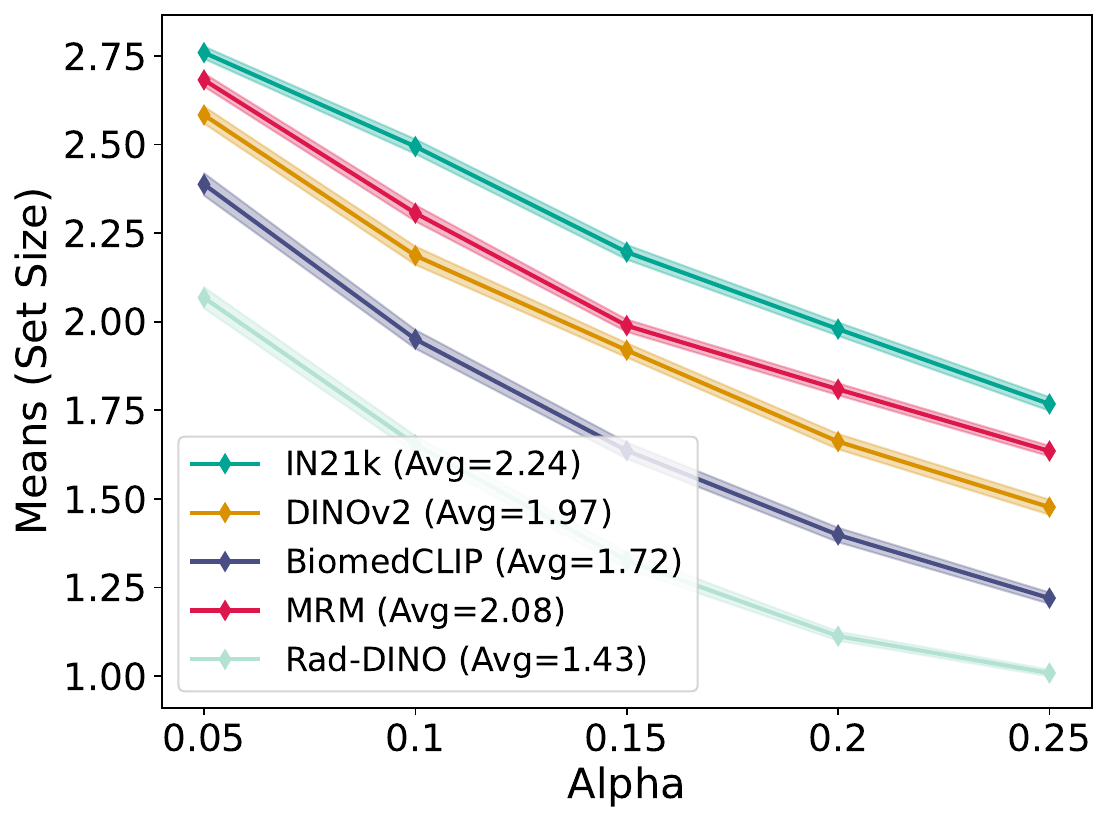}
    \includegraphics[height=0.19\textwidth, width=0.23\textwidth]{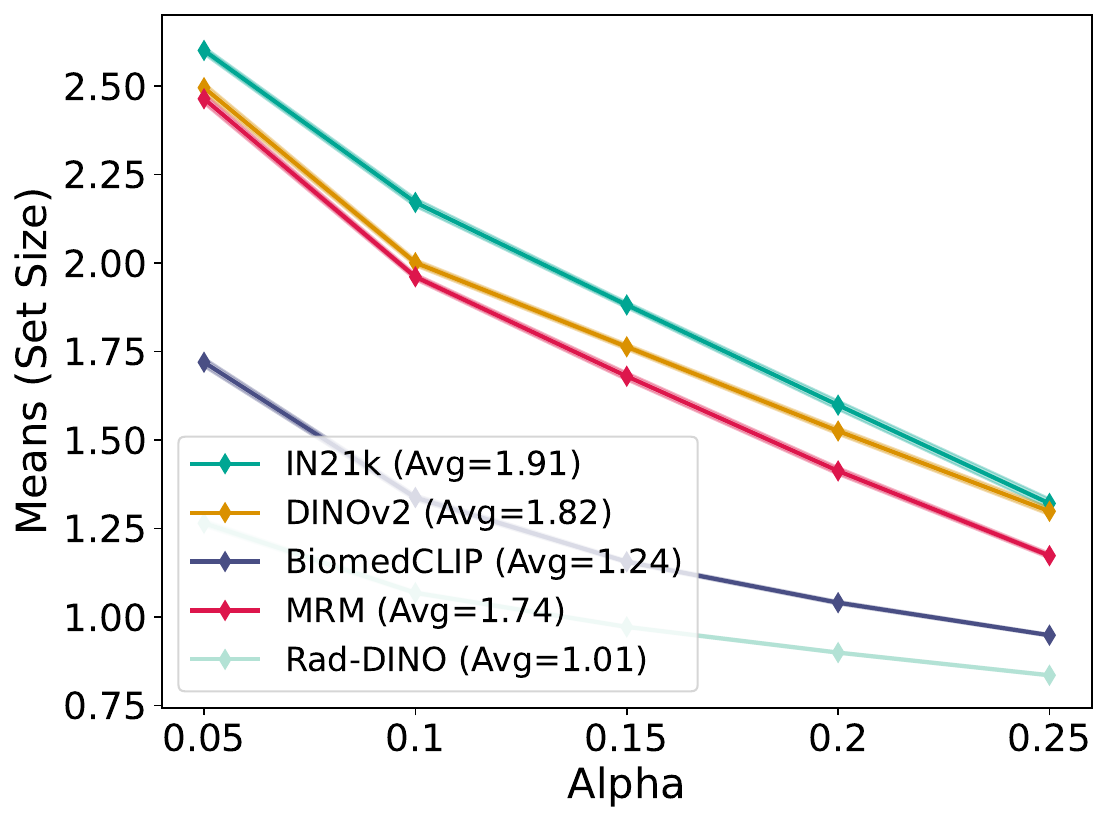}
    \caption{\textbf{Conformal prediction Set Size - RAPS (X-Rays)}}
    \label{fig:raps_cp_xrays_setsize}
\end{figure*}

We additionally present result for conformal prediction set size and coverage for RAPS \citep{angelopoulos2021uncertainty} in \Cref{fig:raps_cp_retina_setsize,fig:raps_cp_histopath_setsize,fig:raps_cp_xrays_setsize,fig:raps_cp_retina_coverage,fig:raps_cp_histopath_coverage,fig:raps_cp_xrays_coverage}, where the experimental result coincides with main conclusion from the result of LACS.

\newpage

\section{Least Ambiguous Set-valued Classifiers}
\label{apd:lac}
\begin{algorithm}
\caption{Conformal Prediction for Classification (LAC)}
\label{alg:1}
\begin{algorithmic}[1]
\REQUIRE model function $f(x)_y$ that generate class probability output, conformal score function $s(x,y)$, significance level $\alpha$, calibration examples $\{(x_1,y_1),...,(x_{n-1},y_{n-1})\}$, new example $x_n$
\ENSURE construct conformal prediction set $C_{\hat{q}}(x)=\{y:s(x,y)\leq\hat{q}\}$, where $\hat{q}$ is a conformal quantile threshold
\STATE Compute conformal scores $s_i=s(x_i,y_i)$ on calibration dataset $\{x_i,y_i\}_{i=1}^{n-1}$
\STATE Compute $q_{level}$ as $\frac{\lceil(n+1)(1-\alpha)\rceil}{n}$ 
\STATE Compute $\hat{q}$ as $q_{level}$ quantile of the calibration scores $s_1,...,s_n$
\STATE Compute conformal prediction set for the new example as $C_{\hat{q}}(x_n)=\{y:s(x_n,y)\leq\hat{q}\}$
\end{algorithmic}
\end{algorithm}

The algorithm first compute some conformal score (nonconformity measure) $s(x,y)$ to quantify the difference between $x$ and $y$ in some separate calibration set $\{x_i,x_y\}_{i=1}^{n-1}$ to derive the threshold $\hat{q}$. Then, the confidence set is constructed in the way that including all classes $y$ whose scores $s(x_n,y)$ are not larger than $\hat{q}$ for a new sample $x_n$. This ensures that the probability of the true label not being in the set is controlled by $\alpha$. Thus, this algorithm essentially matches the conformal prediction framework.

\section{Empirical Coverage}
\begin{figure*}[!ht]
    \centering
    \makebox[\textwidth][l]{%
        \hspace{0.23\textwidth}
        \textbf{Retina} \hspace{0.14\textwidth} \textbf{IDRiD} \hspace{0.14\textwidth} \textbf{APTOS2019}
    } \\[0.2cm]
    \includegraphics[height=0.19\textwidth, width=0.23\textwidth]{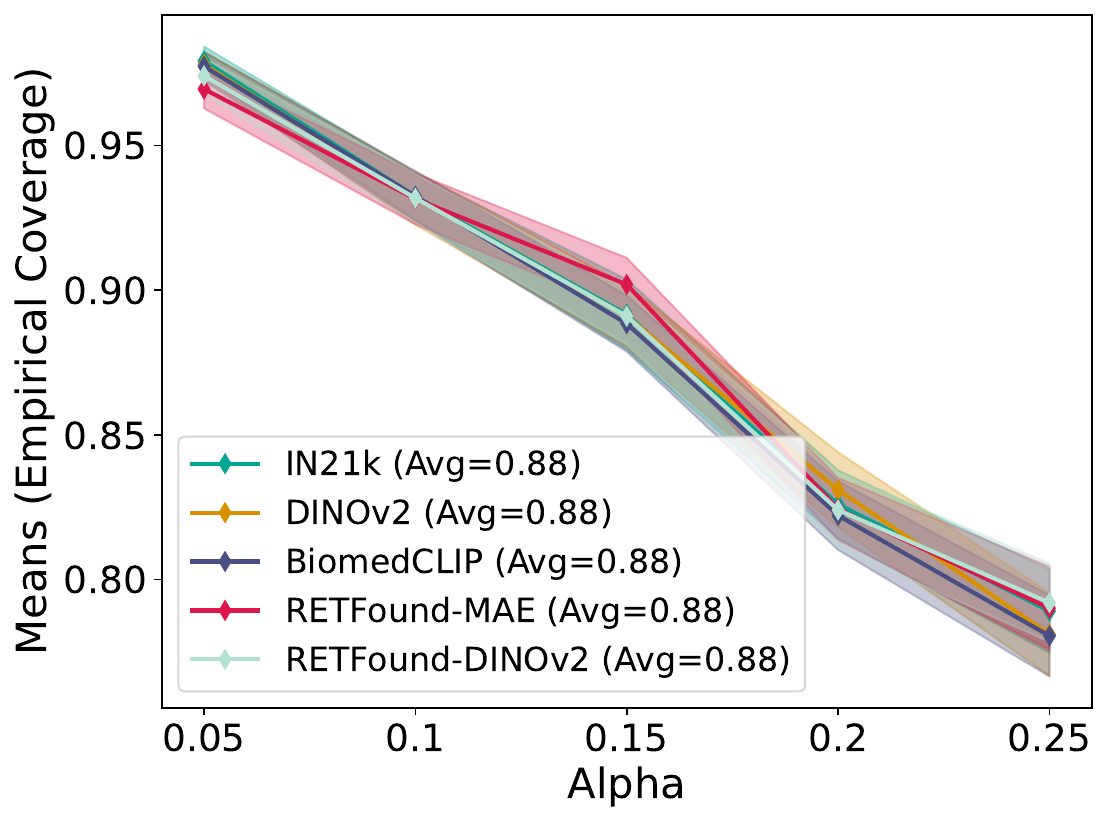} 
    \includegraphics[height=0.19\textwidth, width=0.23\textwidth]{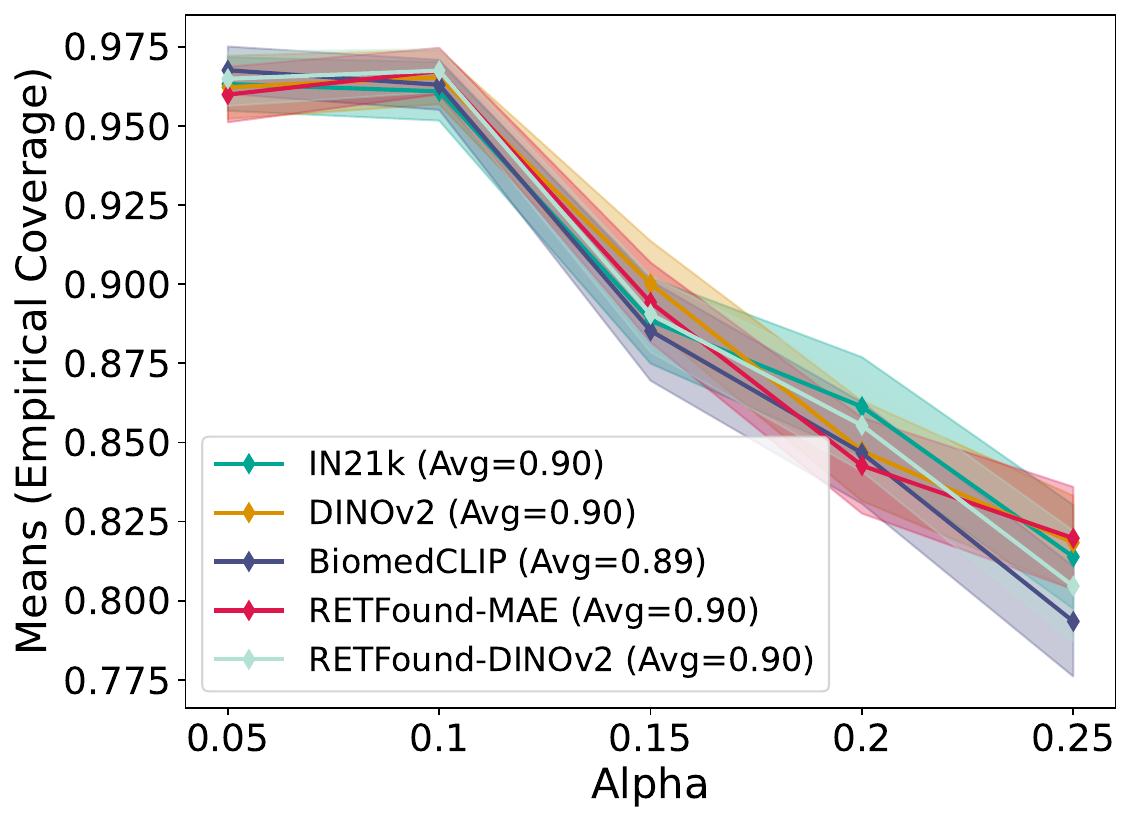}
    \includegraphics[height=0.19\textwidth, width=0.23\textwidth]{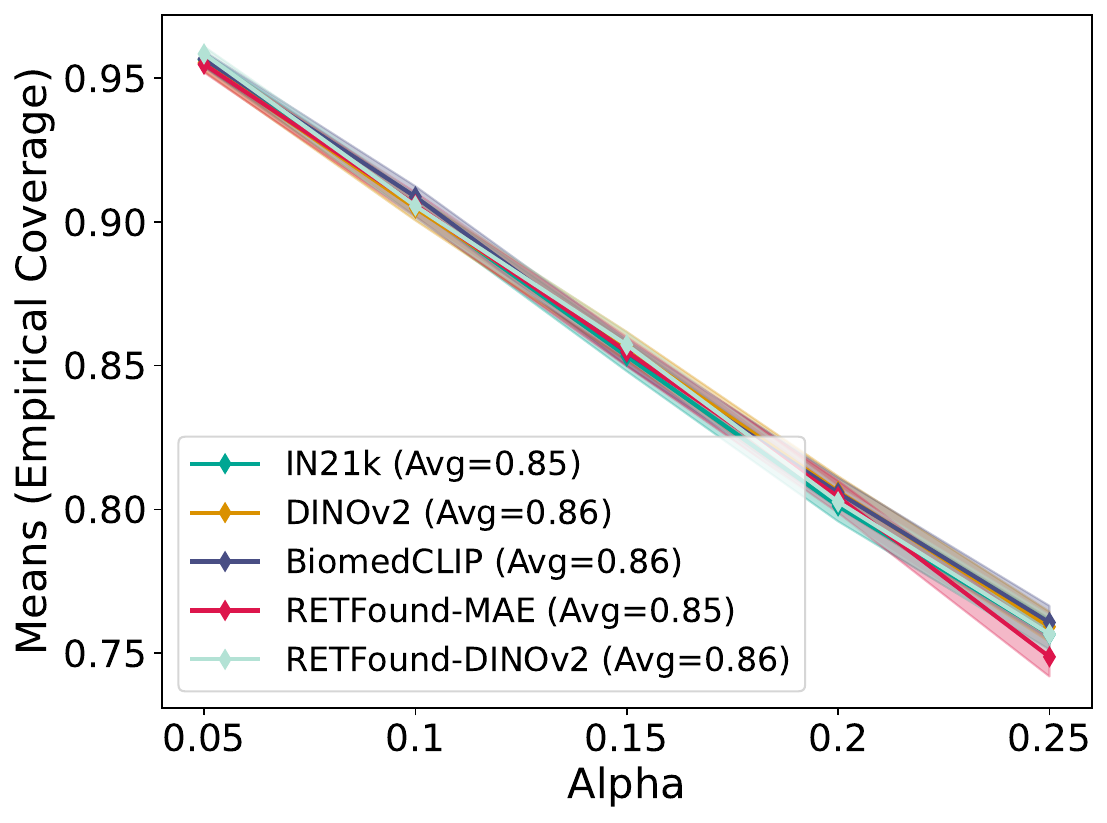}\\[0.2cm]
    \makebox[\textwidth][l]{%
        \hspace{0.21\textwidth}
        \textbf{Retina (T)} \hspace{0.11\textwidth} \textbf{IDRiD (T)} \hspace{0.10\textwidth} \textbf{APTOS2019 (T)}
    } \\[0.2cm]
    \includegraphics[height=0.19\textwidth, width=0.23\textwidth]{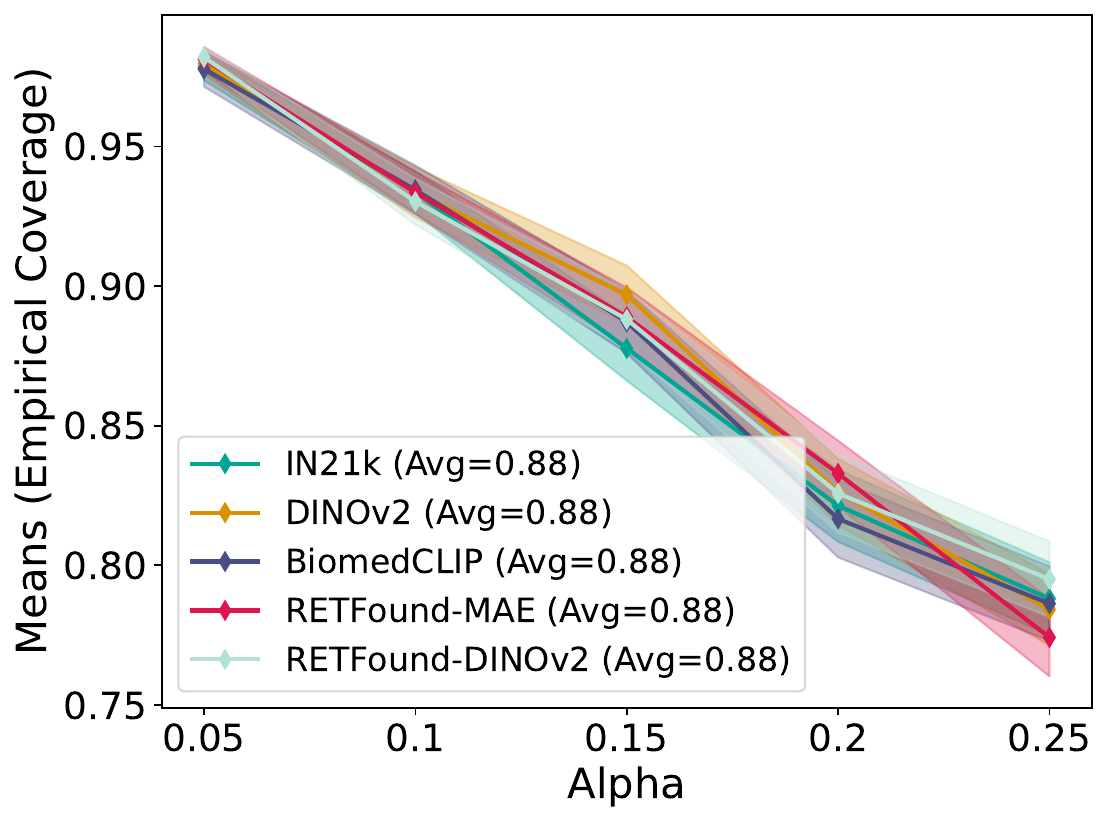} 
    \includegraphics[height=0.19\textwidth, width=0.23\textwidth]{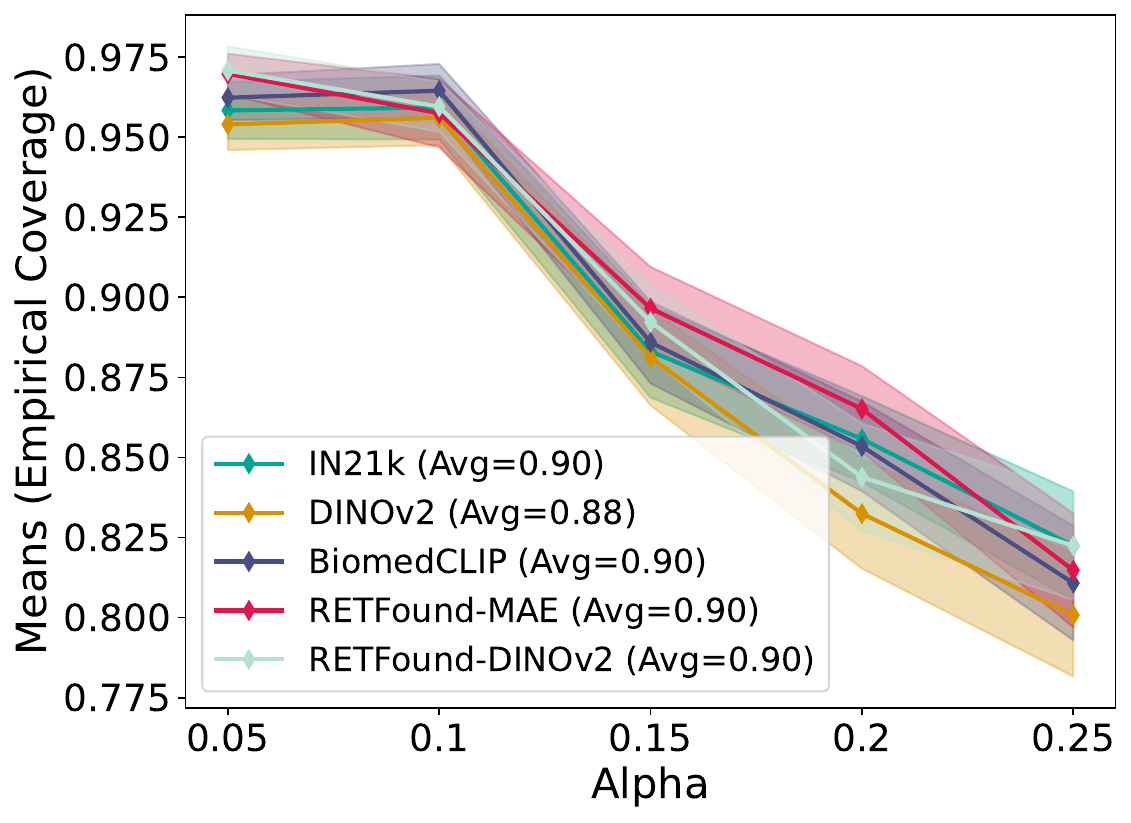}
    \includegraphics[height=0.19\textwidth, width=0.23\textwidth]{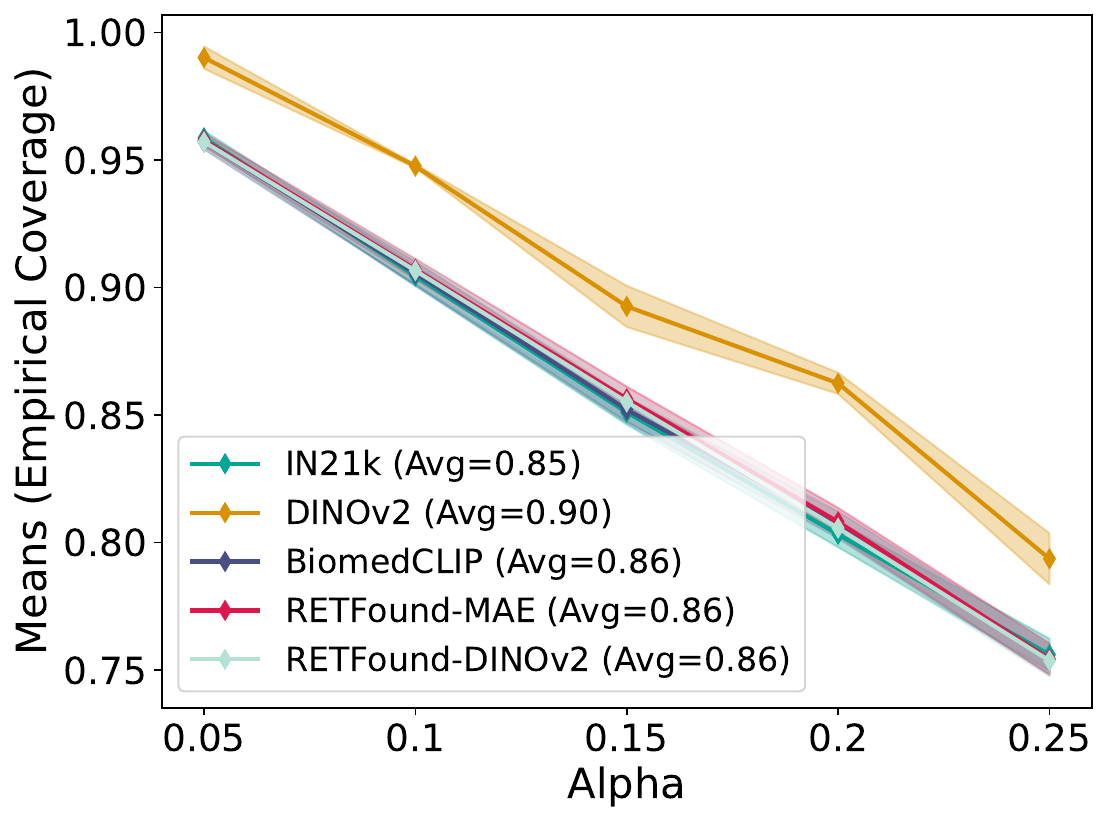}\\[0.2cm]
    \makebox[\textwidth][l]{%
        \hspace{0.20\textwidth}
        \textbf{Retina (LS)} \hspace{0.10\textwidth} \textbf{IDRiD (LS)} \hspace{0.08\textwidth} \textbf{APTOS2019 (LS)}
    } \\[0.2cm]
    \includegraphics[height=0.19\textwidth, width=0.23\textwidth]{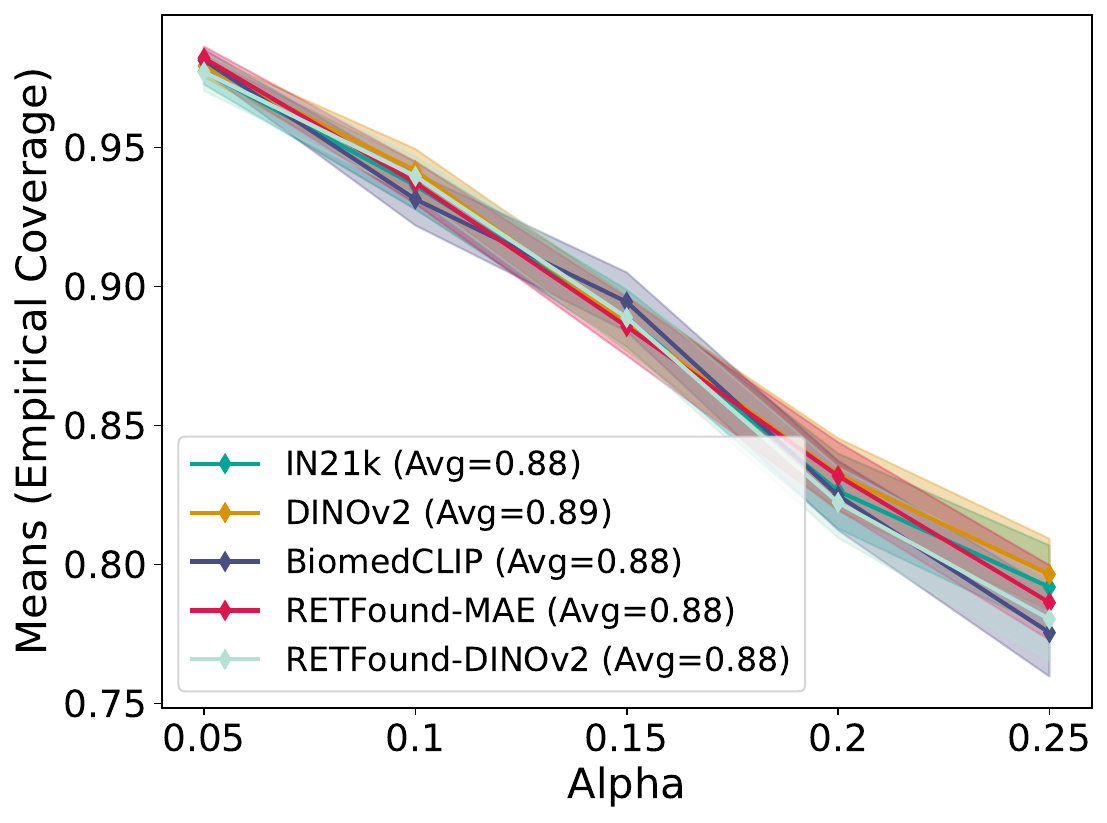} 
    \includegraphics[height=0.19\textwidth, width=0.23\textwidth]{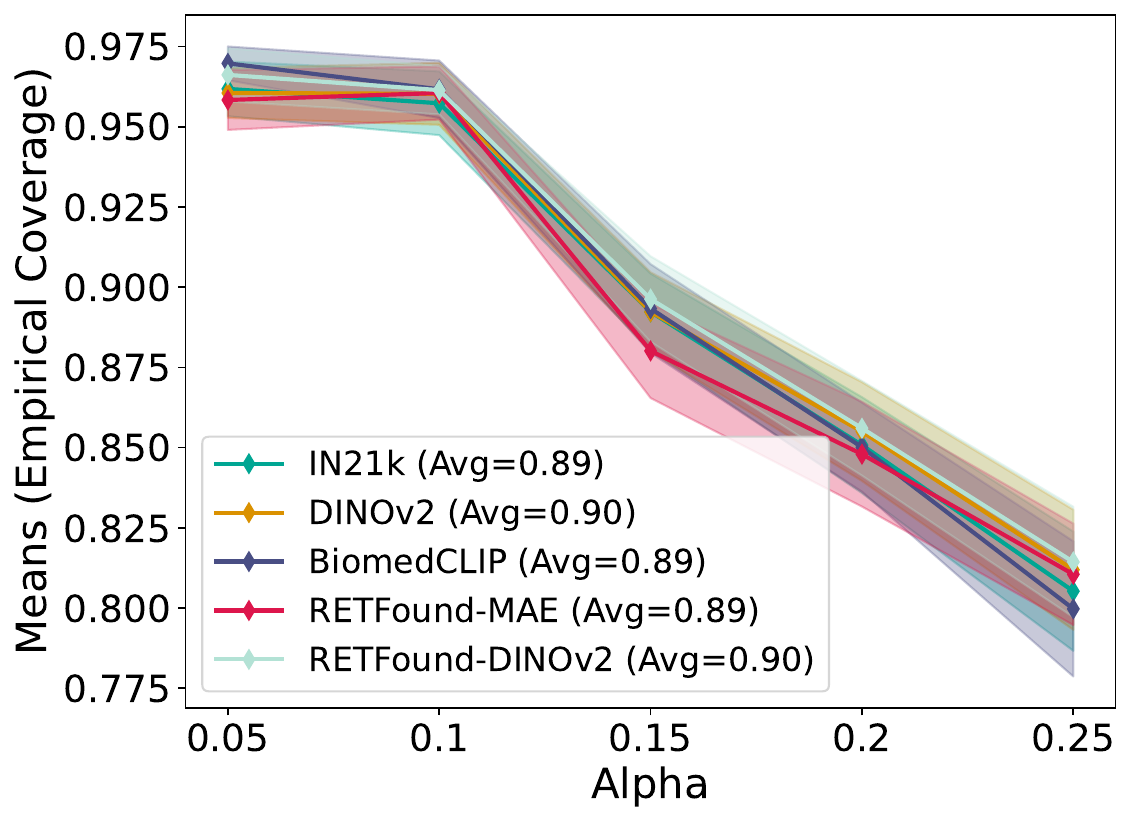}
    \includegraphics[height=0.19\textwidth, width=0.23\textwidth]{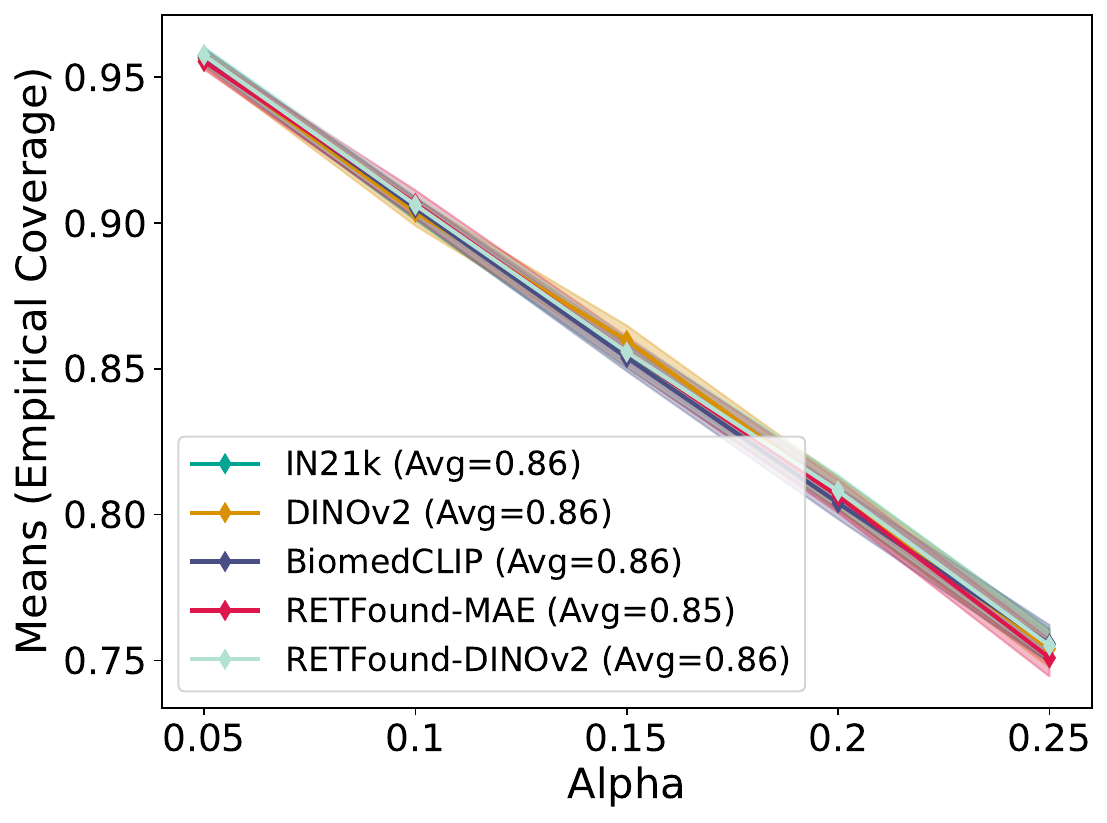}
    \caption{\textbf{Conformal prediction Empirical Coverage (Retina)}}
    \label{fig:cp_retina_coverage}
\end{figure*}

\begin{figure*}[!ht]
    \centering
    \makebox[\textwidth][l]{%
        \hspace{0.21\textwidth}
        \textbf{CRC100K} \hspace{0.13\textwidth} \textbf{TCGA} \hspace{0.15\textwidth} \textbf{BraTS}
    } \\[0.2cm]
    \includegraphics[height=0.19\textwidth, width=0.23\textwidth]{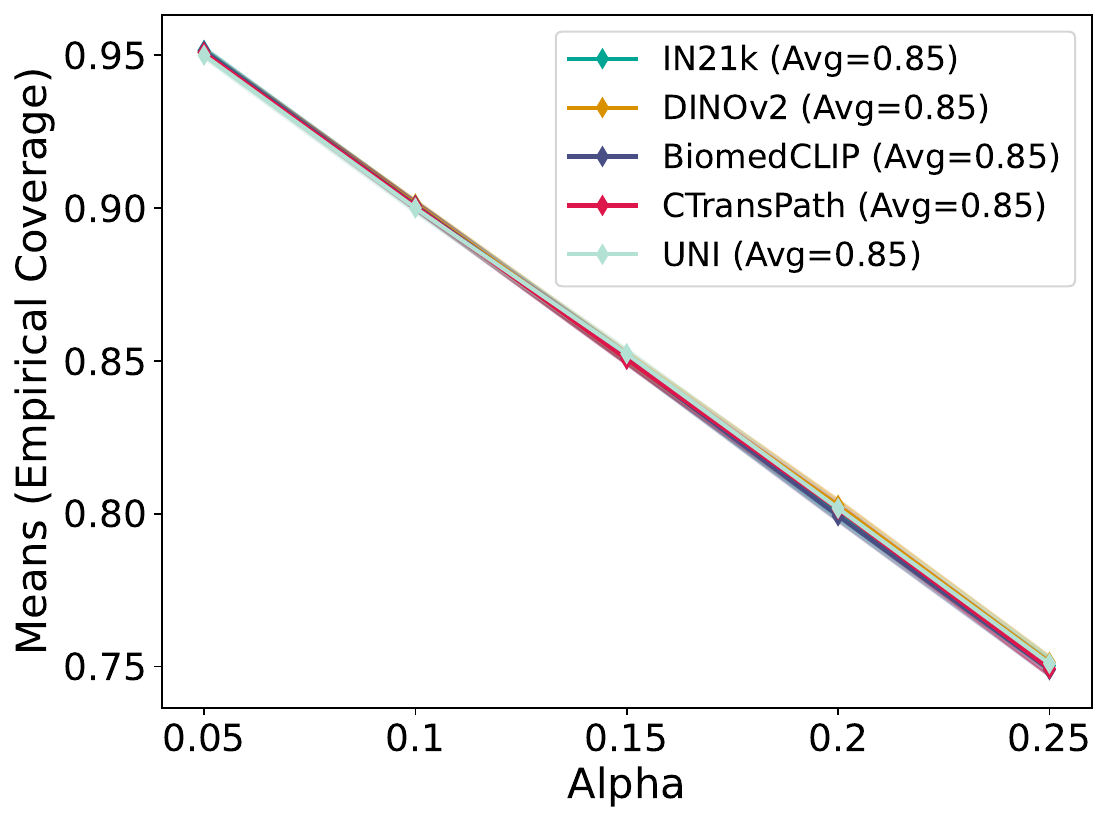} 
    \includegraphics[height=0.19\textwidth, width=0.23\textwidth]{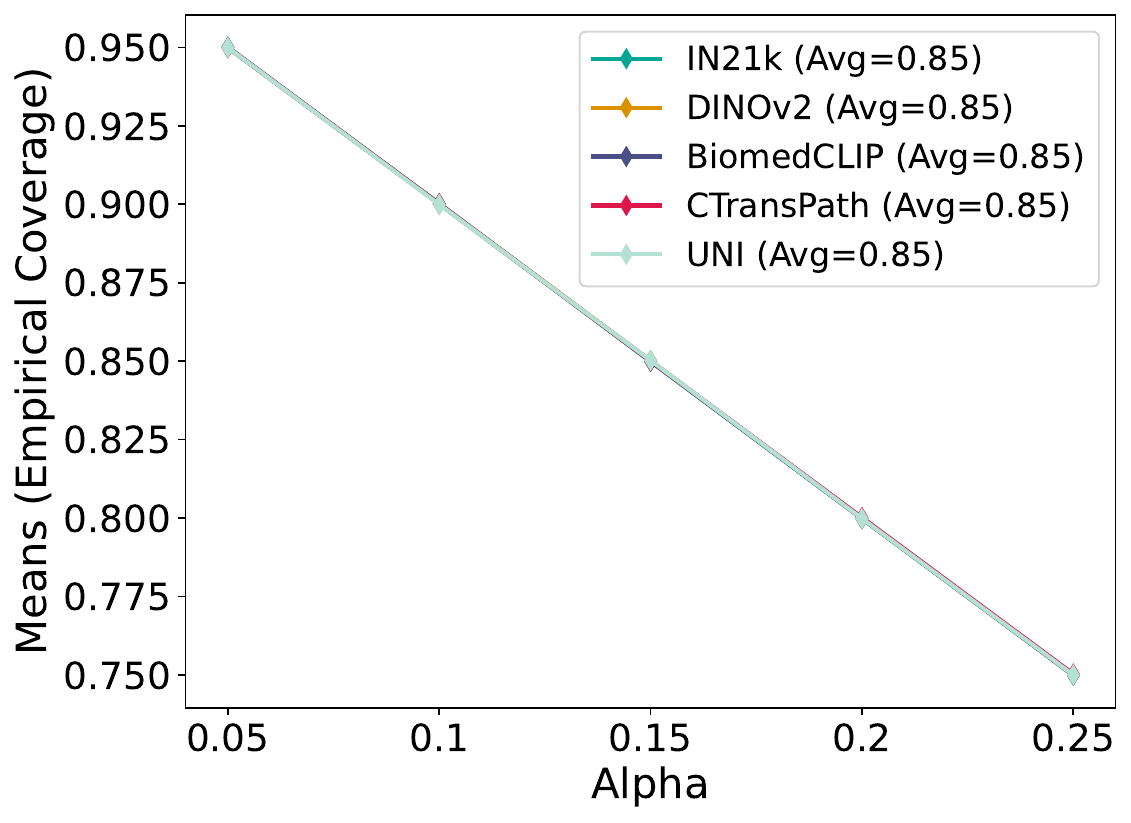}
    \includegraphics[height=0.19\textwidth, width=0.23\textwidth]{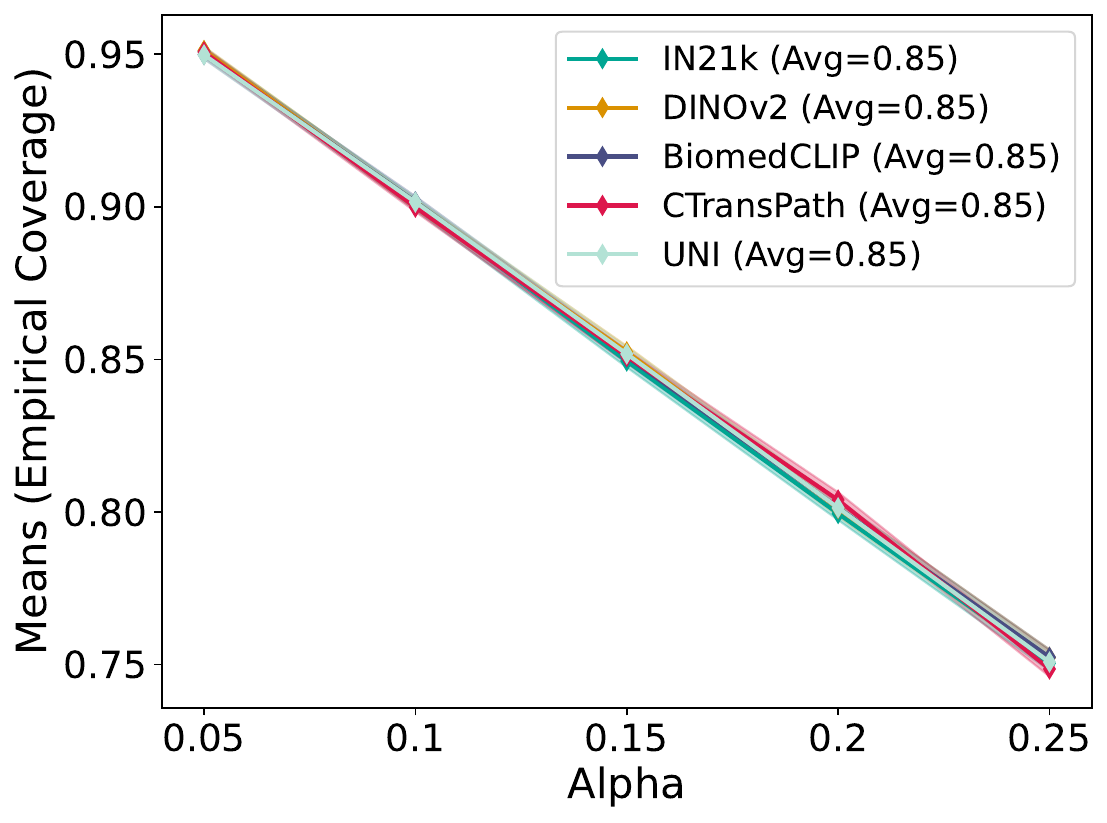}\\[0.2cm]
    \makebox[\textwidth][l]{%
        \hspace{0.20\textwidth}
        \textbf{CRC100K (T)} \hspace{0.08\textwidth} \textbf{TCGA (T)} \hspace{0.11\textwidth} \textbf{BraTS (T)}
    } \\[0.2cm]
    \includegraphics[height=0.19\textwidth, width=0.23\textwidth]{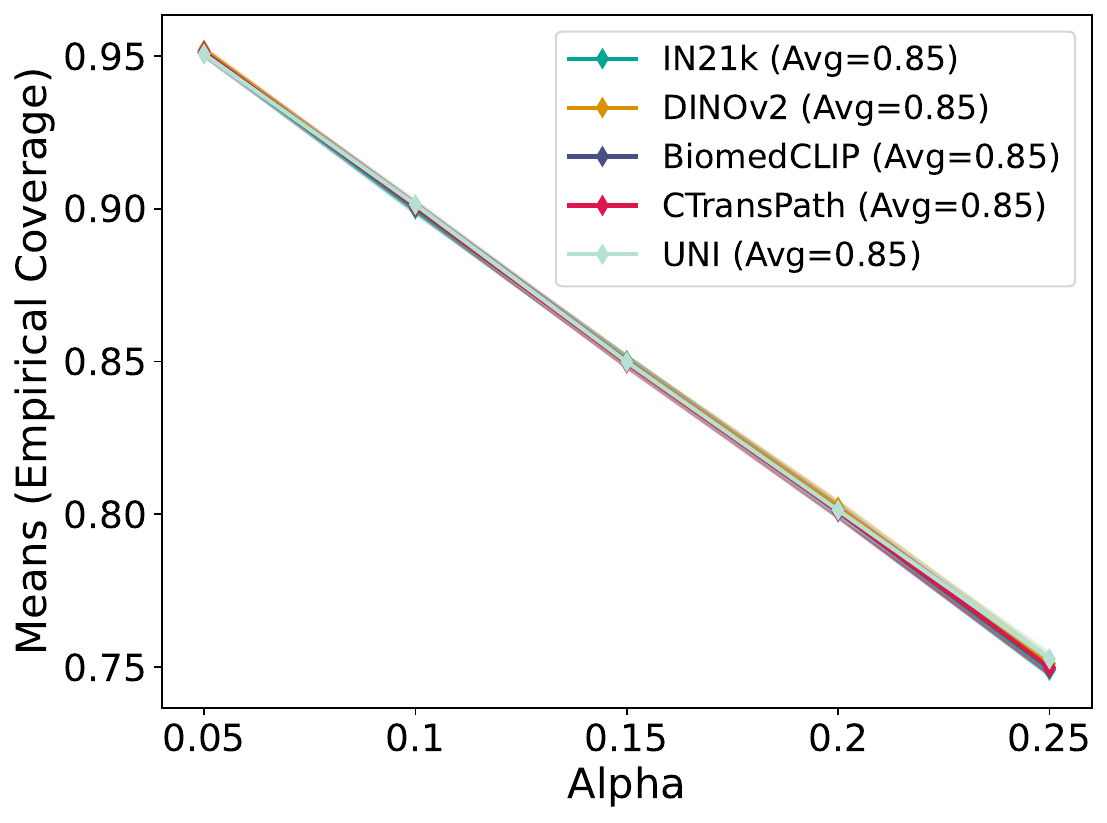} 
    \includegraphics[height=0.19\textwidth, width=0.23\textwidth]{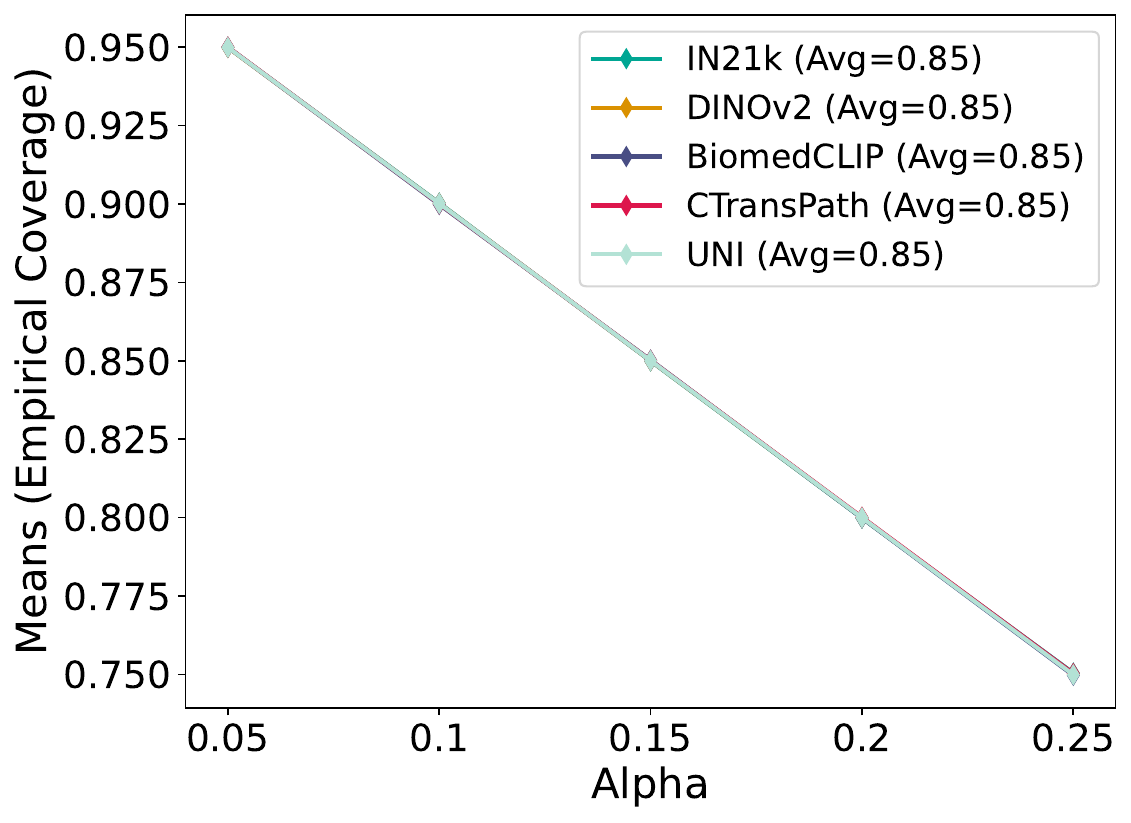}
    \includegraphics[height=0.19\textwidth, width=0.23\textwidth]{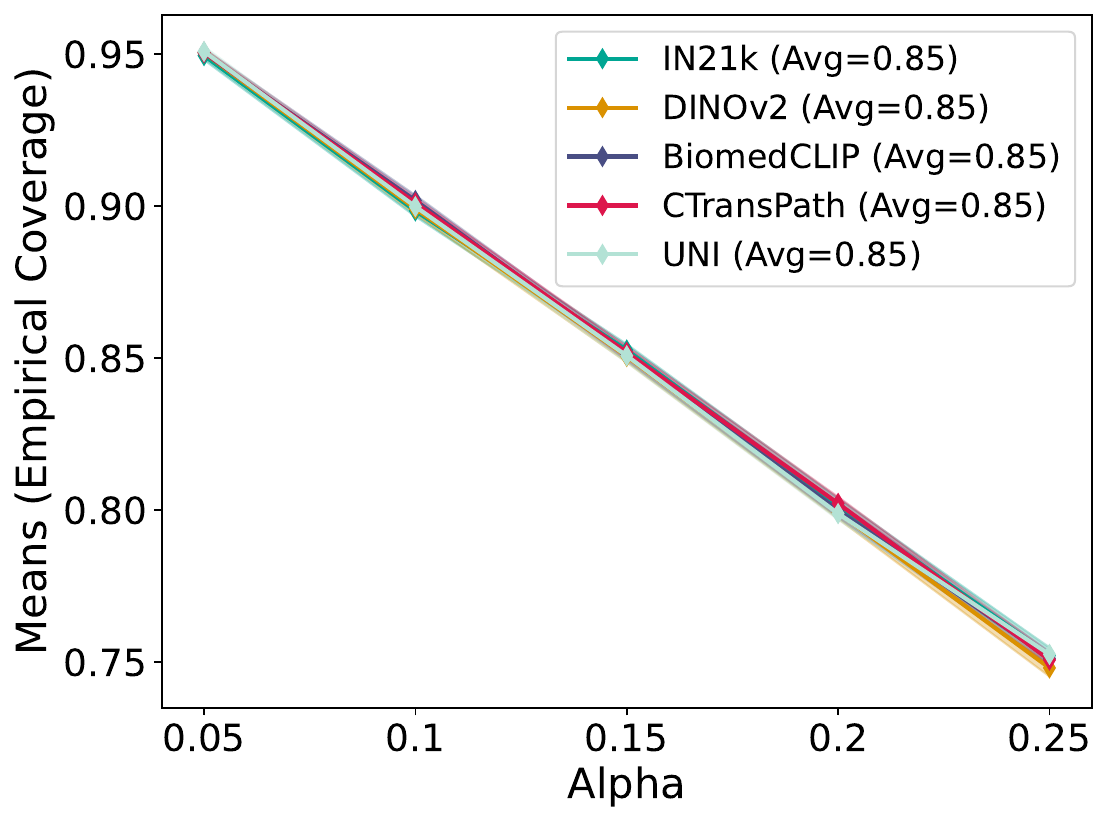}
    \makebox[\textwidth][l]{%
        \hspace{0.19\textwidth}
        \textbf{CRC100K (LS)} \hspace{0.07\textwidth} \textbf{TCGA (LS)} \hspace{0.09\textwidth} \textbf{BraTS (LS)}
    } \\[0.2cm]
    \includegraphics[height=0.19\textwidth, width=0.23\textwidth]{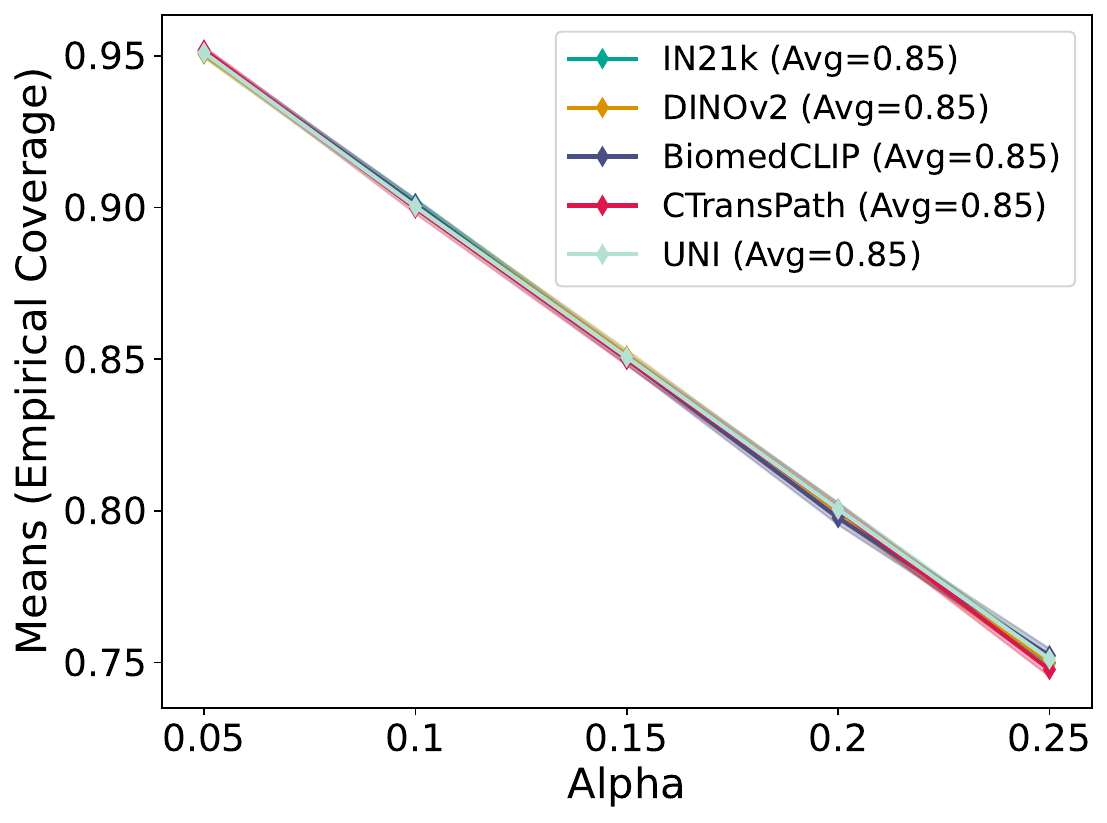} 
    \includegraphics[height=0.19\textwidth, width=0.23\textwidth]{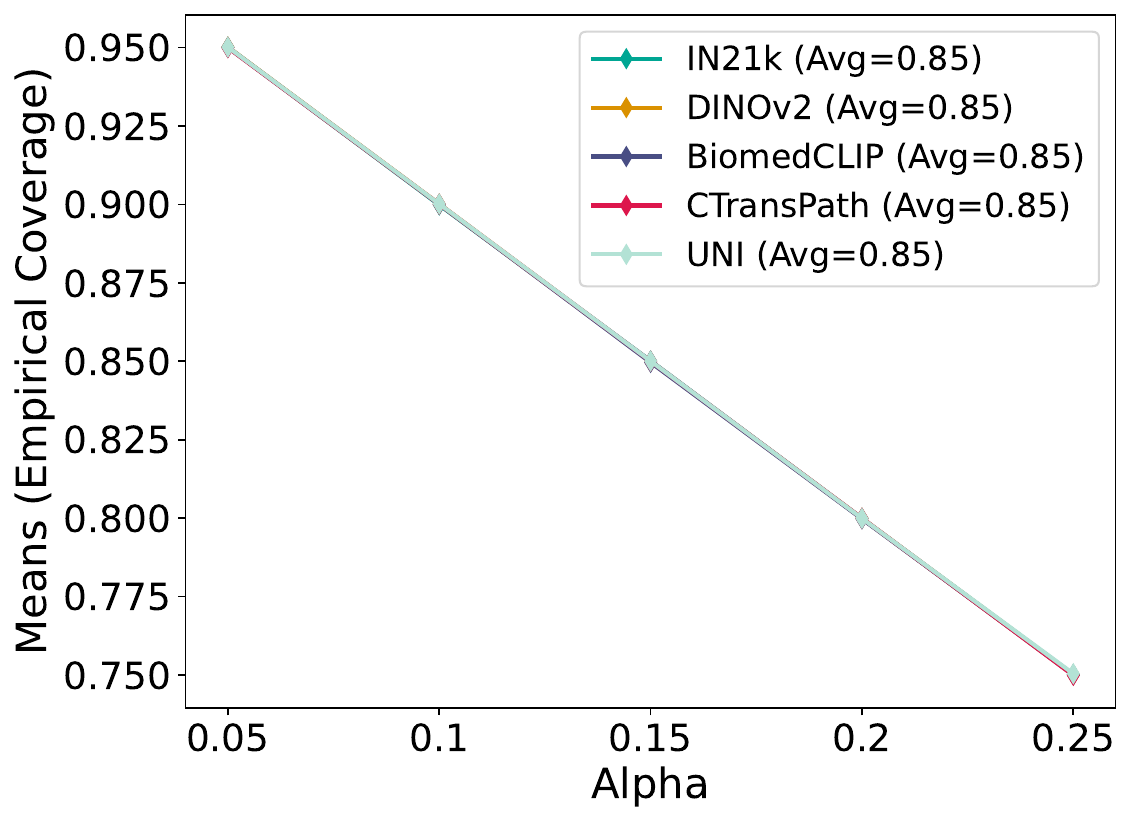}
    \includegraphics[height=0.19\textwidth, width=0.23\textwidth]{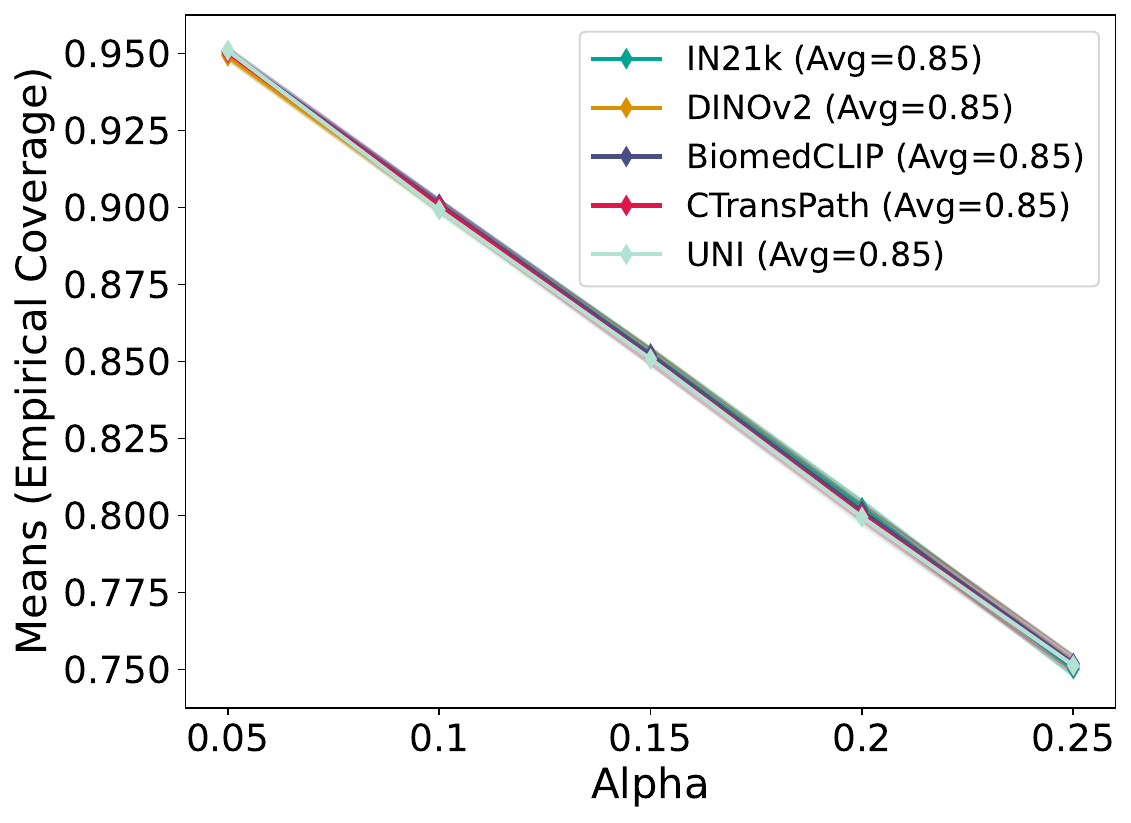}
    \caption{\textbf{Conformal prediction Empirical Coverage (Histopathology)}}
    \label{fig:cp_histopath_coverage}
\end{figure*}

\begin{figure*}[!ht]
    \centering
    \makebox[\textwidth][l]{%
        \hspace{0.24\textwidth}
        \textbf{RSNA} \hspace{0.13\textwidth} \textbf{POLCOVID} \hspace{0.090\textwidth} \textbf{COVID-Rad}
    } \\[0.2cm]
    \includegraphics[height=0.19\textwidth, width=0.23\textwidth]{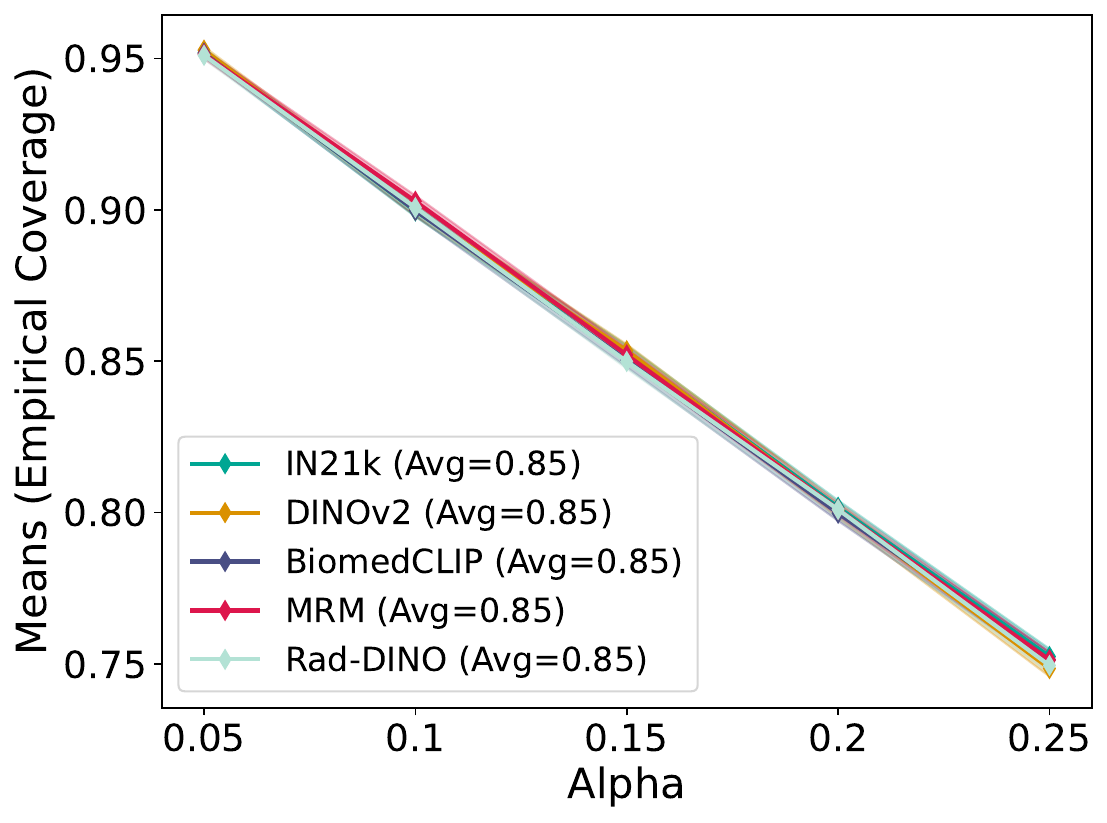} 
    \includegraphics[height=0.19\textwidth, width=0.23\textwidth]{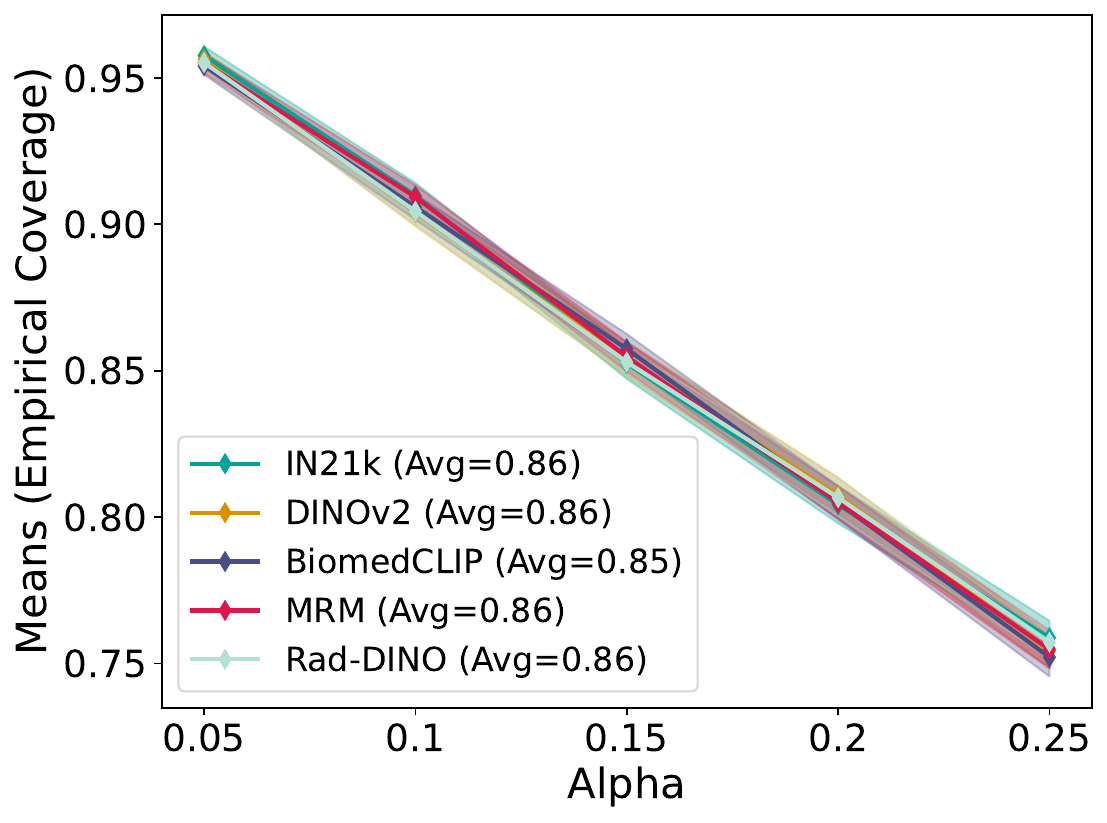}
    \includegraphics[height=0.19\textwidth, width=0.23\textwidth]{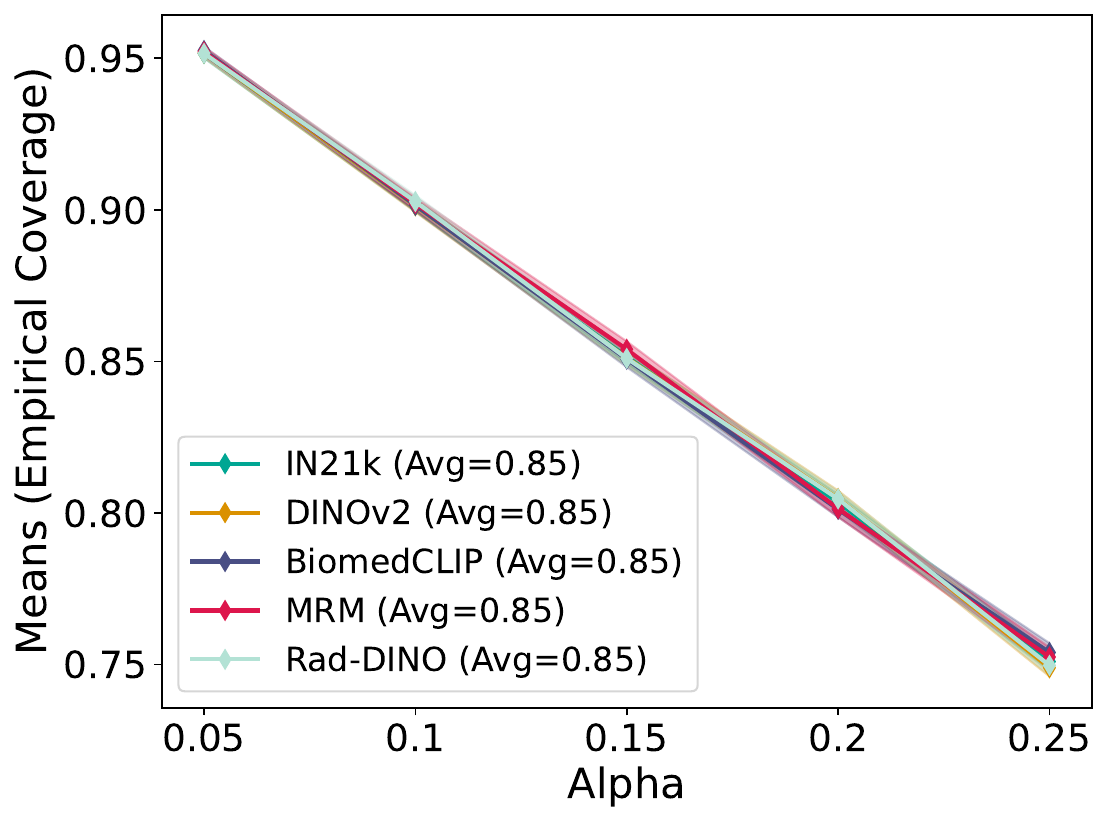}
    \makebox[\textwidth][l]{%
        \hspace{0.22\textwidth}
        \textbf{RSNA (T)} \hspace{0.09\textwidth} \textbf{POLCOVID (T)} \hspace{0.04\textwidth} \textbf{COVID-Rad (T)}
    } \\[0.2cm]
    \includegraphics[height=0.19\textwidth, width=0.23\textwidth]{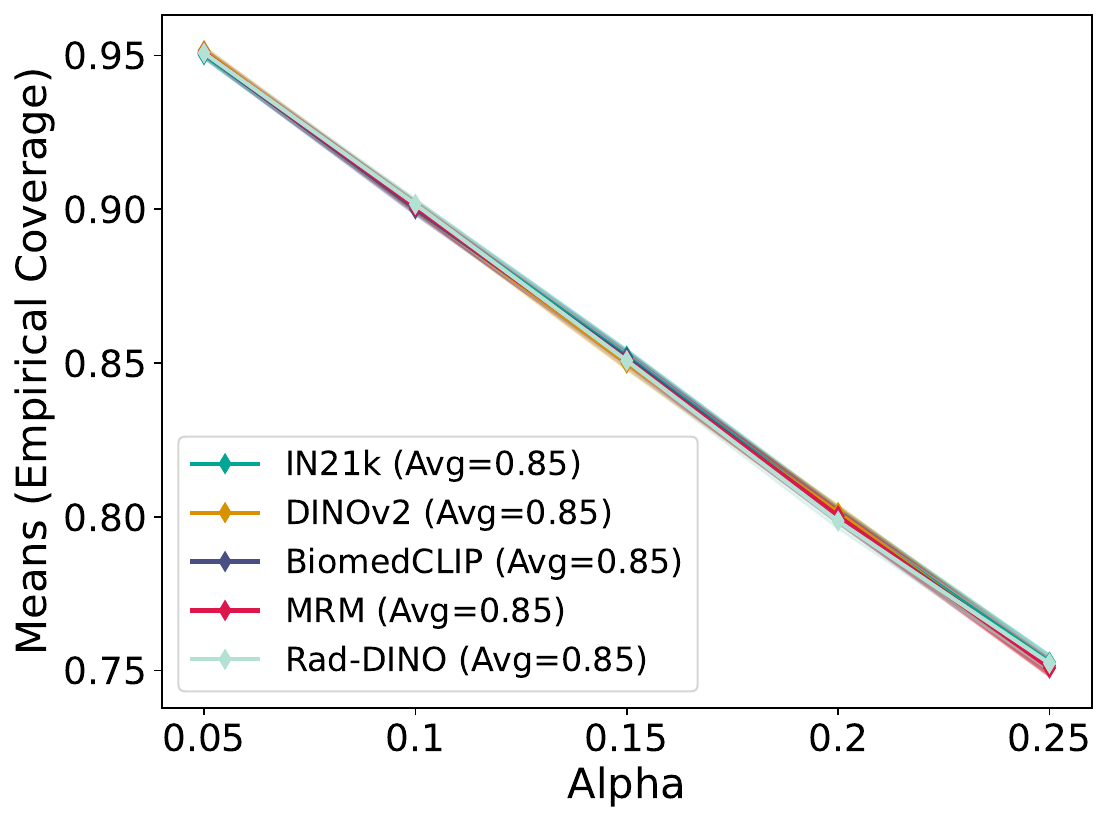} 
    \includegraphics[height=0.19\textwidth, width=0.23\textwidth]{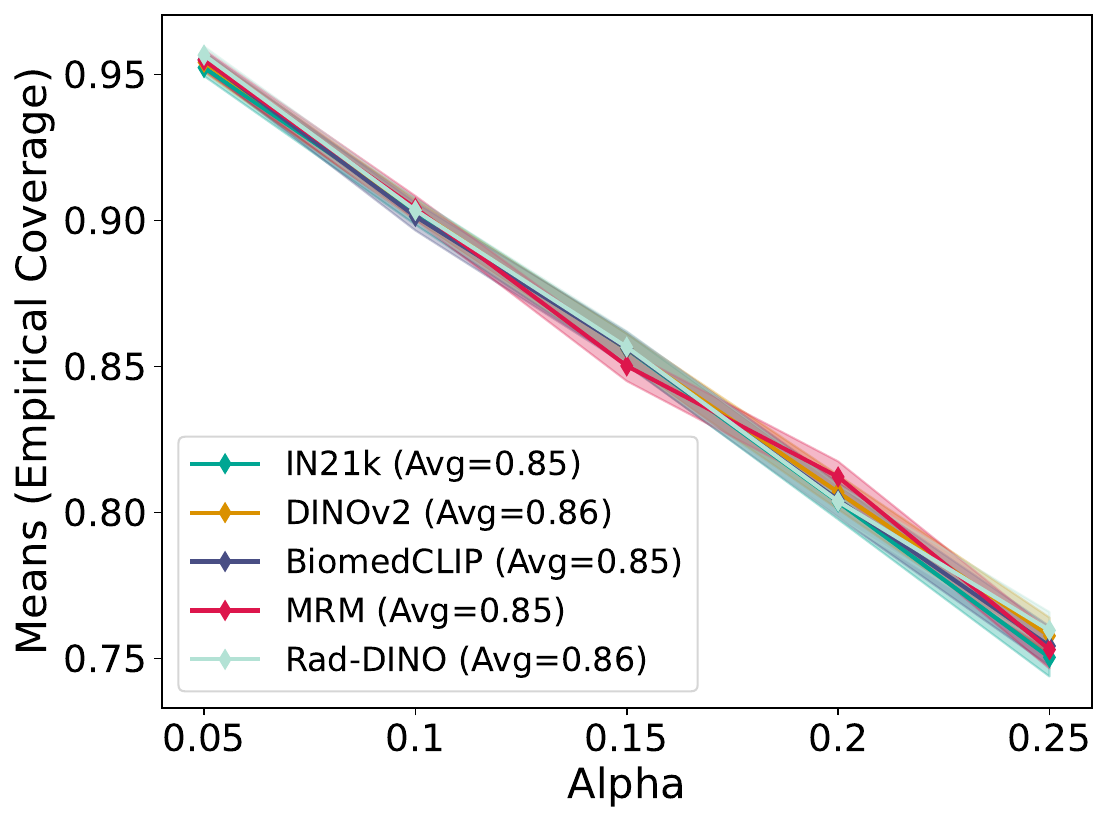}
    \includegraphics[height=0.19\textwidth, width=0.23\textwidth]{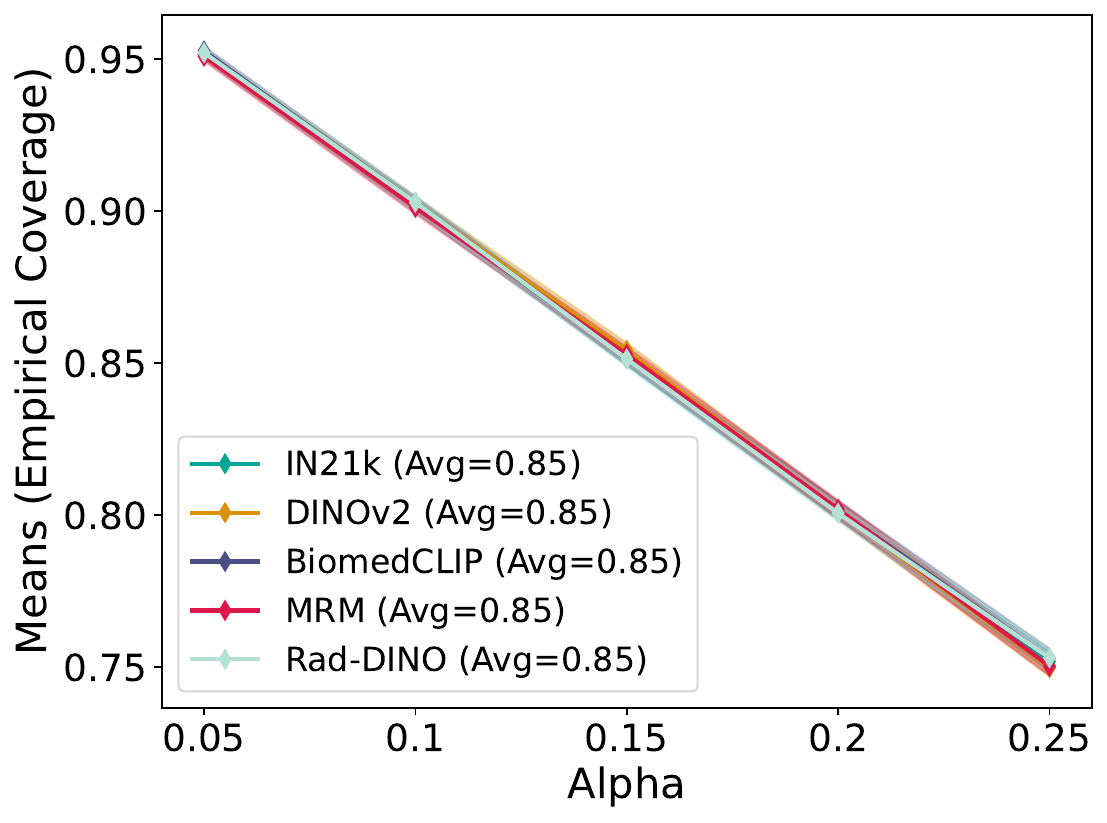}
    \makebox[\textwidth][l]{%
        \hspace{0.21\textwidth}
        \textbf{RSNA (LS)} \hspace{0.08\textwidth} \textbf{POLCOVID (LS)} \hspace{0.03\textwidth} \textbf{COVID-Rad (LS)}
    } \\[0.2cm]
    \includegraphics[height=0.19\textwidth, width=0.23\textwidth]{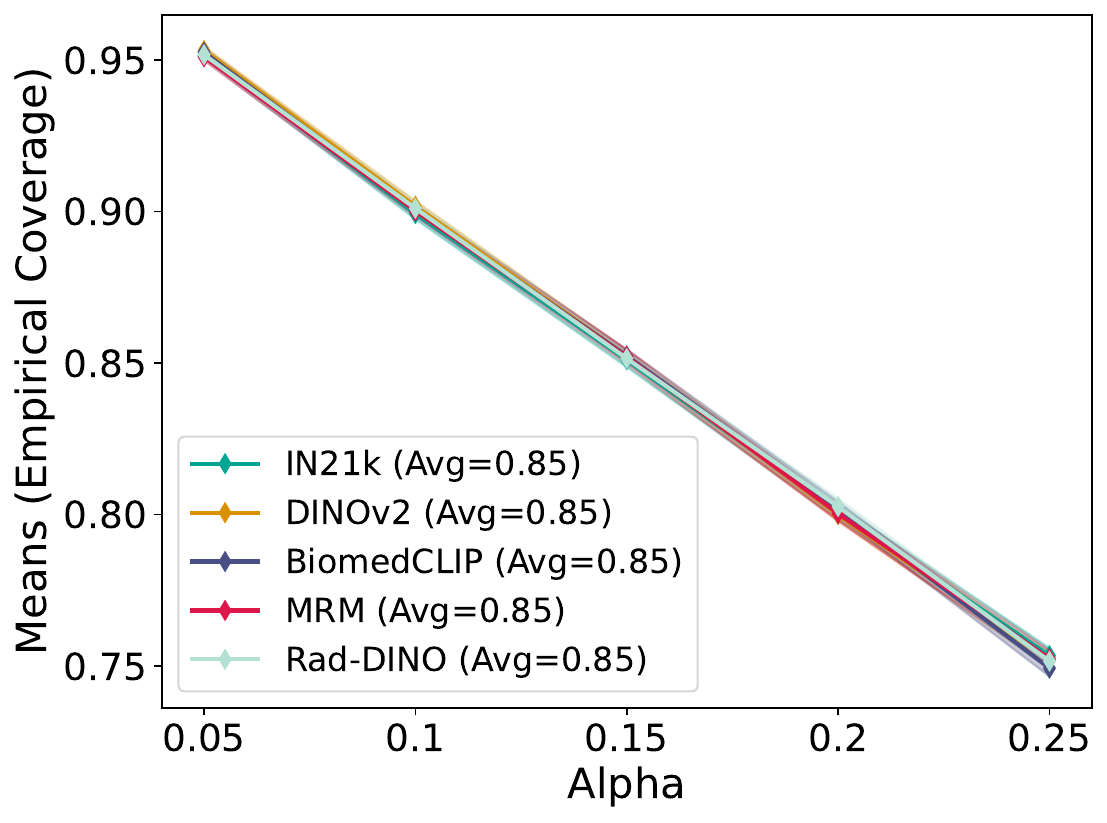} 
    \includegraphics[height=0.19\textwidth, width=0.23\textwidth]{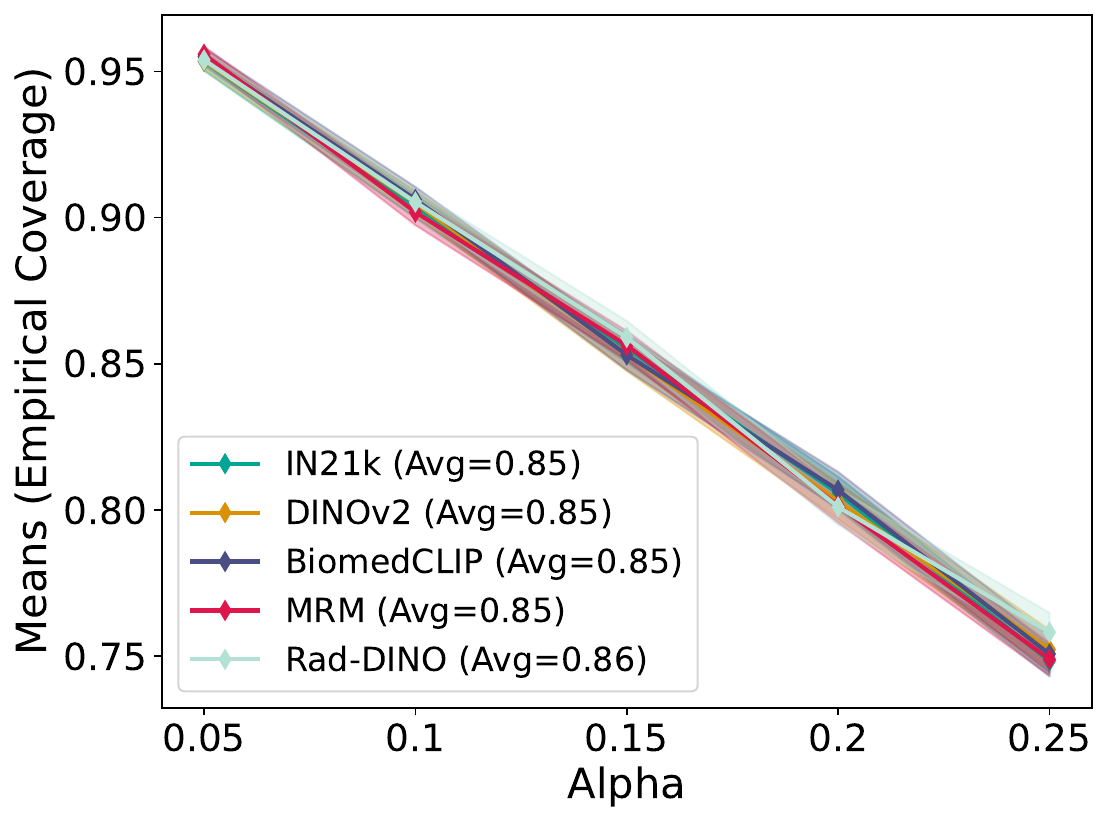}
    \includegraphics[height=0.19\textwidth, width=0.23\textwidth]{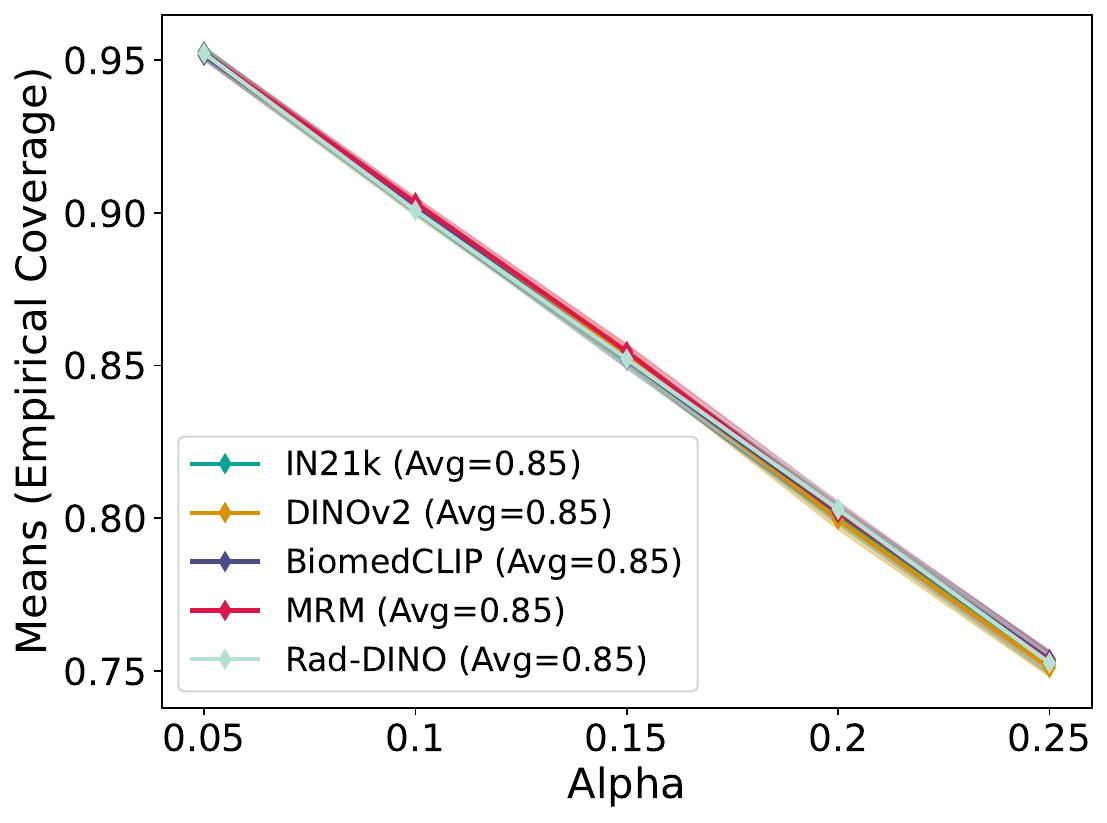}
    \caption{\textbf{Conformal prediction Empirical Coverage (X-Rays)}}
    \label{fig:cp_xray_coverage}
\end{figure*}

\begin{figure*}[!ht]
    \centering
    \makebox[\textwidth][l]{%
        \hspace{0.23\textwidth}
        \textbf{Retina} \hspace{0.14\textwidth} \textbf{IDRiD} \hspace{0.14\textwidth} \textbf{APTOS2019}
    } \\[0.2cm]
    \includegraphics[height=0.19\textwidth, width=0.23\textwidth]{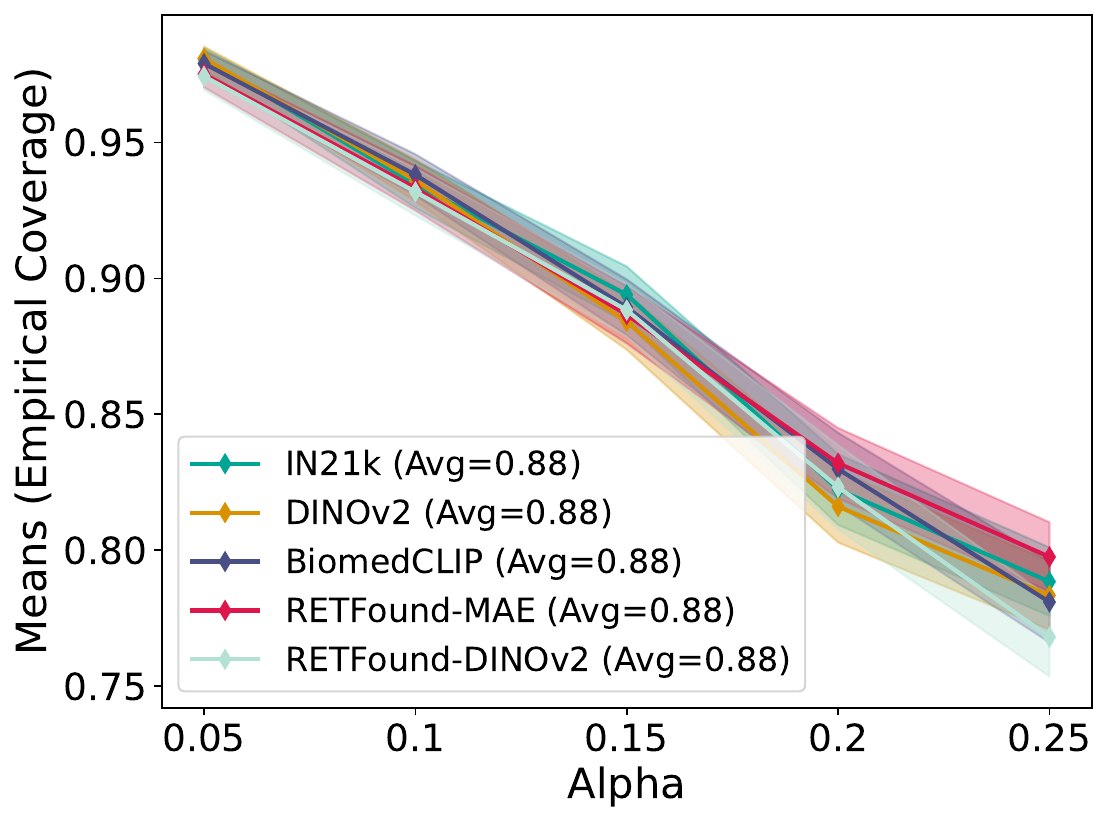} 
    \includegraphics[height=0.19\textwidth, width=0.23\textwidth]{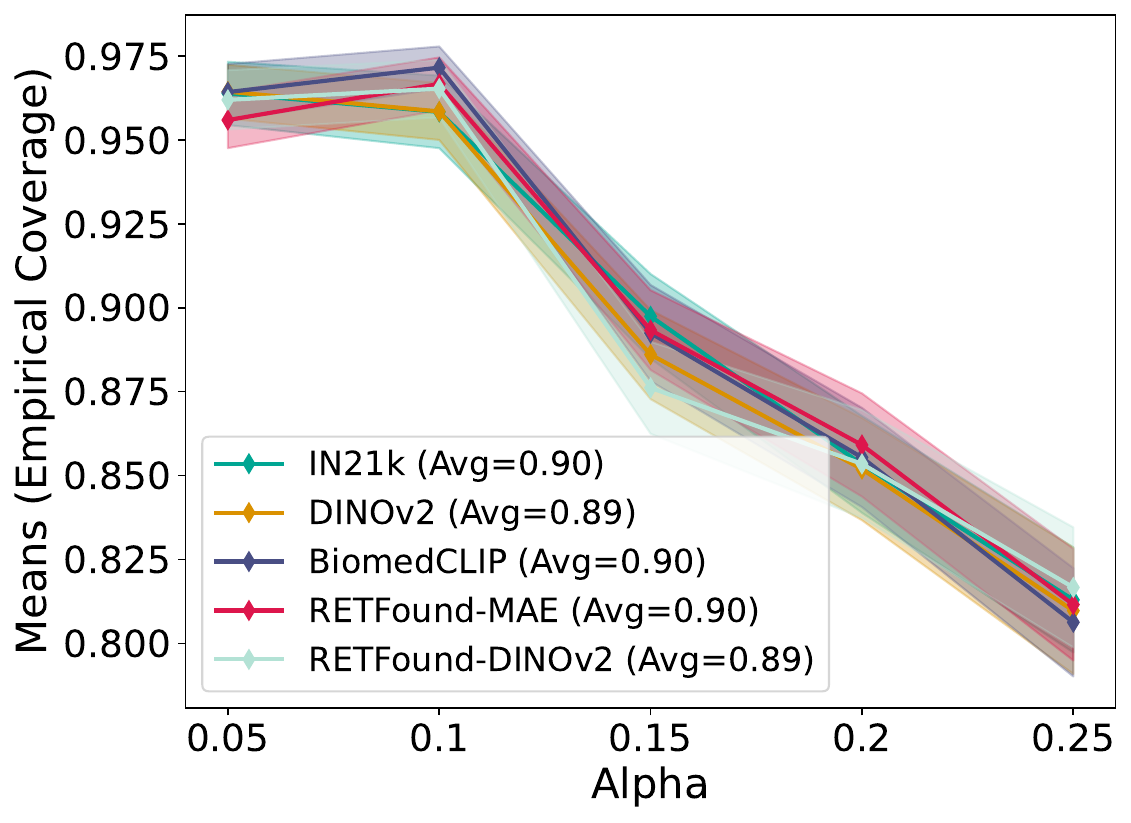}
    \includegraphics[height=0.19\textwidth, width=0.23\textwidth]{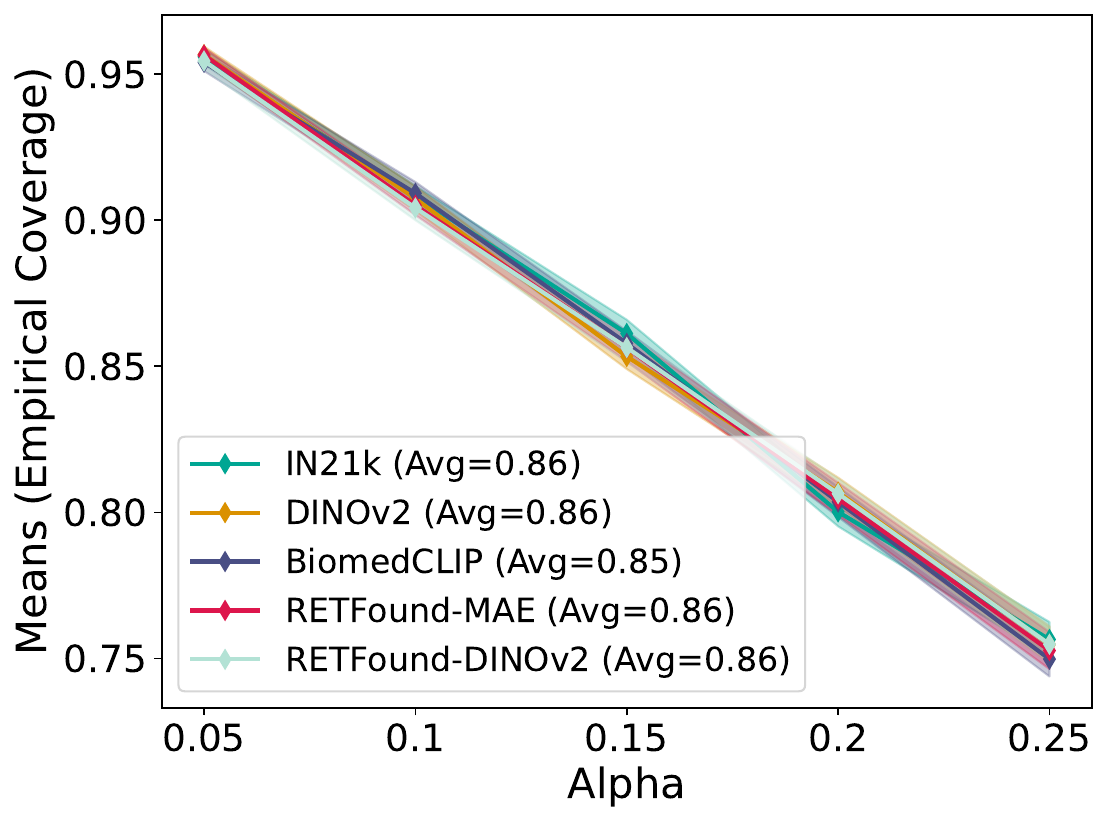}\\[0.2cm]
    \makebox[\textwidth][l]{%
        \hspace{0.21\textwidth}
        \textbf{Retina (T)} \hspace{0.11\textwidth} \textbf{IDRiD (T)} \hspace{0.10\textwidth} \textbf{APTOS2019 (T)}
    } \\[0.2cm]
    \includegraphics[height=0.19\textwidth, width=0.23\textwidth]{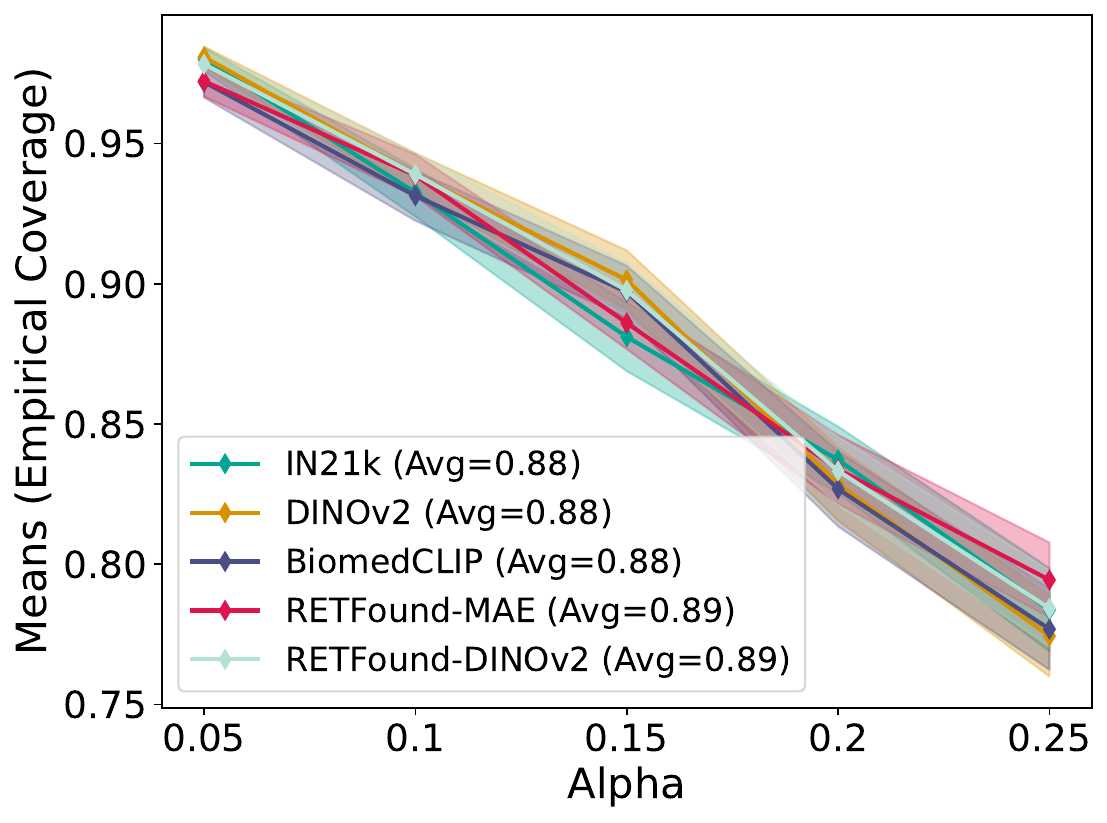} 
    \includegraphics[height=0.19\textwidth, width=0.23\textwidth]{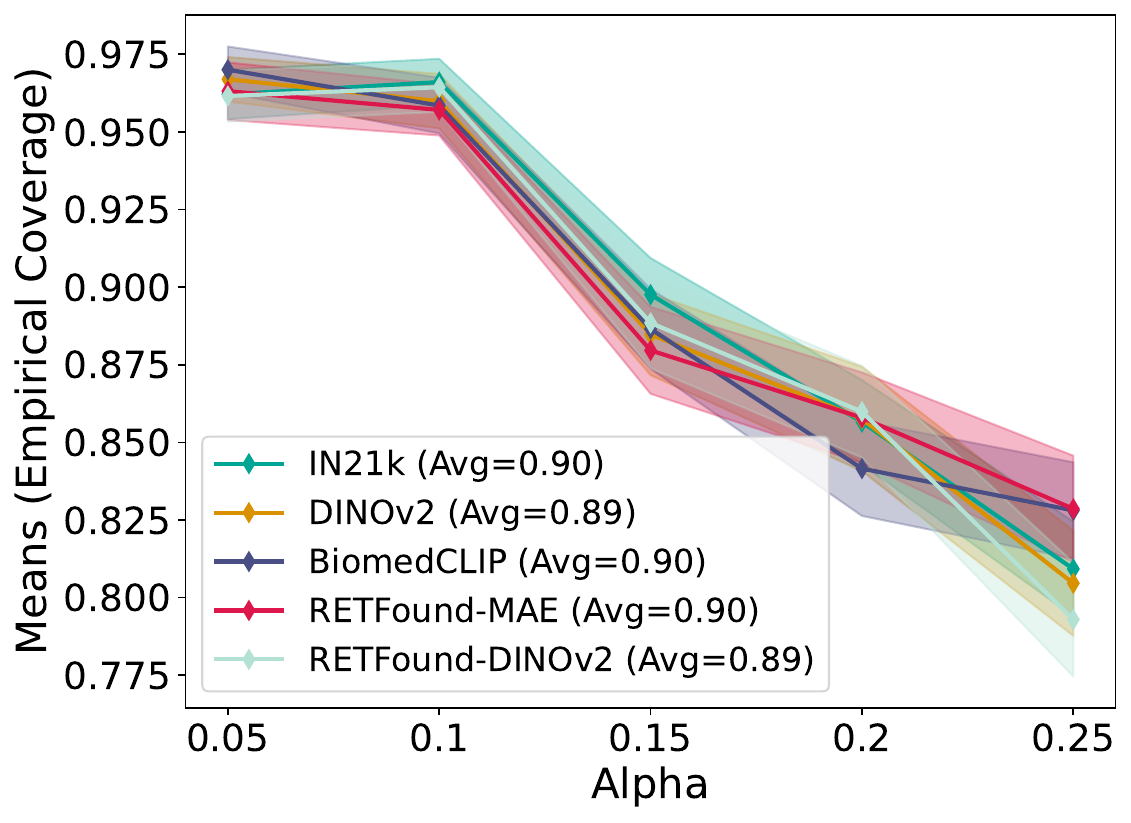}
    \includegraphics[height=0.19\textwidth, width=0.23\textwidth]{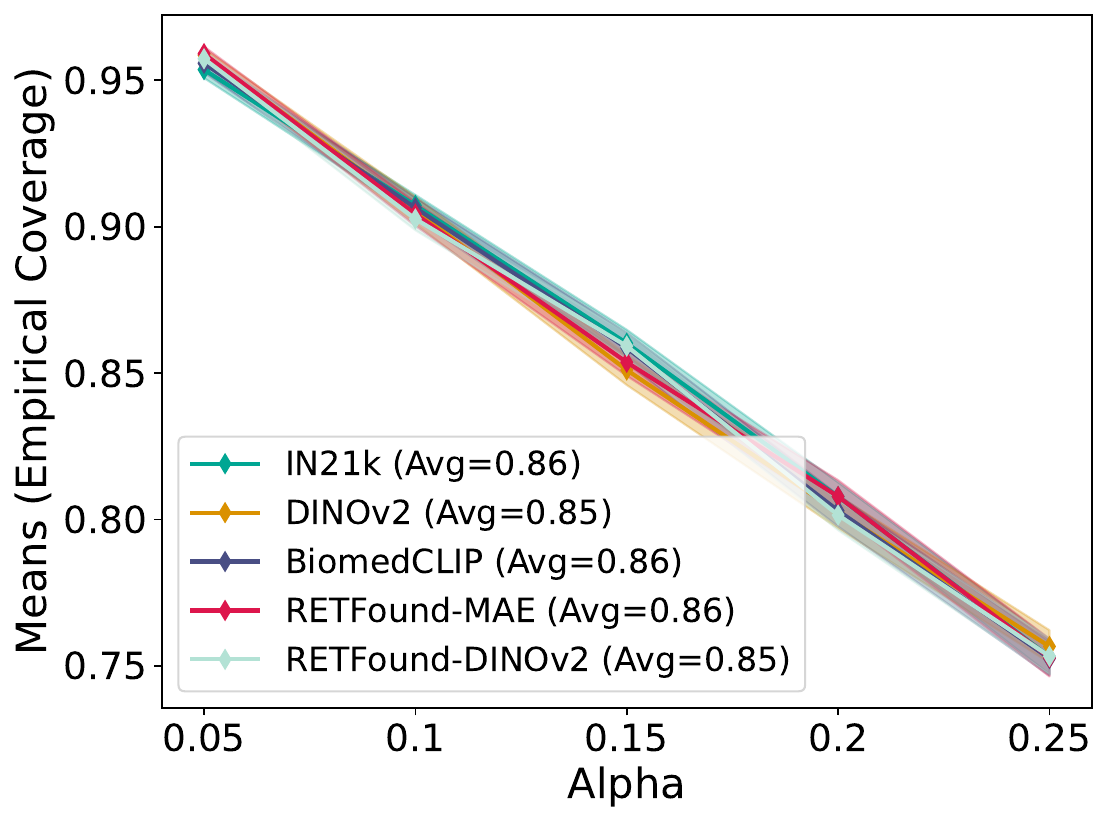}\\[0.2cm]
    \makebox[\textwidth][l]{%
        \hspace{0.20\textwidth}
        \textbf{Retina (LS)} \hspace{0.10\textwidth} \textbf{IDRiD (LS)} \hspace{0.08\textwidth} \textbf{APTOS2019 (LS)}
    } \\[0.2cm]
    \includegraphics[height=0.19\textwidth, width=0.23\textwidth]{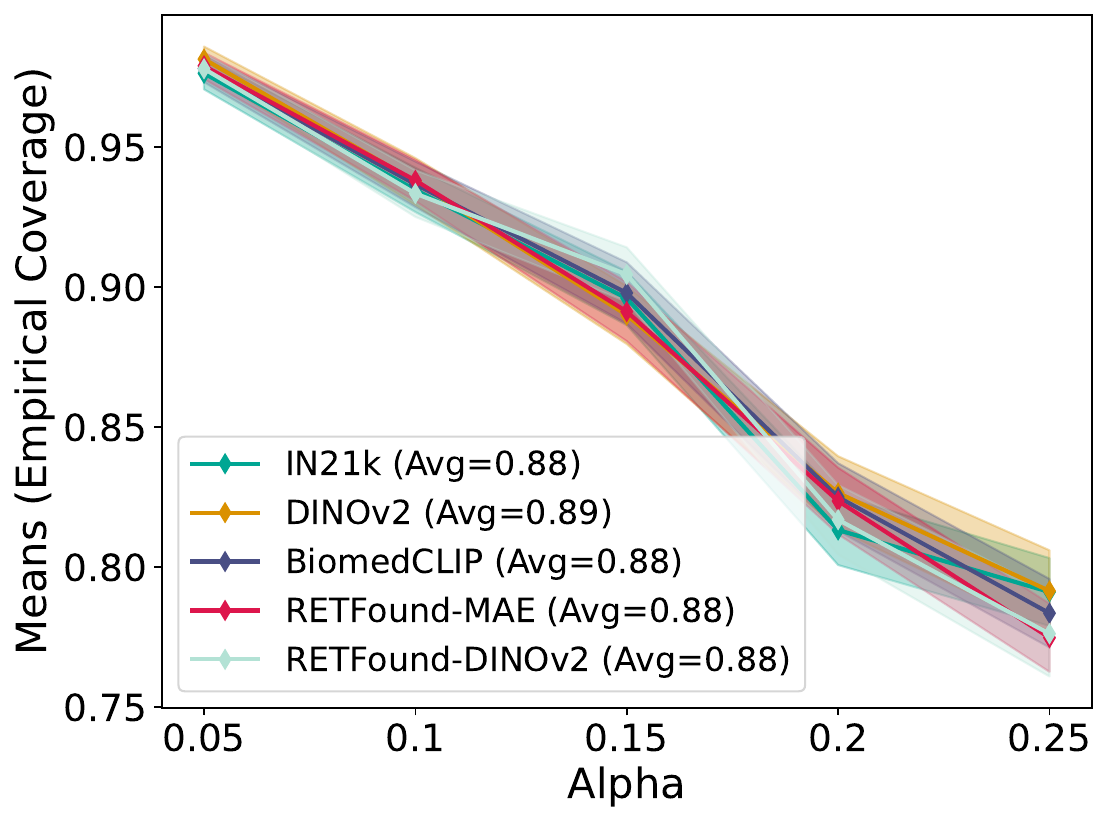} 
    \includegraphics[height=0.19\textwidth, width=0.23\textwidth]{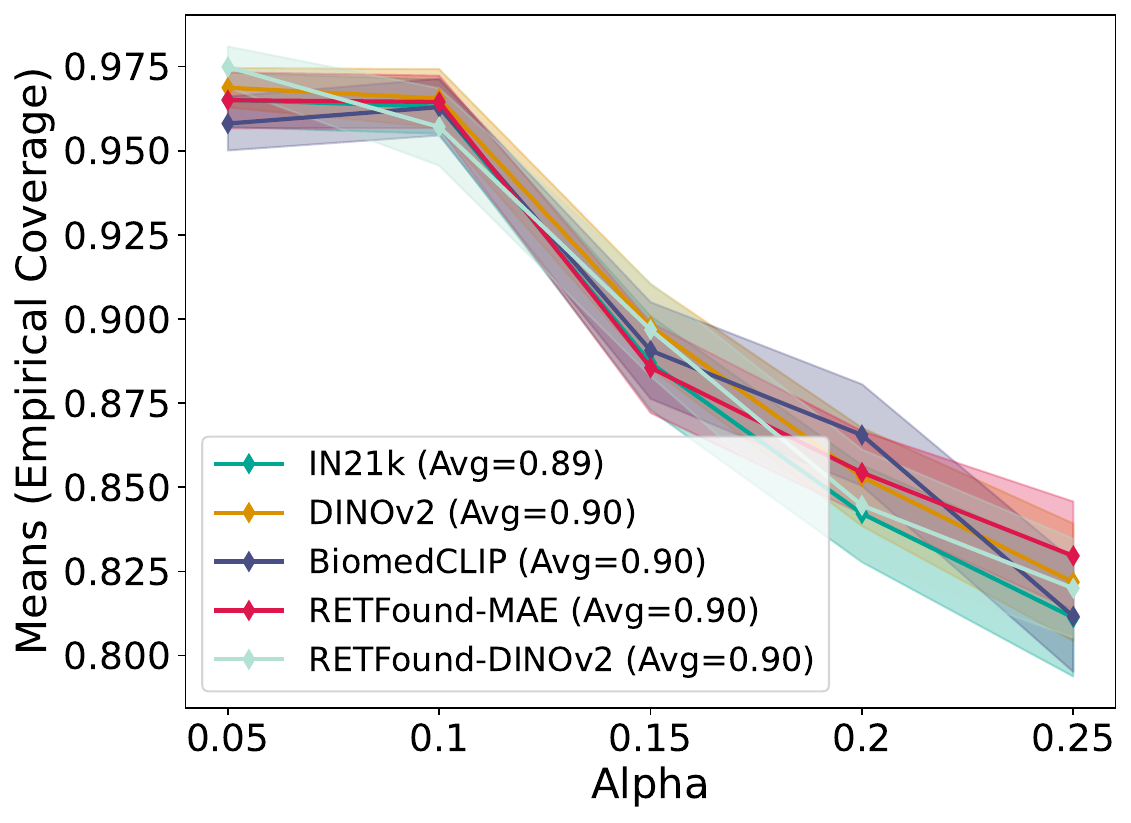}
    \includegraphics[height=0.19\textwidth, width=0.23\textwidth]{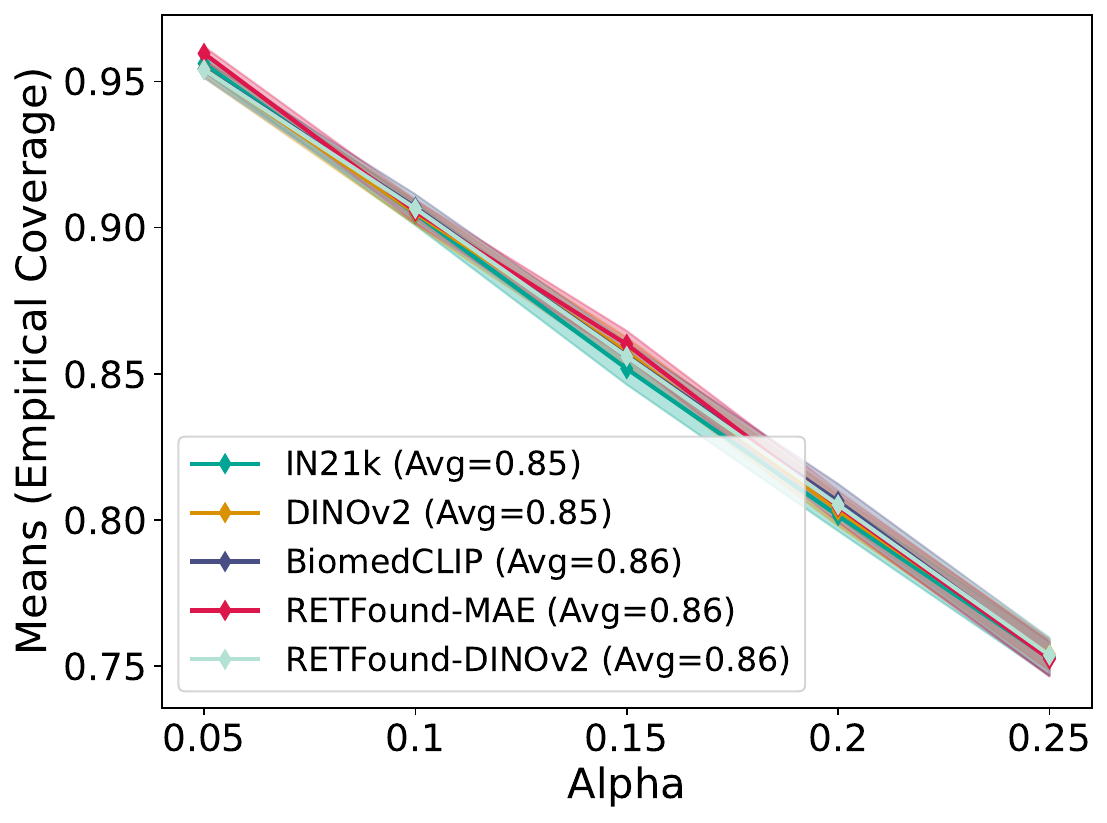}
    \caption{\textbf{Conformal prediction Empirical Coverage - RAPS (Retina)}}
    \label{fig:raps_cp_retina_coverage}
\end{figure*}

\begin{figure*}[!ht]
    \centering
    \makebox[\textwidth][l]{%
        \hspace{0.21\textwidth}
        \textbf{CRC100K} \hspace{0.13\textwidth} \textbf{TCGA} \hspace{0.15\textwidth} \textbf{BraTS}
    } \\[0.2cm]
    \includegraphics[height=0.19\textwidth, width=0.23\textwidth]{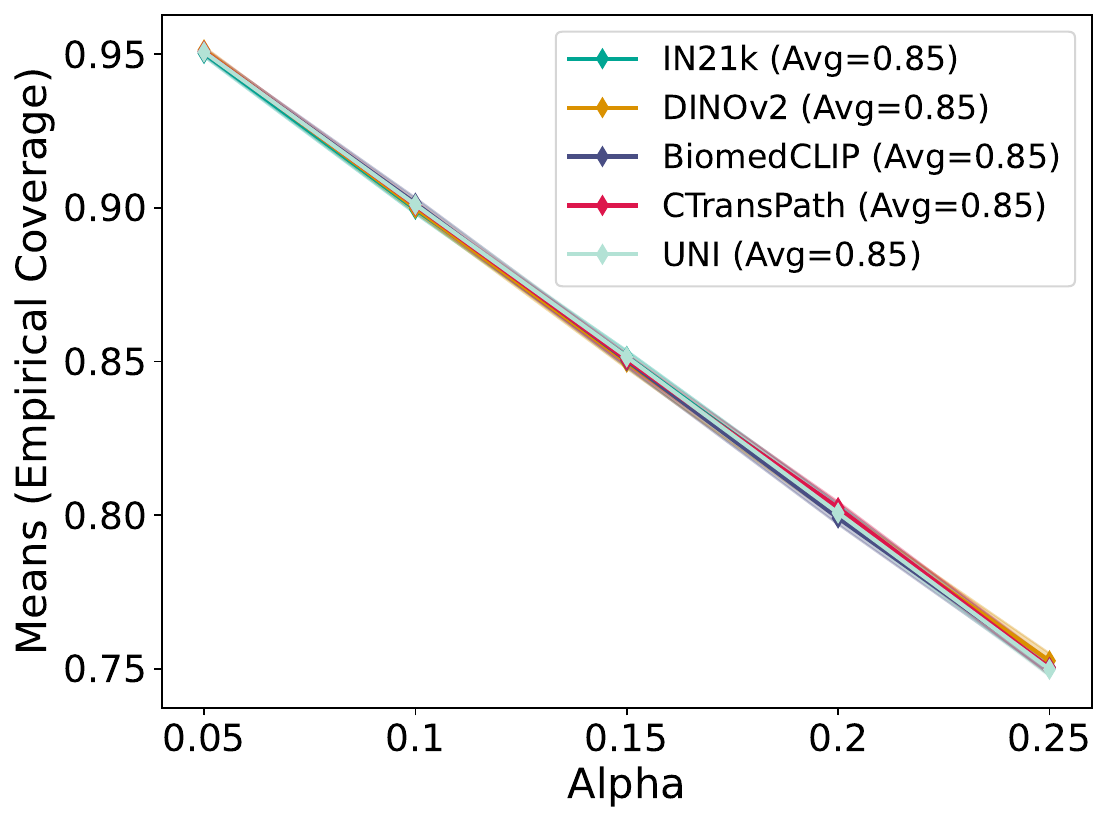} 
    \includegraphics[height=0.19\textwidth, width=0.23\textwidth]{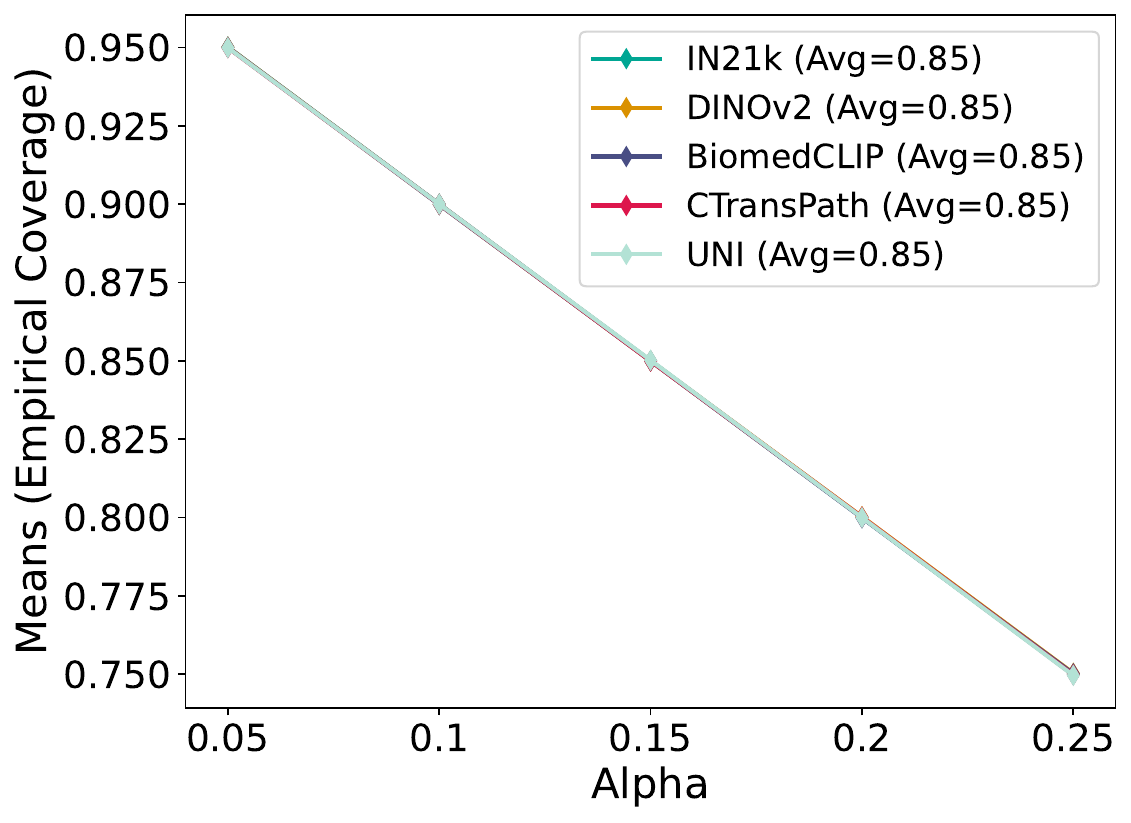}
    \includegraphics[height=0.19\textwidth, width=0.23\textwidth]{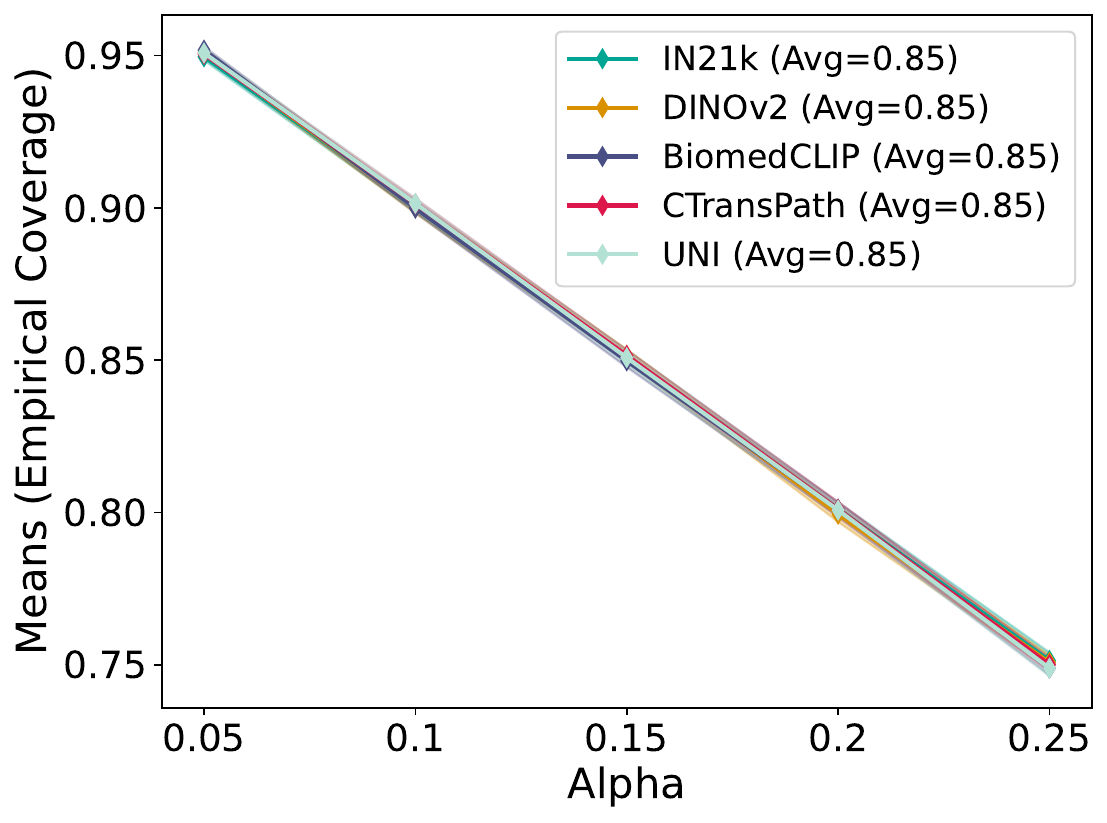}\\[0.2cm]
    \makebox[\textwidth][l]{%
        \hspace{0.20\textwidth}
        \textbf{CRC100K (T)} \hspace{0.08\textwidth} \textbf{TCGA (T)} \hspace{0.11\textwidth} \textbf{BraTS (T)}
    } \\[0.2cm]
    \includegraphics[height=0.19\textwidth, width=0.23\textwidth]{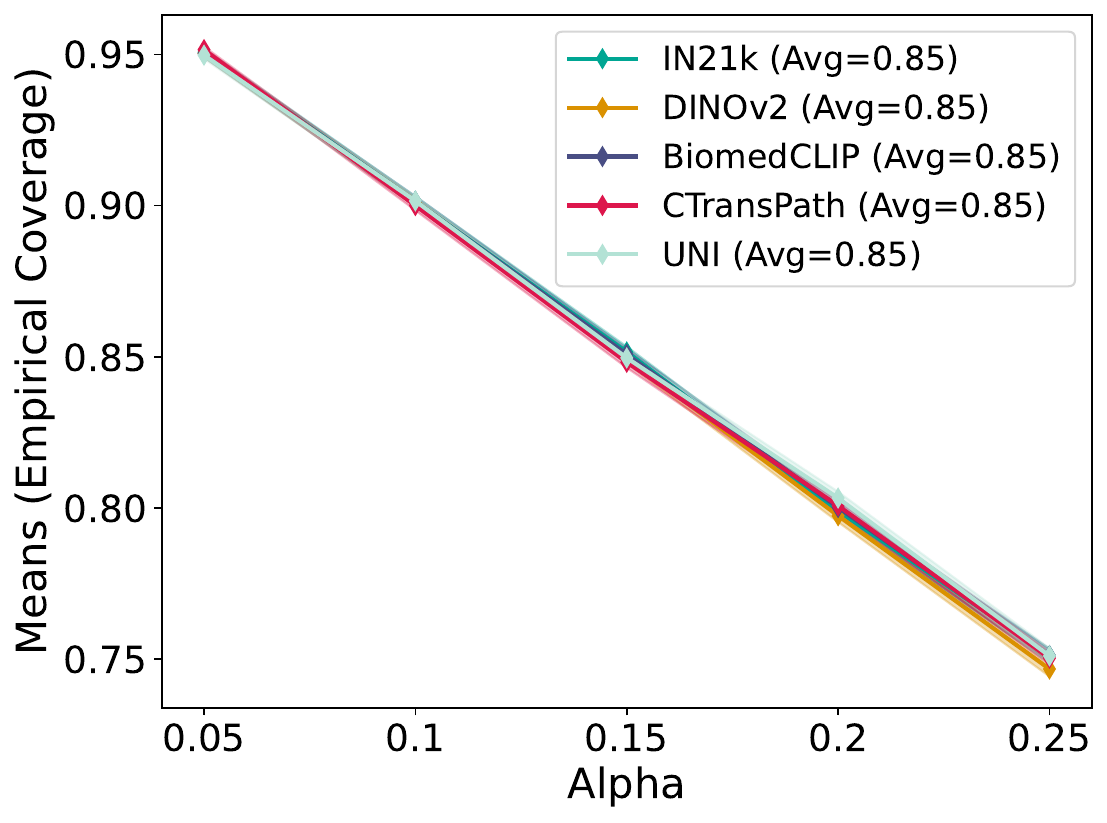} 
    \includegraphics[height=0.19\textwidth, width=0.23\textwidth]{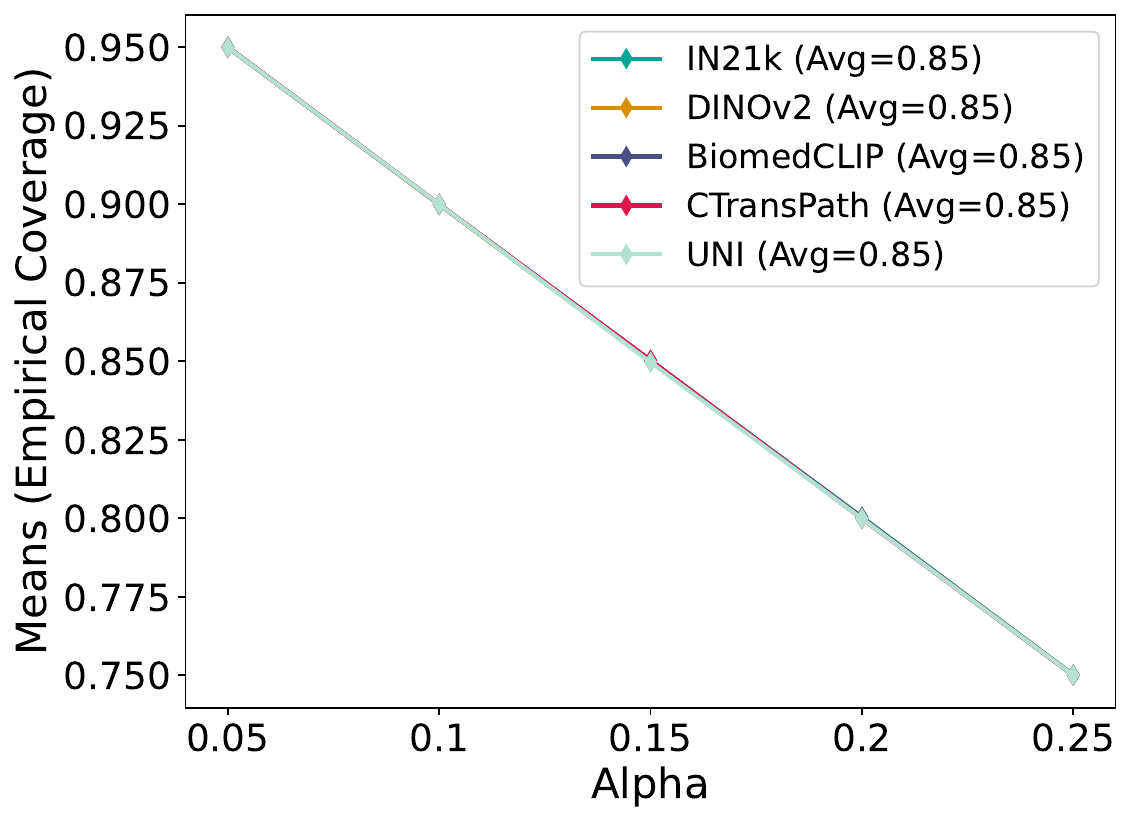}
    \includegraphics[height=0.19\textwidth, width=0.23\textwidth]{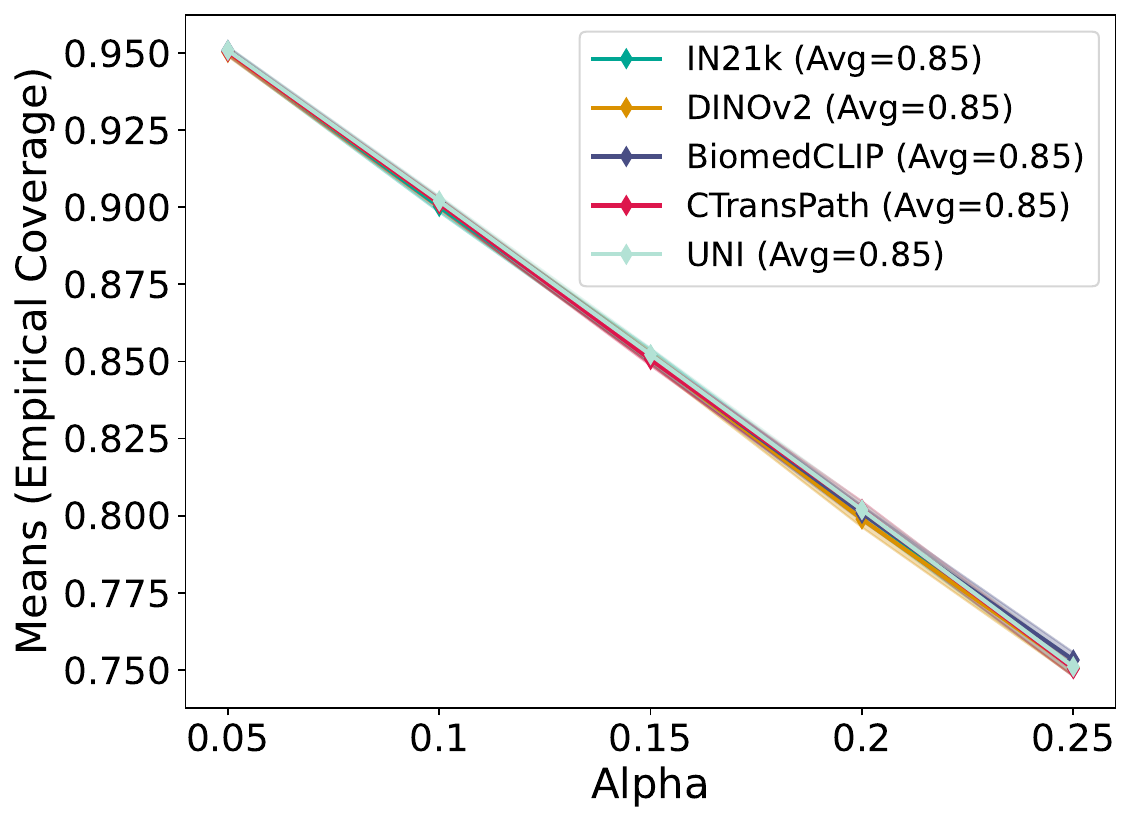}
    \makebox[\textwidth][l]{%
        \hspace{0.19\textwidth}
        \textbf{CRC100K (LS)} \hspace{0.07\textwidth} \textbf{TCGA (LS)} \hspace{0.09\textwidth} \textbf{BraTS (LS)}
    } \\[0.2cm]
    \includegraphics[height=0.19\textwidth, width=0.23\textwidth]{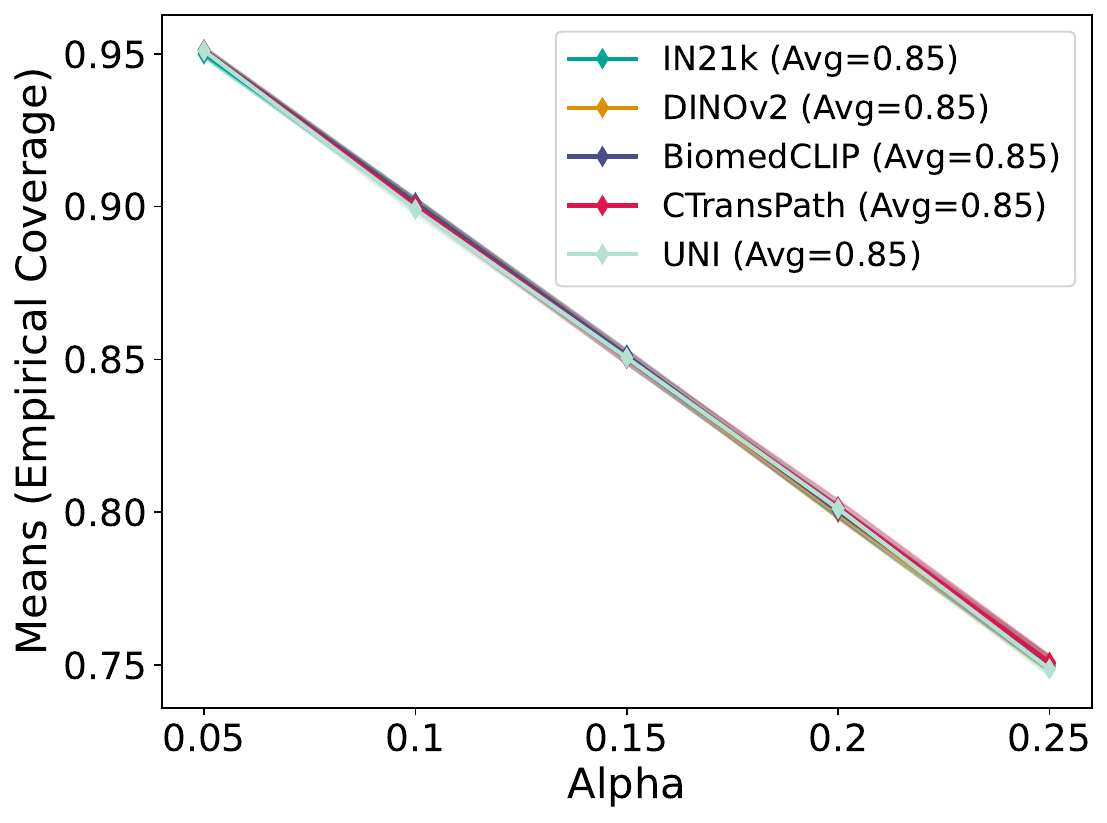} 
    \includegraphics[height=0.19\textwidth, width=0.23\textwidth]{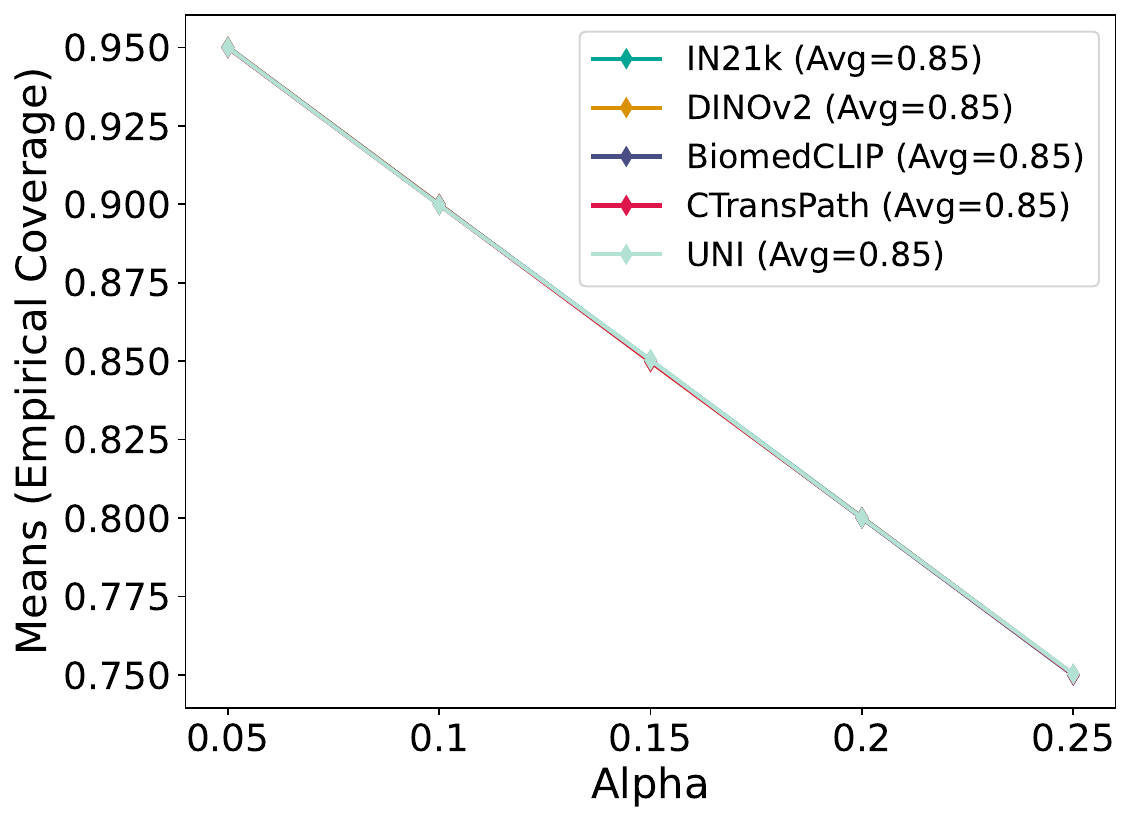}
    \includegraphics[height=0.19\textwidth, width=0.23\textwidth]{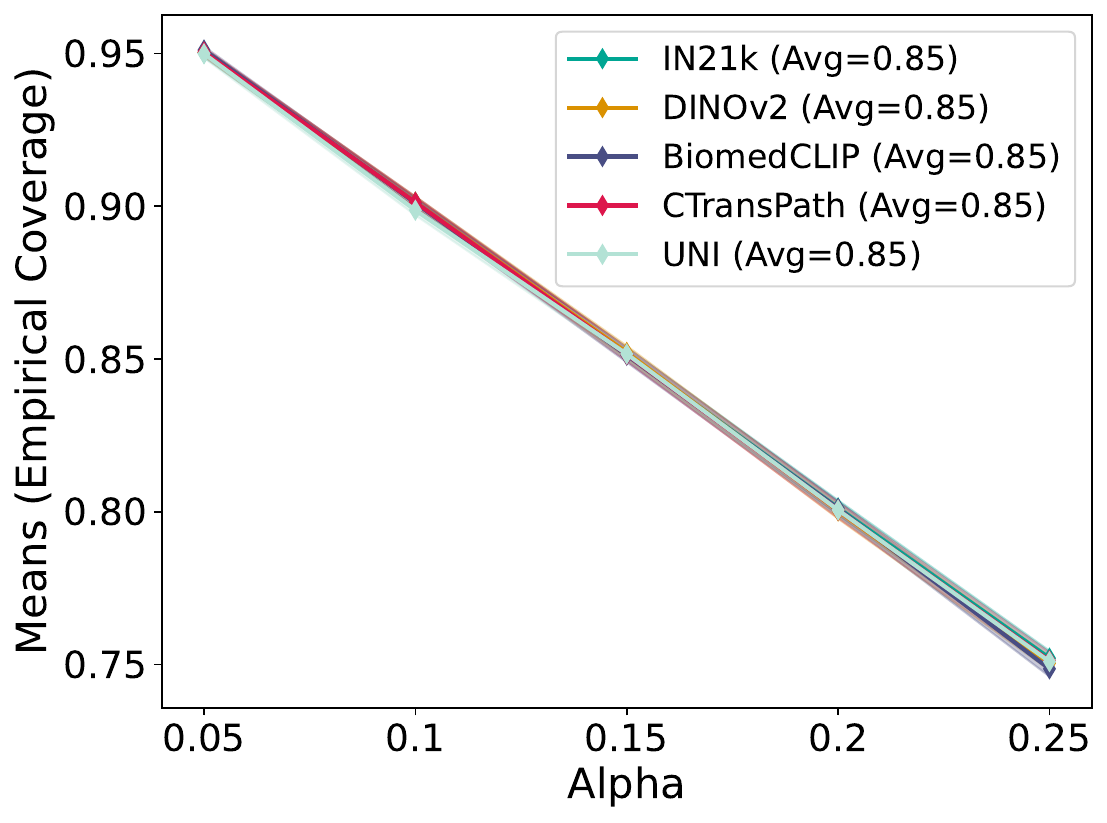}
    \caption{\textbf{Conformal prediction Empirical Coverage - RAPS (Histopathology)}}
    \label{fig:raps_cp_histopath_coverage}
\end{figure*}

\begin{figure*}[!ht]
    \centering
    \makebox[\textwidth][l]{%
        \hspace{0.24\textwidth}
        \textbf{RSNA} \hspace{0.12\textwidth} \textbf{POLCOVID} \hspace{0.09\textwidth} \textbf{COVID-Rad}
    } \\[0.2cm]
    \includegraphics[height=0.19\textwidth, width=0.23\textwidth]{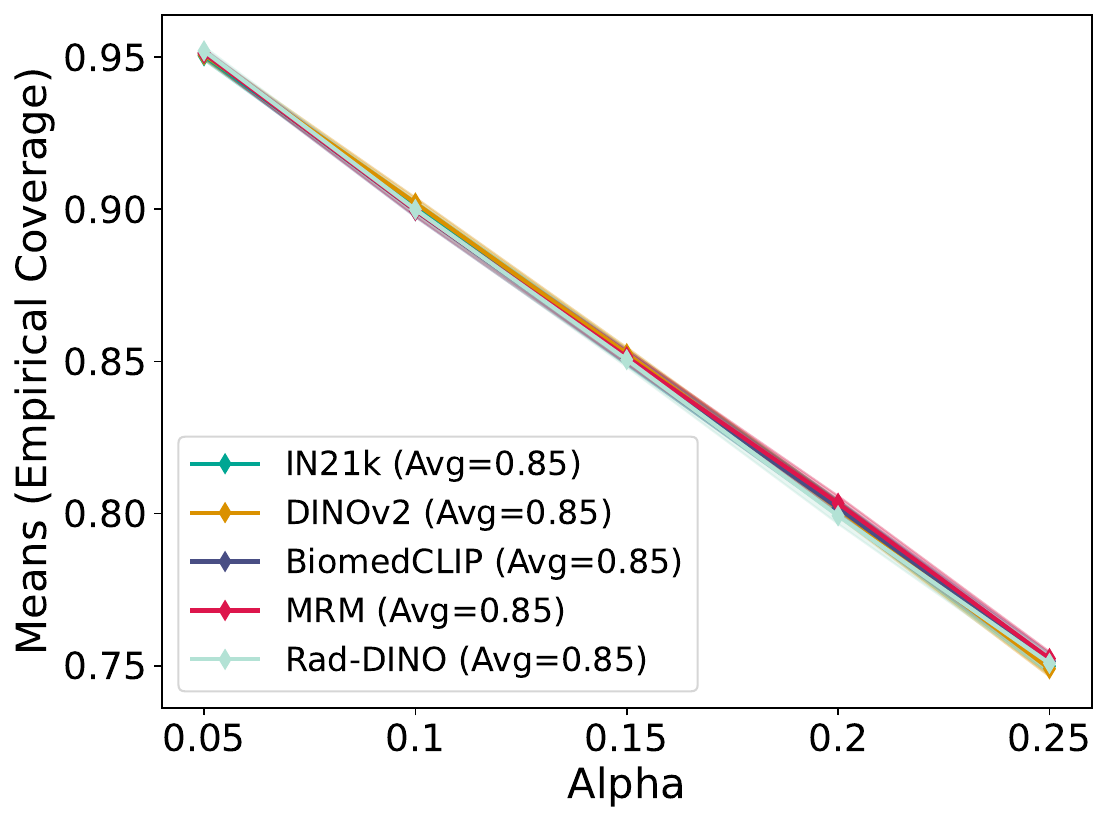} 
    \includegraphics[height=0.19\textwidth, width=0.23\textwidth]{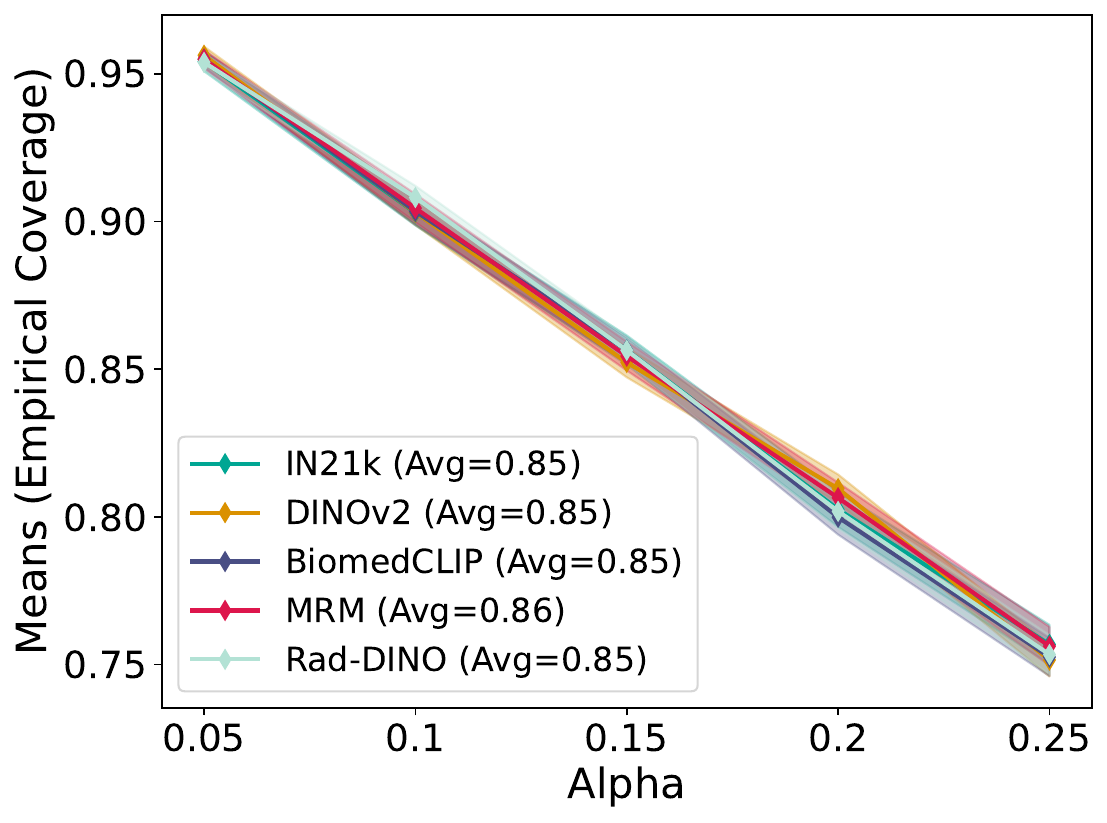}
    \includegraphics[height=0.19\textwidth, width=0.23\textwidth]{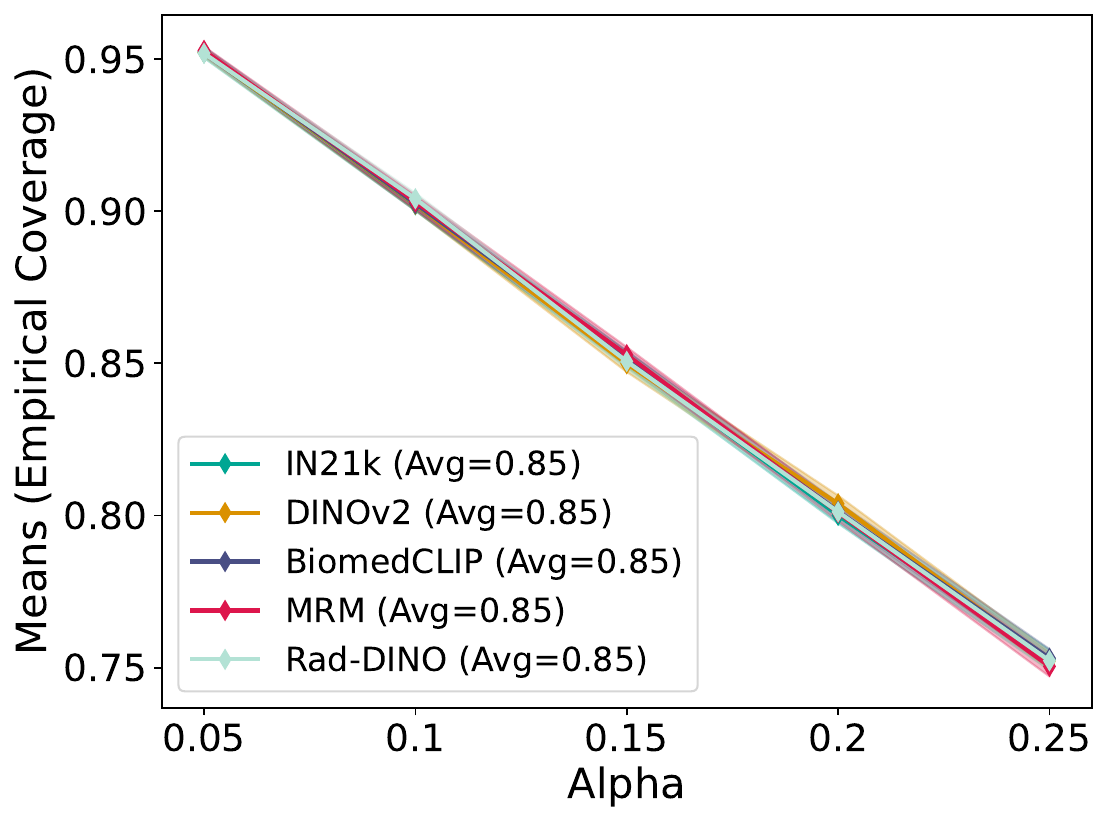}
    \makebox[\textwidth][l]{%
        \hspace{0.22\textwidth}
        \textbf{RSNA (T)} \hspace{0.09\textwidth} \textbf{POLCOVID (T)} \hspace{0.05\textwidth} \textbf{COVID-Rad (T)}
    } \\[0.2cm]
    \includegraphics[height=0.19\textwidth, width=0.23\textwidth]{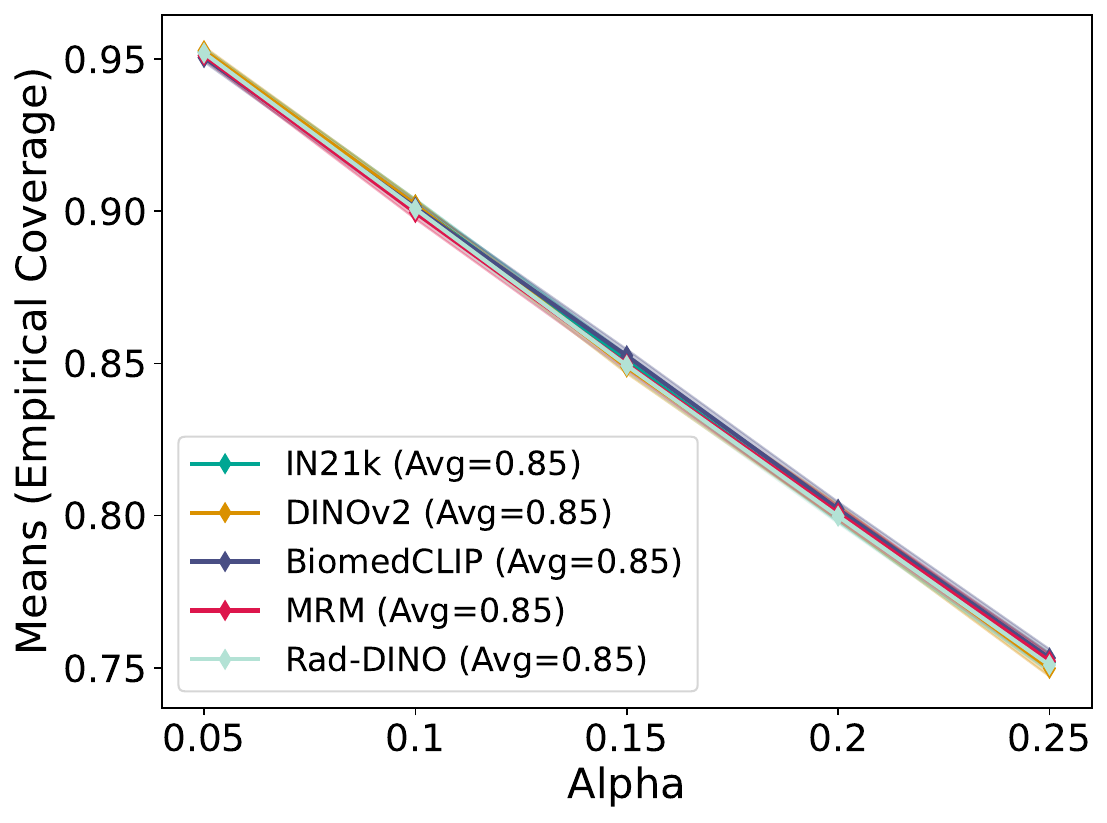} 
    \includegraphics[height=0.19\textwidth, width=0.23\textwidth]{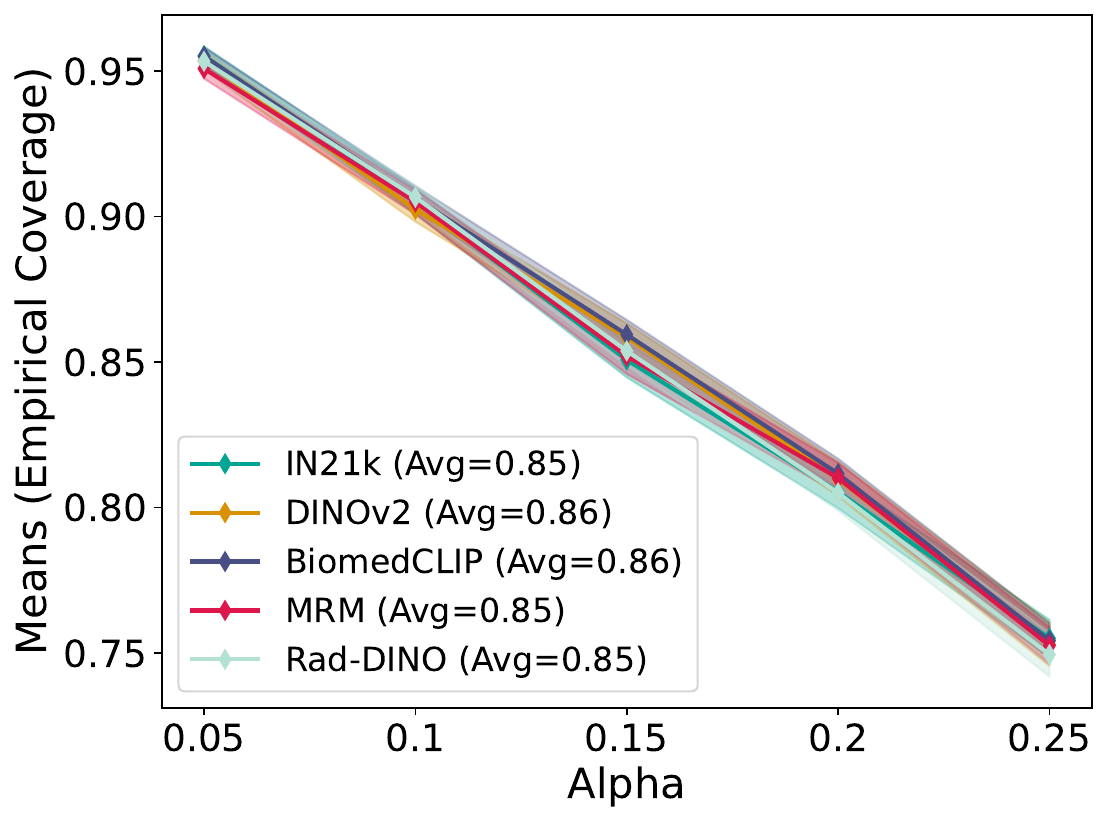}
    \includegraphics[height=0.19\textwidth, width=0.23\textwidth]{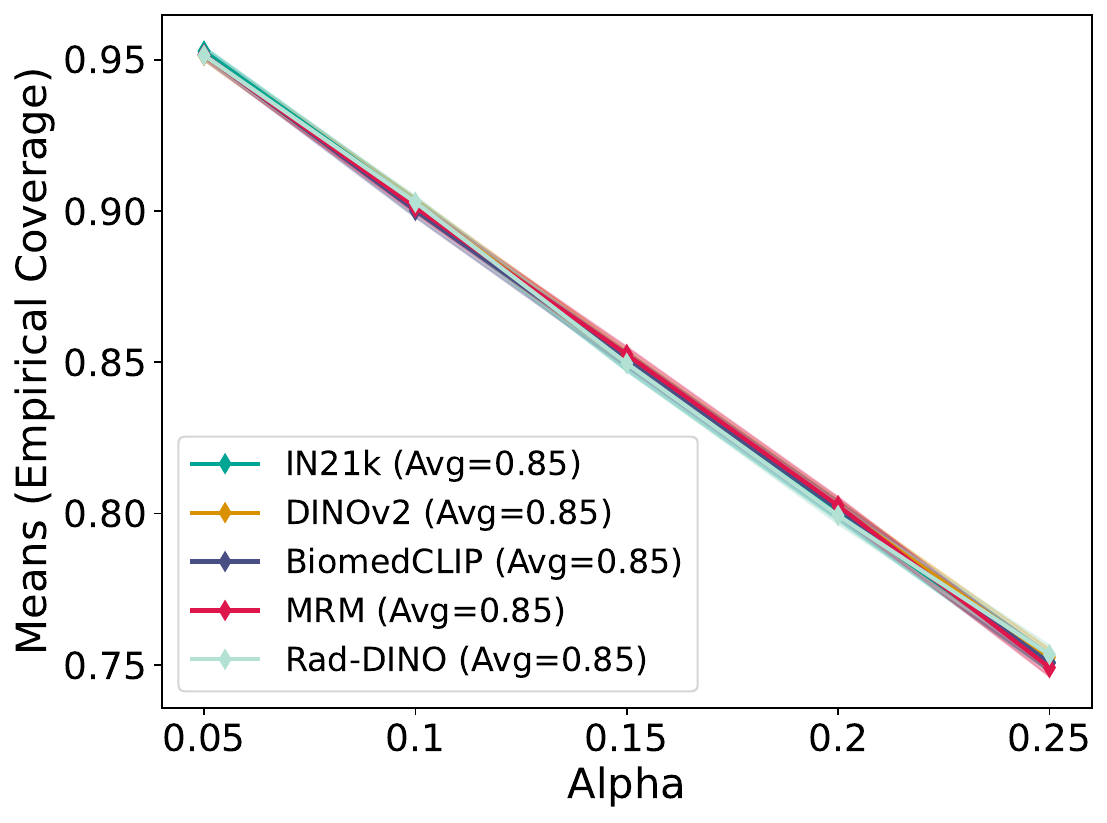}
    \makebox[\textwidth][l]{%
        \hspace{0.21\textwidth}
        \textbf{RSNA (LS)} \hspace{0.08\textwidth} \textbf{POLCOVID (LS)} \hspace{0.04\textwidth} \textbf{COVID-Rad (LS)}
    } \\[0.2cm]
    \includegraphics[height=0.19\textwidth, width=0.23\textwidth]{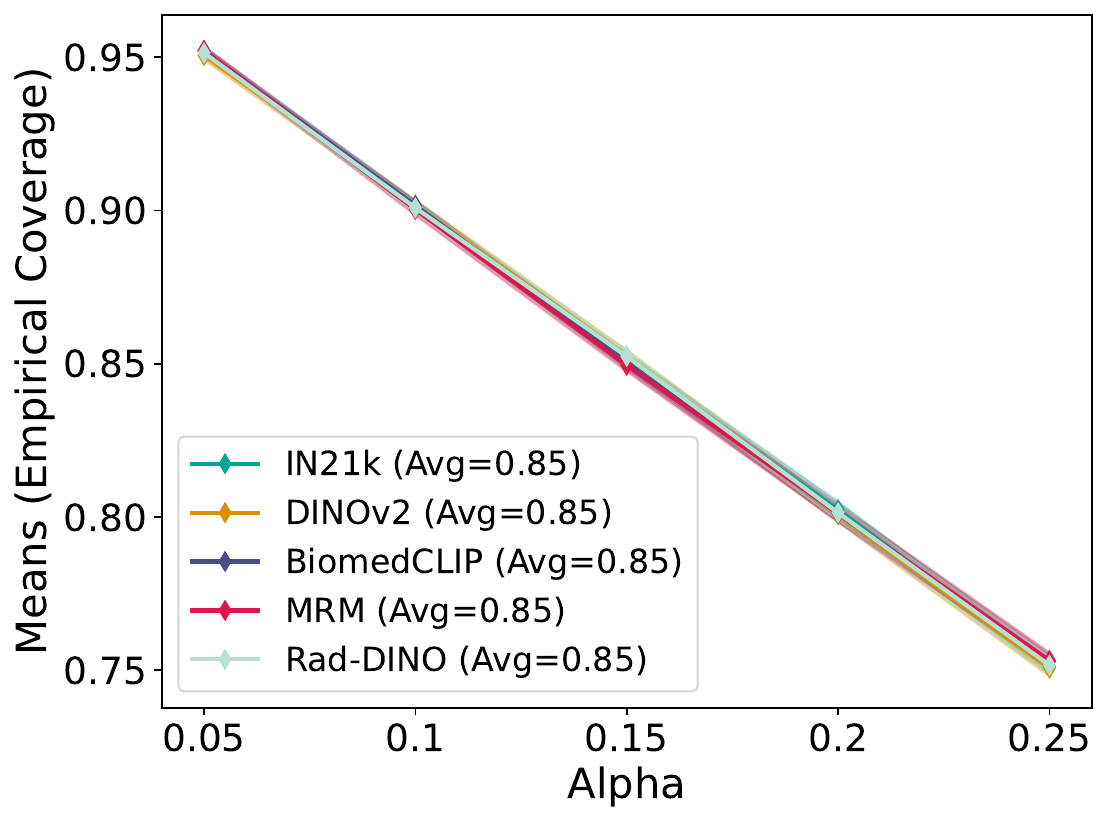} 
    \includegraphics[height=0.19\textwidth, width=0.23\textwidth]{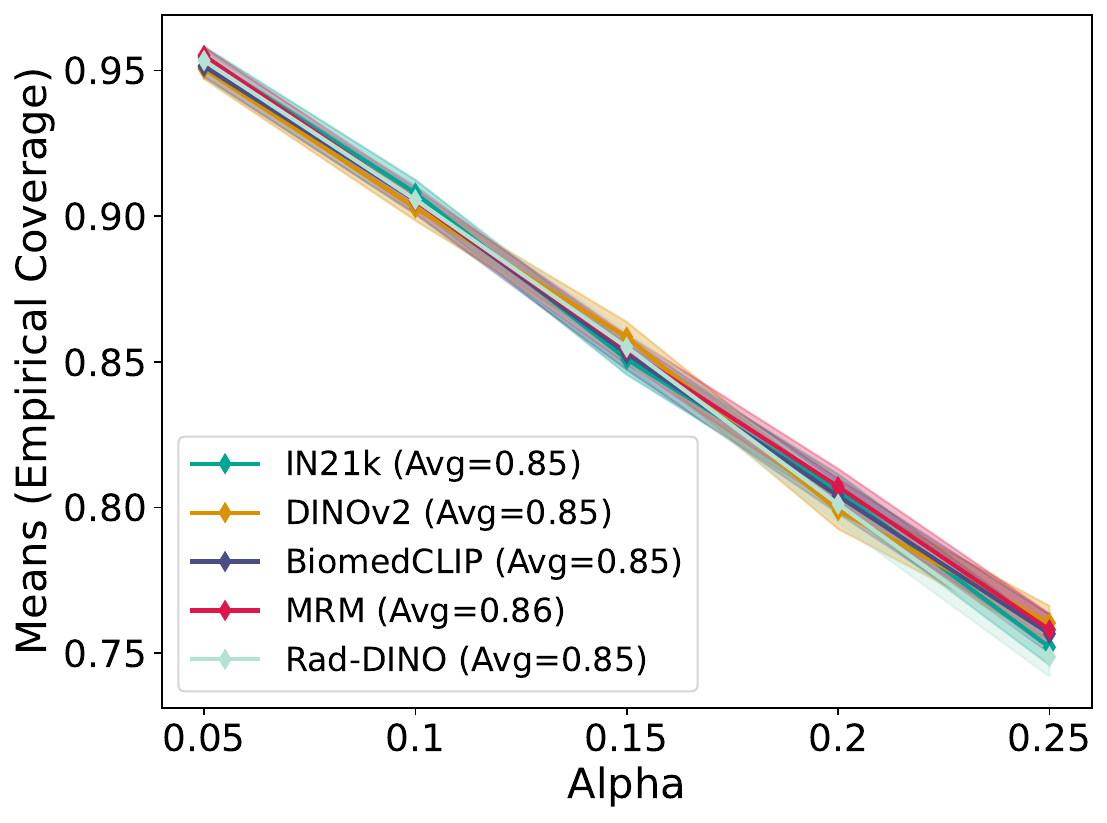}
    \includegraphics[height=0.19\textwidth, width=0.23\textwidth]{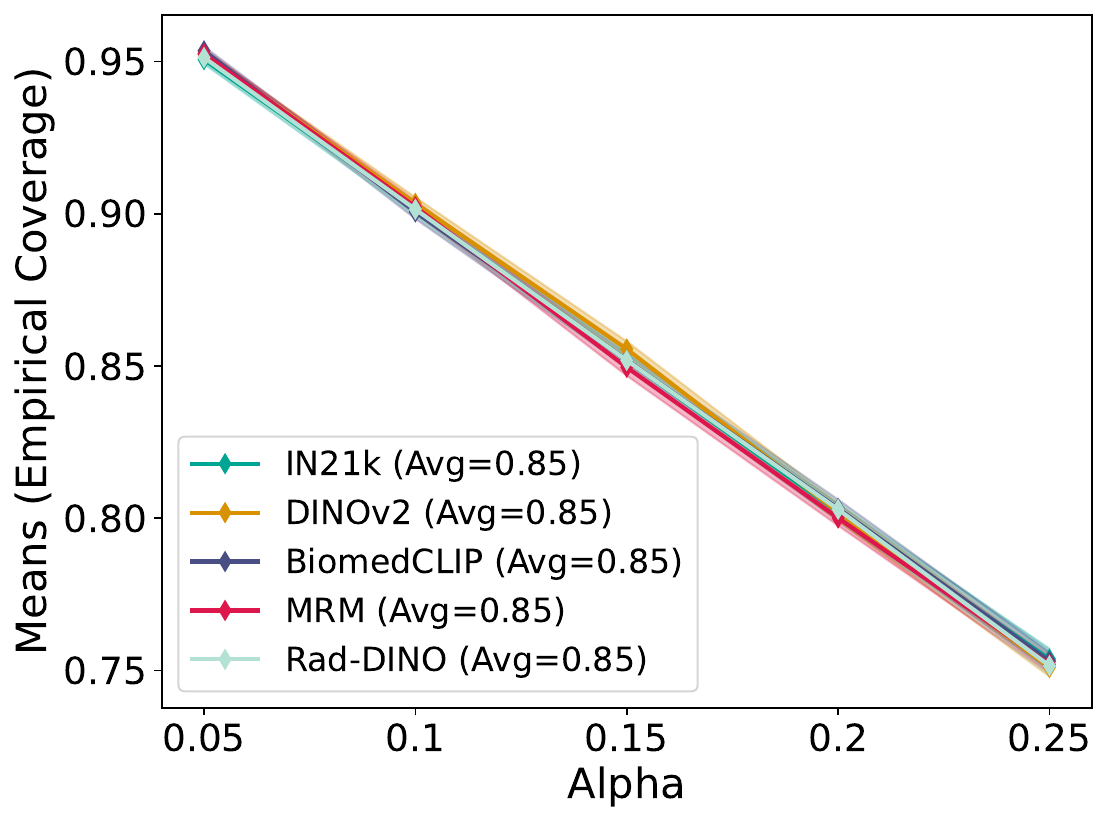}
    \caption{\textbf{Conformal prediction Empirical Coverage - RAPS (MRI)}}
    \label{fig:raps_cp_xrays_coverage}
\end{figure*}

We show model empirical coverage in \Cref{fig:raps_cp_retina_coverage,fig:raps_cp_histopath_coverage,fig:raps_cp_xrays_coverage,fig:cp_retina_coverage,fig:cp_histopath_coverage,fig:cp_xray_coverage}, where the result shows that the conformal predictors can achieve at least $1-\alpha$ coverage in all cases, indicating that they achieve expected coverage.

\clearpage
\section{Point Prediction Uncertainty Results}
We present complete point prediction uncertainty results in \Cref{tab:retina_uncertainty,tab:histopathology_uncertainty,tab:xray_uncertainty}, with ECE, Brier, and NLL. We show that model with better performance metrics (more accurate) is more likely have smaller conformal confidence set ({\bf T} stands for temperature scaling and {\bf LS} stands for label smoothing).
\begin{table*}[!ht]
\centering
\caption{Retina Datasets }
\resizebox{0.42\textwidth}{!}{%
\begin{tabular}{@{}p{3.0cm}llll@{}}
\toprule
\textbf{Model} & \textbf{ECE} $\downarrow$ & \textbf{Brier} $\downarrow$ & \textbf{NLL} $\downarrow$ &    \\ 
\midrule

\multicolumn{4}{c}{\textbf{Retina}}  \\ 
\midrule
ImageNet21k & 7.52  & 0.53  & 0.97 &    \\ 
DINOv2      & 15.38 & 0.58  & 1.09 &    \\ 
BioMedCLIP  & 8.41  & \underline{0.52}  & \textbf{0.93} &   \\ 
RETFound-MAE    & \textbf{4.38} & 0.67 & 1.26 &   \\ 
RETFound-DINOv2    & 8.63 & \textbf{0.52} & 0.97 &   \\ 
\midrule

\multicolumn{4}{c}{\textbf{Retina (T)}}  \\ 
\midrule
ImageNet21k & 6.06  & 0.65  & 1.21 &   \\ 
DINOv2      & 12.87  & 0.65  & 1.20 &   \\ 
BioMedCLIP  & 6.50  & 0.65  & 1.20 &   \\ 
RETFound-MAE    & \textbf{4.25}  & 0.72  & 1.33 &   \\  
RETFound-DINOv2    & 9.79 & \textbf{0.52} & \textbf{0.96} &   \\ 
\midrule

\multicolumn{4}{c}{\textbf{Retina (LS)}}  \\ 
\midrule
ImageNet21k & 9.25  & 0.53 & 0.98 &   \\ 
DINOv2      & 10.15 & 0.58 & 1.09 &   \\ 
BioMedCLIP  & 9.52 & \underline{0.52} & \textbf{0.94} &   \\ 
RETFound-MAE    & \textbf{4.28} & 0.67 & 1.26 &   \\ 
RETFound-DINOv2    & 8.60 & \textbf{0.52} & 0.97 &   \\ 



\midrule

\multicolumn{4}{c}{\textbf{IDRiD}}   \\ 
\midrule
ImageNet21k & 11.45 & 0.64  & 1.20 &  \\ 
DINOv2      & 9.38 & 0.64  & 1.23 &   \\
BioMedCLIP  & 10.22 & 0.63  & 1.19 &   \\ 
RETFound-MAE    & \textbf{4.84} & 0.73  & 1.43 &   \\ 
RETFound-DINOv2    & 8.79 & \textbf{0.62} & \textbf{1.19} &   \\ 
\midrule

\multicolumn{4}{c}{\textbf{IDRiD (T)}}   \\ 
\midrule
ImageNet21k & 7.50  & 0.74  & 1.46 &   \\ 
DINOv2      & 7.51  & 0.78  & 1.55 &   \\ 
BioMedCLIP  & 8.55  & 0.75  & 1.48 &   \\ 
RETFound-MAE    & \textbf{5.15}  & 0.78  & 1.55 &   \\ 
RETFound-DINOv2    & 7.28 & \textbf{0.62} & \textbf{1.20} &   \\ 
\midrule

\multicolumn{4}{c}{\textbf{IDRiD (LS)}}   \\ 
\midrule
ImageNet21k & 5.95 & 0.74  & 1.45 &   \\ 
DINOv2      & 8.78  & 0.65  & 1.24 &   \\ 
BioMedCLIP  & 6.96  & 0.63  & \underline{1.19} &   \\ 
RETFound-MAE    & \textbf{5.51} & 0.73  & 1.43 &   \\ 
RETFound-DINOv2    & 8.11 & \textbf{0.62} & \textbf{1.19} &   \\ 
\midrule

\multicolumn{4}{c}{\textbf{APTOS2019}}   \\ 
\midrule
ImageNet21k & 12.44 & 0.60 & 1.18 &   \\ 
DINOv2      & 8.26 & 0.36 & 0.73 &   \\ 
BioMedCLIP  & 4.13 & 0.32 & 0.63 &   \\ 
RETFound-MAE    &  7.56 & 0.46 & 0.97 &   \\ 
RETFound-DINOv2    & \textbf{1.95} & \textbf{0.28} & \textbf{0.55} &   \\ 
\midrule

\multicolumn{4}{c}{\textbf{APTOS2019 (T)}}   \\ 
\midrule
ImageNet21k & 7.92 & 0.58 & 1.17 &   \\ 
DINOv2      & 5.86 & 0.67 & 1.31 &   \\ 
BioMedCLIP  & 2.97 & 0.32 & 0.63 &   \\ 
RETFound-MAE    & 7.69 & 0.43 & 0.91 &   \\ 
RETFound-DINOv2    & \textbf{1.92} & \textbf{0.28} & \textbf{0.55} &   \\ 
\midrule

\multicolumn{4}{c}{\textbf{APTOS2019 (LS)}}   \\ 
\midrule
ImageNet21k & 11.74 & 0.60 & 1.19 &   \\ 
DINOv2      & 10.91 & 0.37 & 0.75 &   \\ 
BioMedCLIP  & 6.64 & 0.32 & 0.65 &   \\ 
RETFound-MAE    & 5.63 & 0.47 & 0.98 &   \\ 
RETFound-DINOv2    & \textbf{4.96}  & \textbf{0.28} & \textbf{0.56} &   \\ 

\bottomrule
\end{tabular}
}
\label{tab:retina_uncertainty}
\end{table*}

\begin{table*}[!ht]
\centering
\caption{Histopathology Datasets }
\resizebox{0.42\textwidth}{!}{
\begin{tabular}{@{}p{2.5cm}llll@{}}
\toprule
\textbf{Model} & \textbf{ECE} $\downarrow$ & \textbf{Brier} $\downarrow$ & \textbf{NLL} $\downarrow$ &  \\ 
\midrule

\multicolumn{4}{c}{\textbf{CRC100K}}  \\ 
\midrule
ImageNet21k & 19.91  & 0.51  & 1.37 &    \\ 
DINOv2      & 6.18   & 0.23  & 0.46 &    \\ 
BioMedCLIP  & 10.20  & 0.33  & 0.73 &    \\ 
CTransPath  & \textbf{2.96} & \underline{0.20}  & 0.40 &    \\ 
UNI         & 3.24 & \textbf{0.20} & \textbf{0.35} &    \\ 
\midrule

\multicolumn{4}{c}{\textbf{CRC100K (T)}}  \\ 
\midrule
ImageNet21k & 8.16  & 0.46  & 0.90 &    \\ 
DINOv2      & 4.53  & 0.23  & 0.44 &   \\ 
BioMedCLIP  & 7.84  & 0.32  & 0.67 &    \\ 
CTransPath  & \textbf{2.69}  & 0.22  & 0.45 &   \\ 
UNI         & 3.68  & \textbf{0.20}  & \textbf{0.35} &   \\  
\midrule

\multicolumn{4}{c}{\textbf{CRC100K (LS)}}  \\ 
\midrule
ImageNet21k & 8.75  & 0.46  & 0.91 &    \\ 
DINOv2      & 10.18 & 0.22  & 0.46 &    \\ 
BioMedCLIP  & \textbf{6.33}  & 0.33  & 0.71 &     \\ 
CTransPath  & 12.92 & 0.21  & \underline{0.45} &    \\ 
UNI         & 7.98  & \textbf{0.20}  & \textbf{0.45} &  \\  
\midrule

\multicolumn{4}{c}{\textbf{TCGA}}   \\ 
\midrule
ImageNet21k & 5.53  & 0.44  & 1.13 &   \\ 
DINOv2      & 1.47  & 0.41  & 1.02 &   \\ 
BioMedCLIP  & 2.52  & 0.46  & 1.18 &   \\ 
CTransPath  & \textbf{1.43} & 0.34  & 0.81 &   \\ 
UNI         & 6.98 & \textbf{0.27} & \textbf{0.72} &   \\ 
\midrule

\multicolumn{4}{c}{\textbf{TCGA (T)}}  \\ 
\midrule
ImageNet21k & 3.19  & 0.44  & 1.11 &    \\  
DINOv2      & 0.67  & 0.41  & 1.02 &    \\ 
BioMedCLIP  & \textbf{0.42}  & 0.46  & 1.17 &    \\ 
CTransPath  & 0.93  & 0.33  & 0.81 &     \\ 
UNI         & 2.22  & \textbf{0.26}  & \textbf{0.63} &    \\  
\midrule

\multicolumn{4}{c}{\textbf{TCGA (LS)}}  \\ 
\midrule
ImageNet21k & \textbf{2.83}  & 0.44  & 1.11 &   \\  
DINOv2      & 5.76  & 0.41  & 1.04 &    \\ 
BioMedCLIP  & 4.28  & 0.47  & 1.19 &   \\ 
CTransPath  & 7.94  & 0.35  & 0.86 &    \\ 
UNI         & 3.56  & \textbf{0.26}  & \textbf{0.65} &    \\  
\midrule

\multicolumn{4}{c}{\textbf{BraTS-Path}}  \\ 
\midrule
ImageNet21k & 13.76 & 0.28  & 2.39 &     \\ 
DINOv2      & 2.05  & 0.19  & 0.38 &    \\ 
BioMedCLIP  & 4.25  & 0.23  & 0.46 &    \\ 
CTransPath  & 7.88 & 0.20  & 0.39 &    \\ 
UNI         & \textbf{1.59} & \textbf{0.10} & \textbf{0.19} &    \\ 
\midrule

\multicolumn{4}{c}{\textbf{BraTS-Path (T)}}  \\ 
\midrule
ImageNet21k & 3.98  & 0.19  & 0.40 &    \\ 
DINOv2      & \textbf{1.28}  & 0.19  & 0.38 &  \\ 
BioMedCLIP  & 3.81  & 0.23  & 0.45 &   \\ 
CTransPath  & 2.85  & 0.19  & 0.36 &     \\ 
UNI         & 1.42  & \textbf{0.10}  & \textbf{0.19} &    \\  
\midrule

\multicolumn{4}{c}{\textbf{BraTS-Path (LS)}}   \\ 
\midrule
ImageNet21k & 4.61  & 0.18  & 0.37 &    \\ 
DINOv2      & 8.00  & 0.21  & 0.42 &     \\ 
BioMedCLIP  & 4.71  & 0.22  & 0.44 &     \\ 
CTransPath  & 12.81 & 0.22  & 0.45 &     \\ 
UNI         & \textbf{3.96}  & \textbf{0.10}  & \textbf{0.22} &    \\  
\bottomrule
\end{tabular}
}
\label{tab:histopathology_uncertainty}
\end{table*}

\begin{table*}[!ht]
\centering

\caption{Chest X-Rays Datasets }
\resizebox{0.42\textwidth}{!}{
\begin{tabular}{@{}p{2.5cm}llll@{}}
\toprule
\textbf{Model} & \textbf{ECE} $\downarrow$ & \textbf{Brier} $\downarrow$ & \textbf{NLL} $\downarrow$ &  \\ 
\midrule

\multicolumn{4}{c}{\textbf{RSNA-Pneumonia}}  \\ 
\midrule
ImageNet21k  & 3.28 & 0.27 & 0.43 &    \\
DINOv2       & 3.26 & 0.24 & 0.38 &    \\
BioMedCLIP   & 2.81 & 0.28 & 0.43 &    \\
MRM  & 2.14 & 0.26 & 0.41 &   \\
Rad-DINO  & \textbf{1.29} & \textbf{0.23} & \textbf{0.36} &    \\
\midrule

\multicolumn{4}{c}{\textbf{RSNA-Pneumonia (T)}}  \\ 
\midrule
ImageNet21k  & 3.00 & 0.27 & 0.43 &    \\
DINOv2       & 2.98 & 0.28 & 0.43 &    \\
BioMedCLIP   & \textbf{0.76} & \underline{0.23} & 0.37 &    \\
MRM  & 1.72 & 0.26 & 0.40 &   \\
Rad-DINO  & 0.78 & \textbf{0.23} & \textbf{0.36} &    \\ 
\midrule

\multicolumn{4}{c}{\textbf{RSNA-Pneumonia (LS)}}   \\ 
\midrule
ImageNet21k  & 3.32 & 0.28 & 0.43 &    \\
DINOv2       & 3.92 & 0.28 & 0.44 &    \\
BioMedCLIP   & 1.92 & 0.24 & 0.37 &    \\
MRM  & 3.23 & 0.26 & 0.41 &   \\
Rad-DINO  & \textbf{1.25} & \textbf{0.23} & \textbf{0.36} &    \\ 
\midrule

\multicolumn{4}{c}{\textbf{POLCOVID}}  \\ 
\midrule
ImageNet21k  & 22.38 & 0.62 & 1.04 &    \\
DINOv2       & 4.58 & 0.47 & 0.80 &    \\
BioMedCLIP   & \textbf{3.15} & 0.39 & 0.68 &    \\
MRM  & 7.20 & 0.54 & 0.91 &    \\
Rad-DINO  & 3.43 & \textbf{0.29} & \textbf{0.52} &    \\ 
\midrule

\multicolumn{4}{c}{\textbf{POLCOVID (T)}}  \\ 
\midrule
ImageNet21k  & 9.98 & 0.56 & 0.94 &    \\
DINOv2       & 3.36 & 0.47 & 0.79 &    \\
BioMedCLIP   & \textbf{2.36} & 0.39 & 0.68 &    \\
MRM  & 7.87 & 0.53 & 0.90 &   \\
Rad-DINO  & 2.52 & \textbf{0.28} & \textbf{0.52} &    \\ 
\midrule

\multicolumn{4}{c}{\textbf{POLCOVID (LS)}}  \\ 
\midrule
ImageNet21k  & 19.69 & 0.60 & 1.01 &    \\
DINOv2       & 4.42 & 0.60 & 1.01 &    \\
BioMedCLIP   & \textbf{4.31} & 0.39 & 0.68 &    \\
MRM  & 6.54 & 0.54 & 0.91 &   \\
Rad-DINO  & 5.27 & \textbf{0.29} & \textbf{0.53} &    \\ 
\midrule

\multicolumn{4}{c}{\textbf{COVID-Rad}}  \\ 
\midrule
ImageNet21k  & 7.93 & 0.41 & 0.72 &    \\
DINOv2       & 7.31 & 0.43 & 0.75 &    \\
BioMedCLIP   & \textbf{1.09} & 0.19 & 0.34 &    \\
MRM  & 6.72 & 0.37 & 0.67 &   \\
Rad-DINO  & 1.48 & \textbf{0.10} & \textbf{0.18} &    \\
\midrule

\multicolumn{4}{c}{\textbf{COVID-Rad (T)}}  \\ 
\midrule
ImageNet21k  & 7.67 & 0.40 & 0.72 &    \\
DINOv2       & 6.09 & 0.43 & 0.75 &    \\
BioMedCLIP   & 0.98 & 0.19 & 0.34 &    \\
MRM  & 7.28 & 0.37 & 0.66 &   \\
Rad-DINO  & \textbf{0.40} & \textbf{0.10} & \textbf{0.18} &    \\ 
\midrule

\multicolumn{4}{c}{\textbf{COVID-Rad (LS)}}   \\ 
\midrule
ImageNet21k  & 7.89 & 0.41 & 0.72 &    \\
DINOv2       & 6.20 & 0.43 & 0.76 &    \\
BioMedCLIP   & \textbf{4.20} & 0.19 & 0.36 &   \\
MRM  & 7.32 & 0.37 & 0.67 &   \\
Rad-DINO  & 5.26 & \textbf{0.10} & \textbf{0.21} &    \\
\bottomrule
\end{tabular}
}
\label{tab:xray_uncertainty}
\end{table*}

\clearpage
\section{Model Performance Results}
\label{apd:model_perf}
We present detailed model performance results with accuracy (Acc), balanced accuracy (BAcc), Area Under Receiver Operating Characteristic Curve (AUROC), and Area Under Precision-Recall Curve (AUPRC) in \Cref{tab:combined_retina_perf,tab:combined_histopathology_perf,tab:combined_xray_perf}. All results are reported with mean and $95\%$ confidence intervals. The result indicates that the experimented calibration methods do not vary the model accuracy performance much ({\bf T} stands for temperature scaling and {\bf LS} stands for label smoothing).
\begin{table*}[!ht]
\centering
\caption{Retina Datasets Performance Evaluation}
\label{tab:combined_retina_perf}
\resizebox{0.72\textwidth}{!}{%
\begin{tabular}{@{}>{\centering\arraybackslash}p{3.5cm}cccc@{}}
\toprule
\textbf{Model} & \textbf{Acc $\uparrow$} & \textbf{BAcc $\uparrow$} & \textbf{AUROC $\uparrow$} & \textbf{AUPRC $\uparrow$} \\
\hline

\multicolumn{5}{c}{\textbf{Retina}} \\
\hline
ImageNet21k    & 85.61 $\pm$1.58 & 60.36 $\pm$2.38 & 79.50 $\pm$2.69 & 59.66 $\pm$4.55 \\
DINOv2         & 85.70 $\pm$1.41 & 58.56 $\pm$1.77 & 73.49 $\pm$3.03 & 53.47 $\pm$3.75 \\
BioMedCLIP     & 85.83 $\pm$1.47 & 63.14 $\pm$2.09 & \textbf{80.11} $\pm$2.62 & 59.53 $\pm$3.91 \\
RETFound-MAE       & 84.96 $\pm$1.23 & 56.76 $\pm$1.73 & 72.60 $\pm$3.03 & 51.96 $\pm$3.77 \\
RETFound-DINOv2       & \textbf{87.13} $\pm$1.60 & \textbf{67.83} $\pm$2.59 & 77.11$\pm$3.57 & \textbf{59.97} $\pm$5.03 \\
\hline

\multicolumn{5}{c}{\textbf{Retina (T)}} \\
\hline
ImageNet21k    & 86.18 $\pm$1.25 & 62.76 $\pm$2.16 & \textbf{81.66} $\pm$2.27 & \textbf{61.73} $\pm$4.75 \\
DINOv2         & 85.66 $\pm$1.37 & 62.05 $\pm$2.01 & 81.69 $\pm$2.83 & 60.17 $\pm$4.43 \\
BioMedCLIP     & 86.01 $\pm$2.16 & 63.44 $\pm$3.40 & 80.20 $\pm$2.27 & 59.69 $\pm$4.14 \\
RETFound-MAE       & 83.24 $\pm$1.45 & 59.13 $\pm$2.23 & 68.63 $\pm$3.25 & 48.07 $\pm$3.77 \\
RETFound-DINOv2       & \textbf{87.17} $\pm$1.47 & \textbf{67.94} $\pm$2.54 & 77.33 $\pm$2.69 & 60.32 $\pm$3.90\\
\hline

\multicolumn{5}{c}{\textbf{Retina (LS)}} \\
\hline
ImageNet21k    & 85.60 $\pm$1.60 & 60.70 $\pm$2.43 & \textbf{81.44} $\pm$2.27 & \textbf{61.63} $\pm$4.20 \\
DINOv2         & 85.64 $\pm$1.57 & 58.99 $\pm$1.83 & 73.19 $\pm$2.61 & 52.65 $\pm$4.12 \\
BioMedCLIP     & 86.60 $\pm$1.30 & 64.05 $\pm$1.98 & 80.44 $\pm$2.27 & 60.03 $\pm$3.61 \\
RETFound-MAE       & 83.19 $\pm$1.19 & 60.02 $\pm$1.58 & 68.40 $\pm$3.12 & 47.99 $\pm$2.70 \\
RETFound-DINOv2       & \textbf{86.87} $\pm$1.84 & \textbf{67.63} $\pm$2.21 & 77.13 $\pm$3.26 & 59.94 $\pm$4.38 \\
\hline

\multicolumn{5}{c}{\textbf{IDRiD}} \\
\hline
ImageNet21k    & \textbf{84.65}$\pm$2.30 & \textbf{59.48}$\pm$2.42 & \textbf{80.12}$\pm$3.54 &  \textbf{67.11}$\pm$5.35 \\
DINOv2         & 81.18 $\pm$1.61 & 55.29 $\pm$2.04 & 80.09 $\pm$2.09 & 52.70 $\pm$3.61 \\
BioMedCLIP     & 84.63$\pm$2.13 & 58.38$\pm$2.76 & 73.80$\pm$3.64 &  56.84$\pm$4.29 \\
RETFound-MAE       & 84.96 $\pm$1.53 & 56.78 $\pm$1.91 & 72.78 $\pm$2.78 & 52.34 $\pm$4.06 \\
RETFound-DINOv2       & 81.29 $\pm$1.62 & 55.44 $\pm$1.95 & 80.28 $\pm$2.83 & 52.91$\pm$3.30 \\
\hline

\multicolumn{5}{c}{\textbf{IDRiD (T)}} \\
\hline
ImageNet21k    & \textbf{85.78}$\pm$2.44 & 58.97 $\pm$2.85 & \textbf{80.09}$\pm$2.95 &  \textbf{66.31}$\pm$4.88 \\
DINOv2         & 73.44 $\pm$1.79 & 52.21 $\pm$1.21 & 56.64 $\pm$2.62 & 24.30 $\pm$1.26 \\
BioMedCLIP     & 84.67$\pm$2.23 & 58.23$\pm$2.85 & 74.07$\pm$3.55 &  56.89$\pm$3.94 \\
RETFound-MAE       & 84.25 $\pm$2.56 & 57.23 $\pm$2.15 & 71.31$\pm$3.99 & 50.80 $\pm$3.65 \\
RETFound-DINOv2       & 81.76 $\pm$1.24 & \textbf{61.70} $\pm$2.14 & 74.92 $\pm$2.02 & 47.20 $\pm$2.80 \\
\hline

\multicolumn{5}{c}{\textbf{IDRiD (LS)}} \\
\hline
ImageNet21k    & \textbf{85.10} $\pm$1.60 & 57.68 $\pm$2.35 & \textbf{79.56} $\pm$3.82 & \textbf{65.60} $\pm$5.99 \\
DINOv2         & 81.02 $\pm$1.77 & 56.09 $\pm$1.52 & 80.28 $\pm$2.17 & 52.10 $\pm$3.90 \\
BioMedCLIP     & 84.68 $\pm$2.33 & 58.41 $\pm$2.55 & 74.71 $\pm$4.34 & 57.16 $\pm$4.82 \\
RETFound-MAE       & 84.33 $\pm$2.39 & 52.34 $\pm$2.35 & 70.56 $\pm$4.40 & 50.04 $\pm$4.11 \\
RETFound-DINOv2       & 81.74 $\pm$1.77 & \textbf{61.73} $\pm$2.21 & 74.96 $\pm$2.19 & 46.55 $\pm$3.21 \\
\hline

\multicolumn{5}{c}{\textbf{APTOS2019}} \\
\hline
ImageNet21k    & 86.28 $\pm$1.14 & 51.96 $\pm$1.11 & 73.00 $\pm$3.07 & 37.87 $\pm$2.93 \\
DINOv2         & 88.89 $\pm$1.80 & 60.42 $\pm$1.66 & 90.40 $\pm$1.73 & 57.10 $\pm$4.38 \\
BioMedCLIP     & 90.33 $\pm$1.85 & 67.89 $\pm$2.44 & 90.64 $\pm$1.74 & 60.98 $\pm$4.62 \\
RETFound-MAE       & 87.97 $\pm$1.72 & 57.22 $\pm$0.95 & 85.00 $\pm$1.79 & 49.80 $\pm$3.79 \\
RETFound-DINOv2       & \textbf{90.58} $\pm$1.36 & \textbf{70.89} $\pm$2.87 & \textbf{92.16} $\pm$1.79 & \textbf{64.65} $\pm$5.03 \\
\hline

\multicolumn{5}{c}{\textbf{APTOS2019 (T)}} \\
\hline
ImageNet21k    & 86.46 $\pm$1.22 & 51.96 $\pm$0.96 & 73.12 $\pm$3.76 & 37.92 $\pm$3.70 \\
DINOv2         & 87.39 $\pm$1.17 & 51.10 $\pm$1.02 & 60.13 $\pm$2.35 & 23.76 $\pm$1.21 \\
BioMedCLIP     & 90.44 $\pm$1.45 & 68.15 $\pm$3.15 & 90.77 $\pm$1.68 & 61.39 $\pm$4.45 \\
RETFound-MAE       & 88.04 $\pm$1.70 & 57.17 $\pm$0.76 & 84.84 $\pm$2.27 & 49.58 $\pm$4.12 \\
RETFound-DINOv2       & \textbf{90.67} $\pm$1.51 & \textbf{71.09} $\pm$3.72 & \textbf{92.34} $\pm$1.70 & \textbf{65.18} $\pm$6.63 \\
\hline

\multicolumn{5}{c}{\textbf{APTOS2019 (LS)}} \\
\hline
ImageNet21k    & 86.44 $\pm$1.35 & 52.02 $\pm$1.15 & 73.12 $\pm$3.20 & 37.96 $\pm$3.33 \\
DINOv2         & 88.80 $\pm$1.49 & 60.15 $\pm$1.65 & 90.46 $\pm$1.68 & 57.24 $\pm$4.17 \\
BioMedCLIP     & 90.46 $\pm$1.78 & 67.68 $\pm$3.39 & 90.63 $\pm$2.12 & 61.02 $\pm$5.12 \\
RETFound-MAE       & 87.74 $\pm$1.67 & 57.26 $\pm$0.89 & 84.62 $\pm$1.86 & 48.51 $\pm$3.54 \\
RETFound-DINOv2       & \textbf{90.90} $\pm$2.17 & \textbf{71.49} $\pm$3.78 & \textbf{92.06} $\pm$1.47 & \textbf{64.81} $\pm$5.16 \\

\bottomrule
\end{tabular}
}
\end{table*}

\begin{table*}[!ht]
\centering
\caption{Histopathology Datasets Performance Evaluation}
\label{tab:combined_histopathology_perf}
\resizebox{0.7\textwidth}{!}{%
\begin{tabular}{@{}>{\centering\arraybackslash}p{2.5cm}cccc@{}}
\toprule
\textbf{Model} & \textbf{Acc $\uparrow$} & \textbf{BAcc $\uparrow$} & \textbf{AUROC $\uparrow$} & \textbf{AUPRC $\uparrow$} \\
\hline

\multicolumn{5}{c}{\textbf{CRC100K}} \\
\hline
ImageNet21k & 68.97$\pm$1.83 & 65.85$\pm$1.42 & 95.28$\pm$0.40 & 80.72$\pm$1.20 \\
DINOv2      & 84.82$\pm$1.23 & 82.51$\pm$1.22 & 98.69$\pm$0.15 & 86.76$\pm$1.26 \\
BioMedCLIP  & 77.44$\pm$1.14 & 74.99$\pm$1.36 & 97.71$\pm$0.30 & 85.52$\pm$1.18 \\
CTransPath  & 86.12$\pm$1.02 & 84.30$\pm$0.94 & \textbf{99.38}$\pm$0.10 & 93.68$\pm$0.89 \\
UNI         & \textbf{86.78}$\pm$0.99 & \textbf{87.34}$\pm$0.89 & 98.99$\pm$0.20 & \textbf{94.09}$\pm$0.89 \\
\hline

\multicolumn{5}{c}{\textbf{CRC100K (T)}} \\
\hline
ImageNet21k & 68.90 $\pm$1.43 & 65.76 $\pm$1.36 & 95.34 $\pm$0.39 & 81.46 $\pm$1.20 \\
DINOv2      & 84.73 $\pm$1.14 & 82.44 $\pm$1.36 & 98.70 $\pm$0.15 & 86.70 $\pm$1.25 \\
BioMedCLIP  & 77.26 $\pm$1.30 & 74.82 $\pm$1.21 & 97.70 $\pm$0.24 & 85.43 $\pm$1.14 \\
CTransPath  & 85.38 $\pm$1.01 & 83.48 $\pm$1.17 & \textbf{99.13} $\pm$0.11 & 91.82 $\pm$0.94 \\
UNI         & \textbf{86.91} $\pm$1.03 & \textbf{87.44} $\pm$1.03 & 99.00 $\pm$0.20 & \textbf{94.12} $\pm$0.91 \\
\hline

\multicolumn{5}{c}{\textbf{CRC100K (LS)}} \\
\hline
ImageNet21k & 71.36 $\pm$1.37 & 67.80 $\pm$1.38 & 95.31 $\pm$0.65 & 82.32 $\pm$1.22 \\
DINOv2      & \textbf{88.18} $\pm$1.22 & \textbf{85.83} $\pm$1.26 & 98.65 $\pm$0.19 & 89.37 $\pm$0.99 \\
BioMedCLIP  & 78.11 $\pm$1.34 & 75.49 $\pm$1.40 & 97.14 $\pm$0.32 & 85.22 $\pm$1.08 \\
CTransPath  & 84.92 $\pm$1.22 & 83.03 $\pm$1.32 & \textbf{99.03} $\pm$0.13 & 90.91 $\pm$1.10 \\
UNI         & 85.86 $\pm$1.12 & 85.35 $\pm$0.75 & 98.90 $\pm$0.17 & \textbf{94.84} $\pm$0.80 \\

\hline

\multicolumn{5}{c}{\textbf{TCGA}} \\
\hline
ImageNet21k & 67.69$\pm$1.32 & 58.16$\pm$2.21 & 96.04$\pm$0.46 & 63.16$\pm$2.09 \\
DINOv2      & 70.02$\pm$1.22 & 59.97$\pm$1.71 & 96.62$\pm$0.34 & 65.79$\pm$1.26 \\
BioMedCLIP  & 65.52$\pm$1.16 & 56.08$\pm$1.51 & 95.80$\pm$0.56 & 61.16$\pm$1.89 \\
CTransPath  & 75.89$\pm$1.17 & 65.61$\pm$1.71 & 97.71$\pm$0.39 & 72.50$\pm$1.52 \\
UNI         & \textbf{81.64}$\pm$0.95 & \textbf{74.07}$\pm$1.46 & \textbf{98.57}$\pm$0.24 & \textbf{80.44}$\pm$1.50 \\
\hline

\multicolumn{5}{c}{\textbf{TCGA (T)}} \\
\hline
ImageNet21k & 67.71 $\pm$1.23 & 58.03 $\pm$1.42 & 96.06 $\pm$0.43 & 62.95 $\pm$1.51 \\
DINOv2      & 69.98 $\pm$1.47 & 59.86 $\pm$1.78 & 96.63 $\pm$0.35 & 65.70 $\pm$1.69 \\
BioMedCLIP  & 65.46 $\pm$1.29 & 56.10 $\pm$1.84 & 95.77 $\pm$0.40 & 61.19 $\pm$1.50 \\
CTransPath  & 75.88 $\pm$1.24 & 65.37 $\pm$1.80 & 97.71 $\pm$0.32 & 72.30 $\pm$1.94 \\
UNI         & \textbf{81.70} $\pm$1.09 & \textbf{74.14} $\pm$2.04 & \textbf{98.59} $\pm$0.34 & \textbf{80.51} $\pm$1.45 \\
\hline

\multicolumn{5}{c}{\textbf{TCGA (LS)}} \\
\hline
ImageNet21k & 67.71 $\pm$0.98 & 56.40 $\pm$1.93 & 95.49 $\pm$0.55 & 62.09 $\pm$1.69 \\
DINOv2      & 69.85 $\pm$0.92 & 58.34 $\pm$1.34 & 96.21 $\pm$0.29 & 65.03 $\pm$1.46 \\
BioMedCLIP  & 65.46 $\pm$1.19 & 54.14 $\pm$1.35 & 95.21 $\pm$0.52 & 60.08 $\pm$1.48 \\
CTransPaths & 75.40 $\pm$0.86 & 63.35 $\pm$1.65 & 97.41 $\pm$0.42 & 70.99 $\pm$1.61 \\
UNI         & \textbf{81.90} $\pm$1.07 & \textbf{73.36} $\pm$1.55 & \textbf{98.16} $\pm$0.33 & \textbf{79.80} $\pm$1.43 \\

\hline

\multicolumn{5}{c}{\textbf{BraTS-Path}} \\
\hline
ImageNet21k & 87.32$\pm$1.18 & 87.33$\pm$1.16 & 98.36$\pm$0.24 & 93.61$\pm$0.93 \\
DINOv2      & 86.94$\pm$0.95 & 86.93$\pm$0.87 & 98.29$\pm$0.29 & 93.51$\pm$0.85 \\
BioMedCLIP  & 83.95$\pm$1.42 & 83.95$\pm$1.37 & 97.79$\pm$0.27 & 91.53$\pm$1.00 \\
CTransPath  & 87.43$\pm$1.10 & 87.43$\pm$1.09 & 98.53$\pm$0.25 & 94.11$\pm$0.91 \\
UNI         & \textbf{93.74}$\pm$0.73 & \textbf{93.76}$\pm$0.71 & \textbf{99.50}$\pm$1.74 & \textbf{97.92}$\pm$0.58 \\
\hline

\multicolumn{5}{c}{\textbf{BraTS-Path (T)}} \\
\hline
ImageNet21k & 87.30 $\pm$1.03 & 87.30 $\pm$0.96 & 98.35 $\pm$0.24 & 93.61 $\pm$0.89 \\
DINOv2      & 86.85 $\pm$1.02 & 86.86 $\pm$0.92 & 98.26 $\pm$0.24 & 93.46 $\pm$0.82 \\
BioMedCLIP  & 83.96 $\pm$1.35 & 83.96 $\pm$1.36 & 97.78 $\pm$0.33 & 91.51 $\pm$1.07 \\
CTransPath  & 87.21 $\pm$1.04 & 87.22 $\pm$1.05 & 98.56 $\pm$0.20 & 94.22 $\pm$0.71 \\
UNI         & \textbf{93.72} $\pm$0.77 & \textbf{93.71} $\pm$0.74 & \textbf{99.51} $\pm$0.09 & \textbf{97.95} $\pm$0.38 \\
\hline

\multicolumn{5}{c}{\textbf{BraTS-Path (LS)}} \\
\hline
ImageNet21k & 88.50 $\pm$1.09 & 88.51 $\pm$1.12 & 98.40 $\pm$0.22 & 94.15 $\pm$0.80 \\
DINOv2      & 86.49 $\pm$1.10 & 86.50 $\pm$1.14 & 98.20 $\pm$0.25 & 93.26 $\pm$0.90 \\
BioMedCLIP  & 84.90 $\pm$1.40 & 84.89 $\pm$1.37 & 97.80 $\pm$0.35 & 91.92 $\pm$1.16 \\
CTransPath  & 86.87 $\pm$1.38 & 86.87 $\pm$1.43 & 98.39 $\pm$0.23 & 93.59 $\pm$0.78 \\
UNI         & \textbf{93.60} $\pm$0.70 & \textbf{93.60} $\pm$0.73 & \textbf{99.45} $\pm$0.14 & \textbf{97.87} $\pm$0.37 \\

\bottomrule
\end{tabular}
}
\end{table*}

\begin{table*}[!ht]
\centering
\caption{Chest X-Rays Datasets Performance Evaluation}
\label{tab:combined_xray_perf}
\resizebox{0.7\textwidth}{!}{%
\begin{tabular}{@{}>{\centering\arraybackslash}p{2.5cm}cccc@{}}
\toprule
\textbf{Model} & \textbf{Acc $\uparrow$} & \textbf{BAcc $\uparrow$} & \textbf{AUROC $\uparrow$} & \textbf{AUPRC $\uparrow$} \\
\hline

\multicolumn{5}{c}{\textbf{RSNA-Pneumonia}} \\
\hline
ImageNet21k   & 79.45 $\pm$1.58 & 58.39 $\pm$1.62 & 81.53 $\pm$1.86 & 57.10 $\pm$4.28 \\
DINOv2      & 79.49 $\pm$1.43 & 57.47 $\pm$1.70 & 81.33 $\pm$1.82 & 56.82 $\pm$5.20 \\
BioMedCLIP  & 83.35 $\pm$1.22 & 67.88 $\pm$1.97 & 86.64 $\pm$1.42 & 68.65 $\pm$3.76 \\
MRM   & 81.10 $\pm$1.57 & 60.97 $\pm$1.49 & 84.19 $\pm$1.82 & 64.08 $\pm$3.95 \\
Rad-DINO   & \textbf{83.89} $\pm$1.41 & \textbf{73.40} $\pm$2.45 & \textbf{87.57} $\pm$1.45 & \textbf{68.93} $\pm$4.03 \\
\hline

\multicolumn{5}{c}{\textbf{RSNA-Pneumonia (T)}} \\
\hline
ImageNet21k   & 79.64 $\pm$1.30 & 58.37 $\pm$1.43 & 81.68 $\pm$1.59 & 57.21 $\pm$3.21 \\
DINOv2      & 79.67 $\pm$1.35 & 57.57 $\pm$1.42 & 81.39 $\pm$1.87 & 56.77 $\pm$3.65 \\
BioMedCLIP  & 83.38 $\pm$1.52 & 67.84 $\pm$1.98 & 86.63 $\pm$1.56 & 68.46 $\pm$3.04 \\
MRM   & 81.23 $\pm$1.70 & 61.14 $\pm$1.56 & 84.06 $\pm$1.77 & 63.91 $\pm$4.37 \\
Rad-DINO   & \textbf{83.68} $\pm$1.09 & \textbf{74.21} $\pm$1.55 & \textbf{87.48} $\pm$1.59 & \textbf{68.93} $\pm$3.06 \\
\hline

\multicolumn{5}{c}{\textbf{RSNA-Pneumonia (LS)}} \\
\hline
ImageNet21k   & 79.47 $\pm$1.65 & 58.31 $\pm$1.35 & 80.81 $\pm$1.96 & 55.22 $\pm$4.26 \\
DINOv2      & 79.38 $\pm$1.53 & 57.10 $\pm$1.43 & 81.21 $\pm$1.54 & 56.94 $\pm$3.73 \\
BioMedCLIP  & 83.22 $\pm$1.21 & 67.31 $\pm$1.84 & 86.64 $\pm$1.51 & 68.53 $\pm$4.41 \\
MRM   & 81.51 $\pm$1.39 & 61.73 $\pm$1.39 & 84.50 $\pm$1.68 & 64.43 $\pm$3.67 \\
Rad-DINO   & \textbf{83.88} $\pm$1.29 & \textbf{74.40} $\pm$2.15 & \textbf{87.38} $\pm$1.80 & \textbf{69.08} $\pm$4.25 \\
\hline

\multicolumn{5}{c}{\textbf{POLCOVID}} \\
\hline
ImageNet21k   & 75.96 $\pm$2.26 & 53.85 $\pm$1.20 & 77.92 $\pm$2.48 & 61.60 $\pm$3.30 \\
DINOv2      & 78.87 $\pm$2.68 & 65.37 $\pm$2.87 & 82.69 $\pm$2.52 & 68.73 $\pm$4.16 \\
BioMedCLIP  & 82.25 $\pm$3.11 & 73.81 $\pm$3.41 & 86.03 $\pm$2.42 & 75.11 $\pm$4.30 \\
MRM   & 76.88 $\pm$2.58 & 56.79 $\pm$1.79 & 79.17 $\pm$2.50 & 62.95 $\pm$3.46 \\
Rad-DINO   & \textbf{87.00} $\pm$1.85 & \textbf{81.68} $\pm$2.64 & \textbf{92.63} $\pm$1.56 & \textbf{85.59} $\pm$3.14 \\
\hline

\multicolumn{5}{c}{\textbf{POLCOVID (T)}} \\
\hline
ImageNet21k   & 76.28 $\pm$2.38 & 54.03 $\pm$1.62 & 78.19 $\pm$3.03 & 61.98 $\pm$4.62 \\
DINOv2      & 78.68 $\pm$2.90 & 65.17 $\pm$3.30 & 82.71 $\pm$2.67 & 68.56 $\pm$4.50 \\
BioMedCLIP  & 82.65 $\pm$2.50 & 74.24 $\pm$2.81 & 86.11 $\pm$2.43 & 75.35 $\pm$4.02 \\
MRM   & 76.85 $\pm$2.30 & 56.78 $\pm$1.99 & 78.87 $\pm$2.48 & 62.72 $\pm$3.51 \\
Rad-DINO   & \textbf{86.95} $\pm$2.15 & \textbf{81.41} $\pm$3.25 & \textbf{92.68} $\pm$2.16 & \textbf{85.66} $\pm$4.76 \\
\hline

\multicolumn{5}{c}{\textbf{POLCOVID (LS)}} \\
\hline
ImageNet21k   & 76.70 $\pm$2.51 & 54.92 $\pm$1.56 & 77.75 $\pm$2.64 & 61.52 $\pm$3.41 \\
DINOv2      & 78.82 $\pm$3.13 & 65.00 $\pm$4.02 & 82.62 $\pm$3.26 & 68.63 $\pm$4.67 \\
BioMedCLIP  & 82.35 $\pm$2.66 & 73.85 $\pm$3.36 & 85.98 $\pm$2.90 & 75.03 $\pm$4.37 \\
MRM   & 77.15 $\pm$2.16 & 57.34 $\pm$1.96 & 79.36 $\pm$2.83 & 63.39 $\pm$4.54 \\
Rad-DINO   & \textbf{86.81} $\pm$2.47 & \textbf{81.47} $\pm$3.50 & \textbf{92.51} $\pm$1.88 & \textbf{85.34} $\pm$3.64 \\
\hline

\multicolumn{5}{c}{\textbf{COVID-Rad}} \\
\hline
ImageNet21k   & 87.56 $\pm$0.90 & 66.36 $\pm$1.43 & 93.54 $\pm$0.81 & 85.31 $\pm$1.63 \\
DINOv2      & 87.97 $\pm$0.72 & 65.42 $\pm$1.47 & 92.62 $\pm$0.94 & 
82.95 $\pm$2.09 \\
BioMedCLIP  & 94.63 $\pm$0.60 & 89.04 $\pm$1.29 & 97.44 $\pm$0.47 & 94.33 $\pm$1.12 \\
MRM   & 87.25 $\pm$0.87 & 67.23 $\pm$1.33 & 93.83 $\pm$0.64 & 85.07 $\pm$1.70 \\
Rad-DINO   & \textbf{97.59} $\pm$0.37 & \textbf{95.48} $\pm$0.73 & \textbf{99.20} $\pm$0.22 & \textbf{98.24} $\pm$0.24 \\
\hline

\multicolumn{5}{c}{\textbf{COVID-Rad (T)}} \\
\hline
ImageNet21k   & 87.53 $\pm$1.01 & 66.09 $\pm$1.56 & 93.51 $\pm$0.69 & 85.10 $\pm$1.96 \\
DINOv2      & 87.94 $\pm$0.80 & 65.43 $\pm$1.40 & 92.64 $\pm$0.86 & 82.98 $\pm$1.73 \\
BioMedCLIP  & 94.57 $\pm$0.60 & 88.86 $\pm$1.40 & 97.42 $\pm$0.51 & 94.20 $\pm$1.07 \\
MRM   & 87.21 $\pm$0.87 & 67.14 $\pm$1.27 & 93.75 $\pm$0.86 &  
84.95 $\pm$1.66 \\
Rad-DINO   & \textbf{97.58} $\pm$0.40 & \textbf{95.48} $\pm$0.83 &  \textbf{99.20} $\pm$0.22 & \textbf{98.23} $\pm$0.48 \\
\hline

\multicolumn{5}{c}{\textbf{COVID-Rad (LS)}} \\
\hline
ImageNet21k   & 87.45 $\pm$0.87 & 65.82 $\pm$1.45 & 93.56 $\pm$0.74 & 85.49 $\pm$1.87 \\
DINOv2      & 87.92 $\pm$0.73 & 65.20 $\pm$1.36 & 92.64 $\pm$0.77 & 83.13 $\pm$2.29 \\
BioMedCLIP  & 94.54 $\pm$0.76 & 88.66 $\pm$1.30 & 97.42 $\pm$0.50 & 94.19 $\pm$1.07 \\
MRM   & 87.40 $\pm$0.93 & 67.93 $\pm$1.32 & 93.86 $\pm$0.65 & 85.12 $\pm$1.58 \\
Rad-DINO   & \textbf{97.57} $\pm$0.43 & \textbf{95.47} $\pm$0.98 & \textbf{99.20} $\pm$0.16 & \textbf{98.22} $\pm$0.42 \\

\bottomrule
\end{tabular}
}
\end{table*}
\clearpage
\section{Label Smoothing}
\label{apd:label_smoothing}
For label $y$, model probability output $p$, a specified smoothing parameter $\varepsilon\in[0,1]$, number of classes $K$ and class index $j$, the loss of label smoothing can be derived as
\begin{align*}
    \mathcal{L}&=-\sum_{j=1}^{K}y'_j\log p_j\\
    &=-((1-\epsilon+\frac{\epsilon}{K})\log p_k+\sum_{j\neq k}\frac{\epsilon}{K}\log p_j)\\
    &=-(1-\epsilon)\log p_k-\frac{\epsilon}{K}\sum_{j=1}^{K}\log p_j\\
    &=\mathcal{L}_{CE}+\frac{\epsilon}{K}\sum_{j=1}^{K}(-\log p_j)\\
    &=\mathcal{L}_{CE}+\epsilon\mathbb{E}_\textbf{u}[-\log p_j]\\
    &(\mathbb{E}_\textbf{u}[-\log p_j]=\frac{1}{K}\sum_{j=1}^{K}(-\log p_j)\\
    &\text{  for uniform distribution $\textbf{u}$ with $u_j=\frac{1}{K}$})
\end{align*}



\end{document}